\documentclass[10pt,twocolumn,letterpaper]{article}

\usepackage[pagenumbers]{cvpr} % To force page numbers, e.g. for an arXiv version

\usepackage[dvipsnames]{xcolor}
\usepackage{multirow}
\usepackage{makecell}
\usepackage[accsupp]{axessibility} % Improves PDF readability for those with visual impairments.

\definecolor{cvprblue}{rgb}{0.21,0.49,0.74}
\usepackage[pagebackref,breaklinks,colorlinks,citecolor=cvprblue]{hyperref}

\newcommand{\methodname}{2S-UDF}
\graphicspath{{media}}
\newtheorem{theorem}{Theorem}

\title{\methodname{}: A Novel Two-stage UDF Learning Method for Robust Non-watertight Model Reconstruction from Multi-view Images
} %

\author {
\textbf{Junkai Deng}\textsuperscript{1,2}
\quad
\textbf{Fei Hou}\textsuperscript{1,2}\thanks{Corresponding author}
\quad
\textbf{Xuhui Chen}\textsuperscript{1,2}
\quad
\textbf{Wencheng Wang}\textsuperscript{1,2}
\quad
\textbf{Ying He}\textsuperscript{3} \\ %
\textsuperscript{1}State Key Laboratory of Computer Science, Institute of Software, Chinese Academy of Sciences \\
\textsuperscript{2}University of Chinese Academy of Sciences \\
\textsuperscript{3}School of Computer Science and Engineering, Nanyang Technological University \\
{\tt\small \{dengjk, houfei, chenxh, whn\}@ios.ac.cn} 
\quad
{\tt\small yhe@ntu.edu.sg}
\quad
}

\begin{document}
\maketitle

\begin{abstract}
Recently, building on the foundation of neural radiance field, various techniques have emerged to learn unsigned distance fields (UDF) to reconstruct 3D non-watertight models from multi-view images. Yet, a central challenge in UDF-based volume rendering is formulating a proper way to convert unsigned distance values into volume density, ensuring that the resulting weight function remains unbiased and sensitive to occlusions. Falling short on these requirements often results in incorrect topology or large reconstruction errors in resulting models. This paper addresses this challenge by presenting a novel two-stage algorithm, \methodname{}, for learning a high-quality UDF from multi-view images. Initially, the method applies an easily trainable density function that, while slightly biased and transparent, aids in coarse reconstruction. The subsequent stage then refines the geometry and appearance of the object to achieve a high-quality reconstruction by directly adjusting the weight function used in volume rendering to ensure that it is unbiased and occlusion-aware. Decoupling density and weight in two stages makes our training stable and robust, distinguishing our technique from existing UDF learning approaches. Evaluations on the DeepFashion3D, DTU, and BlendedMVS datasets validate the robustness and effectiveness of our proposed approach. In both quantitative metrics and visual quality, the results indicate our superior performance over other UDF learning techniques in reconstructing 3D non-watertight models from multi-view images. Our code is available at \url{https://bitbucket.org/jkdeng/2sudf/}.
\end{abstract}    
\section{Introduction}
\begin{figure}[t]
    \centering
    \begin{tabular}{cccc}
        GT & Ours & NeuralUDF & NeUDF \\
        \includegraphics[width=.6in]{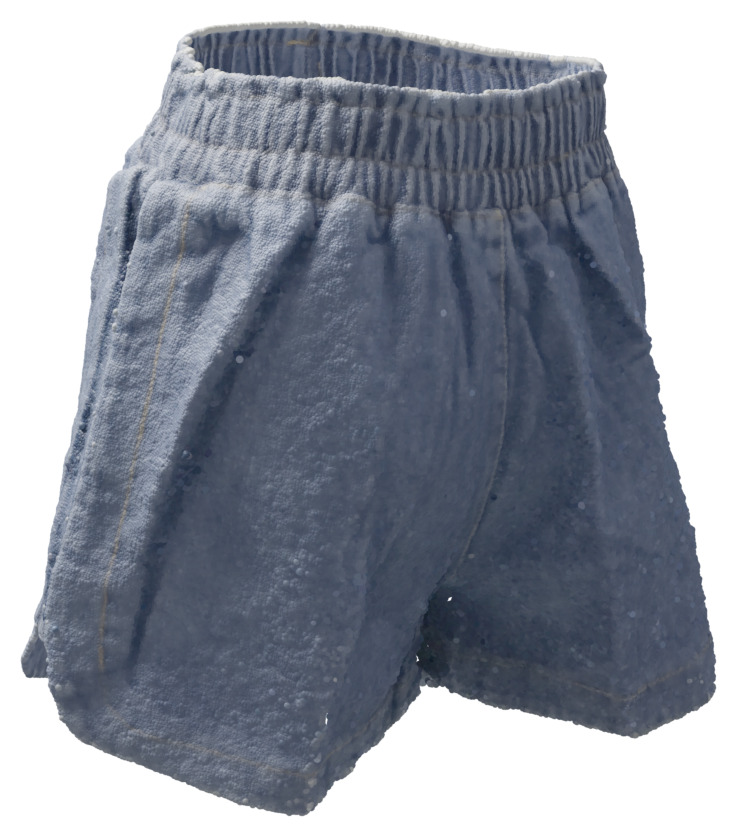} & \includegraphics[width=.6in]{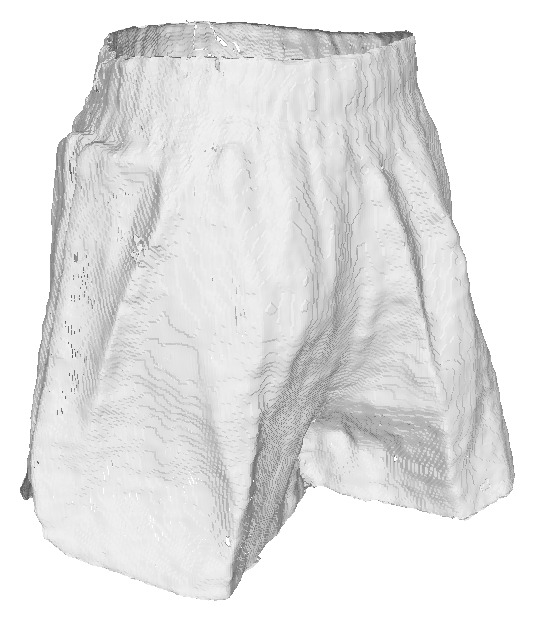} & \includegraphics[width=.6in]{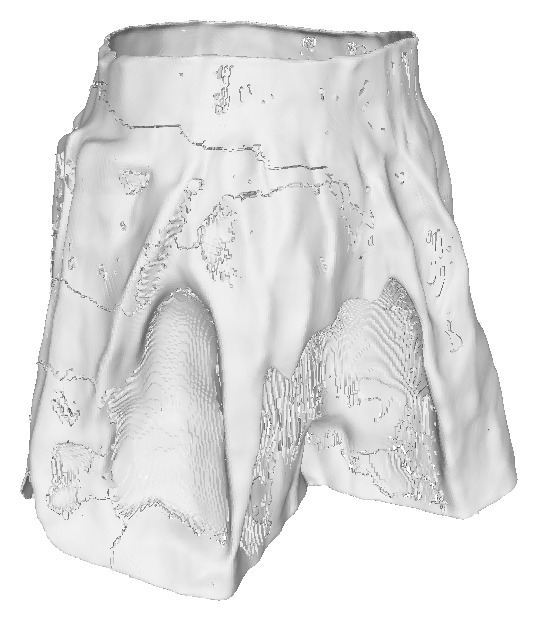} & \includegraphics[width=.6in]{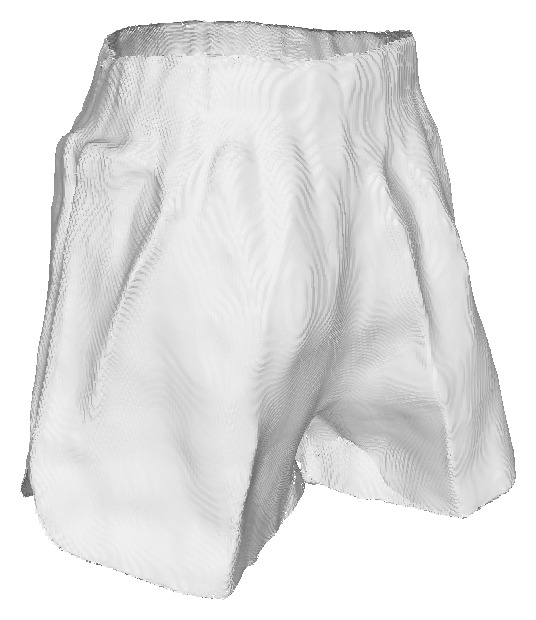} \\
        \includegraphics[width=.6in]{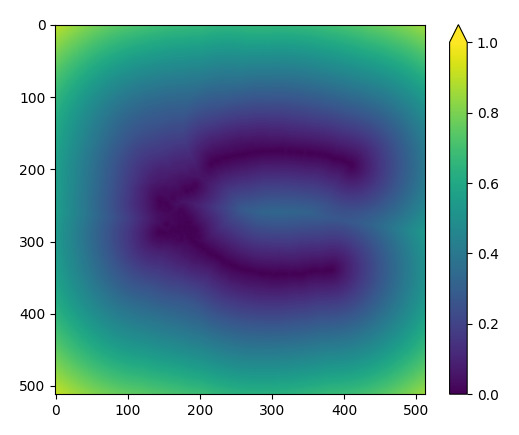} & \includegraphics[width=.6in]{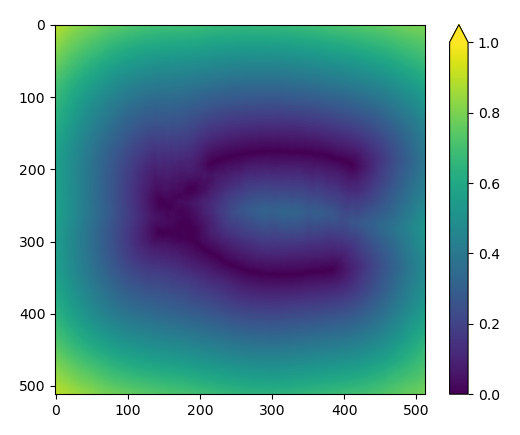} & \includegraphics[width=.6in]{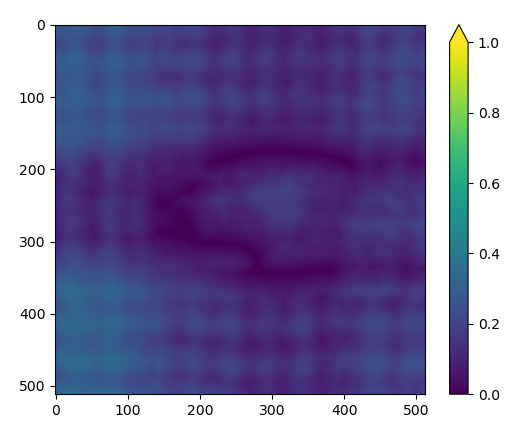} & \includegraphics[width=.6in]{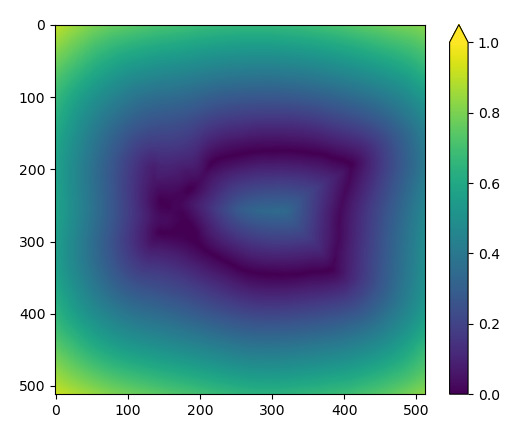} \\
    \end{tabular}
    \caption{We learn a UDF from multiview images for non-watertight model reconstruction. As illustrated in the cross sections of learned UDFs, our learned UDF approximates to the ground truth. In contrast, the learned UDF of NeuralUDF~\cite{Long2023} is choppy leading to significant artifacts, e.g., unexpected pit. The learned UDF of NeUDF~\cite{Liu2023NeUDF} is almost closed struggling to generate open surface.}
    \label{fig:teaser}
\end{figure}

As the success of neural radiance field (NeRF)~\cite{Mildenhall2020}, numerous volume rendering based 3D modeling methods are proposed to learn signed distance fields (SDF) for 3D model reconstruction from multi-view images~\cite{Yariv2021,Wang2021,Wang2022HFNeuSIS,Darmon2022}. 
These approaches map signed distance value to a density function, thereby enabling the use of volume rendering to learn an implicit SDF representation. To calculate pixel colors, they compute the weighted sum of radiances along each light ray. 
Achieving an accurate surface depiction requires the density function to meet three essential criteria. Firstly, the weights, which are derived from the density function, must reach their maximum value when the distance is zero, ensuring unbiasedness. Secondly, as a ray traverses through the surface, the accumulated density should tend towards infinity, rendering the surface opaque — a property referred to as occlusion-awareness. Finally, the density function should be bounded to prevent numerical issues.
The popular SDF approaches, such as NeuS~\cite{Wang2021} and VolSDF~\cite{Yariv2021}, adopt an S-shaped density function that meets all these requirements.

While SDF-based methods excel at reconstructing watertight models, they have limitations in representing open models. This is due to the intrinsic nature of SDF, which differentiates between the interior and exterior of a model, thus failing to accommodate open boundaries. Recent advances have attempted to mitigate this constraint by employing unsigned distance fields (UDF)~\cite{Long2023,Liu2023NeUDF,Meng2023}. 
Unlike signed distance fields, UDFs have non-negative distance values, making them suitable for representing non-watertight models. 
However, learning a UDF from multi-view images is a challenging task since the gradients of the UDF are unstable due to directional changes near the zero level-set, making it difficult to train the neural network. Another major challenge lies in formulating a UDF-induced density function that can simultaneously meet the above-mentioned three requirements. Unlike SDFs, UDFs cannot distinguish between the front and back of a surface based on distance values, thus, directly using an S-shaped density function is off the table. Opting for a bell-shaped density function brings its own issues. It is impossible for these integrations to approach infinity, so as to be occlusion-aware, unless the density becomes boundless at zero distance values. These conflicting requirements make UDF learning a non-trivial task, forcing existing methods to sacrifice at least one of these conditions. As shown in Figure~\ref{fig:teaser}, the existing methods NeuralUDF~\cite{Long2023} and NeUDF~\cite{Liu2023NeUDF} result in either choppy or nearly closed UDFs.

As designing a UDF-induced density function that simultaneously fulfills the three aforementioned conditions remains an unresolved challenge, we propose a novel approach that learns a UDF from multi-view images in two separate stages. In the first stage, we apply an easily trainable but slightly biased and transparent density function for coarse reconstruction.
Such a UDF, although being approximate, provides an important clue so that we can determine where to truncate the light rays. 
This accounts for the occlusion effect, where points behind the surface are not visible and should not contribute to the output color. With \textit{truncated} light rays, we are able to derive the weights from UDF directly bypassing the density function, to further refine the geometry and appearance in the second stage.
Our two-stage learning method, called \methodname{}, leads to an unbiased and occlusion-aware weight function. Furthermore, by sidestepping density function learning in Stage 2, we effectively bypass the challenges associated with ensuring its boundedness. This strategy enhances the numerical stability of our method.
Evaluations on benchmark datasets DeepFashion3D~\cite{Zhu2020} and DTU~\cite{Jensen2014} show that \methodname{} outperforms existing UDF learning methods in terms of both reconstruction accuracy and visual quality. Additionally, we observe that the training stability of \methodname{} is notably superior compared to other UDF learning neural networks.

\section{Related Work}
\textbf{3D Reconstruction from Multi-View Images.} Surface reconstruction from multi-view images has been a subject of study for several decades, and can generally be classified into two categories: voxel-based and point-based methods. Voxel-based methods~\cite{Bonet1999,Broadhurst2001,Ji2021,Kar2017,Sun2021} divide the 3D space into voxels and determine which ones belong to the object. These methods can be computationally expensive and may not be suitable for reconstructing complex surfaces.
Point-based methods~\cite{Galliani2015,Schonberger2016,Yao2019} use structure-from-motion~\cite{Hartley2004} to calibrate the images and generate a dense point cloud using multi-view stereo~\cite{Furukawa2015}. Finally, surface reconstruction methods (e.g.,  ~\cite{Bernardini1999,Kazhdan2013,Hou2022}) are used to generate a mesh. Since multi-view stereo requires dense correspondences to generate a dense point cloud, which are often difficult to compute, its results often contain various types of artifacts, such as noise, holes, and incomplete structures.

\textbf{Neural Volume Rendering.} Neural network-based 3D surface reconstruction has received attention in recent years with the emergence of neural rendering~\cite{Mildenhall2020}. Several methods have been proposed for volume rendering and surface reconstruction using neural networks. VolSDF~\cite{Yariv2021} uses the cumulative distribution function of Laplacian distribution to evaluate the density function from SDF for volume rendering and surface reconstruction.
NeuS~\cite{Wang2021} adopts an unbiased density function to the first-order approximation of SDFs for more accurate reconstruction.
SparseNeuS~\cite{Long2022SparseNeuS} extends NeuS to use fewer images for reconstruction.
HF-NeuS~\cite{Wang2022HFNeuSIS} improves NeuS by proposing a simplified and unbiased density function and using hierarchical multilayer perceptrons (MLPs) for detail reconstruction.
Geo-NeuS~\cite{Fu2022} incorporates structure-from-motion to add more constraints.
NeuralWarp~\cite{Darmon2022} improves the accuracy by optimizing consistency between warped views of different images.
PET-NeuS~\cite{Wang_2023_CVPR_PETNeuS} further improves the accuracy by introducing tri-planes into the SDF prediction module, incorporating with MLP.
All these methods learn SDFs, which can only reconstruct watertight models.
Recently, Long \emph{et al}. proposed NeuralUDF~\cite{Long2023} for learning UDF for reconstructing open models.
It adapts the S-shaped density function for learning SDF to UDFs by introducing an indicator function.
However, the indicator function is complicated to learn, and also introduces biases.
Liu \emph{et al}. proposed NeUDF~\cite{Liu2023NeUDF} adopting a bell-shaped density. However, to make it occlusion-aware, the density has to be unbounded resulting in an improper integral, which reduces accuracy.
Meng \emph{et al}. proposed NeAT~\cite{Meng2023} to learn SDF with validity so as to reconstruct open models from SDF.
However, it needs foreground masks for data.

\textbf{3D Reconstruction from Point Clouds.} There has been recent interest in surface representation using signed distance fields (SDFs) and occupation fields. Several methods have been proposed for learning SDFs~\cite{Park2019,Chabra2020,Sitzmann2020,Ma2021,Wang2022}, while occupation fields have been used in methods such as ~\cite{Mescheder2019,Chibane2020a}.
However, both SDFs and occupation fields can only represent watertight models.
To represent non-watertight models, some methods are proposed to learn UDF from 3D point clouds~\cite{Chibane2020b,Zhao2021,Zhou2022}. Our proposed method also uses UDF for non-watertight models representation, but we learn it directly from multi-view images, which is a challenging problem.

\section{Method}
At the foundation of UDF-based learning approaches is the task of crafting a density function that converts unsigned distance values into volume density, ensuring that the resulting weight function is unbiased and responsive to occlusions.
None of the existing UDF learning methods~\cite{Long2023,Liu2023NeUDF} can simultaneously meet the three critical requirements, i.e., ensuring the density function is bounded, and that the weight function remains both unbiased and occlusion aware.

We tackle these challenges by decoupling the density function and weight function across two stages. In the initial stage (Section~\ref{sec:stage1}), we utilize an easy-to-train, bell-shaped density function (which is inherently bounded) to learn a coarse UDF. While the resulting weight function is not theoretically unbiased or occlusion-aware, we can make it practically usable by choosing a proper parameter. Moving into the second stage (Section~\ref{sec:stage2}), we sidestep the density function entirely, focusing instead on refining the UDF by directly adjusting the weight function within the neural volume rendering framework. Specifically, we truncate light rays after they hit the front side of the object and obtain a weight function that is both unbiased and sensitive to occlusions, without the overhang of density function boundedness concerns. Finally, Section~\ref{sec:training} presents the training details.

\subsection{Stage 1: Coarse UDF Learning via a Simple Density Function}
\label{sec:stage1}

We consider the scenario of a single planar plane $\mathcal{M}$ and a single ray-plane intersection. Inspired by HF-NeuS~\cite{Wang2022HFNeuSIS}, we propose an easy-to-learn density function $\sigma_1$ that maps unsigned distance $f$ to density
\begin{equation}
\sigma_1(f(t)) = \frac{cse^{-sf(t)}}{1+e^{-sf(t)}},\;s>0,\;c>0,
\end{equation}
where $c>0$ is a fixed, user-specified parameter and $s>0$ is a learnable parameter controlling the width of the bell-shaped curve. 
Straightforward calculation shows that the weight function $w_1(f(t))=e^{-\int_{0}^t\sigma_1(f(u))\mathrm{d}u}\sigma_1(f(t))$ is monotonically decreasing behind the plane $\mathcal{M}$ and the maximum value occurs at a point $t^{*}$ in front of $\mathcal{M}$ with an unsigned distance value of $f(t^*)=\frac{1}{s}\ln\frac{c}{\left\lvert\cos (\theta)\right\lvert}, (c>|\cos (\theta)|)$ or $f(t^*)=0, (0<c\leq |\cos (\theta)|)$,
where $\theta$ is the incident angle between the light ray and the surface normal. 
This means that the weight function $w_1$ is not unbiased.
Furthermore, the line integral $\int_{0}^t\sigma_1(f(u))\mathrm{d}u$ does not approach infinity when a light ray passes through the front-most layer of the surface, indicating $w_1$ is only partially occlusion-aware.

While the density function $\sigma_1$ is not perfect in theory, by selecting an appropriate $c$, we can practically minimize bias and enhance opacity. Clearly, a smaller $c$ value decreases $f(t^*)$, thereby reducing bias. To gauge the effect of $c$ on opacity, we now consider the most extreme scenario where the incident light ray is perpendicular to the planar surface $\mathcal{M}$, and assume that the intersection point is located at $t=1$. In such a situation, the unsigned distance function is $f(t)=1-t$ for points in front of $\mathcal{M}$. Since $\sigma_1$ is symmetrical on either side of $\mathcal{M}$, the surface transparency is the square of the transparency of the front side. The theoretic transparency is,
\begin{equation*}
\begin{aligned}
     \left(e^{-\int_0^1\hat{\sigma_1}(f(t))\mathrm{d}t}\right)^2
    &=
    \left[\exp\left(-\int^1_0\frac{cse^{-s(1-t)}}{1+e^{-s(1-t)}}\mathrm{d} t\right)\right]^2\\
    &=\left(\frac{1+e^{-s}}{2}\right)^{2c}.
\end{aligned}
\end{equation*}
Therefore, we should choose a relatively large $c$ to reduce transparency. In our implementation, we set the constant $c=5$ based on the typical value of the learned parameter $s$ which usually ranges between $1000$ and $2000$. Calculations of bias and translucency show that this setting offers a good balance between occlusion-awareness and unbiasedness in the first stage training. Please refer to the supplementary material for a detailed analysis.

\subsection{Stage 2: UDF Refinement through  Weight Adjustment}
\label{sec:stage2}

In this stage, we refine the UDF learned in Stage 1 to improve the quality of geometry and appearance. Unlike Stage 1 and all other UDF-learning methods, inspired by~\cite{Azinovic2022}, we truncate light rays based on the approximated UDF learned in Stage 1 and learn the weight function $w(t)$ directly instead of the density function $\sigma(t)$ to refine the UDF. 

Ideally, for a single ray-plane intersection, we want a bell-shaped function $w(t)$ that attains its maximum at the points with zero distance values, and satisfies partition of unity. Therefore, we adopt the derivative of the sigmoid function as the weight function~\cite{Azinovic2022}, defined as 
\begin{equation}
    w_2(f(t)) = \frac{se^{-sf(t)}}{(1 + e^{-sf(t)})^2} \cdot |\cos(\theta)|.
\end{equation}
with $\theta$ being the incident angle between the light ray and the surface normal.

Intuitively speaking, learning such a  weight function $w_2$ in Stage 2 of our UDF method is similar to learning an S-shaped density function in SDF-based approaches, such as~\cite{Wang2022HFNeuSIS}. As a result, the learning process in Stage 2 is as stable as those SDF approaches. Furthermore, it can totally avoid using the visibility indicator function, which is necessary in NeuralUDF~\cite{Long2023}.

Calculation shows that the weight $w_2$ attains its maximum at zero distance values, therefore it is unbiased. However, if we naively predict the weight function directly, it will not be occlusion-aware, so we introduce the ray truncation. To make $w_2$ occlusion-aware, we can truncate the light rays after they pass through the frontmost layer of the surface, thereby preventing rendering the interior of the object. Note that we do not expect the truncation to be exactly on the frontmost layer of the surface. In fact, as long as it occurs between the frontmost layer and the second layer, we consider the truncation valid. This means that the approximate UDF learned in the first stage, which can capture the main topological features (such as boundaries) and provide a fairly good representation of the target object, is sufficient for us to determine where to cut off the light rays. 

In our implementation, we adopt a simple strategy to determine the truncation point for each light ray. Specifically, the  truncation point of ray $\bf r$ is the first sample point along $\bf r$ such that 
\begin{itemize}
    \item The unsigned distance value at the point is a local maxima. To avoid distance vibration interference, it should be the maximum in a window centered at the point. And
    \item The accumulated weight up to this point is greater than $\delta_{thres}$.
\end{itemize}
The accumulated weight threshold $\delta_{thres}$ is intuitively set to 0.5. This choice is based on the assumption that if the Stage 1 training is performed well enough, the accumulated weights at each sample point along the ray would be either 0 (for not reaching a surface) or 1 (for having intersected with a surface). Hence, we intuitively select 0.5 for $\delta_{thres}$ because it is the midpoint between 0 and 1.
With the cutoff mechanism, only the first ray-surface intersection contributes to the color of the ray, effectively achieving occlusion-awareness. Given these properties, we conclude that,
\begin{theorem}
    The weight $w_2$ with light cutting off is unbiased and occlusion-aware.
\end{theorem}

\begin{figure}[ht]
    \centering
    \includegraphics[width=3.in]{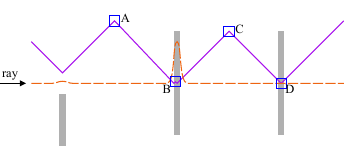}
    \caption{An intuitive illustration of our ray cutting algorithm, best viewed in color and magnified. A ray shoots from left to right, approaching the boundary of the first surface, and going through another two surfaces (gray boxes). The violet solid line represents the UDF values along the ray; the orange dashed line represents the corresponding color weight.
    }
    \label{fig:cutoff-fig}
\end{figure}

Figure~\ref{fig:cutoff-fig} is an intuitive illustration of our Stage 2 weight learning and truncation strategy. The UDF maxima point $A$ in front of the intersection surface would not affect the cutting point selection as the accumulated weight is below $\delta_{thres}$ (0.5). The local maxima $B$ due to UDF oscillation also would not affect it since it's not the maximum in a large enough neighborhood. The light is cut at maxima point $C$, and thus the weight of point $D$ is zero without contributions to the rendering. As illustrated in Figure~\ref{fig:cutoff-fig}, the cutting process is robust against UDF oscillation, open boundaries, and local maxima in front of the intersection surface.

\subsection{Training}
\label{sec:training}

\textbf{Differentiable UDFs.} NeuS uses an MLP network to learn the signed distance function $f$, which is a differentiable function. In contrast, UDF is not differentiable at the zero level set, 
making the network difficult to learn the values and gradients of the UDF close to the zero level set. 

Another crucial requirement is to ensure non-negative values for the computed distances, which seems like a trivial task as one may simply apply absolute value or normalization such as ReLU~\cite{Fukushima1975} to the MLP output. However, applying the absolute value to the distance is not viable due to its non-differentiability at zero. Similarly, normalizing the output value using ReLU is not feasible as it is also non-differentiable at zero and its gradient vanishes for negative inputs. This can be particularly problematic for learning UDFs, since when the MLP returns a negative distance value, the ReLU gradient vanishes, hindering the update of the distance to a positive value in the subsequent iterations.

We add a softplus~\cite{Dugas2000} function after the output layer of the MLP~\cite{Liu2023NeUDF}.
The softplus function is a smooth and differentiable approximation of the ReLU function, which is defined as
$\text{softplus}(x) = \frac{1}{\beta} \ln(1 + e^{\beta x}).$
Softplus has the same shape as ReLU, but it is continuous and differentiable at every point and its gradients do not vanish anywhere. Using the softplus function allows us to ensure that the output of the MLP is non-negative and differentiable, making it suitable for learning the UDF. Similar to NeUDF~\cite{Liu2023NeUDF}, we set $\beta=100$ in our experiments.

\textbf{Loss functions.} 
Following NeuralUDF~\cite{Long2023}, we adopt an iso-surface regularizer to penalize the UDF values of the non-surface points from being zero, therefore encouraging smooth and clean UDFs. The regularization loss is defined as~\cite{Long2023}
\begin{equation*}    \mathcal{L}_{reg}=\frac{1}{MN} \sum_{i, k}\exp{\left(-\tau \cdot f(t_{i,k})\right)},
\end{equation*}
where $\tau$ is a constant scalar that scales the learned UDF values, $M$ is the total number of sampled rays per training iteration, and $N$ is the number of sampled points on a single ray.
$\tau$ is set to 5.0 in the first stage and 50.0 in the second stage.

The value of $s$, which is learnable in our method, significantly affects the quality of the reconstruction. When $s$ is small, it introduces a larger bias and leads to a more blurred output. We observe that $s$ typically converges to a relatively large value between 1000 and 2000, leading to visually pleasing results. However, in rare cases when $s$ stops increasing during training, we apply a penalty to force it to increase. The penalty is defined as follows
\begin{equation*}
    \mathcal{L}_{s} = \frac{1}{M} \sum_{i, k} \frac{1}{s_{i, k}},
\end{equation*}
where $M$ is the number of rays during a training epoch. 
This term $\mathcal{L}_{s}$ aggregates the reciprocals of all $s$ values used for the point $t_{i,k}$ on ray $r_i$. 
Intuitively speaking, it encourages a larger $s$ during the early stage of training. In our implementation, we make this term optional since $s$ generally increases with a decreasing rate during training, and the penalty term is only necessary in rare cases when $s$ stops at a relatively low value.

As in other SDF- and UDF-based methods~\cite{Wang2021,Wang2022HFNeuSIS,Long2023}, we adopt color loss and Eikonal loss in our approach. Specifically, the color loss $\mathcal{L}_{color}$ is the $L_1$ loss between the predicted color and the ground truth color of a single pixel as used in~\cite{Wang2021}. The Eikonal loss $\mathcal{L}_{eik}$ is used to regularize the learned distance field to have a unit gradient~\cite{Gropp2020}. Users may also choose to adopt object masks for supervision as introduced in other SDF- and UDF-based methods~\cite{Wang2021,Long2023}.
Putting it all together, we define the combined loss function as a weighted sum,
\begin{equation*}
    \mathcal{L} = \mathcal{L}_{color} + \lambda_1 \mathcal{L}_{eik} + \lambda_2 \mathcal{L}_{reg} + \lambda_3 \mathcal{L}_{s} \left(+\lambda_m \mathcal{L}_{mask}\right),
\end{equation*}
where $\lambda_1$, $\lambda_2$, $\lambda_3$ and the optional $\lambda_m$ are hyperparameters that control the weight of each loss term.

\section{Experiments}

\emph{Datasets.} To evaluate our method, we use three datasets: DeepFashion3D~\cite{Zhu2020}, DTU~\cite{Jensen2014} and BlendedMVS~\cite{Yao_2020_CVPR}.
The DeepFashion3D dataset consists of clothing models, which are open models with boundaries. As only 3D points are available, we render 72 images of resolution $1024\times 1024$ with a white background from different viewpoints for each model. In addition to DeepFashion3D images rendered by us most of which are texture-less, we also take the image data from NeuralUDF~\cite{Long2023} most of which are texture-rich into our experiments. We call them DF3D\#Ours and DF3D\#NeuralUDF, respectively. The DTU dataset consists of models captured in a studio, all of which are watertight. We use this dataset to validate that our method also works well for watertight models. These datasets have been widely used in previous works such as~\cite{Yariv2021,Wang2021,Wang2022HFNeuSIS}. In our experiments, open models such as in DeepFashion3D are trained without mask supervision; DTU is trained with mask supervision.

\emph{Baselines.} To validate the effectiveness of our method, we compare it with state-of-the-art UDF learning methods: 
NeuralUDF~\cite{Long2023}, NeUDF~\cite{Liu2023NeUDF} and NeAT~\cite{Meng2023}; and SDF learning methods: VolSDF~\cite{Yariv2021} and NeuS~\cite{Wang2021}. 

\subsection{Comparisons on Open Models}
\begin{figure*}[!t]
\centering
\setlength\tabcolsep{0pt}
\begin{tabular}{ccccccccc}
    & Ref. Img. & GT & Ours & VolSDF & NeuS & NeAT & NeuralUDF & NeUDF\\
    \raisebox{.22in}{\#2} & \includegraphics[width=.386in]{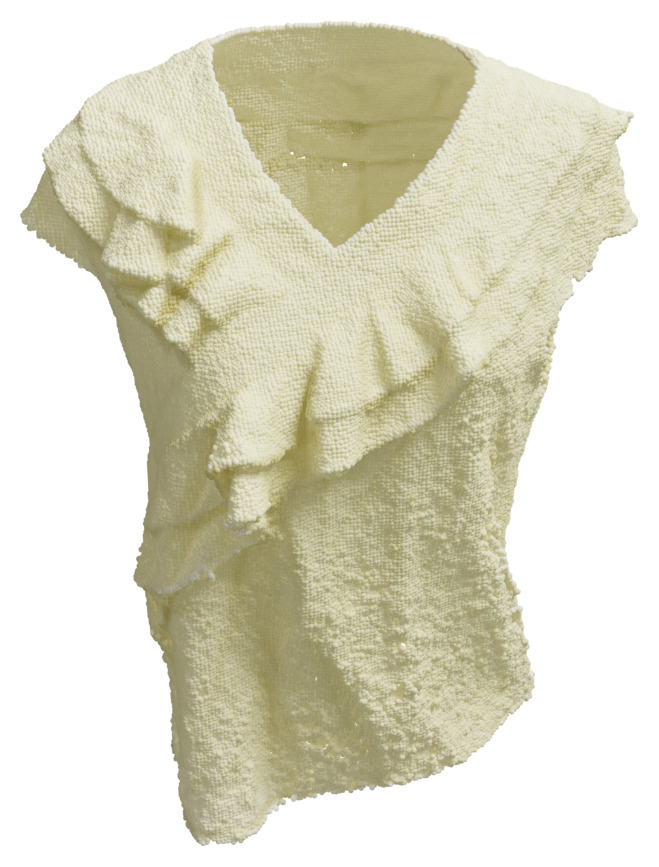}&
    \includegraphics[width=.386in]{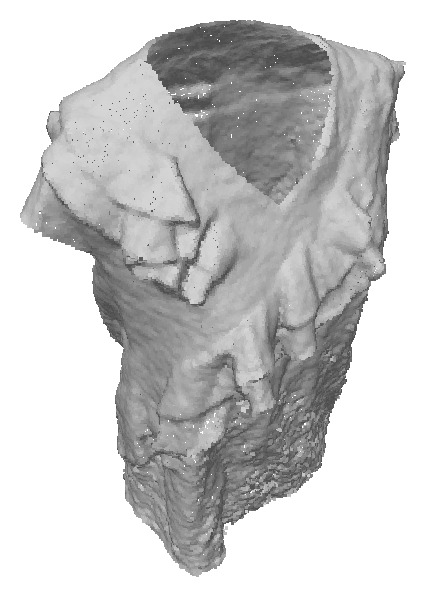}\includegraphics[width=.3in]{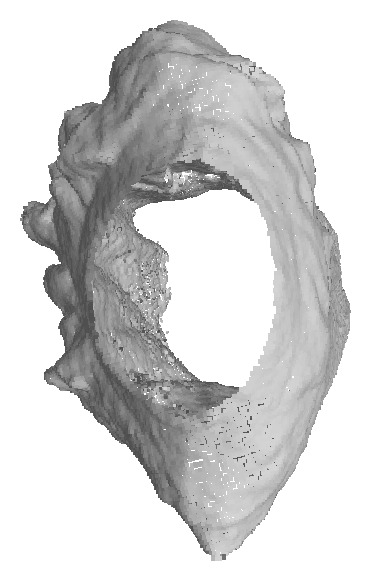}&
    \includegraphics[width=.386in]{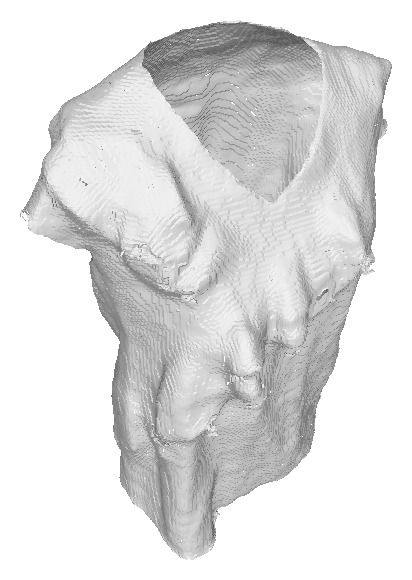}\includegraphics[width=.3in]{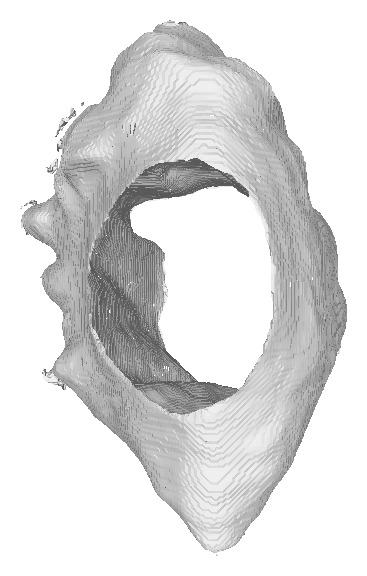}&
    \includegraphics[width=.386in]{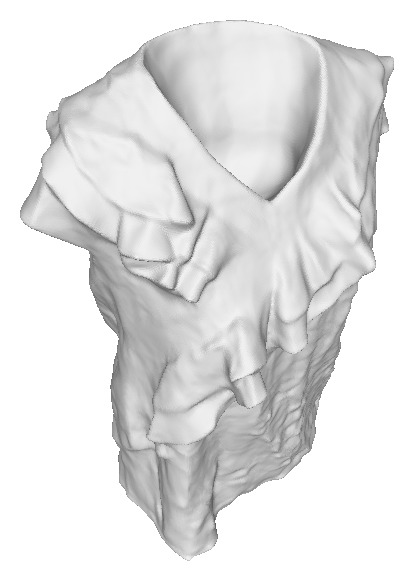}\includegraphics[width=.3in]{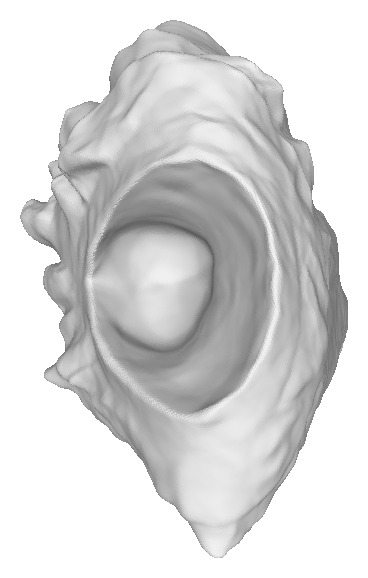}&
    \includegraphics[width=.386in]{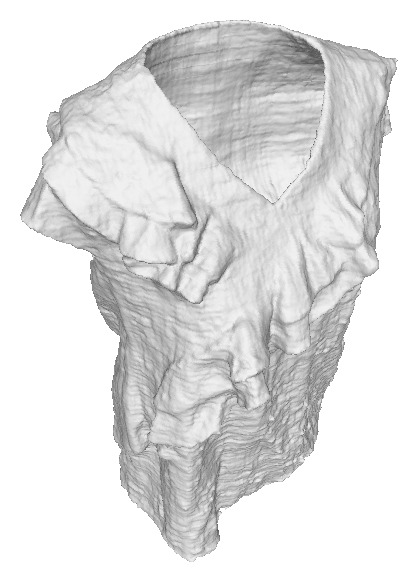}\includegraphics[width=.3in]{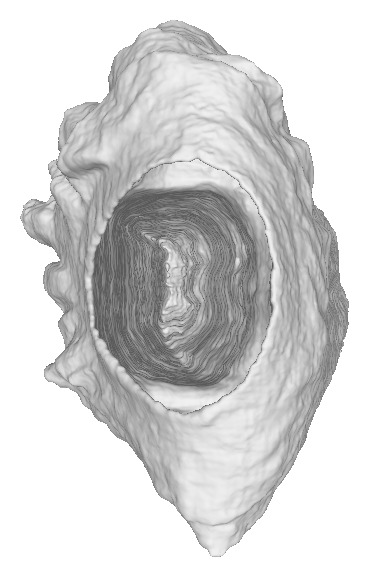}&
    \includegraphics[width=.386in]{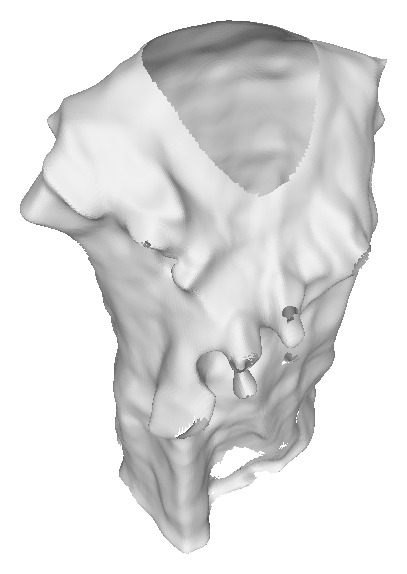}\includegraphics[width=.3in]{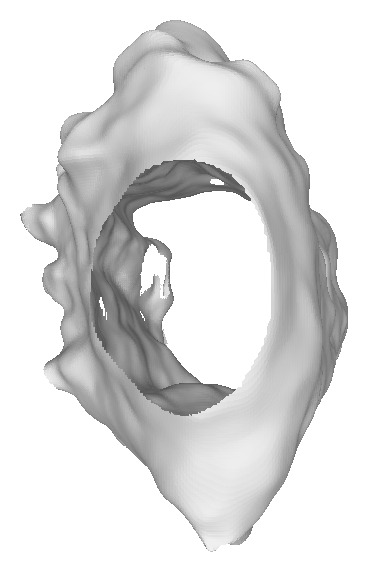}&
    \includegraphics[width=.386in]{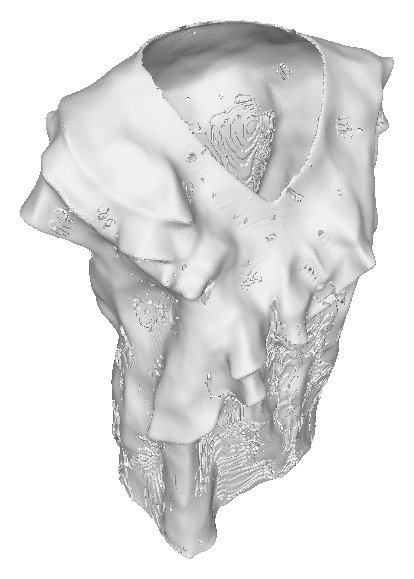}\includegraphics[width=.3in]{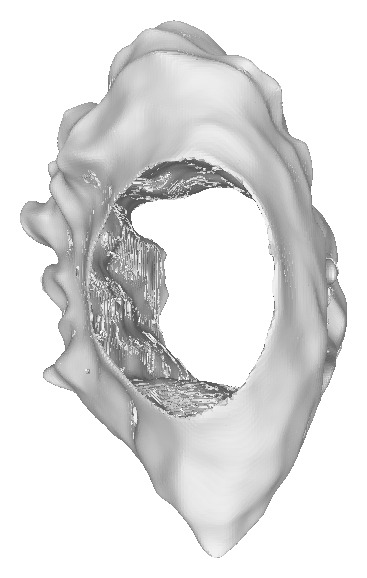}&
    \includegraphics[width=.386in]{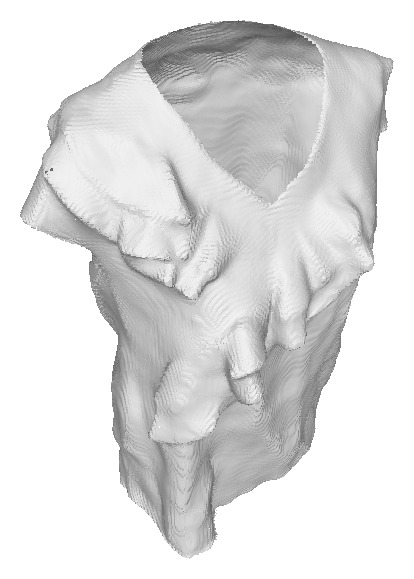}\includegraphics[width=.3in]{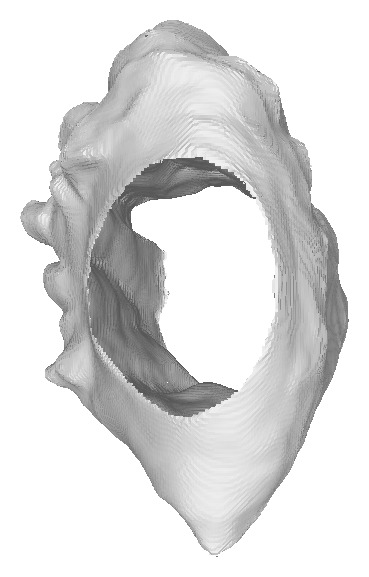}\\
    \raisebox{.22in}{\#3} & \includegraphics[width=.386in]{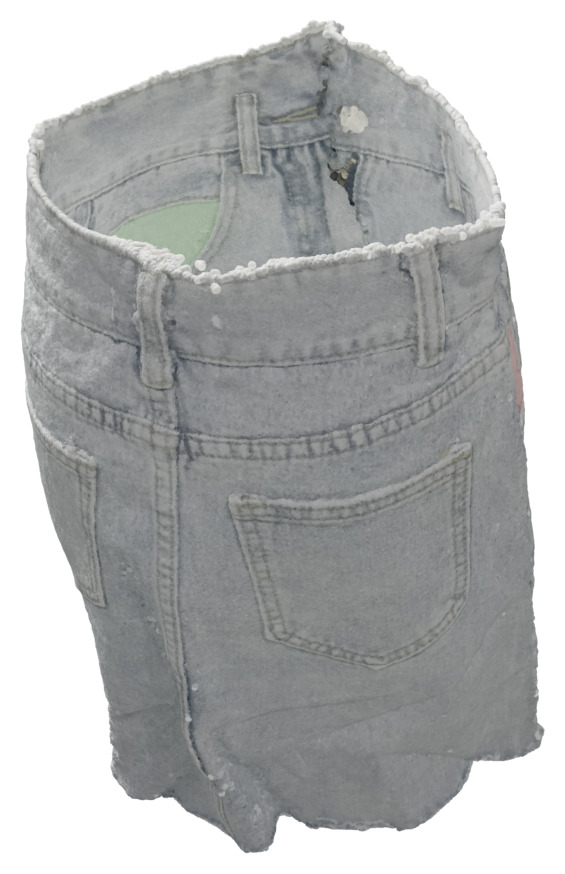}&
    \includegraphics[width=.386in]{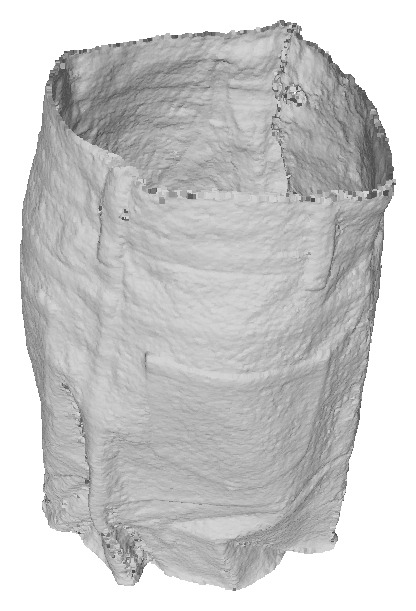}\includegraphics[width=.386in]{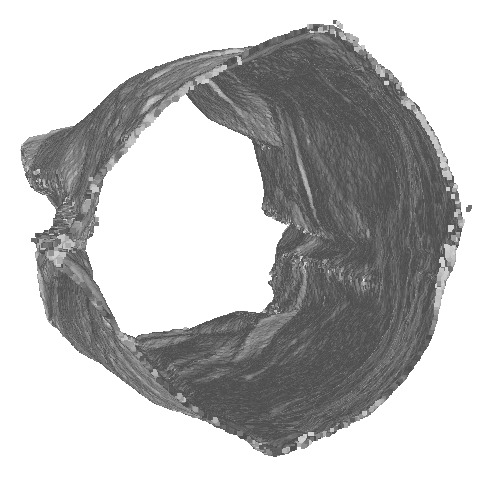}&
    \includegraphics[width=.386in]{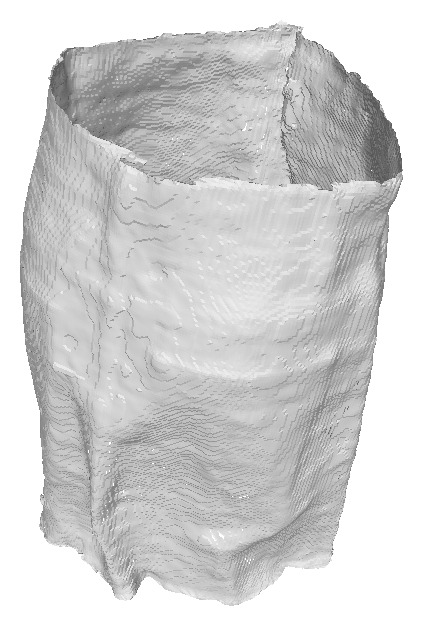}\includegraphics[width=.386in]{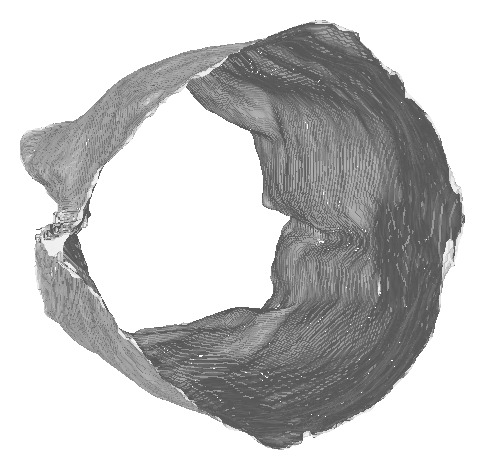}&
    \includegraphics[width=.386in]{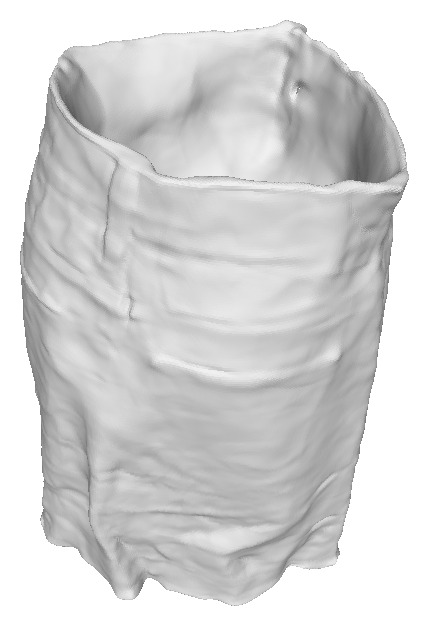}\includegraphics[width=.386in]{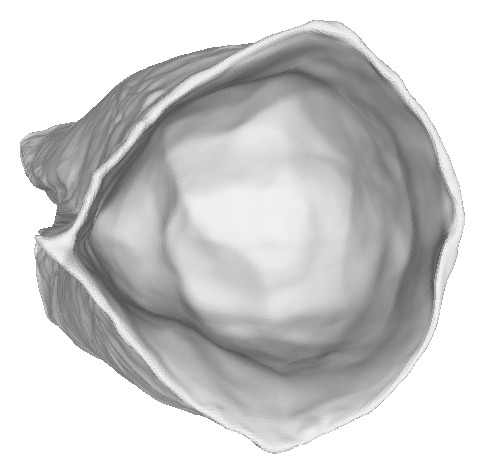}&
    \includegraphics[width=.386in]{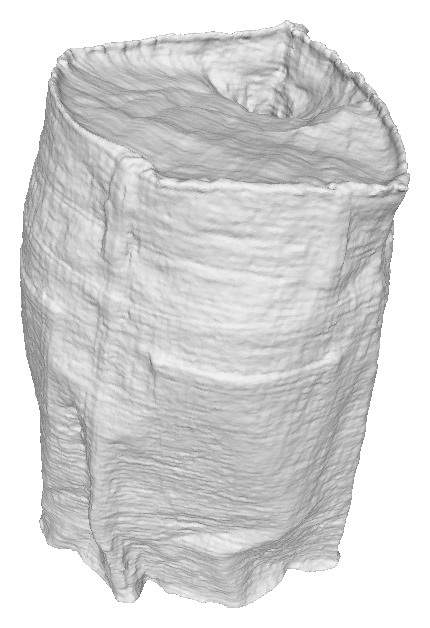}\includegraphics[width=.386in]{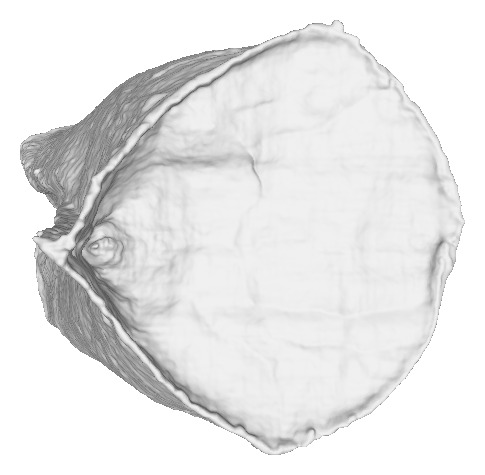}&
    \includegraphics[width=.386in]{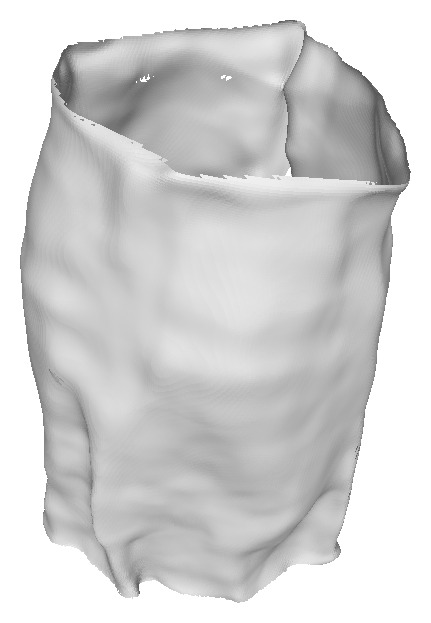}\includegraphics[width=.386in]{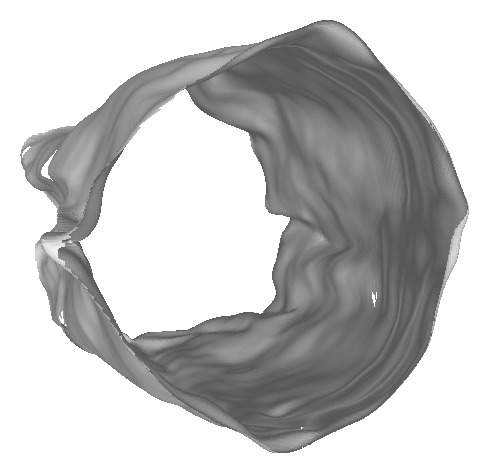}&
    \includegraphics[width=.386in]{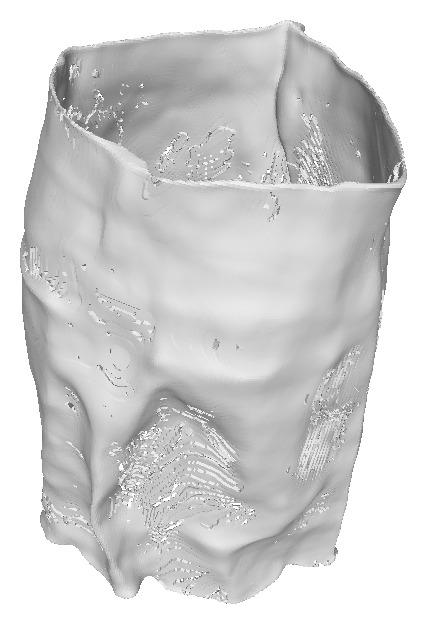}\includegraphics[width=.386in]{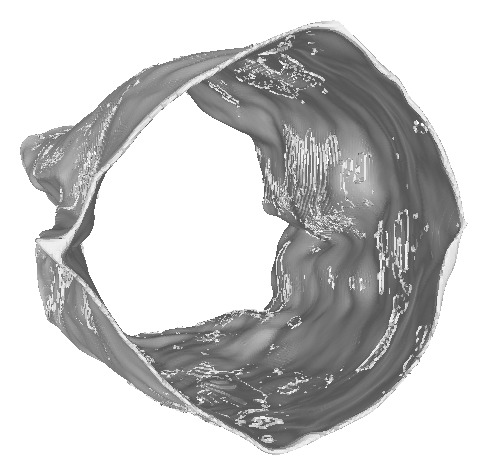}&
    \includegraphics[width=.386in]{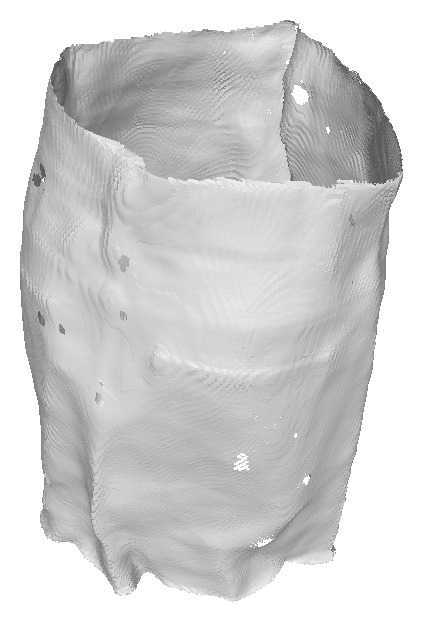}\includegraphics[width=.386in]{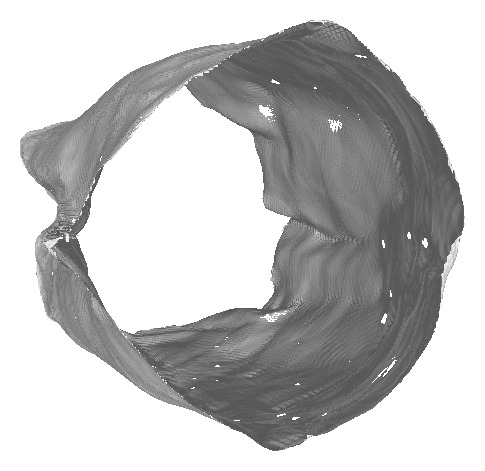}\\
    \raisebox{.22in}{\#4} & \includegraphics[width=.429in]{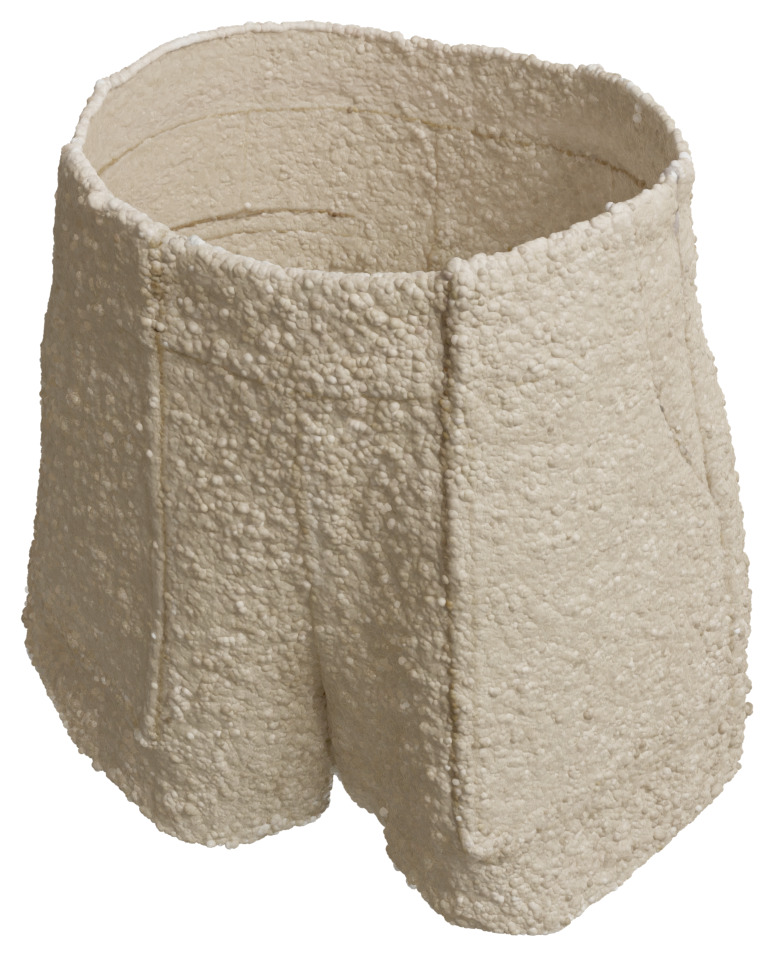}&
    \includegraphics[width=.429in]{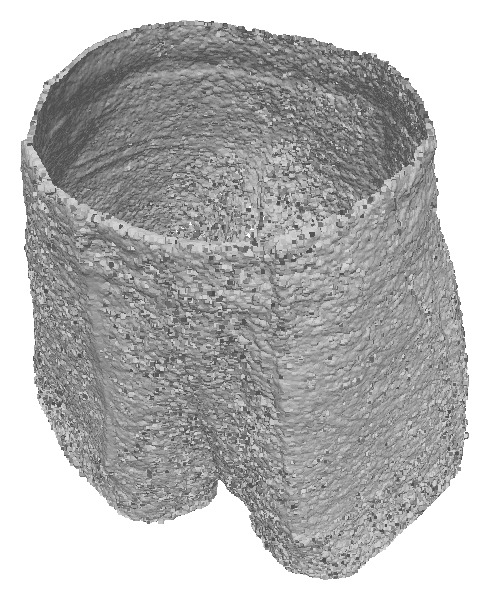}\includegraphics[width=.386in]{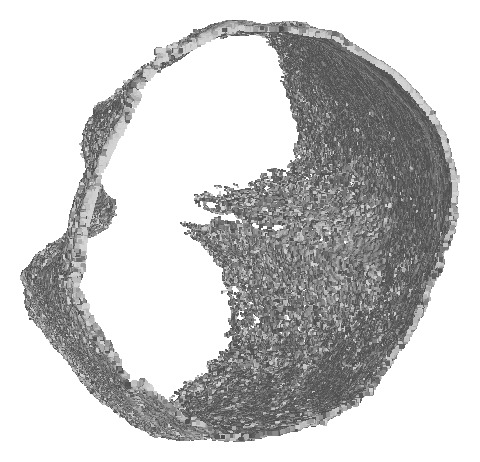}&
    \includegraphics[width=.429in]{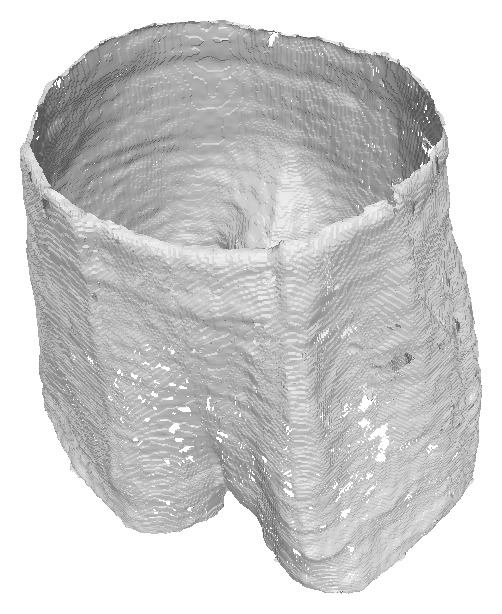}\includegraphics[width=.386in]{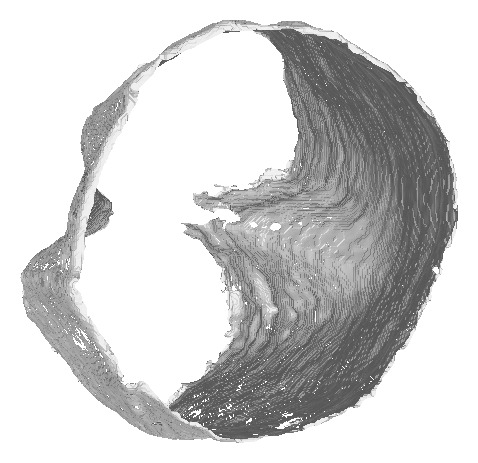}&
    \includegraphics[width=.429in]{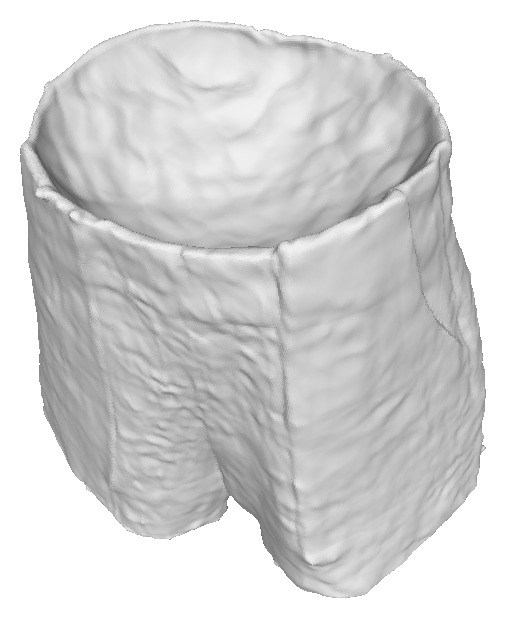}\includegraphics[width=.386in]{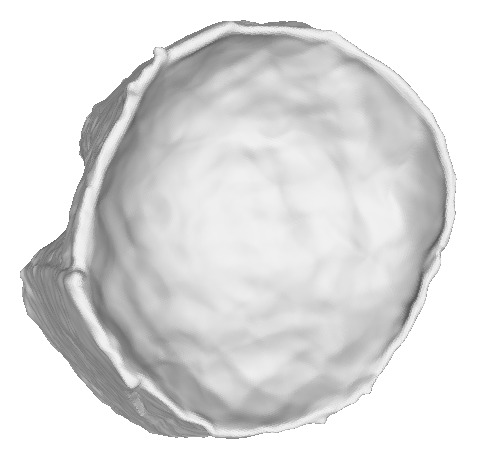}&
    \includegraphics[width=.429in]{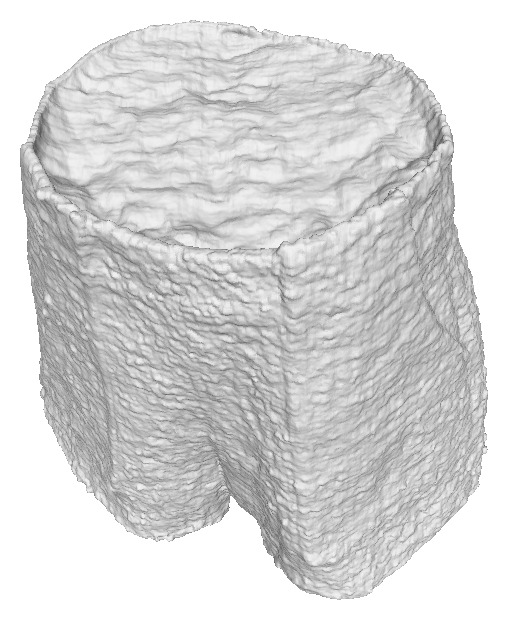}\includegraphics[width=.386in]{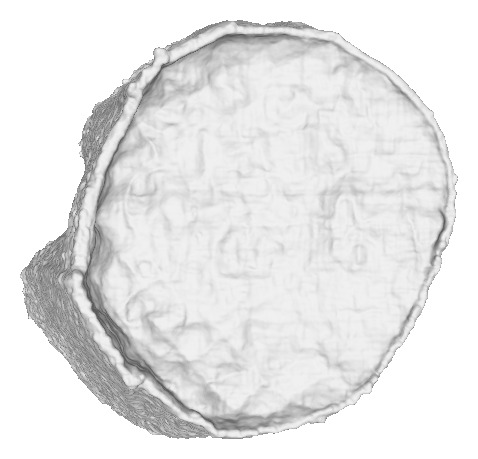}&
    \includegraphics[width=.429in]{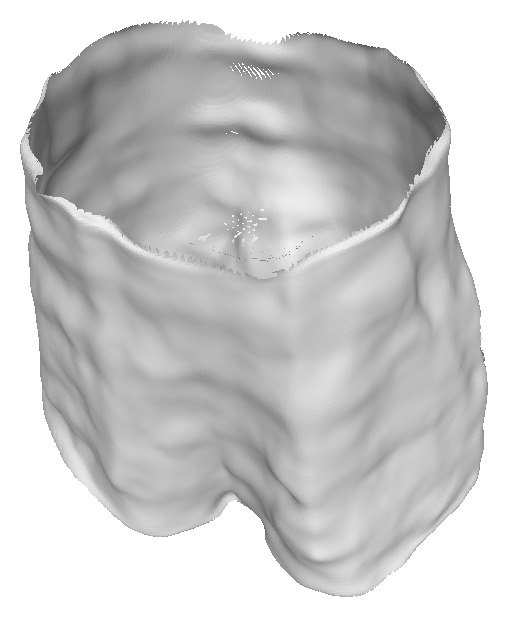}\includegraphics[width=.386in]{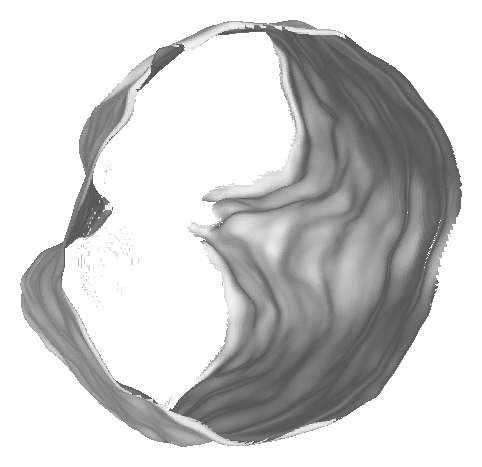}&
    \includegraphics[width=.429in]{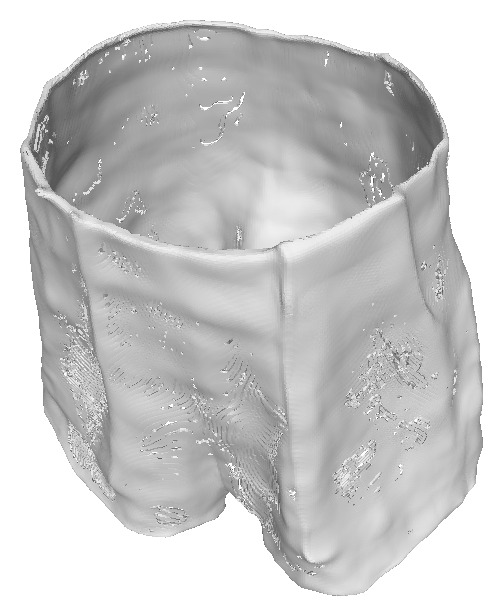}\includegraphics[width=.386in]{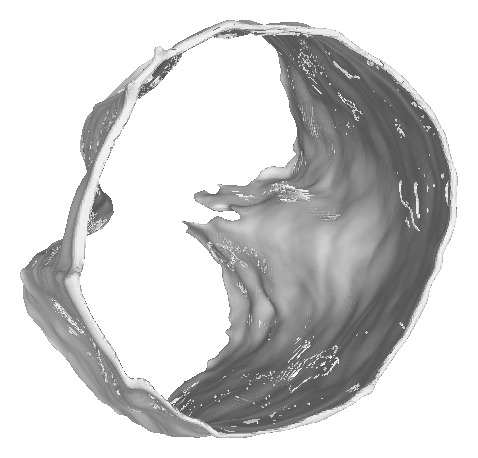}&
    \includegraphics[width=.429in]{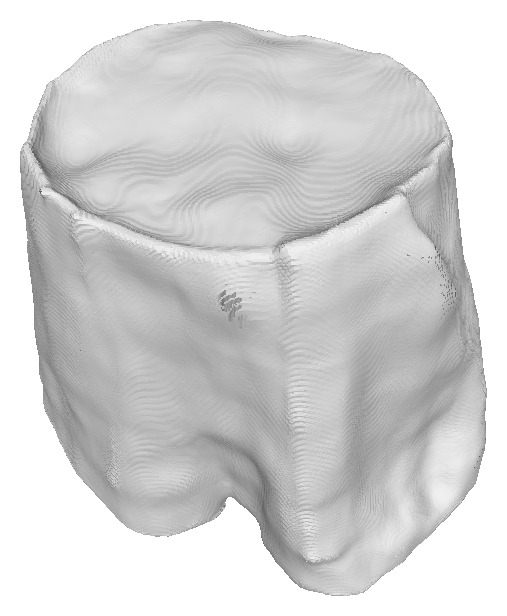}\includegraphics[width=.386in]{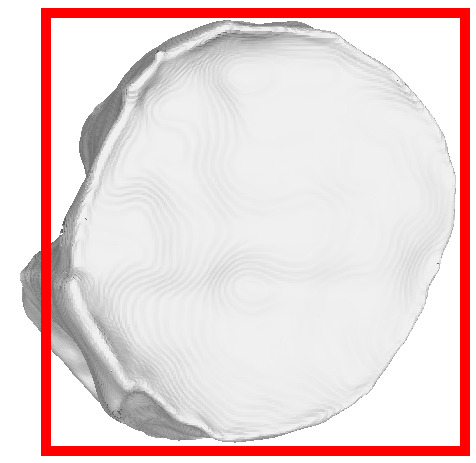}\\
    \raisebox{.22in}{LS-D0} & \includegraphics[width=.386in]{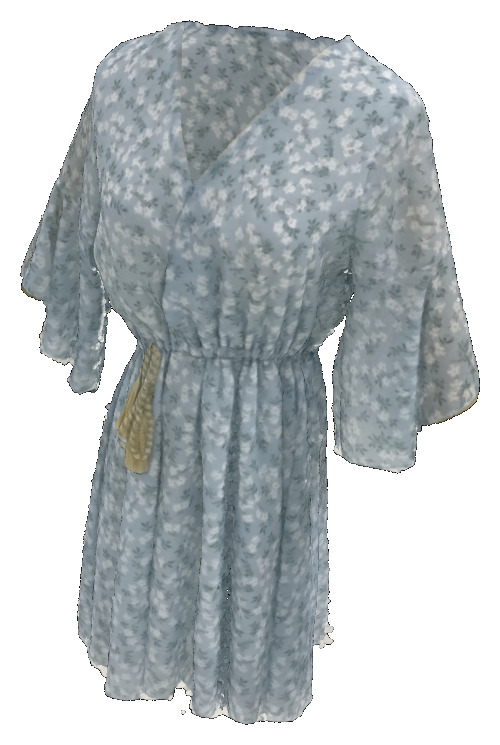}&
    \includegraphics[width=.386in]{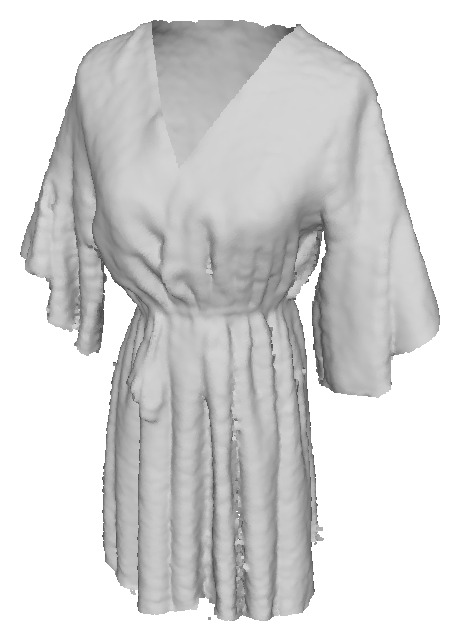}\includegraphics[width=.343in]{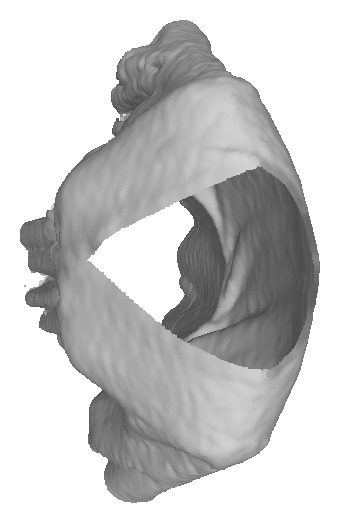}&
    \includegraphics[width=.386in]{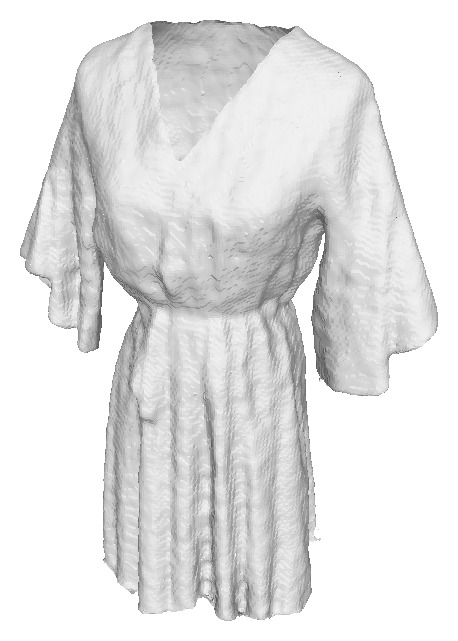}\includegraphics[width=.343in]{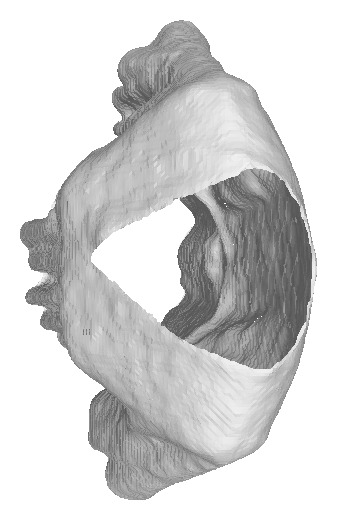}&
    \includegraphics[width=.386in]{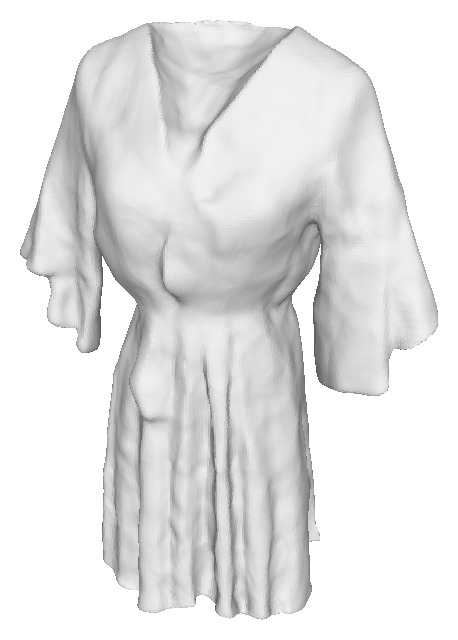}\includegraphics[width=.343in]{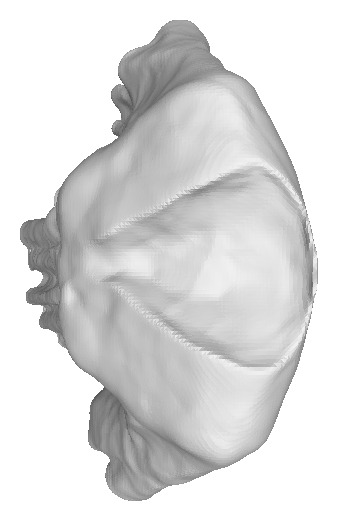}&
    \includegraphics[width=.386in]{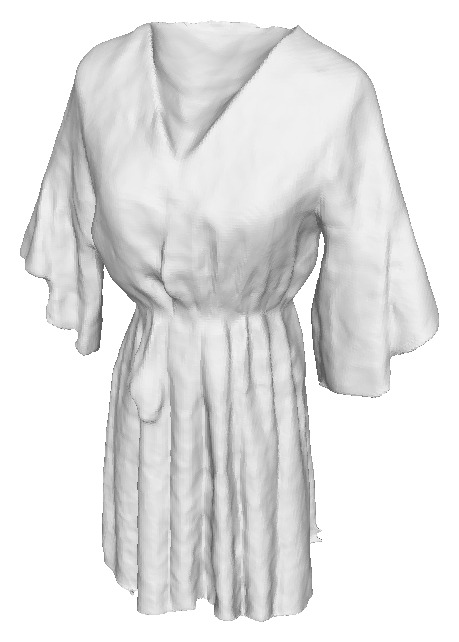}\includegraphics[width=.343in]{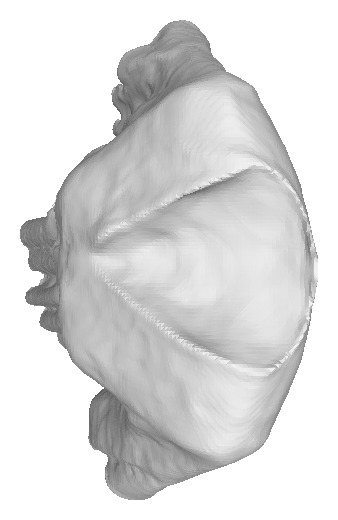}&
    \includegraphics[width=.386in]{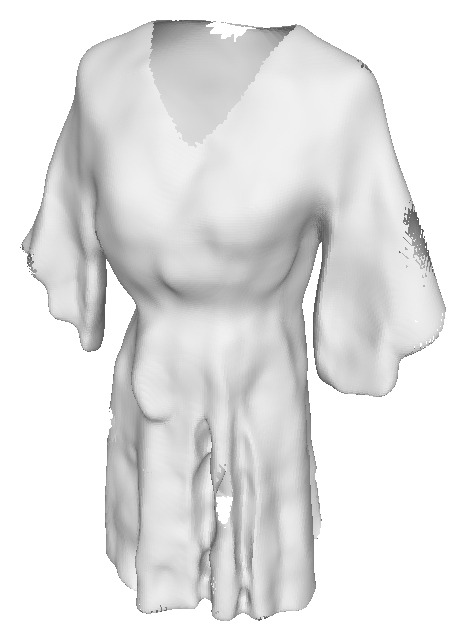}\includegraphics[width=.343in]{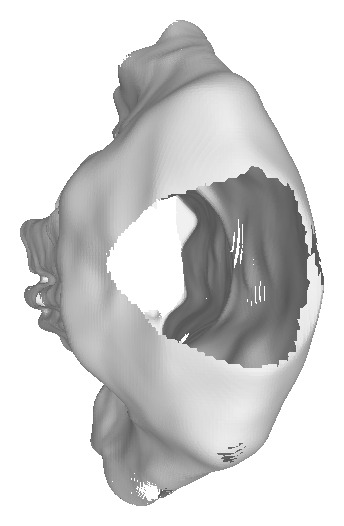}&
    \includegraphics[width=.386in]{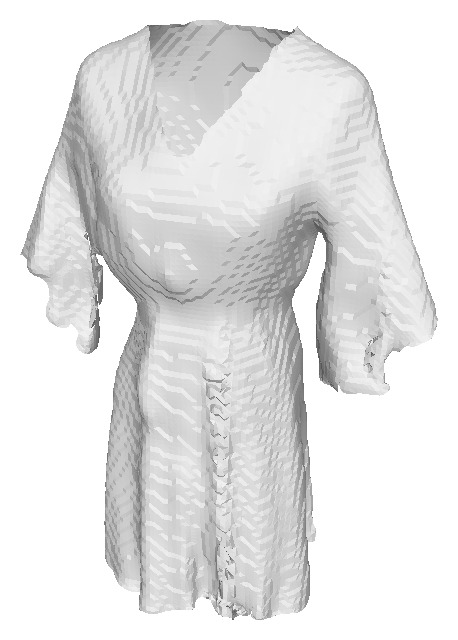}\includegraphics[width=.343in]{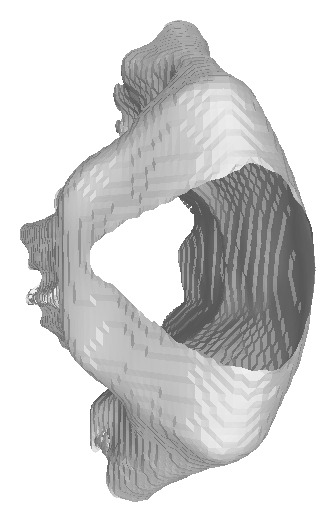}&
    \raisebox{.22in}{N.A.} \\
\end{tabular}
\caption{Visual comparisons on selected models of the DeepFashion3D~\cite{Zhu2020} dataset. The surfaces produced by NeuS and VolSDF are closed watertight models, thereby post-processing is required to remove the unnecessary parts. NeAT can produce open models by learning an SDF and predicting which surfaces in the extracted meshes should be removed, but it needs mask for supervision. NeuralUDF can generate open surfaces, but struggles with textureless inputs, leading to double-layered regions and large reconstruction errors. NeUDF generally performs well, but its training is unstable and may stumble on less distinguished, darker models like LS-D0. In contrast, our \methodname{} consistently delivers effective reconstructions of non-watertight models. See the supplementary material for additional results.}

\label{fig:df3d}
\end{figure*}

\begin{table}[!htbp]
\centering
\setlength\tabcolsep{1.25pt}
\begin{footnotesize}
\begin{tabular}{c|ccccccccc|c}
    \hline
    Method & \#1 & \#2 & \#3 & \#4 & \#5 & \#6 & \#7 & \#8 & \#9 & Mean \\
    \hline
    NeuS & 6.69 & 13.50 & 10.32 & 15.01 & 8.99 & 12.92 & 12.94 & 9.93 & 9.49 & 11.09 \\
    VolSDF & 6.36 & 9.44 & 11.87 & 16.03 & 10.78 & 14.91 & 15.06 & 11.34 & 8.96 & 11.64 \\
    NeAT & 10.54 & 13.89 & 7.30 & 13.12 & 13.18 & 12.44 & 8.22 & 10.30 & 11.33 & 11.15 \\
    NeuralUDF & 6.07 & 11.58 & 7.68 & \underline{10.96} & 11.16 & 9.76 & 6.98 & 6.13 & 6.41 & 8.53 \\
    NeUDF & \textbf{4.39} & \underline{8.29} & \underline{4.94} & 19.56 & \underline{7.52} & \underline{8.18} & \underline{3.81} & \underline{3.81} & \underline{5.76} & \underline{7.36} \\
    Ours & \underline{4.55} & \textbf{5.77} & \textbf{4.27} & \textbf{7.43} & \textbf{6.59} & \textbf{4.77} & \textbf{2.88} & \textbf{3.21} & \textbf{5.73} & \textbf{5.02} \\
    \hline
\end{tabular}
\end{footnotesize}

\begin{footnotesize}
\begin{tabular}{c|ccccccc|c}
\hline
    Method & LS-C0 & SS-D0 & LS-D0 & NS-D1 & LS-C1 & Skirt1 & SS-C0 & Mean \\
    \hline
    NeuS & 3.18 & 4.82 & 5.71 & 2.21 & 3.60 & 2.44 & 5.13 & 3.87 \\
    VolSDF & 5.92 & 4.79 & 5.96 & 4.36 & 8.73 & 7.74 & 8.84 & 6.62 \\
    NeAT & 3.06 & 4.33 & 5.92 & 3.52 & 8.84 & 3.91 & 4.30 & 4.84 \\
    NeuralUDF & \textbf{1.92} & \underline{2.05} & \underline{4.11} & 1.50 & \underline{2.47} & \underline{2.16} & 2.15 & 2.34 \\
    NeUDF & \underline{1.95} & 2.93 & N.A. & \underline{1.48} & 2.66 & 2.74 & \textbf{1.77} & \underline{2.26} \\
    Ours & \textbf{1.92} & \textbf{1.97} & \textbf{2.46} & \textbf{1.47} & \textbf{2.14} & \textbf{1.84} & \underline{1.91} & \textbf{1.96} \\
    \hline
\end{tabular}
\end{footnotesize}

\caption{Chamfer distances ($\times 10^{-3}$) on DF3D\#Ours (top) and DF3D\#NeuralUDF (bottom). NeAT requires mask supervision and others do not need.}
\label{tab:df3d}
\end{table}

We evaluate our method and compare it with baselines using the garments from DeepFashion3D~\cite{Zhu2020}, where the models have multiple open boundaries.  VolSDF and NeuS always close the boundaries since they learn SDFs. 

NeuralUDF, NeUDF and NeAT are designed to learn non-watertight models. NeAT learns SDFs for open models, and requires mask supervision to produce reasonable results, but other methods do not require mask supervision for DeepFashion3D. The released codebase of NeuralUDF indicates that it also has a two-stage training process. We evaluate the results of NeuralUDF at the end of both stages, and present whichever is better.

In contrast, NeuralUDF, NeUDF and our method learn UDFs, which can generate open models. Table~\ref{tab:df3d} shows the Chamfer distances of the results on DeepFashion3D. Some of the Chamfer distances of the compared methods are large because the open holes are closed or the model is over-smoothed, resulting in significant errors.

\begin{figure}[ht]
\centering
\setlength\tabcolsep{1pt}
\begin{tabular}{ccccc}
    & GT & Ours & NeuralUDF & NeUDF \\
    \raisebox{.22in}{\#2} & \includegraphics[width=.7in]{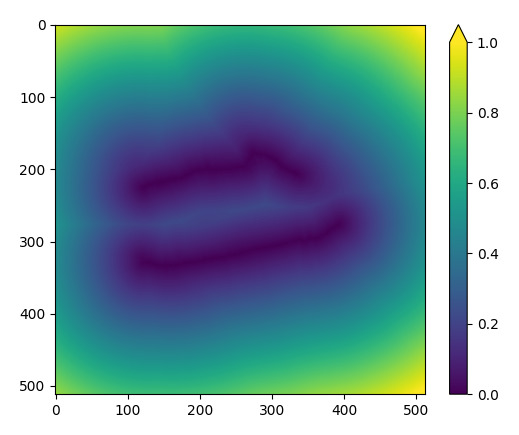} &
    \includegraphics[width=.7in]{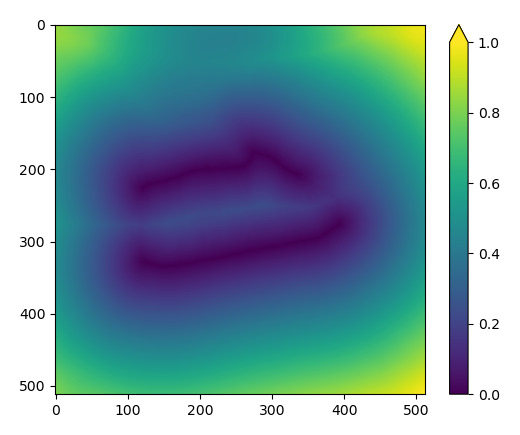} &
    \includegraphics[width=.7in]{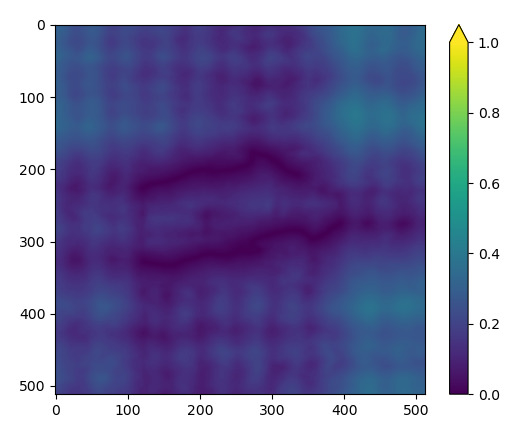} &
    \includegraphics[width=.7in]{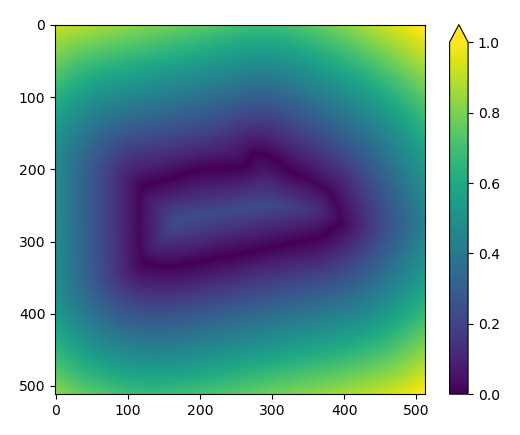} \\
    \raisebox{.22in}{\#3} & \includegraphics[width=.7in]{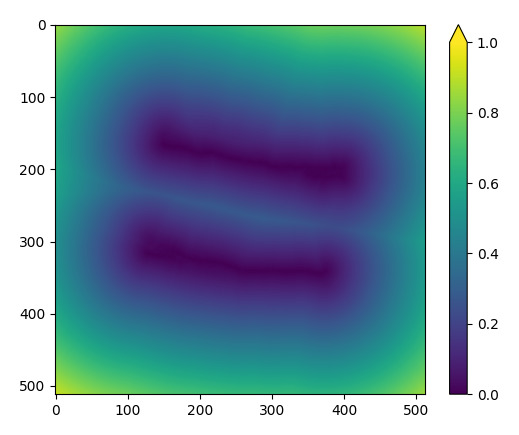} &
    \includegraphics[width=.7in]{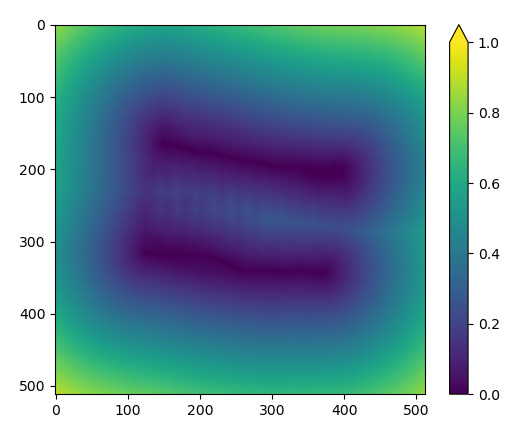} &
    \includegraphics[width=.7in]{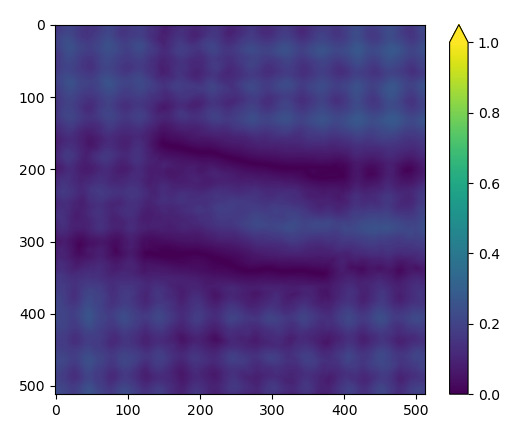} &
    \includegraphics[width=.7in]{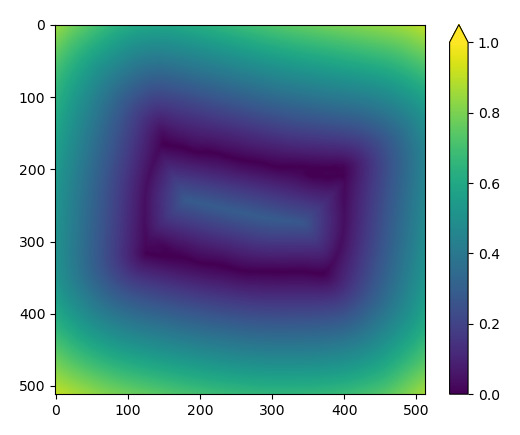} \\
    \raisebox{.22in}{\#4} & \includegraphics[width=.7in]{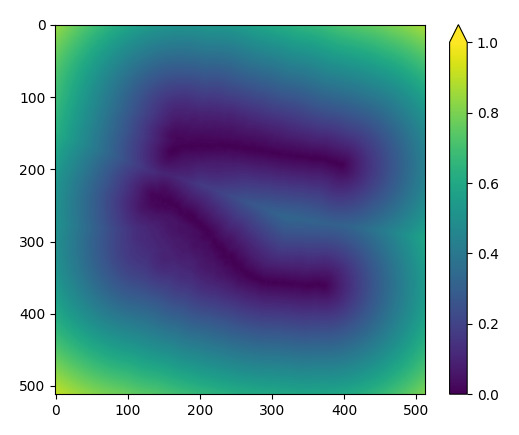} &
    \includegraphics[width=.7in]{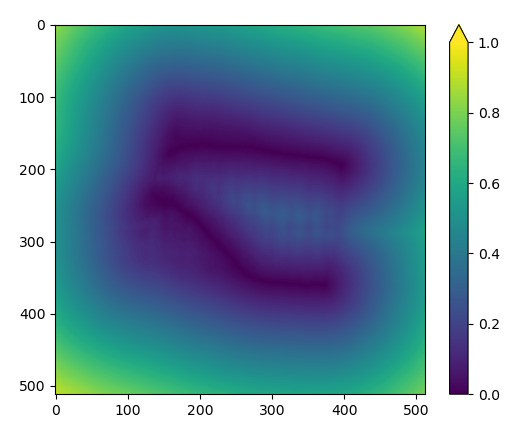} &
    \includegraphics[width=.7in]{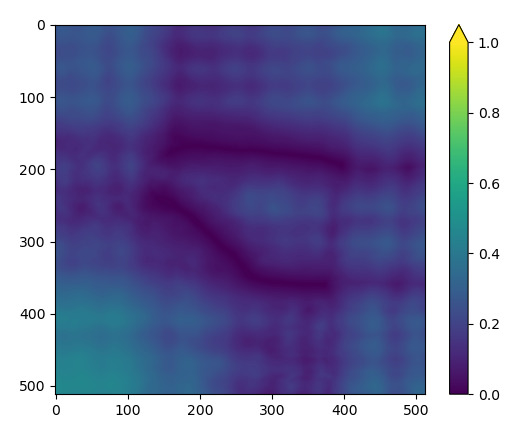} &
    \includegraphics[width=.7in]{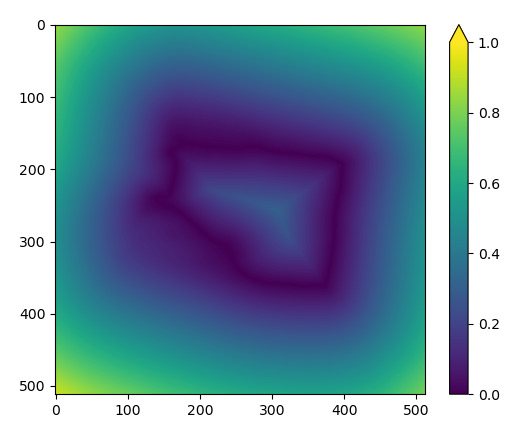} \\
    \raisebox{.22in}{LS-D0} & \includegraphics[width=.7in]{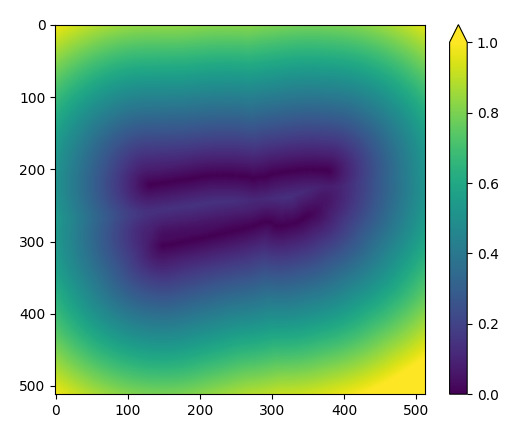} &
    \includegraphics[width=.7in]{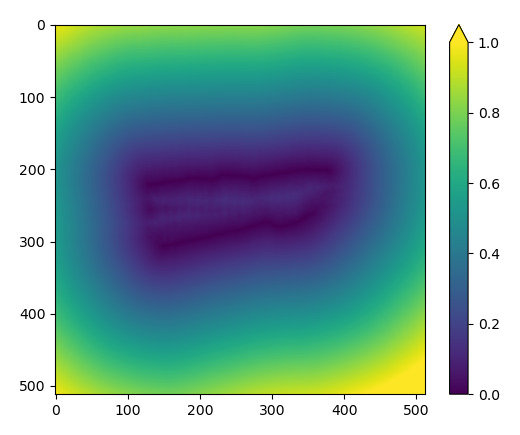} &
    \includegraphics[width=.7in]{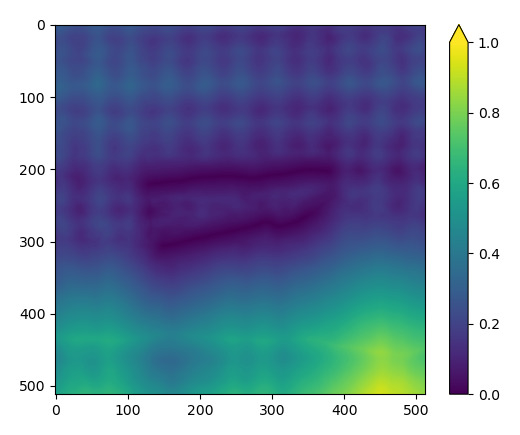} &
    \raisebox{.22in}{N.A.} \\
\end{tabular}
\caption{Visualization of the learned UDFs on cross sections. Compared with the ground truth, our method can learn a UDFs that most closely resemble the ground truth, among our method, NeuralUDF, and NeUDF. NeAT is omitted in this visualization, because it learns SDFs in lieu of UDFs. Note that for LS-D0, NeUDF completely collapses without a reasonable UDF learned.}
\label{fig:vs_udf}
\end{figure}

As demonstrated in Figure~\ref{fig:df3d}, we test various types of garments, some of which have rich textures, while others are nearly a single color. Learning UDFs for textureless models is more challenging since various regions of a model are ambiguous without clear color differences. However, our \methodname{} generates satisfactory results even without masks. Though with mask supervision, the results of NeAT~\cite{Meng2023} are over-smoothed, missing details, resulting in large Chamfer distance errors. NeuralUDF~\cite{Long2023} is unable to properly reconstruct textureless models on most models, possibly due to their complex density function which is difficult to converge. Some of the NeUDF~\cite{Liu2023NeUDF} models become watertight.
To analyze the reasons, we illustrate these UDFs cross sections in Figure~\ref{fig:vs_udf}. To compute the ground truth UDFs, we sample 30,000 points from every input point model and compute the distances to the nearest sample point for every point in a 3D grid of resolution $512\times 512\times 512$. All other UDFs are extracted by querying the distance neural network in a 3D grid of the same resolution. Our learned UDFs resemble the ground truth with little difference. While, the UDFs of NeuralUDF deviate from the ground truth significantly explaining its difficulty to converge. The UDFs of NeUDF are better, but the distances approach to zero around open holes. As a result, it is challenging and tricky to generate non-watertight models and some of them are even closed. NeAT learns SDF, so we do not show their distance fields.

\begin{figure}[!ht]
    \centering
    \begin{scriptsize}
    \setlength\tabcolsep{2pt}
    \begin{tabular}{cccc}
        \multirow{6}{*}{\includegraphics[width=1.35in]{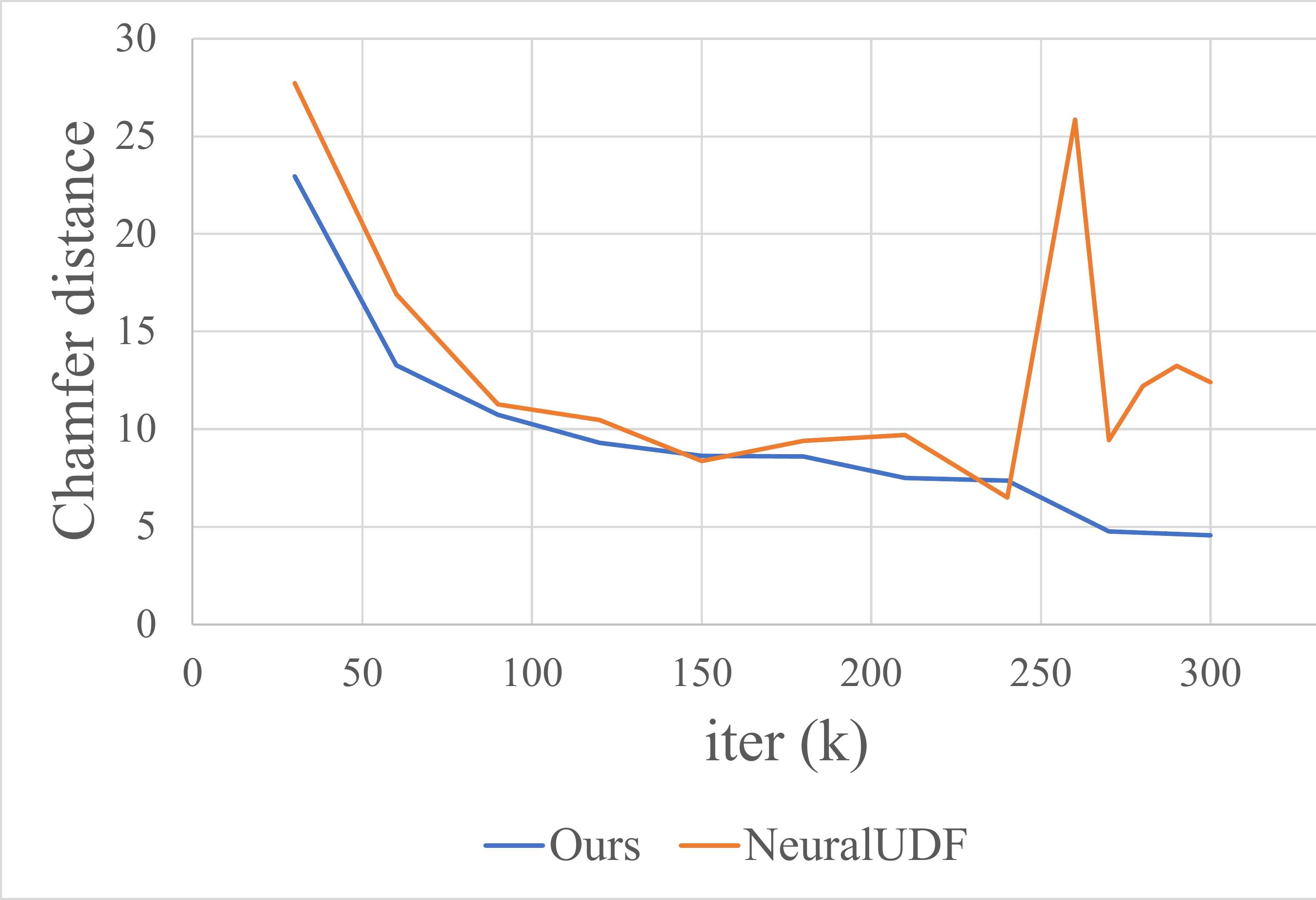}} &
        \includegraphics[width=.6in]{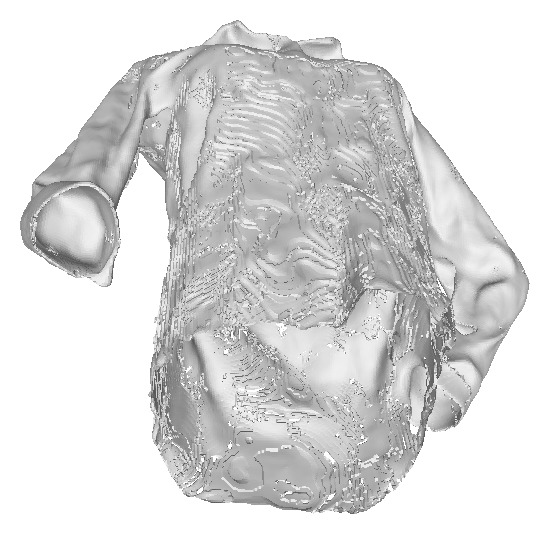} &
        \includegraphics[width=.6in]{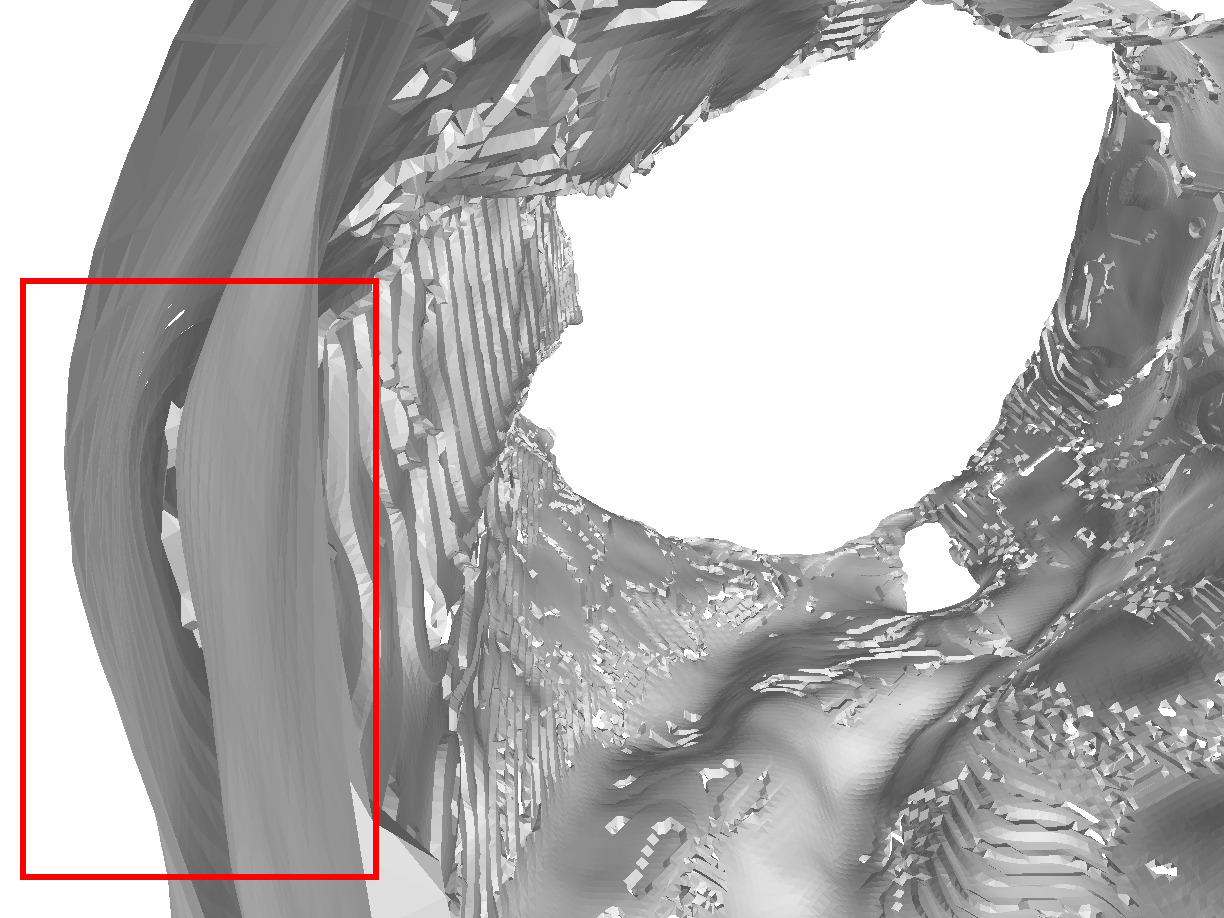} &
        \includegraphics[width=.6in]{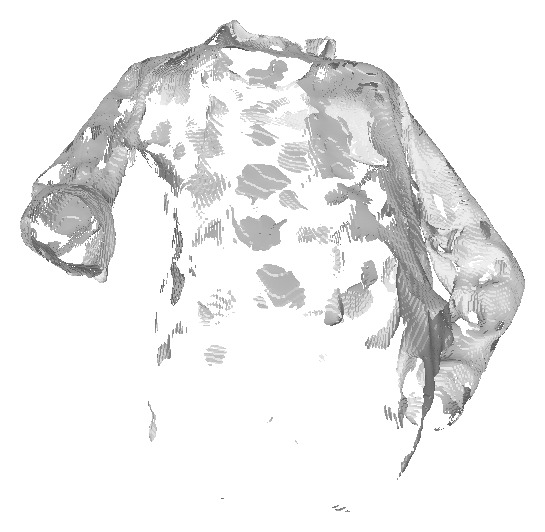} \\
        & \multicolumn{2}{c}{Stage 1} & Stage 2\\
        & \multicolumn{3}{c}{NeuralUDF}\\
        & 
        \includegraphics[width=.6in]{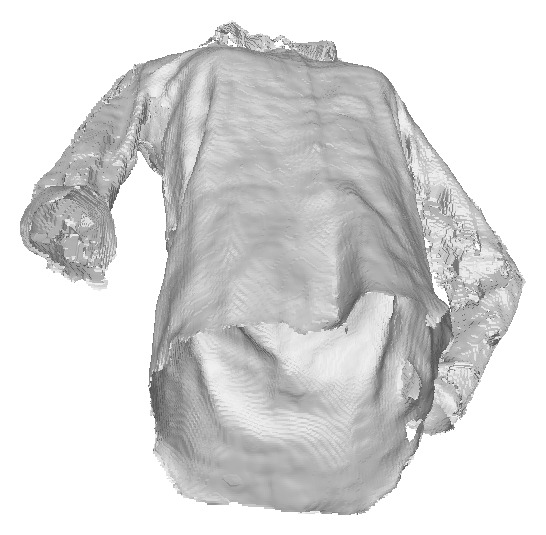} &
        \includegraphics[width=.6in]{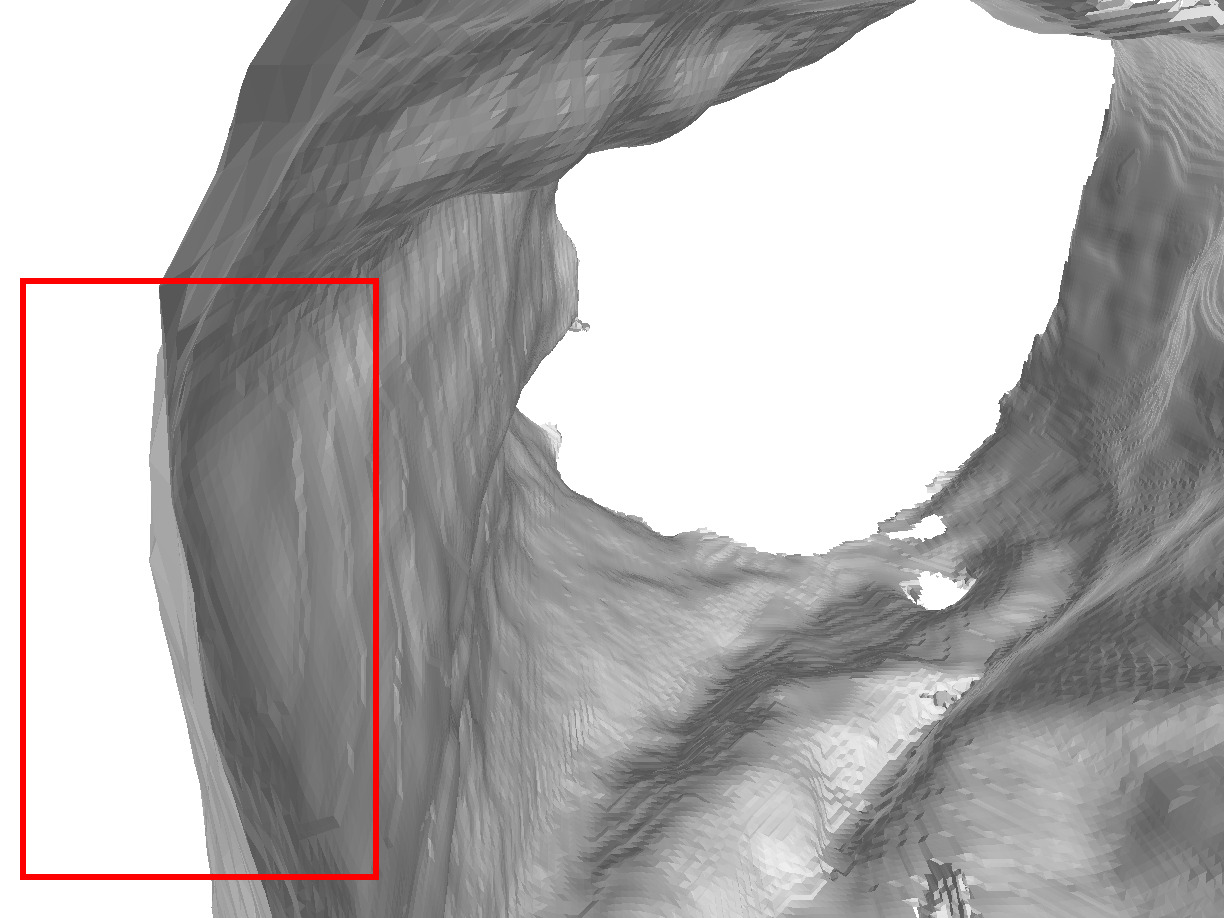} &
        \includegraphics[width=.6in]{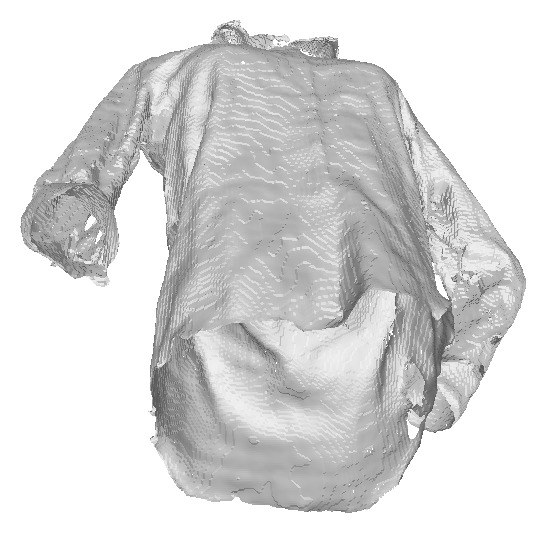} \\
        & \multicolumn{2}{c}{Stage 1} & Stage 2\\
        & \multicolumn{3}{c}{\methodname{}} \\
    \end{tabular}
    \end{scriptsize}
    \caption{Plots of the Chamfer distance throughout the training process. Our method consistently reduces CD across both stages. In contrast, NeuralUDF, which also adopts a two-stage learning strategy, exhibits instability and yields a fragmented output following the second stage. The first-stage output of NeuralUDF, however, contains double-layered regions as marked above. In this figure, both methods start their stage 2 training at 250k iterations.}
    \label{fig:vs_cd}
\end{figure}

As illustrated in Figure~\ref{fig:vs_cd}, perhaps due to the absolute of an MLP for UDF representation, NeuralUDF possibly generates two layers of zero level-sets on both sides of the surface resulting in double-layered regions after Stage 1 learning. However, in its Stage 2 refinement, the surface is crushed into pieces and the Chamfer distance errors surge suddenly.

\begin{figure}[htbp]
\centering
\setlength\tabcolsep{1pt}
\begin{tabular}{cccccc}
    & Ref. & Ours & NeUDF & NeuralUDF & NeAT \\
    \raisebox{.12in}{\rotatebox{90}{rack}} & \includegraphics[width=.58in]{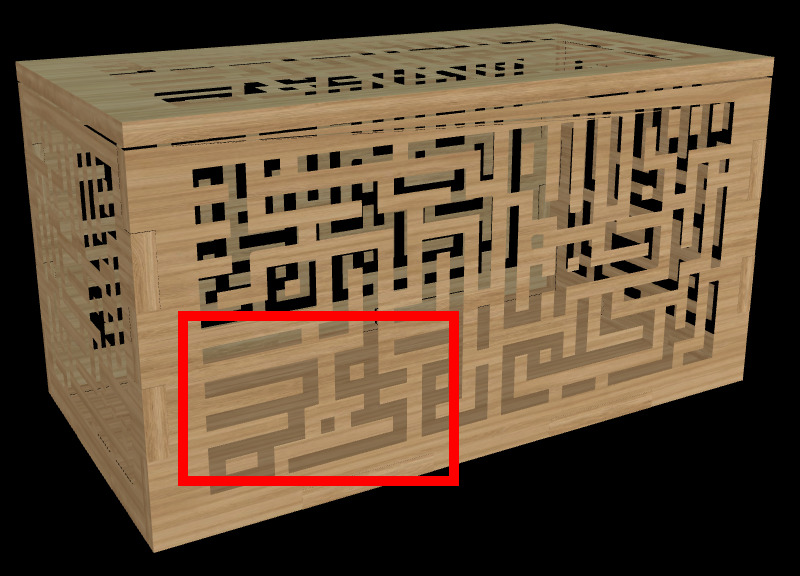} &
    \includegraphics[width=.58in]{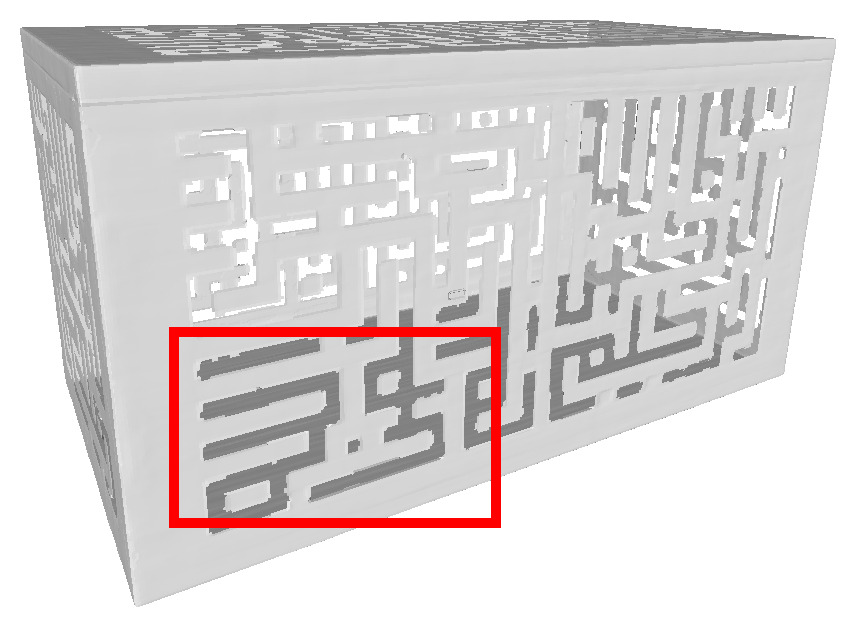} &
    \includegraphics[width=.58in]{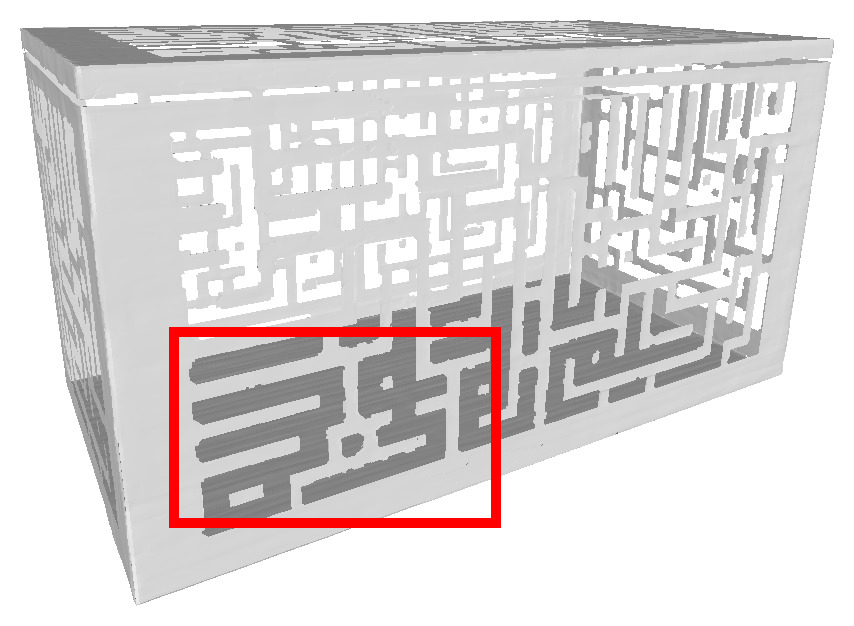} &
    \includegraphics[width=.58in]{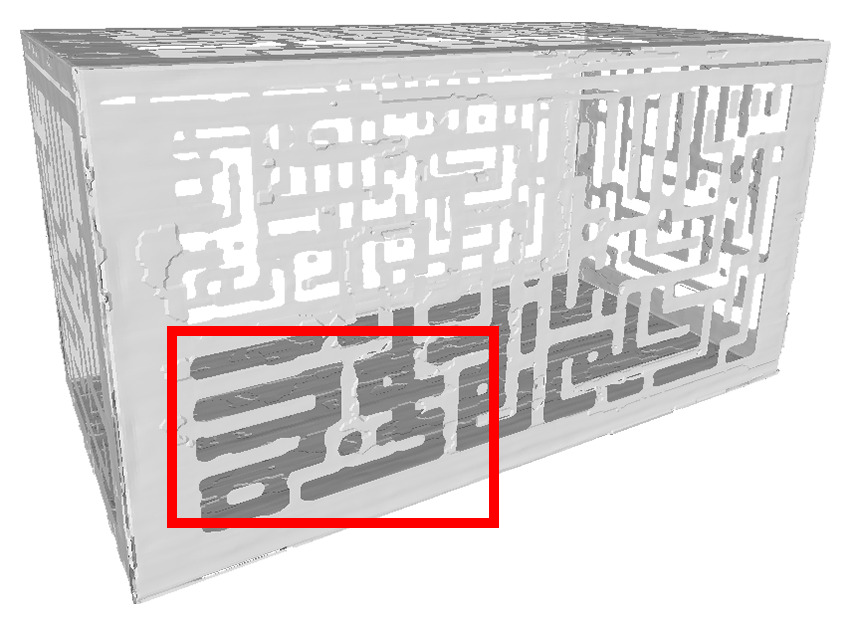} &
    \includegraphics[width=.58in]{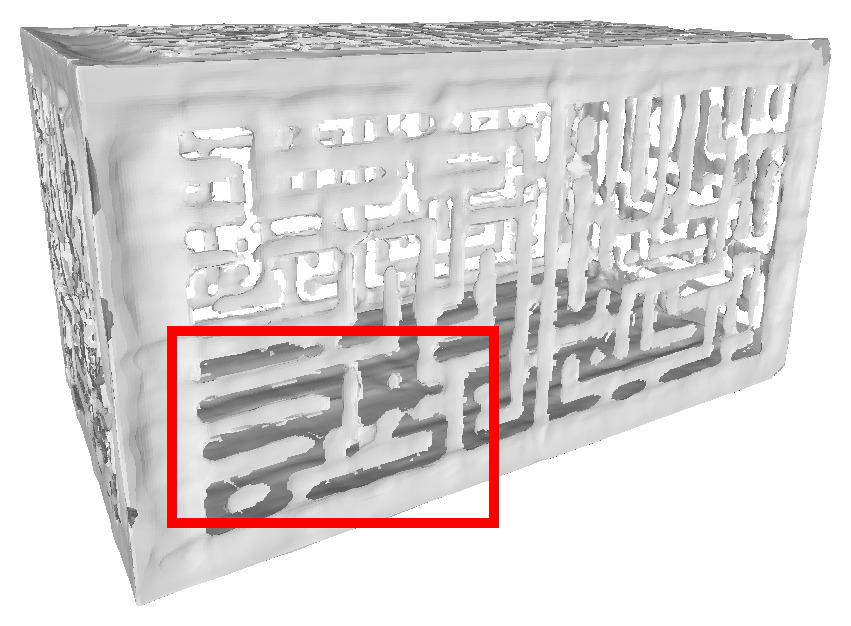} \\
    \raisebox{.14in}{\rotatebox{90}{plant}} & \includegraphics[width=.58in]{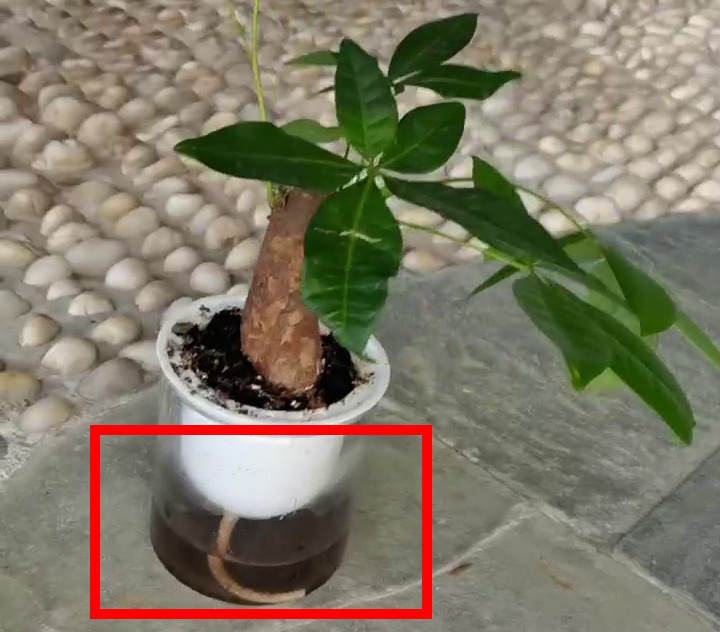} &
    \includegraphics[width=.58in]{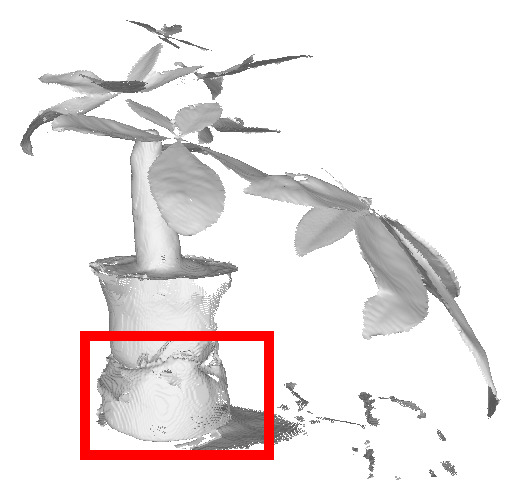} &
    \includegraphics[width=.58in]{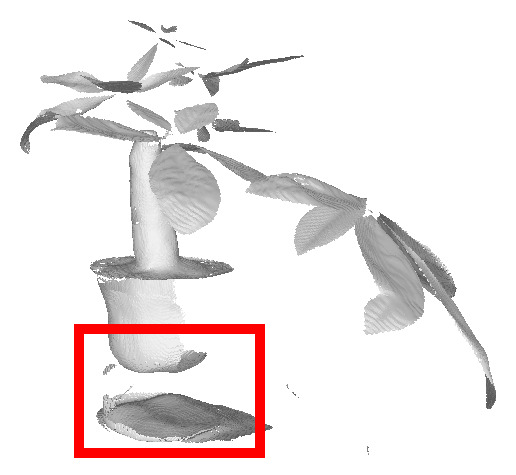} &
    \includegraphics[width=.58in]{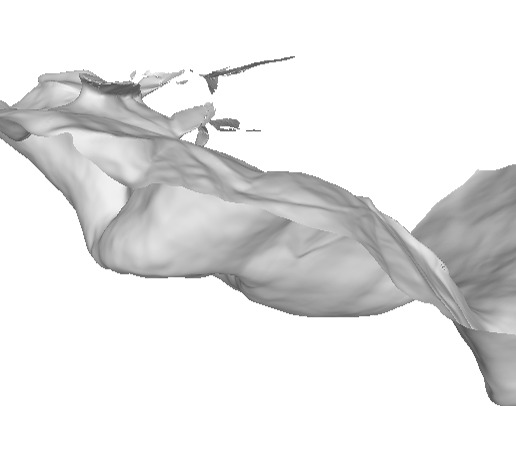} &
    \raisebox{.22in}{N.A.} \\
\end{tabular}
\caption{Qualitative comparisons with NeAT~\cite{Meng2023}, NeuralUDF~\cite{Long2023} and NeUDF~\cite{Liu2023NeUDF} on some example data released by NeUDF~\cite{Liu2023NeUDF}. Note that NeAT cannot reconstruct ``plant'' dataset because the ground truth mask for ``plant'' is unavailable.}
\label{fig:neudf-data}
\end{figure}

In Figure~\ref{fig:neudf-data}, we conduct additional experiments on some open model dataset provided by NeUDF~\cite{Liu2023NeUDF}. For the rack model, the thin structures reconstructed by NeuralUDF~\cite{Long2023} and NeUDF~\cite{Liu2023NeUDF} seem eroded, but ours don't. The thin structures reconstructed by NeAT~\cite{Meng2023} is the closest to the reference image, but the surface is dented inward with visible artifacts due to imperfect SDF validity learning. The plant model does not have an object mask, making NeAT~\cite{Meng2023} impractical for training. NeuralUDF~\cite{Long2023} completely fails to reconstruct a reasonable surface. Between our method and NeUDF~\cite{Liu2023NeUDF} which can reconstruct a sensible model, the flower pot region marked in red is missing in NeUDF but not in ours. These show our method's ability to reconstruct non-watertight models more robustly compared to other methods.

\subsection{Comparisons on Watertight Models}
\label{sec:watertight}

\begin{table}
\centering
\setlength\tabcolsep{1.25pt}
\begin{footnotesize}
\begin{tabular}{c|cccccccccc|c}
    \hline
    Method & 37 & 55 & 65 & 69 & 97 & 105 & 106 & 114 & 118 & 122 & Mean \\
    \hline
    NeuralUDF & 1.18 & \textbf{0.44} & \textbf{0.66} & \textbf{0.67} & \textbf{0.94} & 0.95 & \textbf{0.57} & \textbf{0.37} & \textbf{0.56} & \underline{0.55} & \textbf{0.69} \\
    NeAT & 1.18 & \underline{0.47} & 0.82 & \underline{0.84} & 1.09 & 0.75 & 0.76 & \underline{0.38} & \textbf{0.56} & \underline{0.55} & 0.74 \\
    \hline
    NeUDF & \underline{0.90} & 0.65 & 0.73 & 0.97 & \underline{1.07} & \textbf{0.63} & 0.94 & 0.59 & 0.72 & 0.62 & 0.78 \\
    Ours & \textbf{0.89} & 0.55 & \underline{0.68} & 0.88 & 1.15 & \underline{0.70} & \underline{0.74} & 0.41 & \underline{0.61} & \textbf{0.51} & \underline{0.71} \\
    \hline
\end{tabular}
\end{footnotesize}
\caption{Chamfer distances on DTU dataset.}
\label{tab:dtu}
\end{table}

\begin{figure}
\centering
\setlength\tabcolsep{1pt}
\begin{tabular}{ccccc}
    & 55 & 65 & 118 & 122 \\
    \raisebox{.18in}{\rotatebox{90}{Ref.}} & \includegraphics[width=.75in]{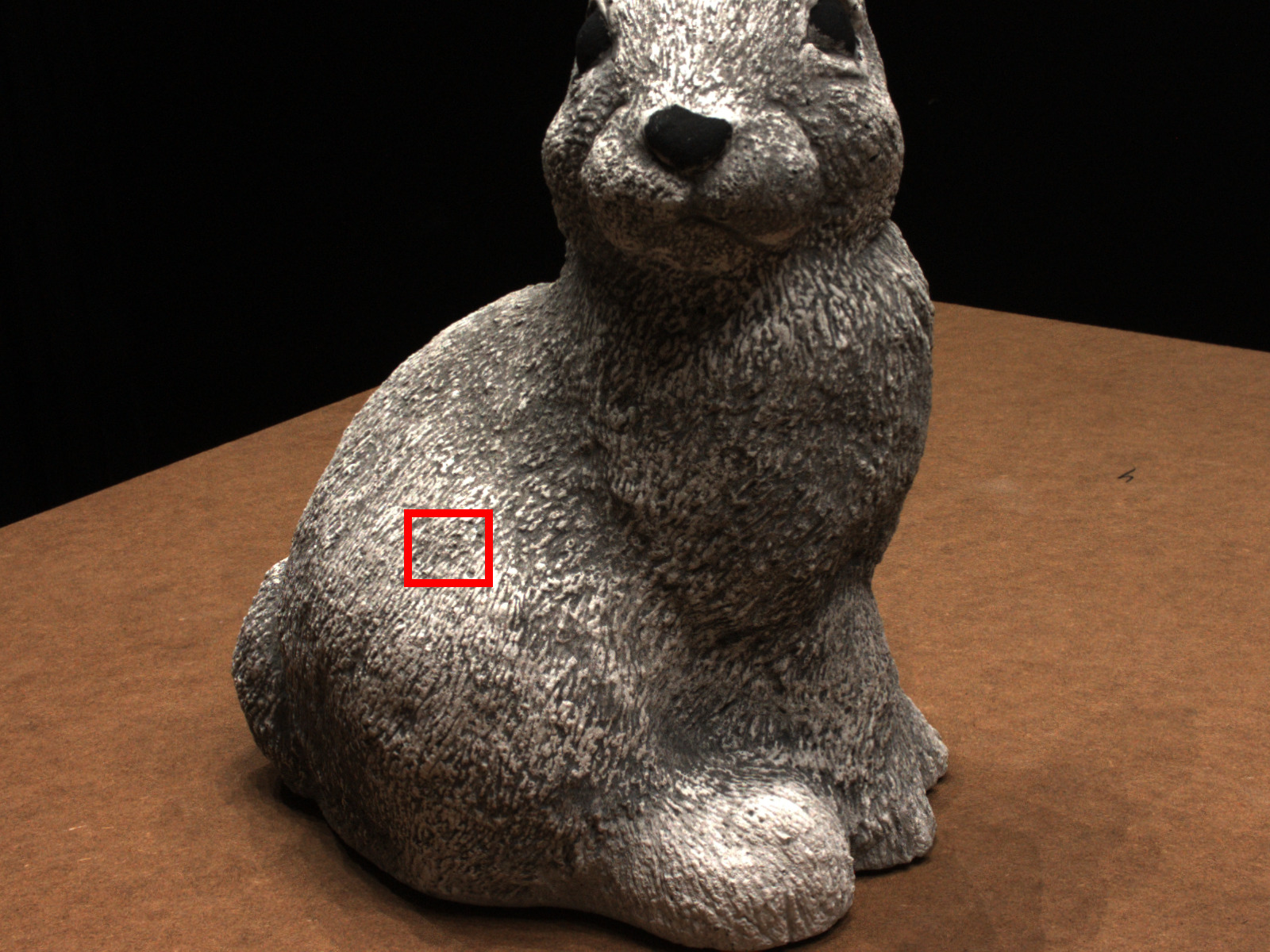} & \includegraphics[width=.75in]{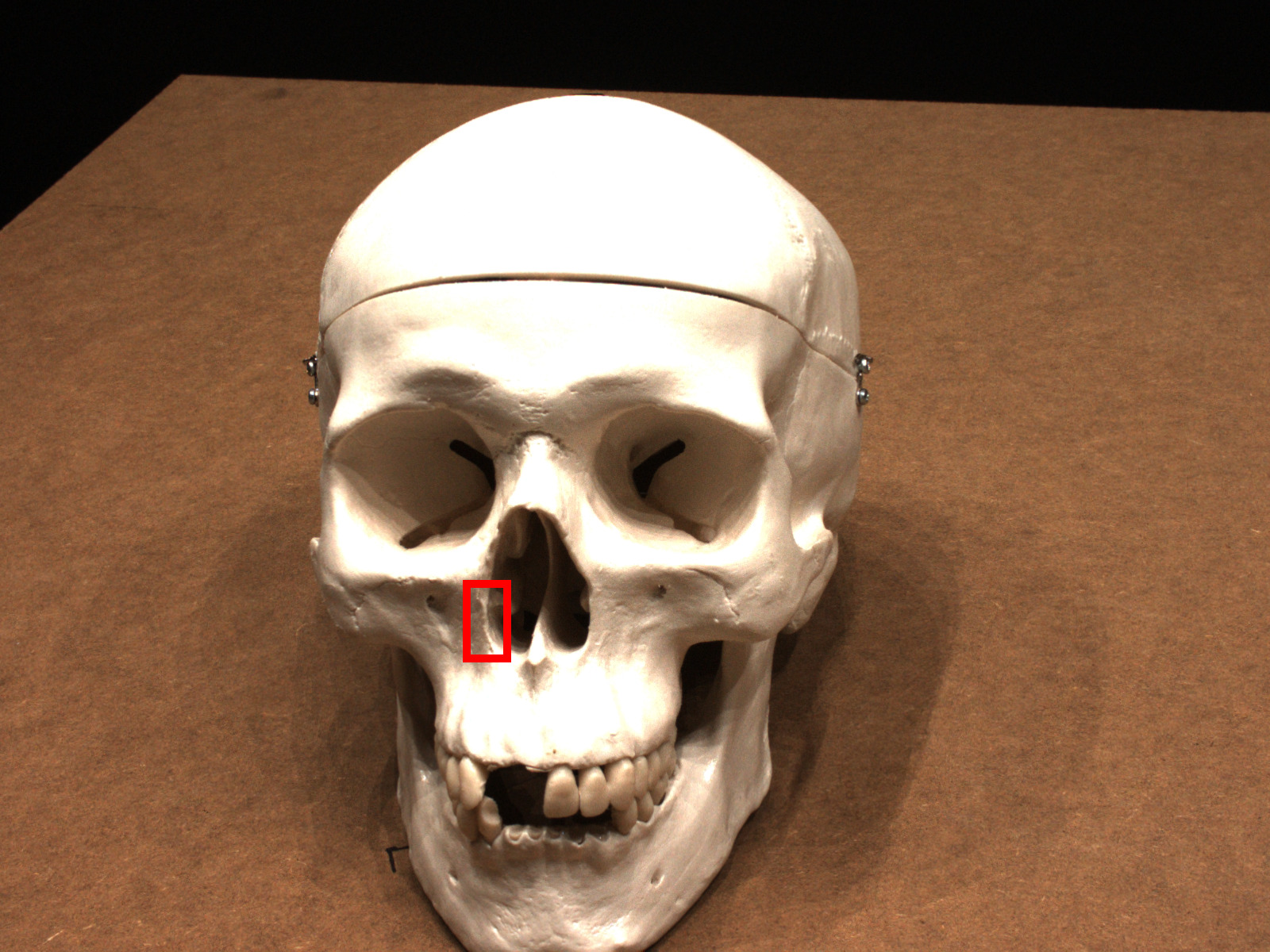} & \includegraphics[width=.75in]{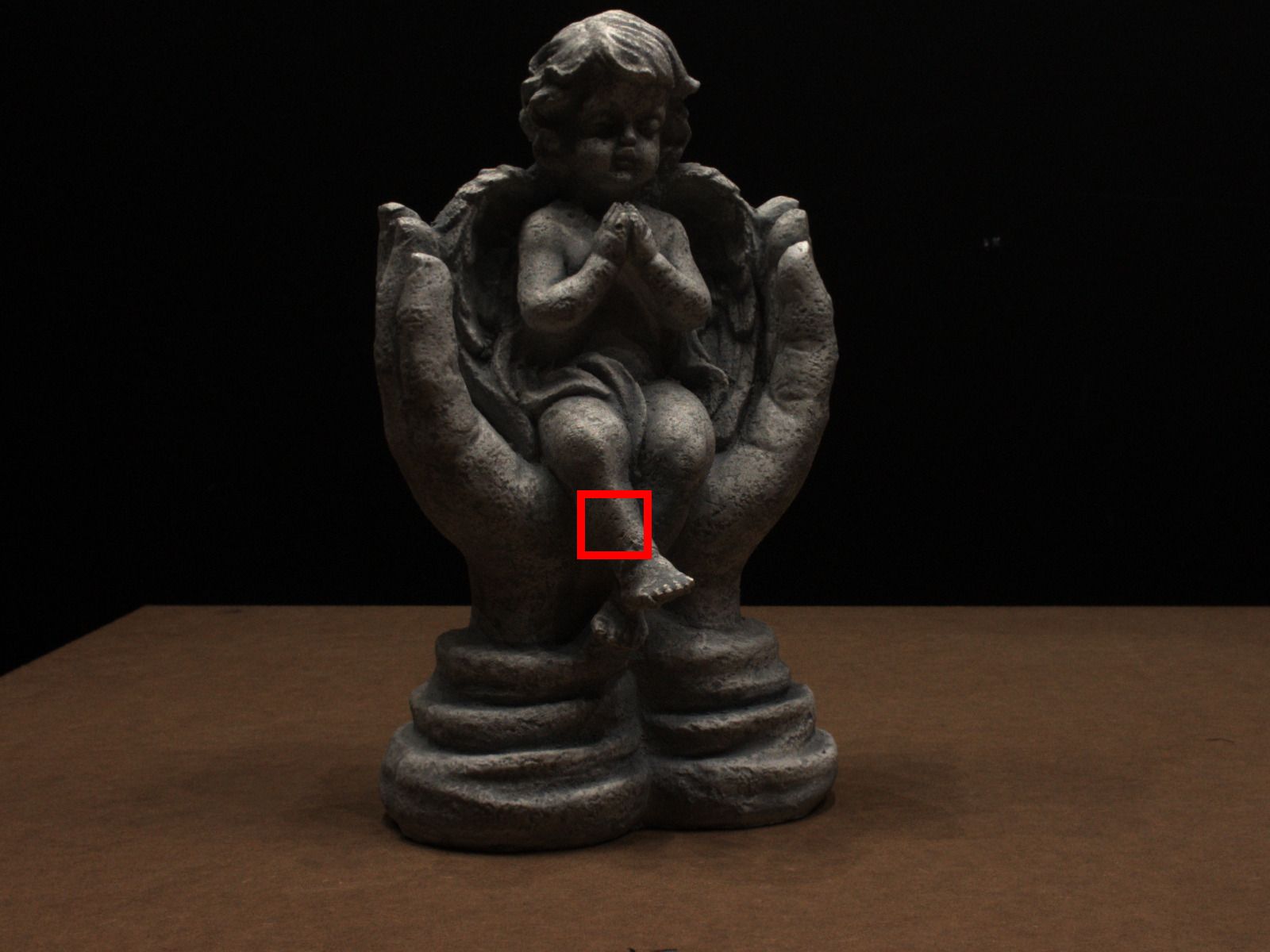} & \includegraphics[width=.75in]{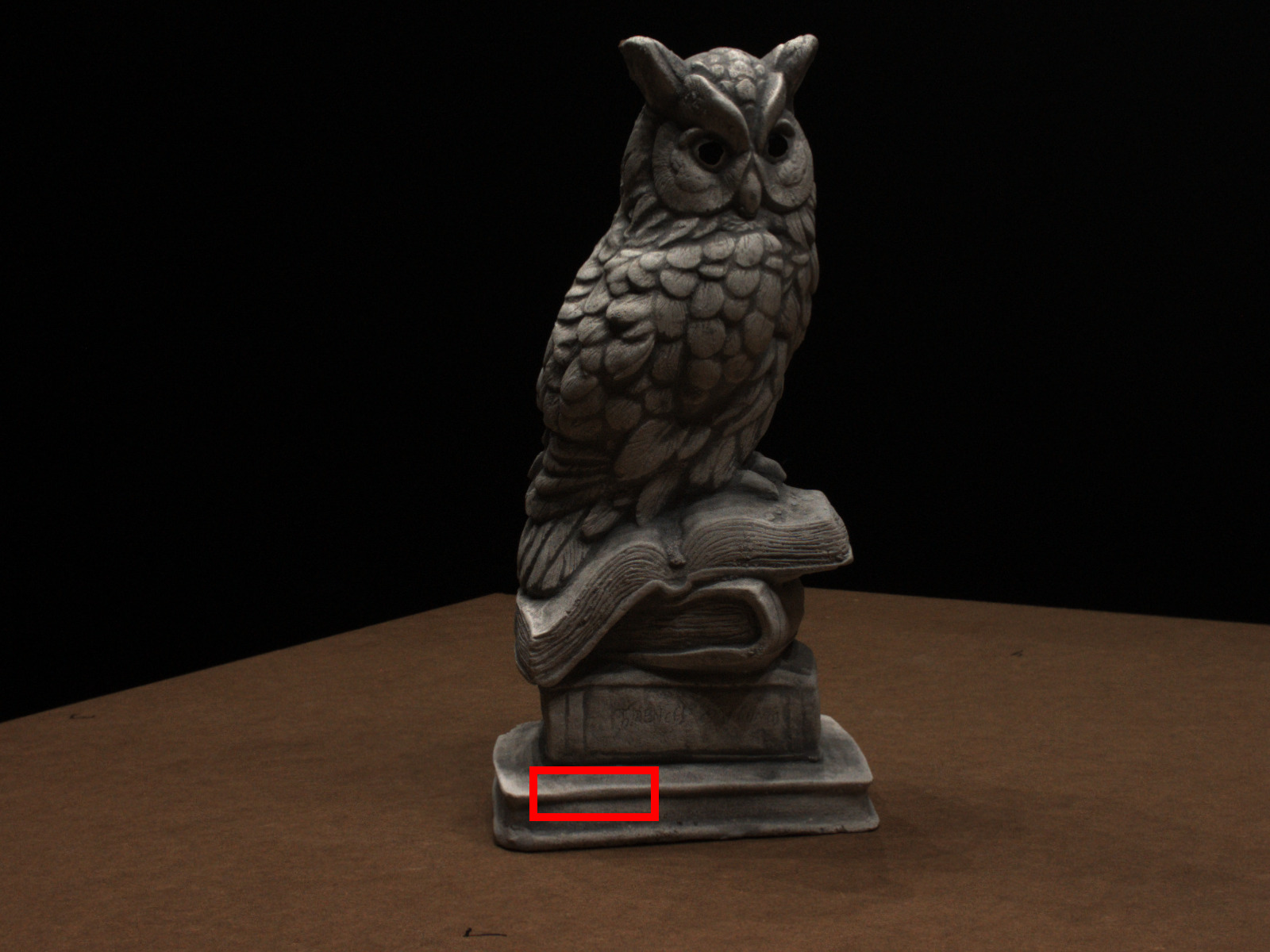} \\
    \raisebox{.18in}{\rotatebox{90}{Ours}} & \includegraphics[width=.375in]{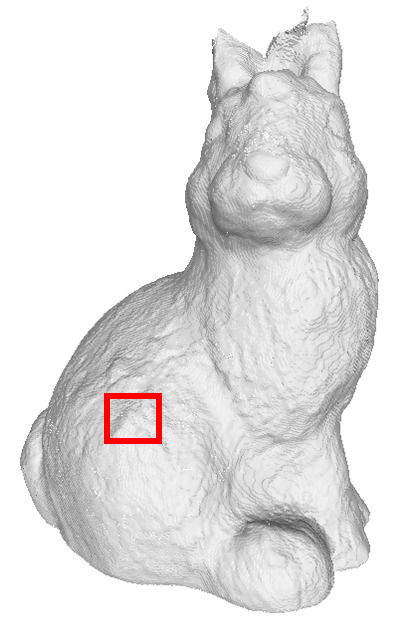}\includegraphics[width=.375in]{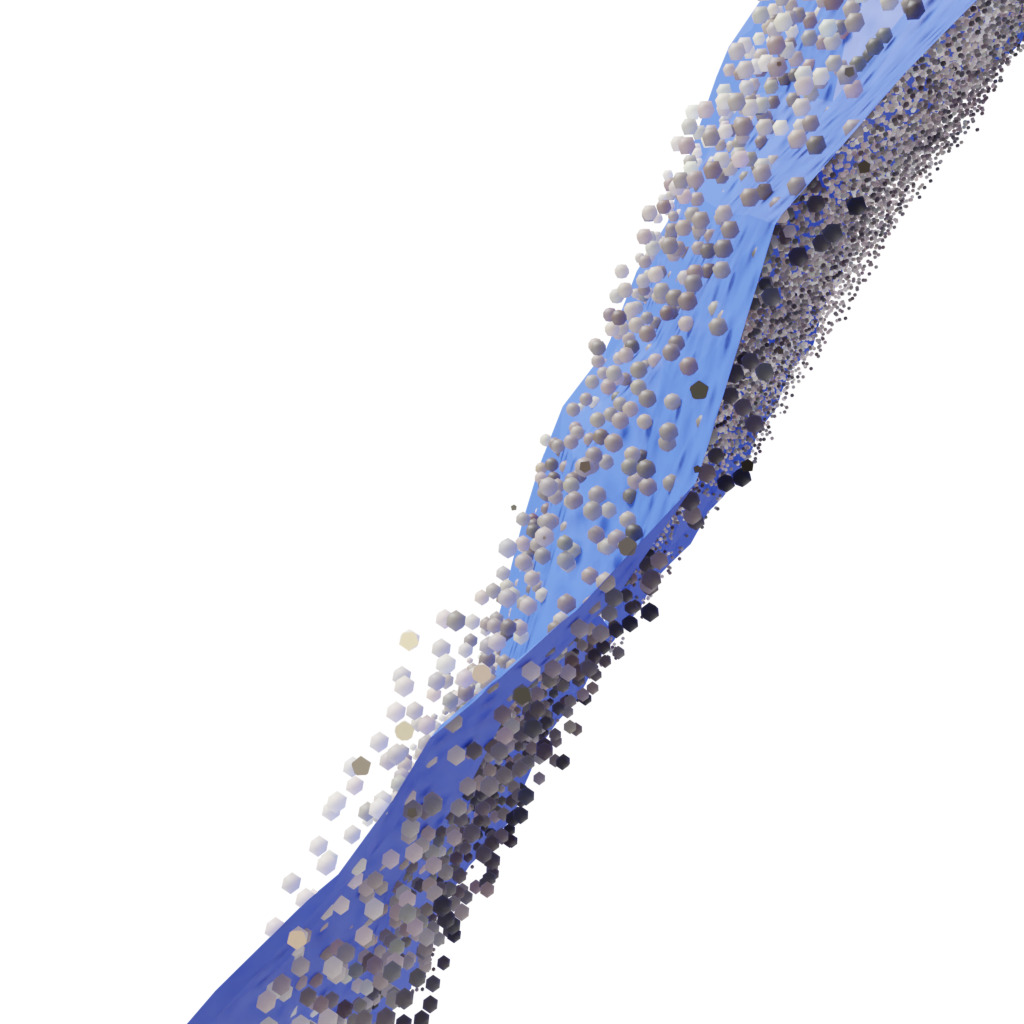} & \includegraphics[width=.375in]{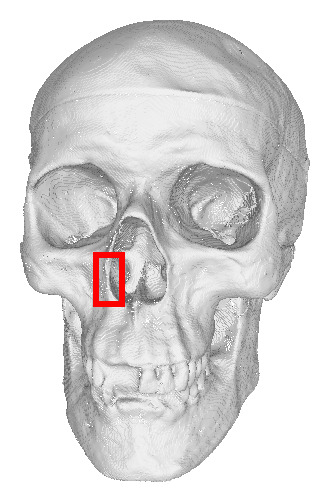}\includegraphics[width=.375in]{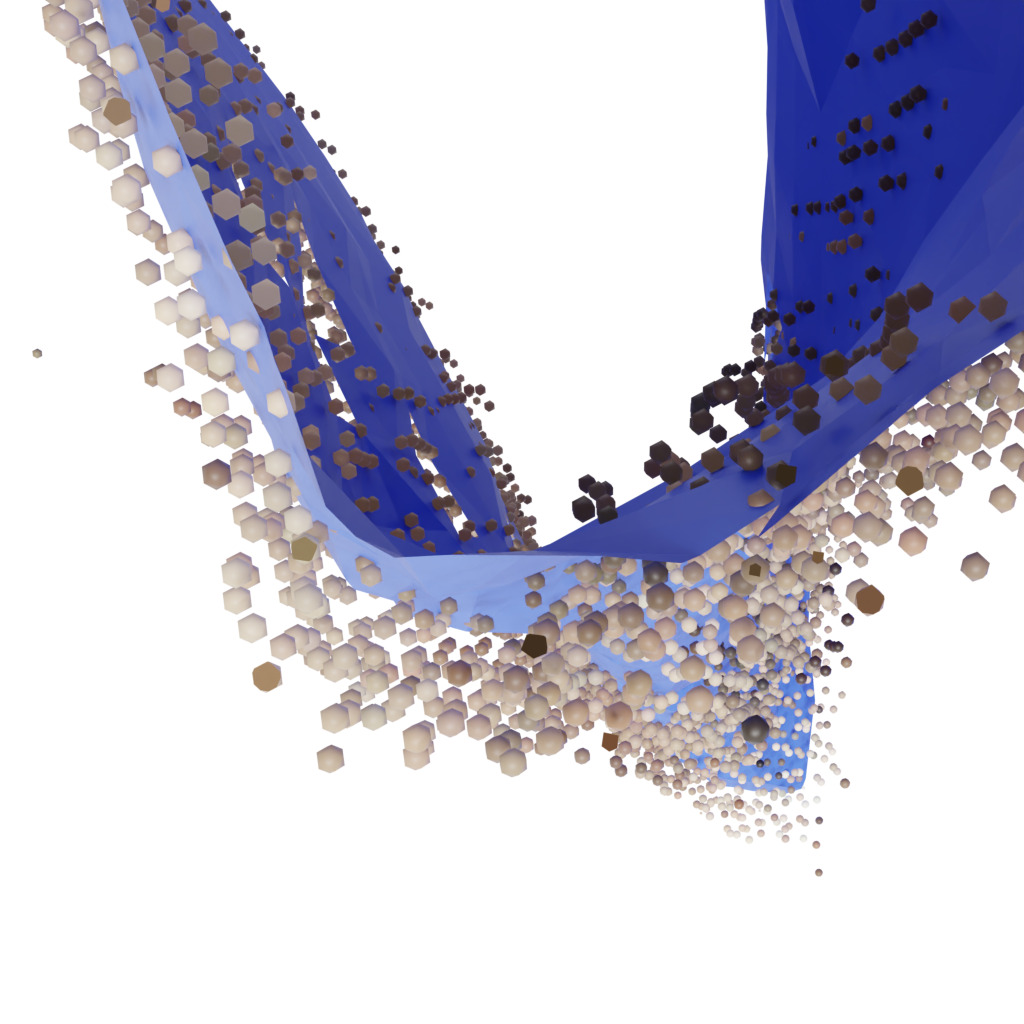} & \includegraphics[width=.342in]{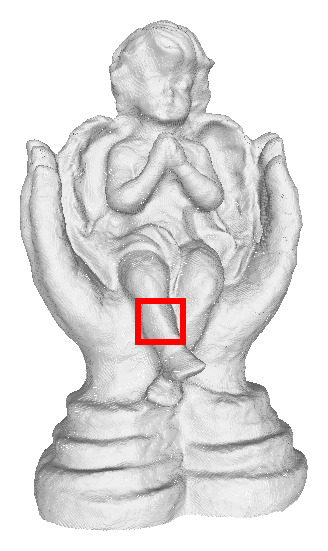}\includegraphics[width=.342in]{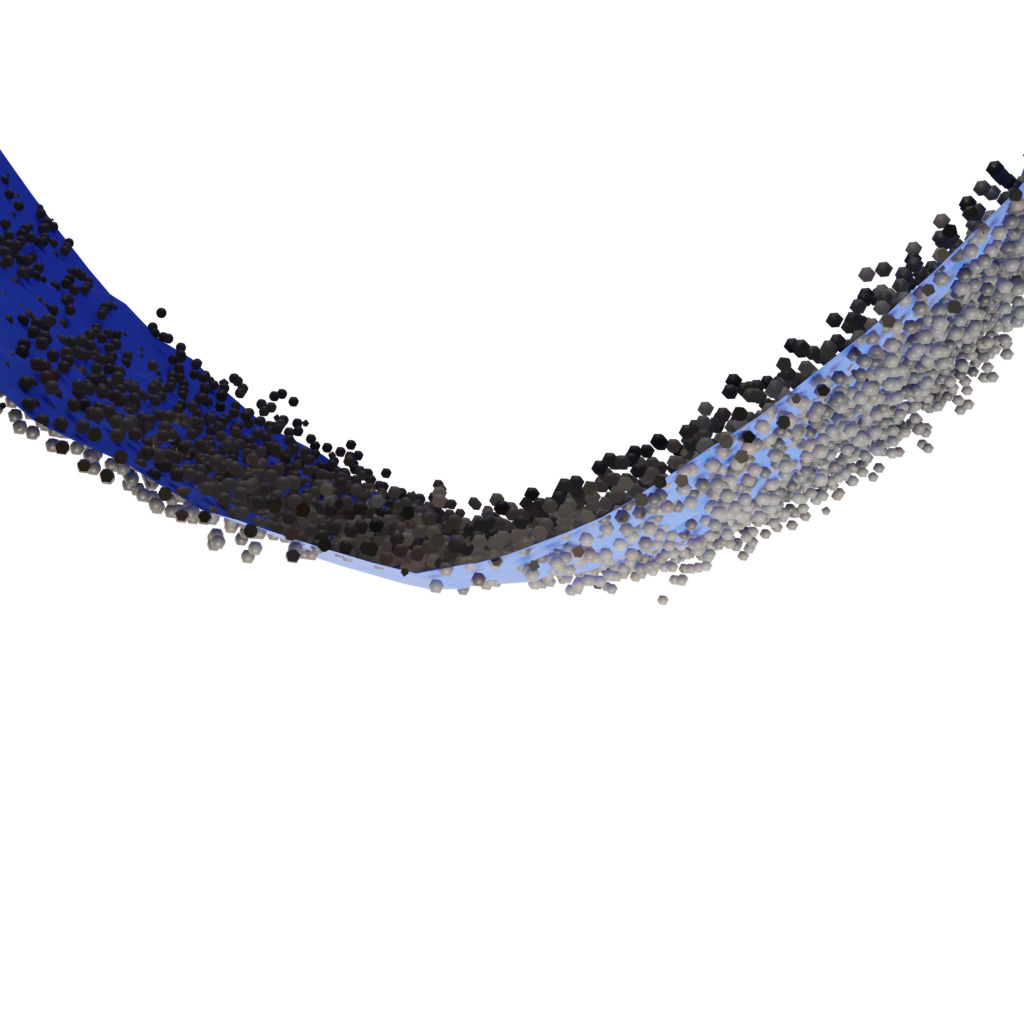} & \includegraphics[width=.292in]{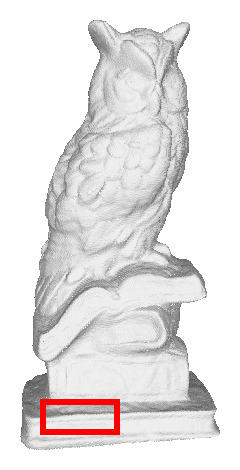}\includegraphics[width=.292in]{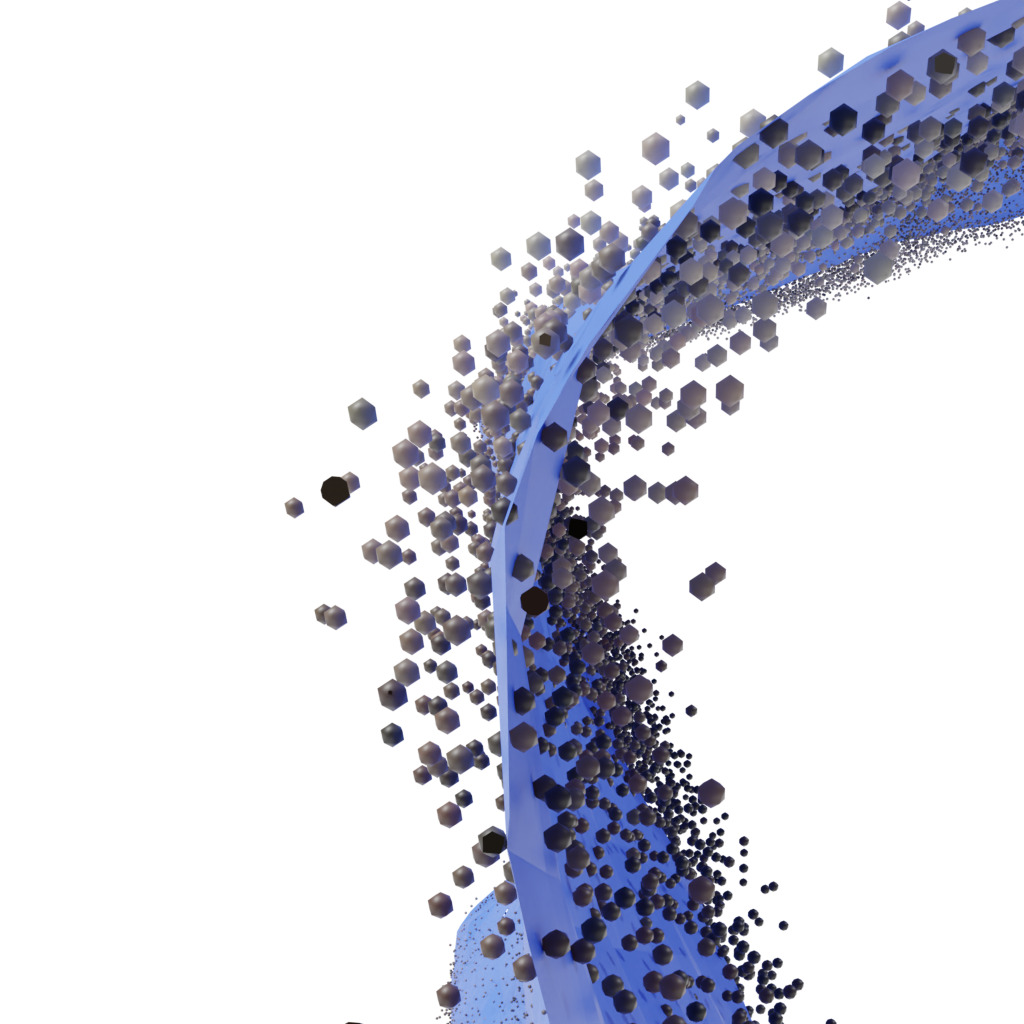} \\
    \raisebox{.1in}{\rotatebox{90}{NeUDF}} & \includegraphics[width=.375in]{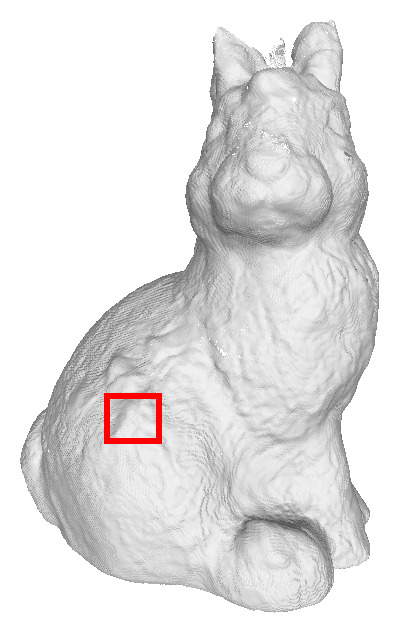}\includegraphics[width=.375in]{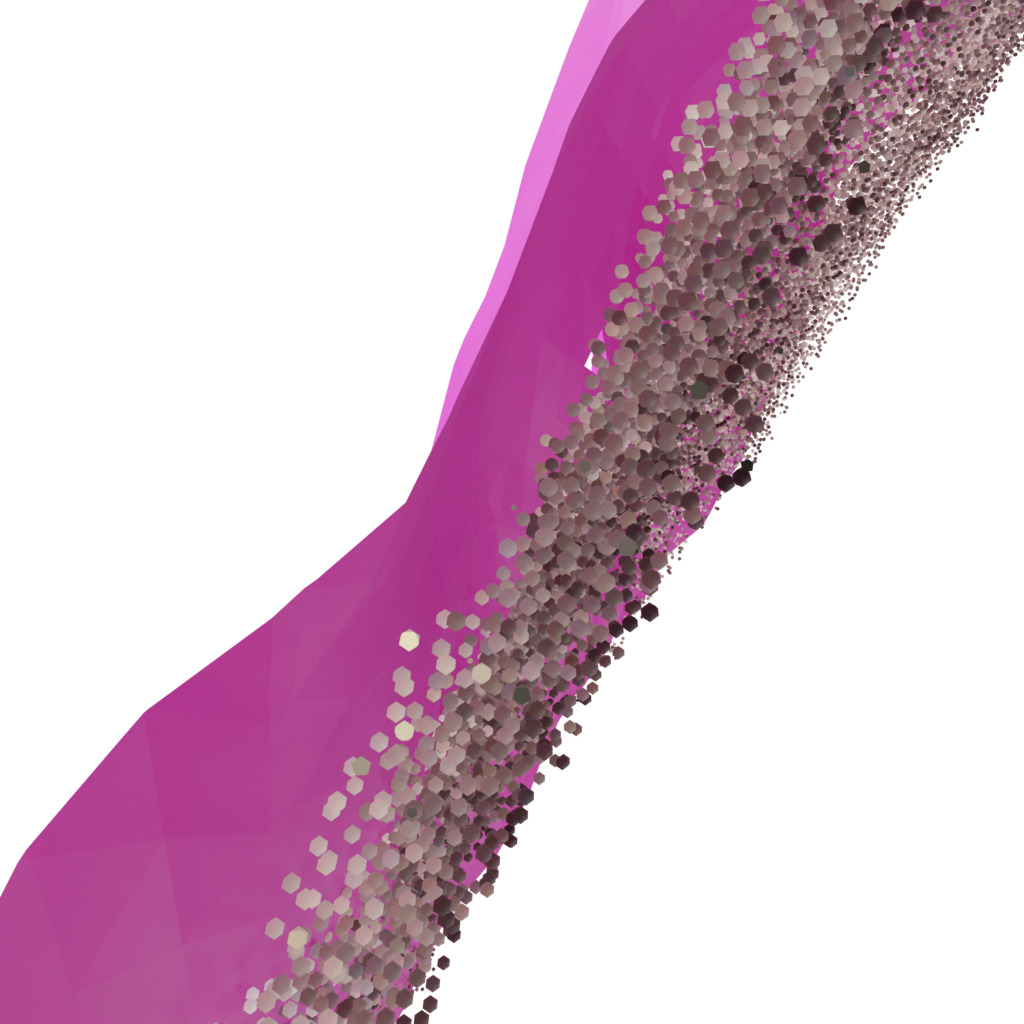} & \includegraphics[width=.375in]{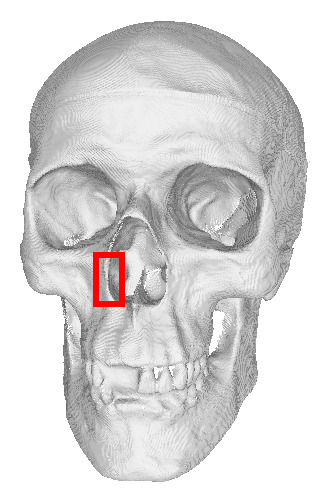}\includegraphics[width=.375in]{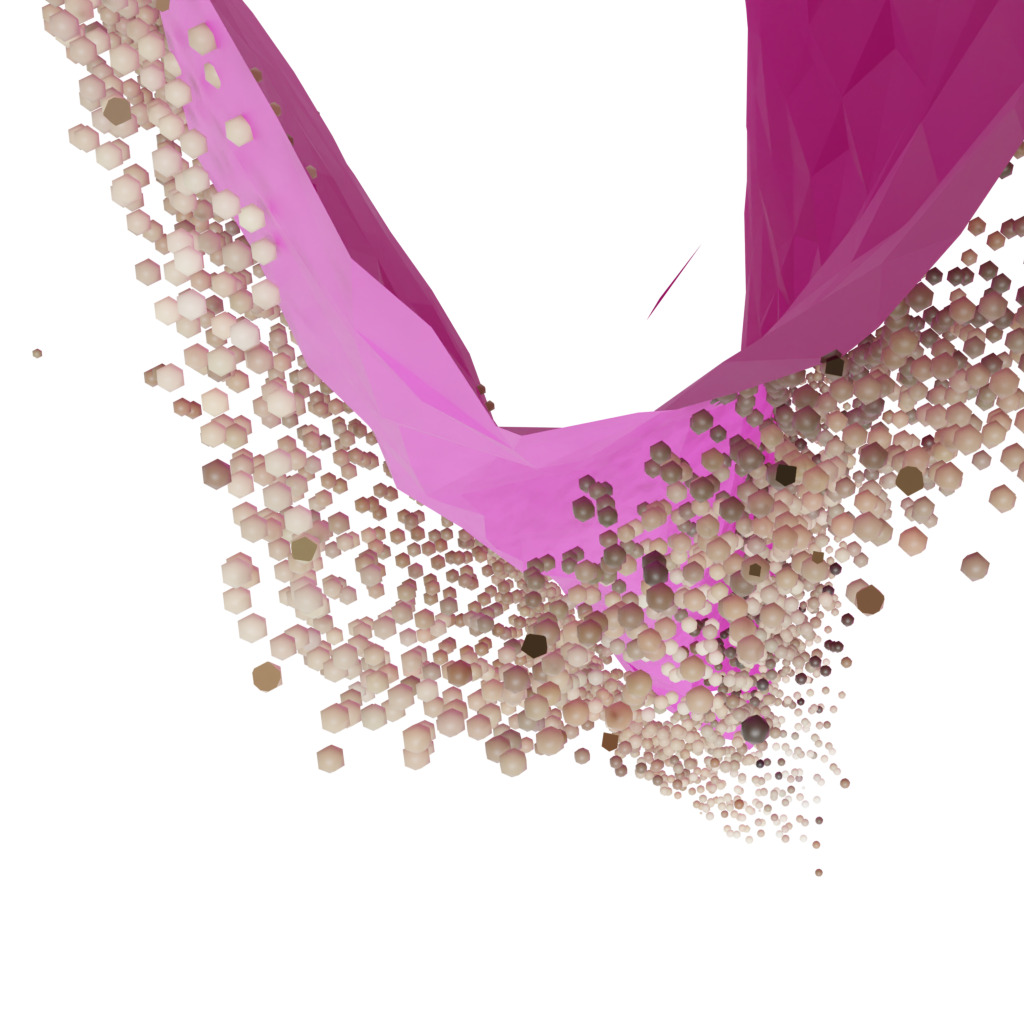} & \includegraphics[width=.342in]{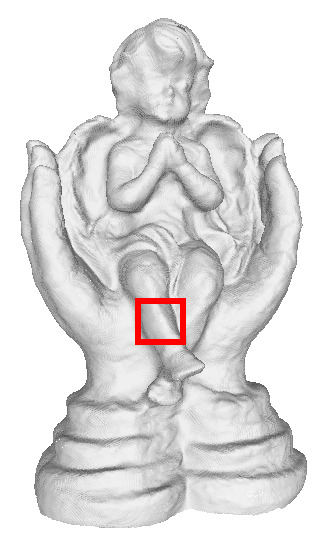}\includegraphics[width=.342in]{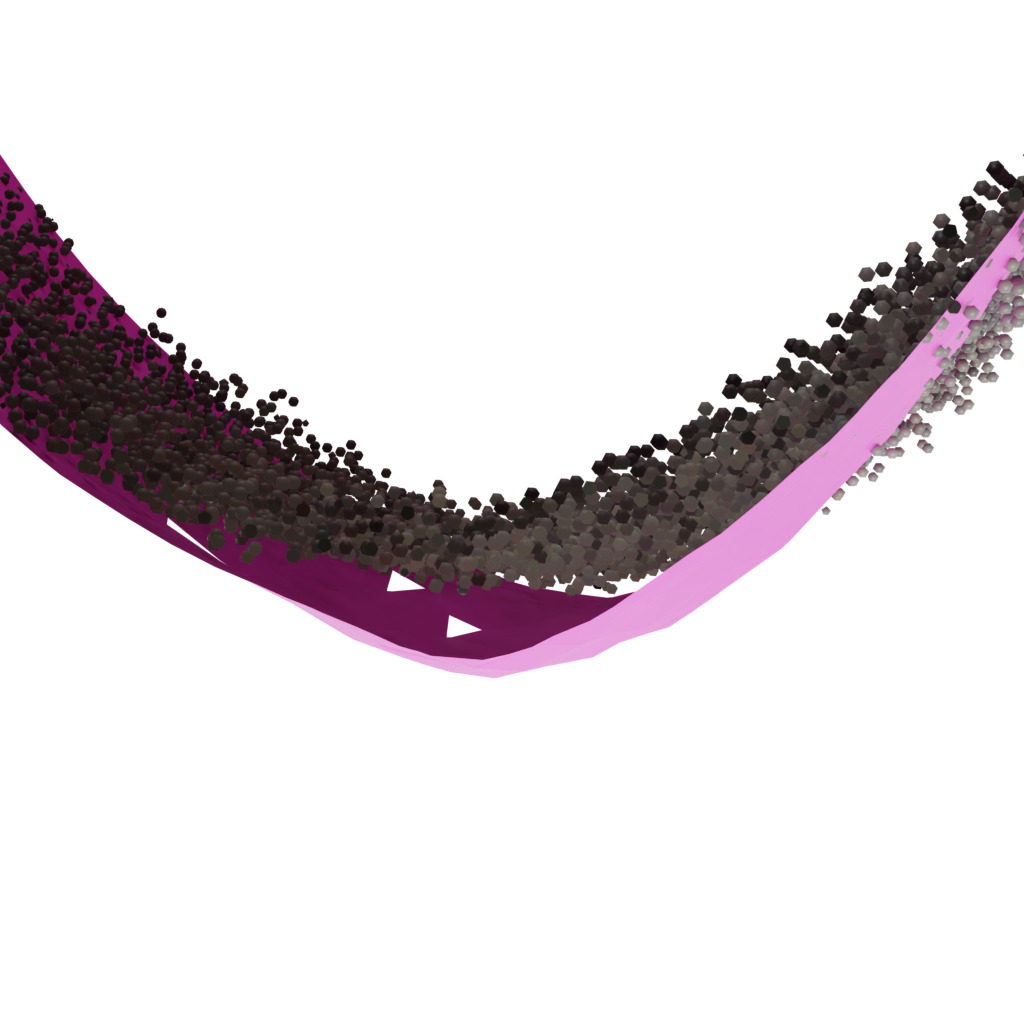} & \includegraphics[width=.292in]{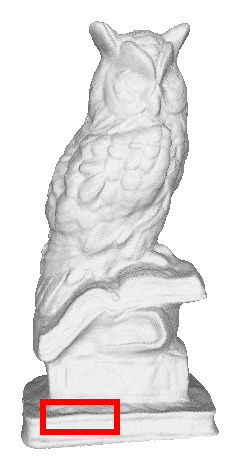}\includegraphics[width=.292in]{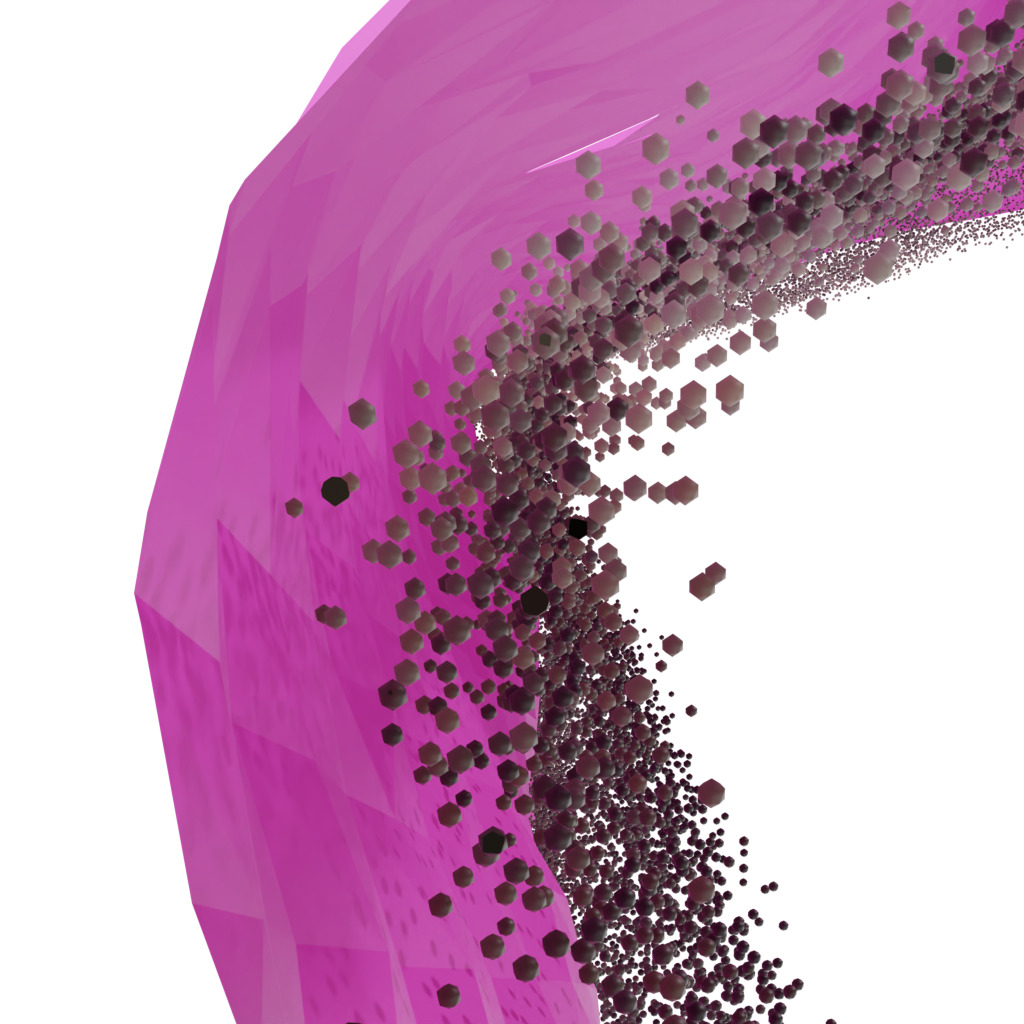} \\
    \scalebox{0.85}{\rotatebox{90}{NeuralUDF}} & \includegraphics[width=.375in]{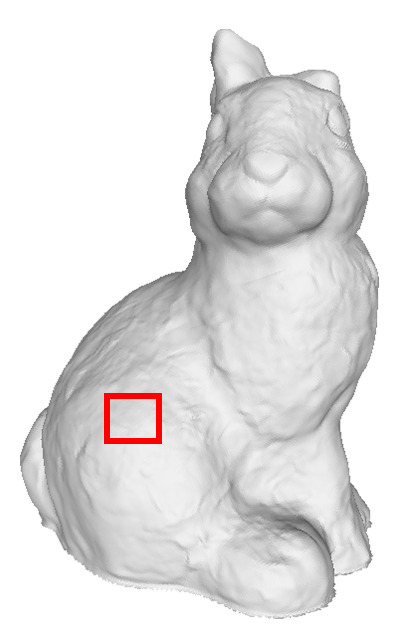}\includegraphics[width=.375in]{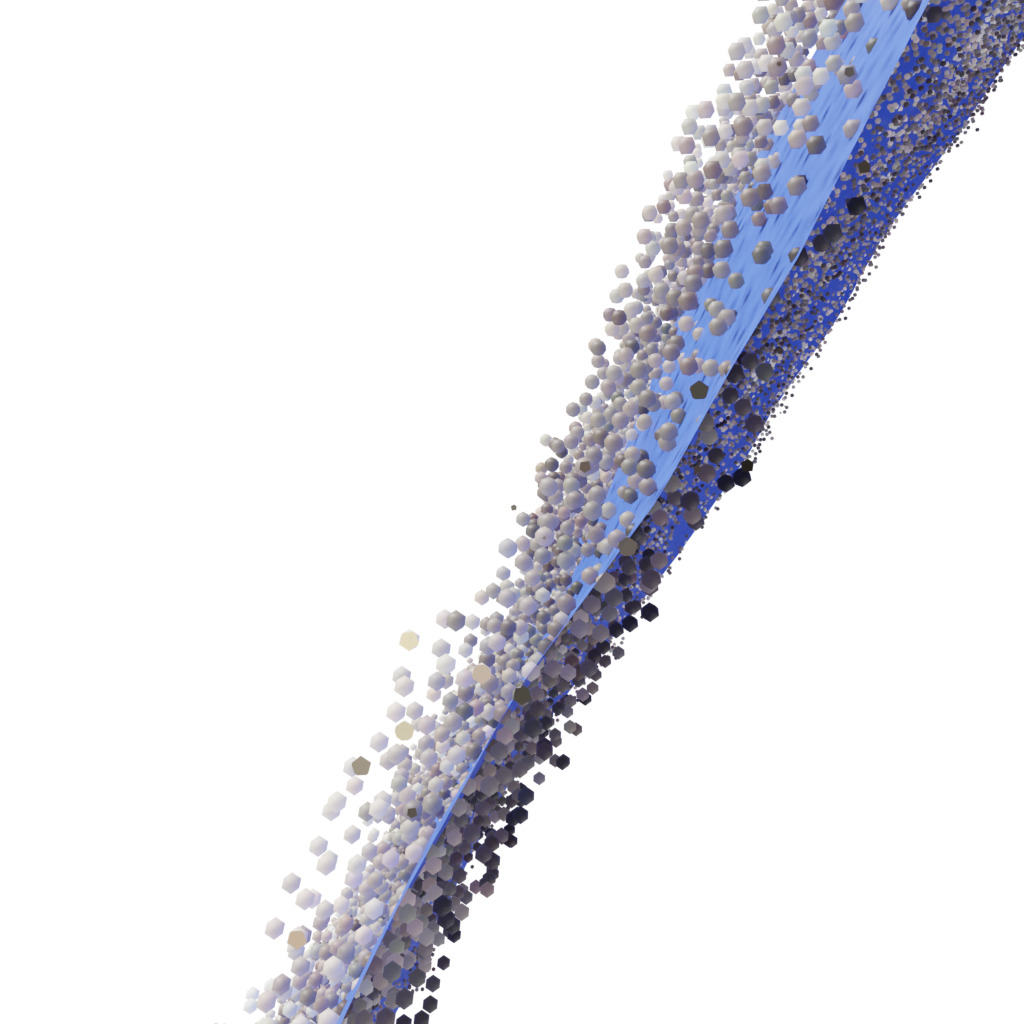} & \includegraphics[width=.375in]{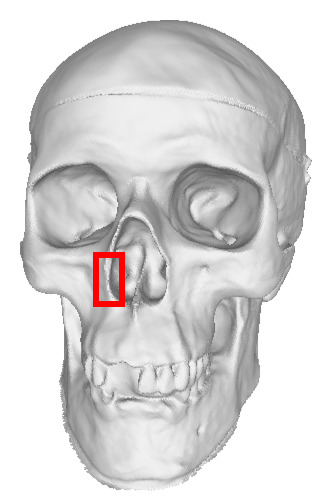}\includegraphics[width=.375in]{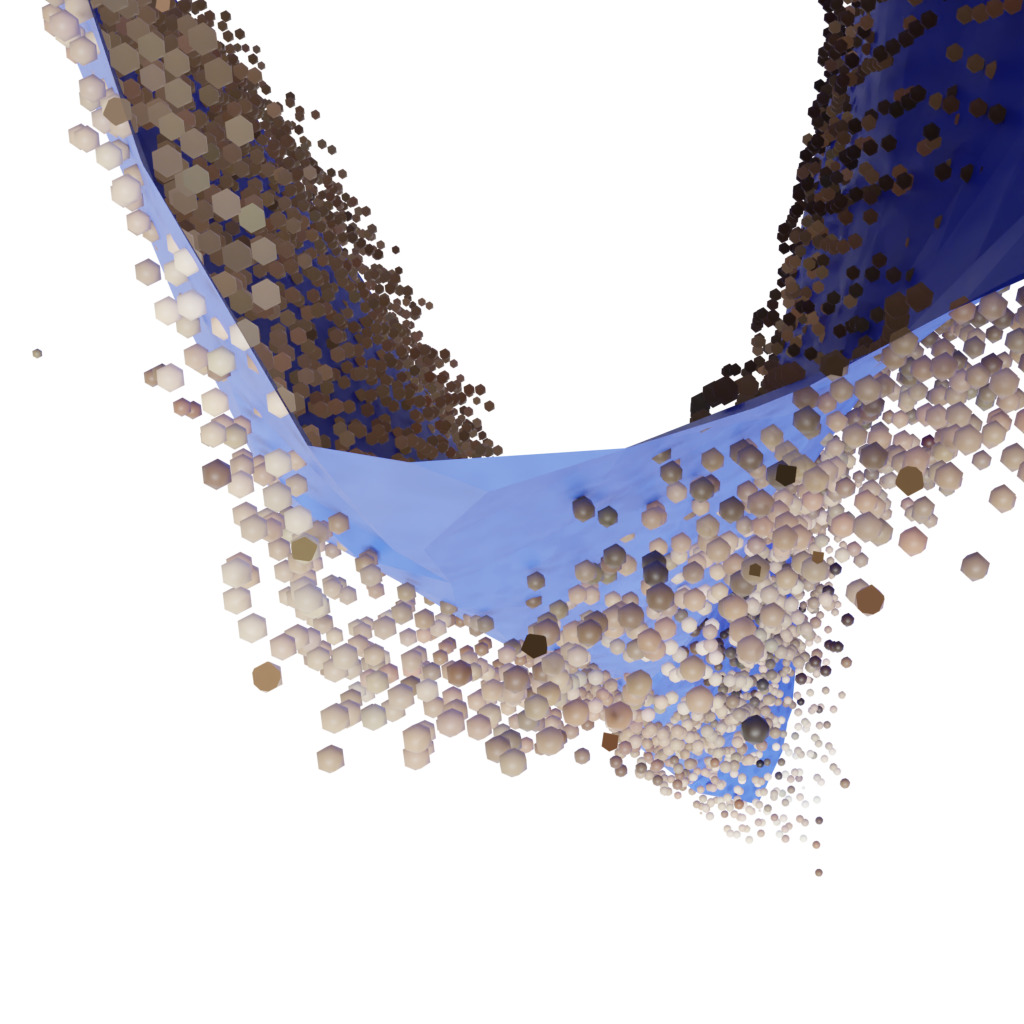} & \includegraphics[width=.342in]{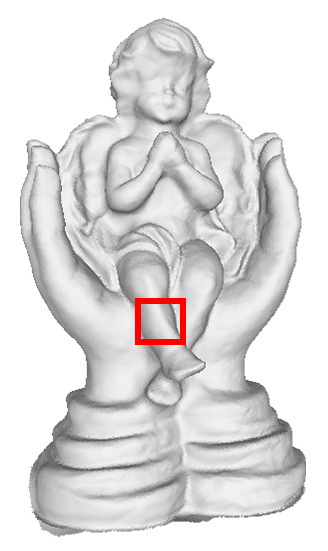}\includegraphics[width=.342in]{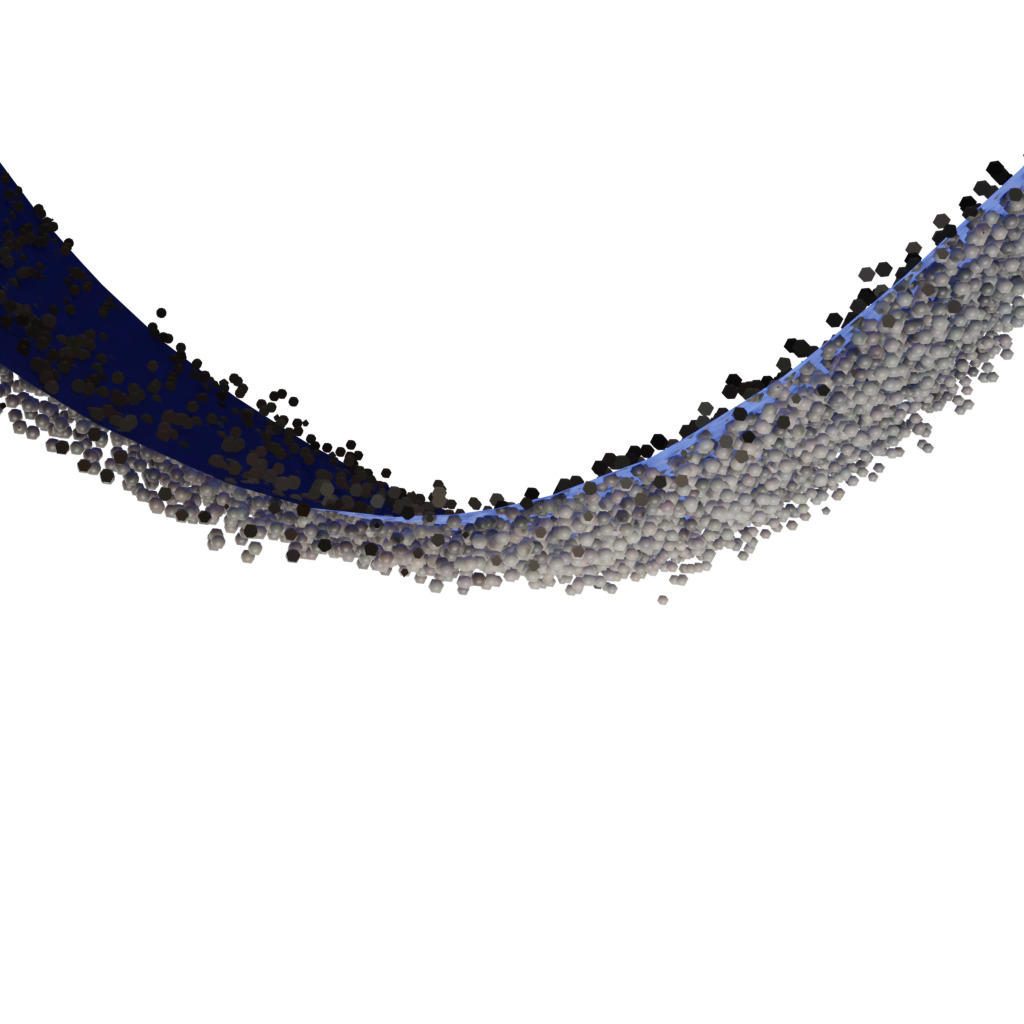} & \includegraphics[width=.292in]{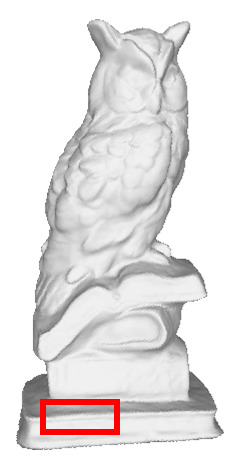}\includegraphics[width=.292in]{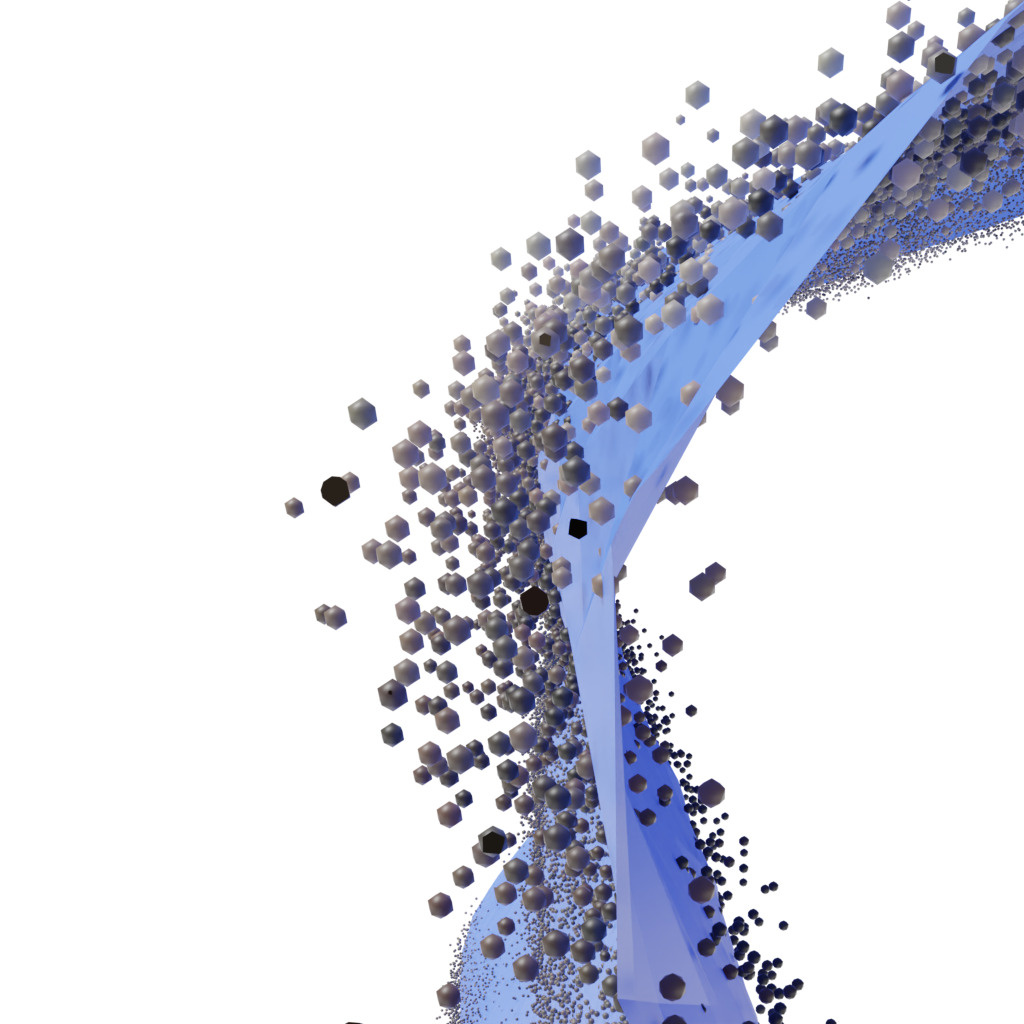} \\
    \raisebox{.18in}{\rotatebox{90}{NeAT}} & \includegraphics[width=.375in]{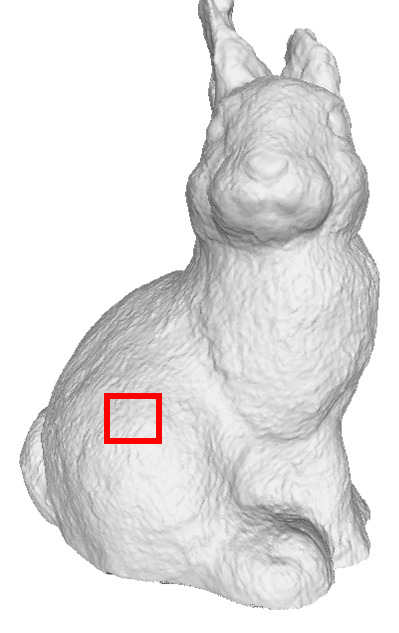}\includegraphics[width=.375in]{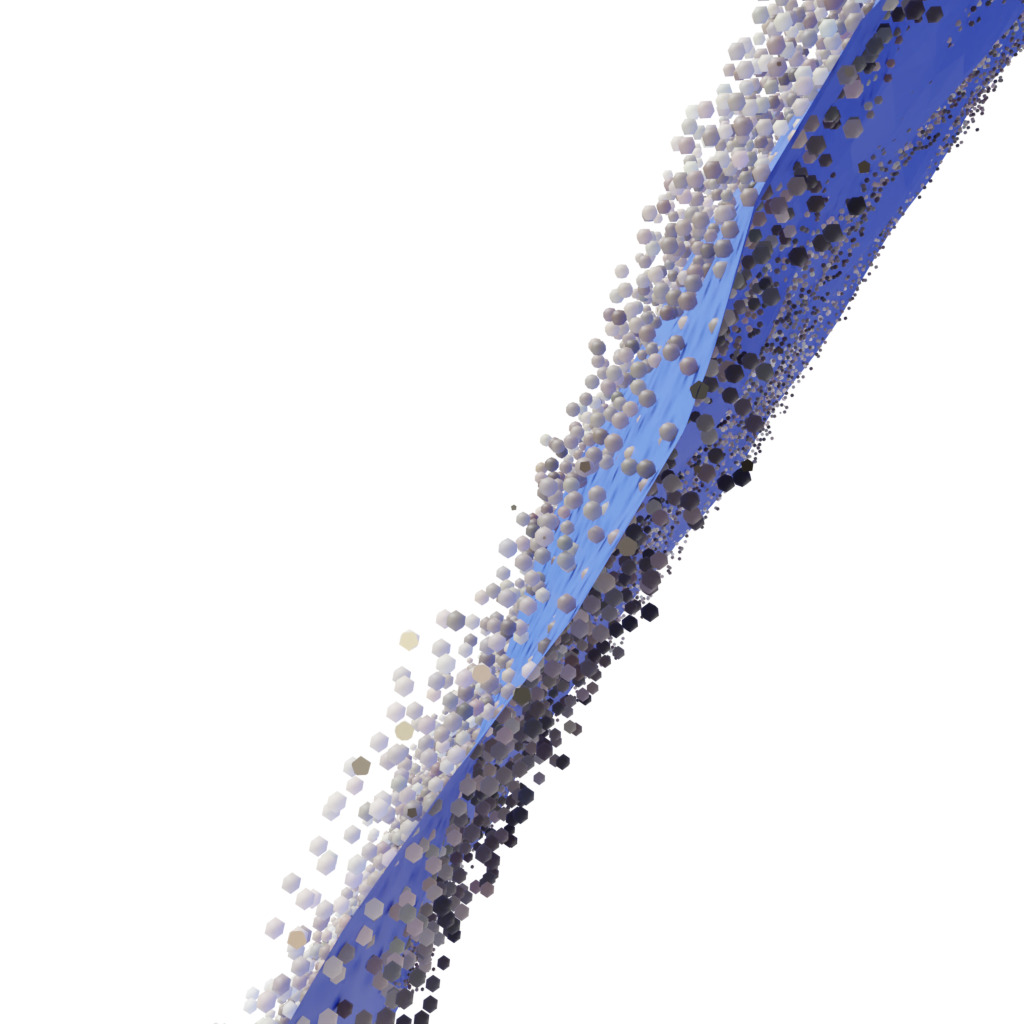} & \includegraphics[width=.375in]{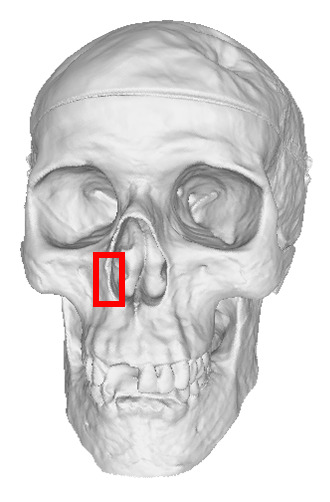}\includegraphics[width=.375in]{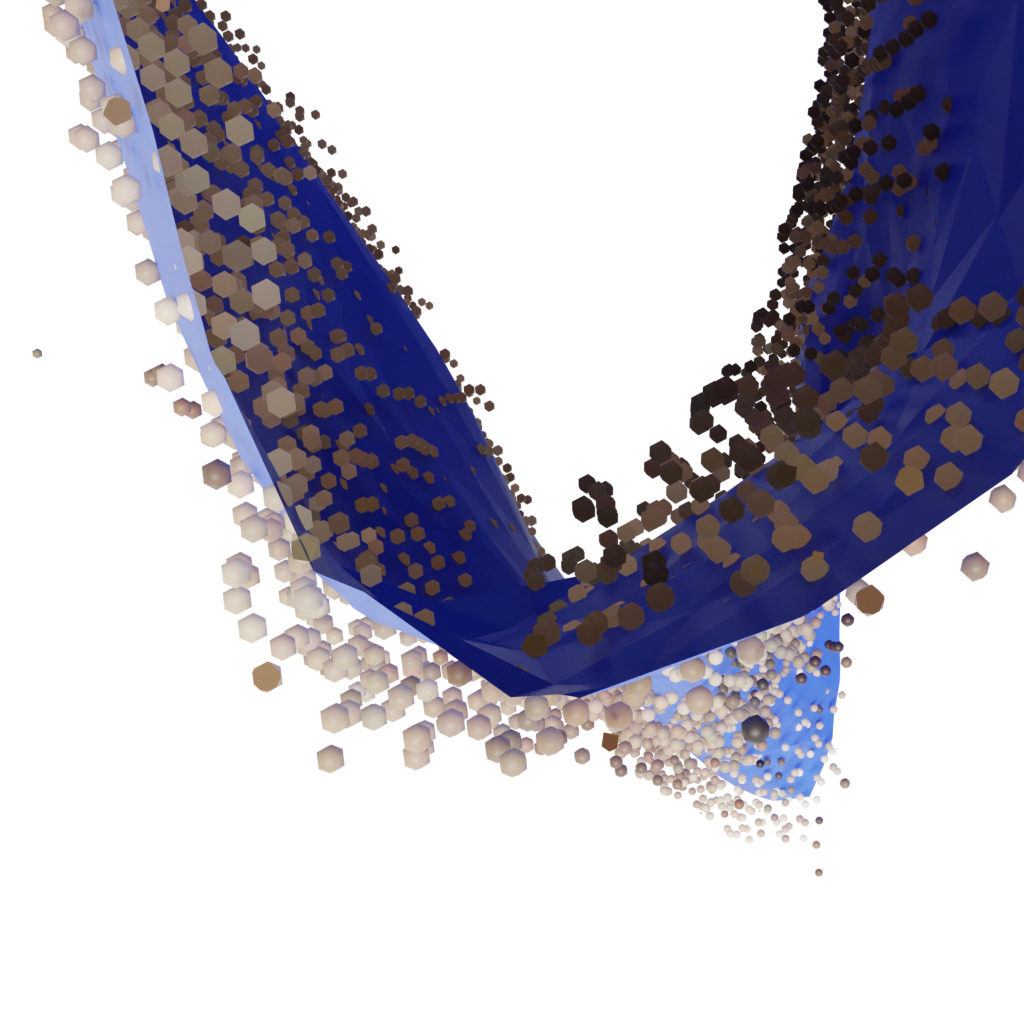} & \includegraphics[width=.342in]{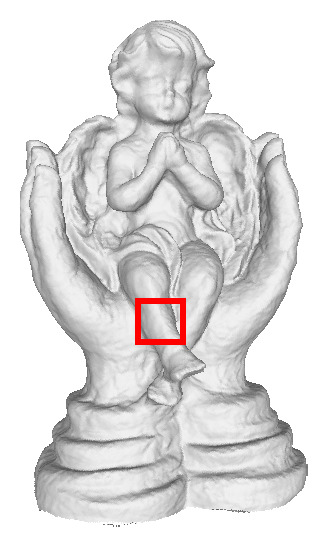}\includegraphics[width=.342in]{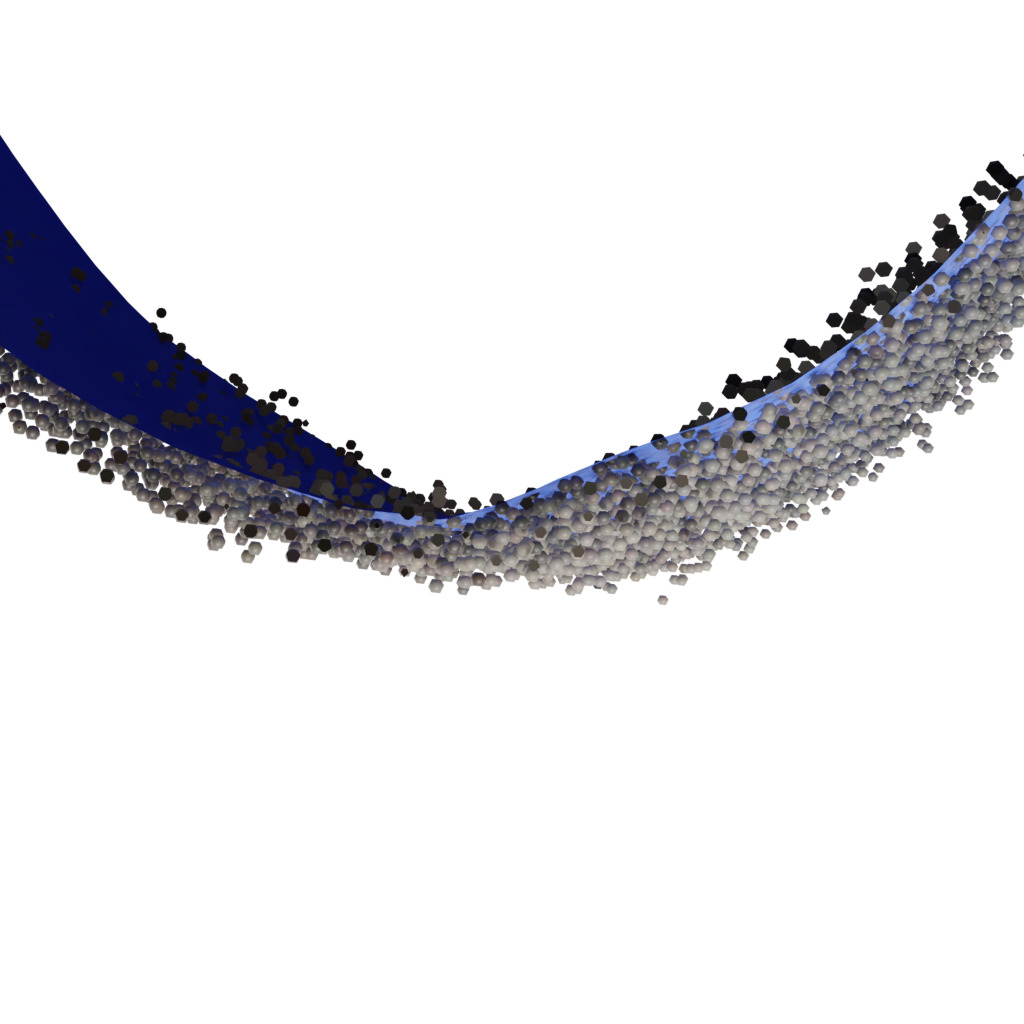} & \includegraphics[width=.292in]{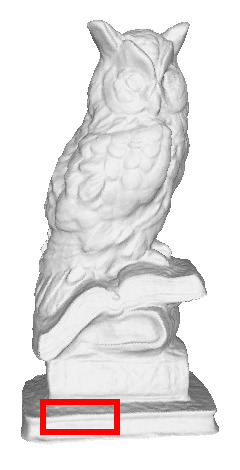}\includegraphics[width=.292in]{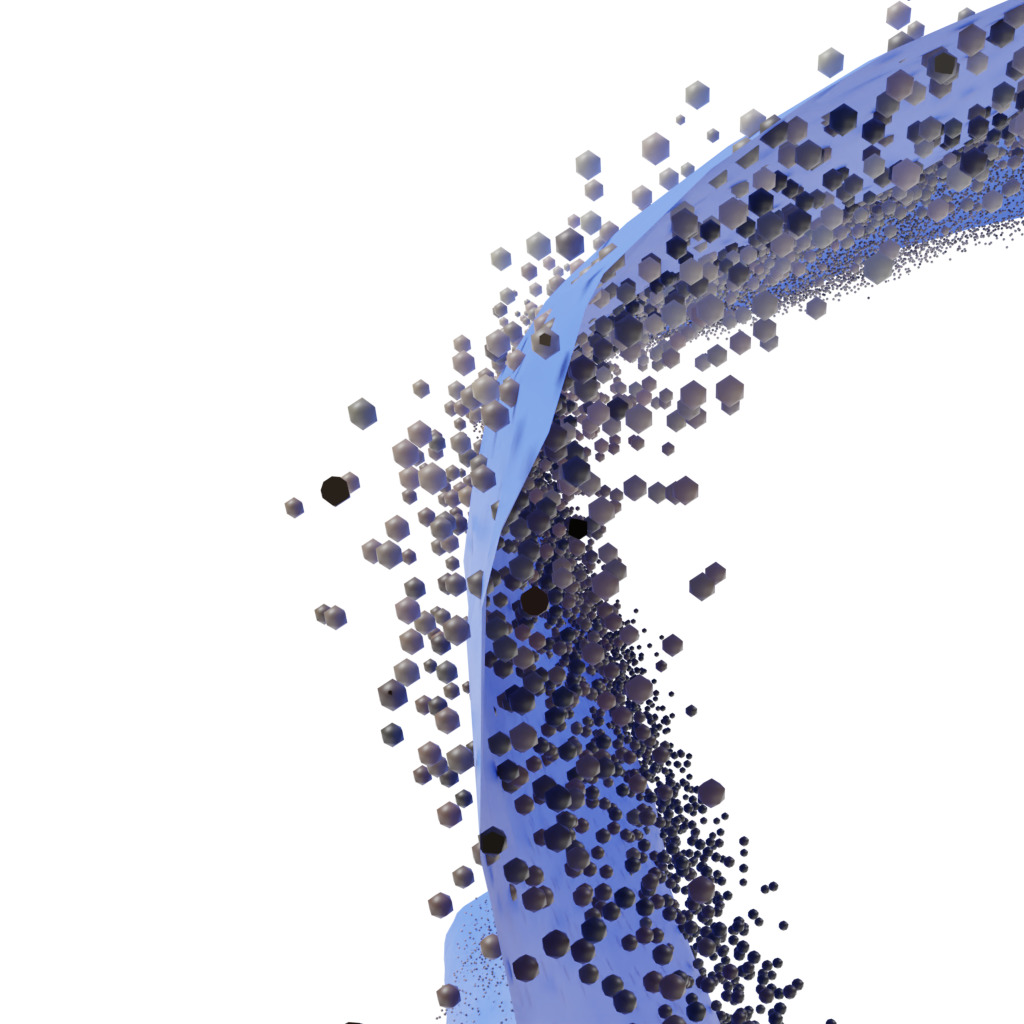} \\
\end{tabular}
\caption{\label{fig:dtu}Qualitative comparisons with NeAT, NeuralUDF and NeUDF on the DTU~\cite{Jensen2014} dataset and close-up comparisons against NeUDF. Our method can reconstruct surfaces closer to the ground truth point clouds in various places such as the marked region, generally improving the reconstruction accuracy of NeUDF by around 10\%, on a par with NeuralUDF and NeAT at the bottom two rows.}
\end{figure}

Other methods can also be used as the first stage of our \methodname{}. We use NeUDF for the first stage training on the DTU dataset~\cite{Jensen2014}. As detailed in Table~\ref{tab:dtu}, we compare the Chamfer distances of the reconstruction results with NeuralUDF, NeAT and NeUDF without our second-stage training. SDFs generally excel at learning watertight models, and it is worth pointing out that NeuralUDF takes the absolute value of the output of MLP as the UDF value of a given point. Therefore for closed models, they can easily learn an SDF and take its absolute value to produce a UDF. NeAT, on the other hand, explicitly learns an SDF. NeUDF and our method truly learn UDFs. While UDF learning is much more complicated than SDF learning because the UDF gradient nearby 0 is blurry and the gradient is not available at 0, our method still improves the reconstruction quality of NeUDF by around 10\% as shown in Figure~\ref{fig:dtu}. We further provide a close-up view of specific parts of the models for detailed comparisons in Figure~\ref{fig:dtu}. These local callouts exhibit the ground truth points located on both sides of our surfaces, whereas most of the points are only on one side of the surfaces of NeUDF. These illustrate our reconstructed surfaces are closer to the ground truth points and thus improving the resulting quality over NeUDF, on a par with NeuralUDF and NeAT.

\subsection{Ablation Studies}
\label{sec:ablation}

In this section, we present main ablation studies. We refer interested readers to the supplementary material for additional ablation studies.

\begin{table}[ht]
    \centering
    \begin{tabular}{c|cccccc}
    \hline
    Method & \#1 & \#7 & \#8 & LS-D0 \\
    \hline
    S1 \& S2 & \textbf{4.55} & \underline{2.88} & \textbf{3.21} & \textbf{2.46} \\
    S1 & 7.22 & \textbf{2.46} & \underline{3.38} & 6.04 &  \\
    S2 & \underline{5.75} & 4.00 & 5.96 & \underline{3.65}  \\
    \hline
    Method & NS-D1 & LS-C1 & DTU 114 & DTU 122 \\
    \hline
    S1 \& S2 & \underline{1.47} & \textbf{2.14} & \textbf{0.41} & \textbf{0.51} \\
    S1 & \textbf{1.46} & 6.23 & \underline{0.59} & 0.62 \\
    S2 & 1.64 & \underline{2.98} & 0.63 & \underline{0.60} \\
    \hline
    \end{tabular}
    \caption{Chamfer distances of models learned by both Stage 1 and 2 (S1 \& S2), only Stage 1 (S1) and only Stage 2 (S2) on selected datasets. Models learned by two stages yield similar Chamfer distances, but when trained with only Stage 1 or Stage 2, the Chamfer distances generally become significantly higher.}
    \label{tab:ablation}
\end{table}

\emph{Effect of the two-stage training.} We conduct an ablation study on the effect of the two-stage learning. We compare the Chamfer distances among both two stages, only Stage 1 and only Stage 2 training, shown in Table~\ref{tab:ablation}. Our results show that two-stage training improves the Chamfer distance (lower is better) compared to training with only Stage 1 or 2, under most circumstances.

It should be noted that training by the second stage from scratch is also capable of generating a generally reasonable result. However, the Chamfer distances, as shown in Table~\ref{tab:ablation}, indicate that its learning ability is limited. Therefore, the second refinement learning stage should cooperate with the first coarse learning stage to generate the best results.

\begin{figure}
    \centering
    \begin{tabular}{ccc}
        \includegraphics[width=.6in]{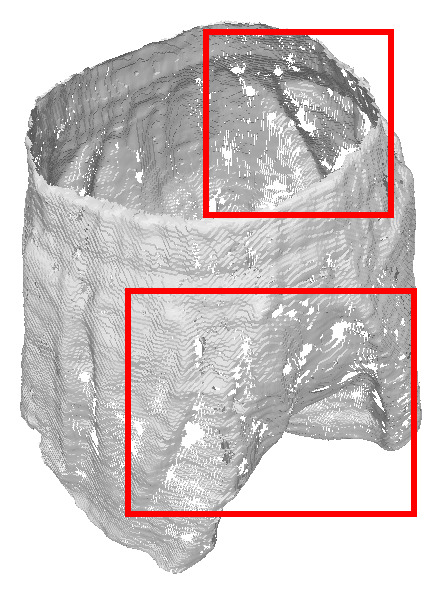} & \includegraphics[width=.6in]{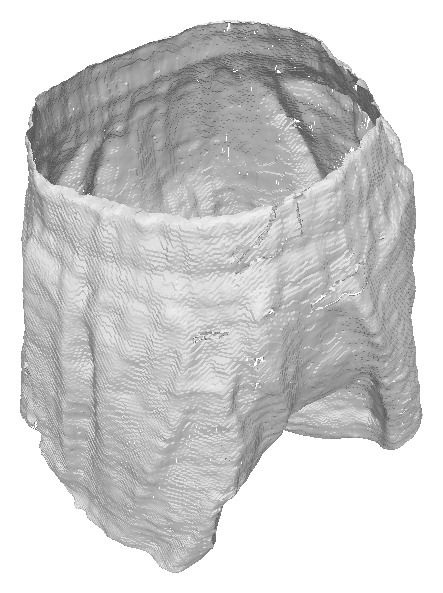} & \includegraphics[width=.6in]{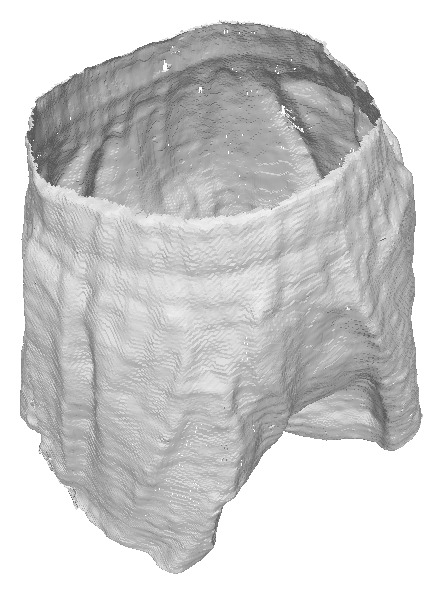} \\
        \includegraphics[width=.6in]{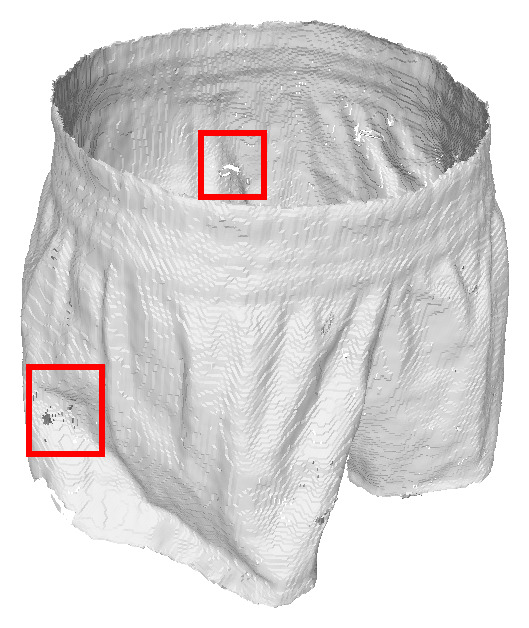} & \includegraphics[width=.6in]{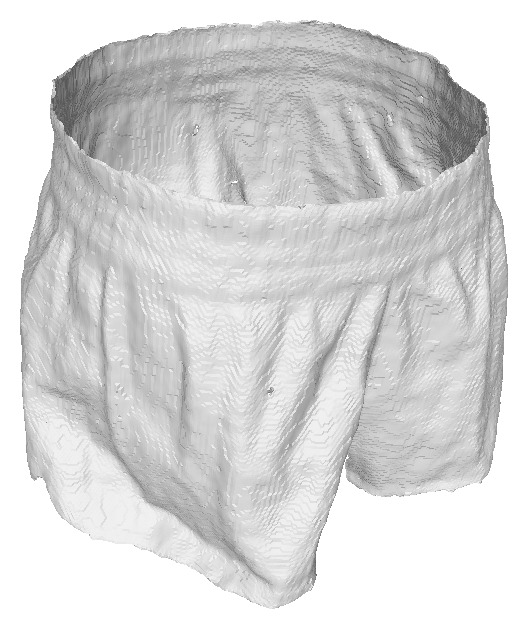} & \includegraphics[width=.6in]{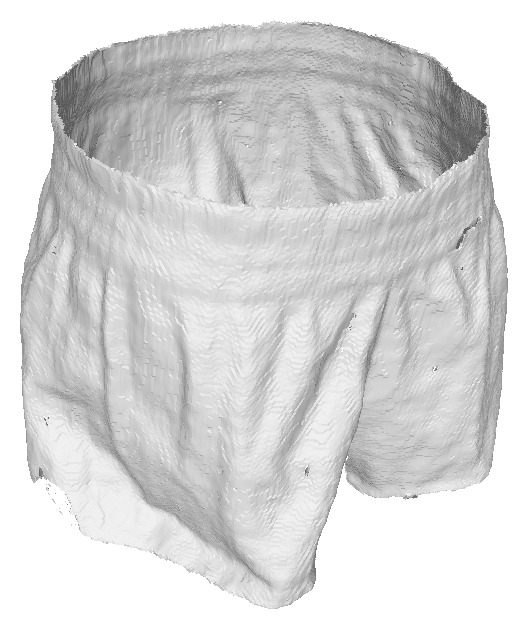} \\
        $\delta_{thres}=0.3$ & $\delta_{thres}=0.5$ & $\delta_{thres}=0.7$ \\
    \end{tabular}
    \caption{Qualitative comparisons on different choices of accumulated weight $\delta_{thres}$. Setting a higher threshold works well few little visual differences; Setting a lower threshold generally works fine, but may introduce more holes in reconstructed meshes.}
    \label{fig:vs_delta_thres}
\end{figure}

\emph{Choice of accumulated weight threshold $\delta_{thres}$.} In Stage 2, being a ray truncate point requires the accumulated weight up until that point to be greater than $\delta_{thres}$, where we intuitively select $\delta_{thres}=0.5$. Figure~\ref{fig:vs_delta_thres} shows the reconstruction results for other choices of $\delta_{thres}$, namely 0.3 and 0.7, respectively. We observe that all threshold choices successfully reconstruct the model. Setting the threshold $\delta_{thres}$ up to 0.7 produces visually similar results. Setting the threshold $\delta_{thres}$ down to 0.3 also works fine generally despite that it may introduce more holes to the reconstructed meshes. We deduce that setting a lower threshold increases the possibility that a ray may be truncated prematurely, leading to less desirable results. Nevertheless, we still have a considerable range of $\delta_{thres}$ from 0.3 to 0.7 without major result regression, indicating that our Stage 2 training exhibits robustness against $\delta_{thres}$.

\subsection{Limitations}
Since the light is cut off after going through a layer of surface, our method relinquishes the ability to model planes with transparency.
Occasionally, due to learning uncertainty, the Chamfer distance may increase slightly in the second stage, but the difference is quite small without visual impact. Overall, the two-stage learning improves the quality significantly.
For watertight models, SDF learning is more suitable than UDF learning, since UDF learning is more complicated than SDF learning.
We still advise using SDF learning, e.g., NeuS~\cite{Wang2021}, HF-NeuS~\cite{Wang2022HFNeuSIS} or PET-NeuS~\cite{Wang_2023_CVPR_PETNeuS}, for watertight model reconstruction.
Also, the mesh extraction of MeshUDF~\cite{Guillard2022} tends to generate holes and ``staircase'' artifacts affecting the mesh reconstruction quality. Adopting a more robust extraction method, e.g., DoubleCoverUDF~\cite{Hou2023}, could alleviate the problem, but we use MeshUDF here for all methods for a fair comparison.

\section{Conclusions}

Overall, \methodname{} offers a promising approach to the problem of reconstructing both open and watertight models from multi-view images. Its advantages over existing methods lie in the use of  a simple and more accurate density function, and a smooth differentiable UDF representation, so that the learned UDF approximates the ground truth as much as possible. A two-stage learning strategy further eliminates bias and improves UDF accuracy.
Results from our experiments on the DeepFashion3D, DTU and BlendedMVS datasets demonstrate the effectiveness of our method, particularly in learning smooth and stably open UDFs revealing the robustness of \methodname{}. Moreover, our method does not rely on object masks for open model reconstruction, making it more practical in real-world applications.
\section*{Acknowledgments}
This project was supported in part by the National Natural Science Foundation of China under Grants (61872347, 62072446), in part by the National Key R\&D Program of China under Grant 2023YFB3002901, in part by the Basic Research Project of ISCAS under Grant ISCAS-JCMS-202303 and in part by the Ministry of Education, Singapore, under its Academic Research Fund Grants (MOE-T2EP20220-0005, RG20/20 \& RT19/22).
{
    \small
    \bibliographystyle{ieeenat_fullname}
    \bibliography{main}
}

\clearpage
\setcounter{page}{1}
\setcounter{section}{0}
\setcounter{figure}{8}
\maketitlesupplementary

\renewcommand{\thesection}{\Alph{section}}

\section{Proof of Theorem 1}
\emph{Proof:}
We consider the scenario where the ray only intersects with the plane once. The weight $w_2$ is the derivative of the logistic sigmoid function scaled by a constant factor $|\cos(\theta)|$, which is a bell-shaped function centered at $f(t^*)=0$ implying that the point $t^*$ is on the surface. Put it in another way, the color weight $w_2$ attains maximum value when the point is on the surface, therefore $w_2$ is unbiased. With the ray truncation mechanism, if there are multiple ray-plane intersections along a single ray, only the first intersection is in effect. Therefore, it is also occlusion-aware.

\section{Bias and Translucency Analysis of Stage 1}
The density function $\sigma_1$ in our Stage 1 coarse training is neither unbiased nor fully opaque, but we select $c=5$ for a good balance. In fact, we can estimate the bias and translucency. For points in front of the surface, the incident angle $\theta$ between the ray and the surface normal is obtuse, so we restrict $\theta$ to the range of $[91^\circ, 180^\circ]$. Assuming $s=1000$, by setting $c=5$, in theory, the offset width between 0.00161 and 0.00566 is obtained relative to the true zero level set, indicating that the maximum relative bias is below 0.5\%. This error level is acceptable for most application scenarios. Moreover, assuming $s=1000$, the surface transparency in the extreme case mentioned above is less than $0.001$. When a ray has a larger incident angle, its transparency becomes even smaller, resulting in an almost opaque density $\sigma_1$. As a result, the weight function $w_1$ is approximately occlusion-aware. Thus, setting the constant $c=5$ offers a good balance between occlusion-awareness and unbiasedness in the first stage training.

\section{Discussion of MeshUDF}
\begin{figure}[ht]
    \centering
    \begin{tabular}{ccc}
        \includegraphics[width=.95in]{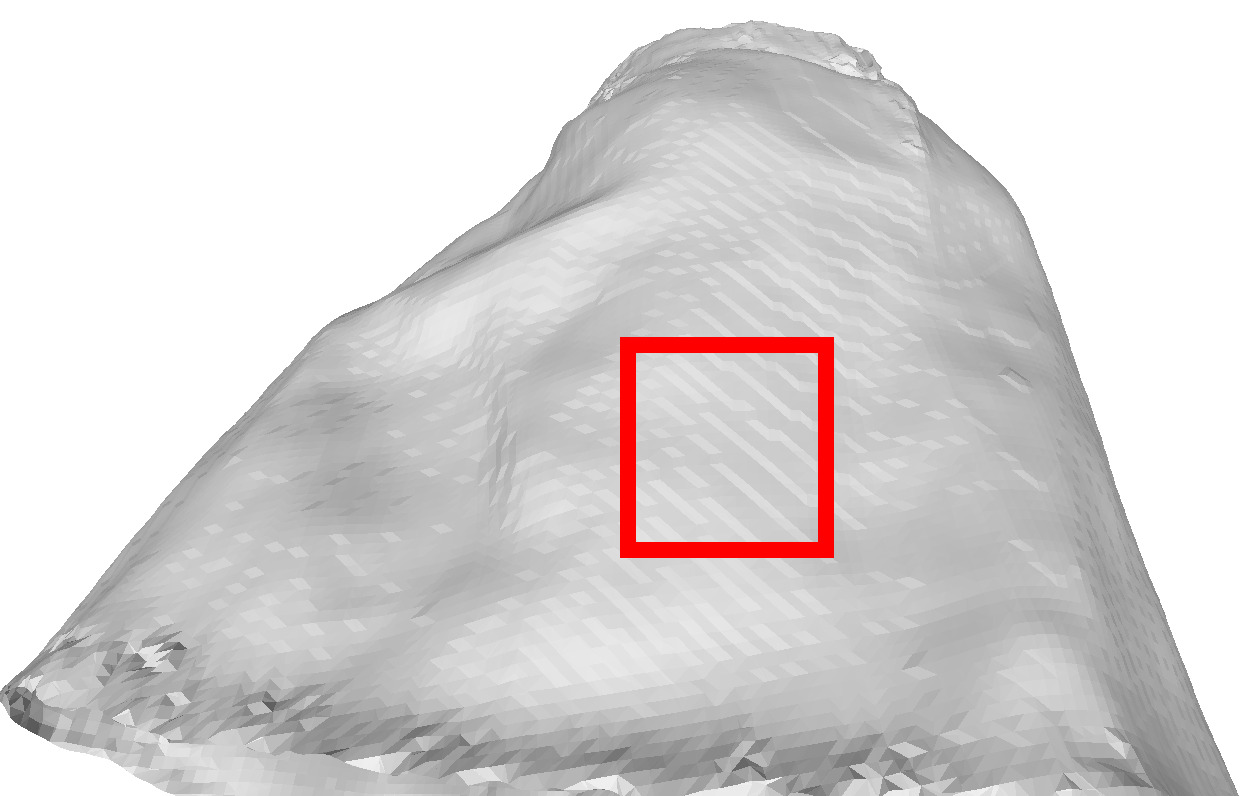} & \includegraphics[width=.95in]{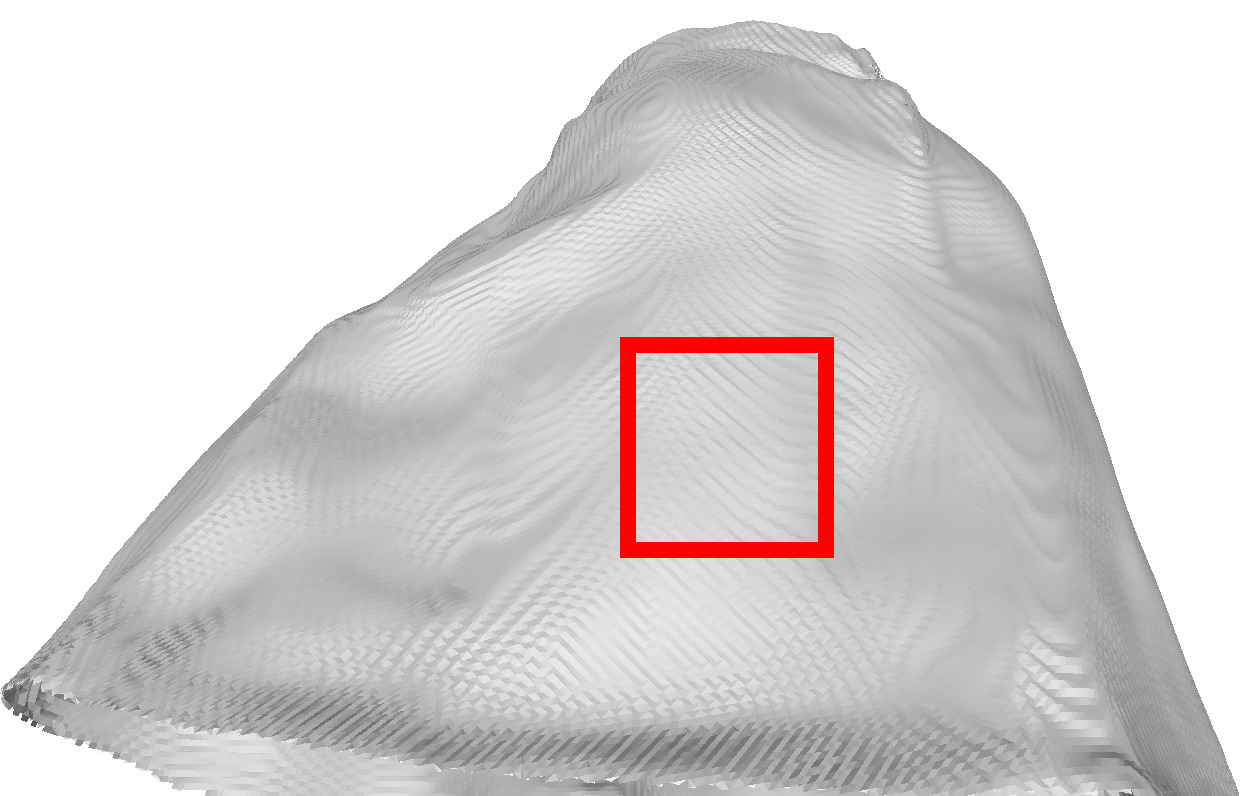} & \includegraphics[width=.95in]{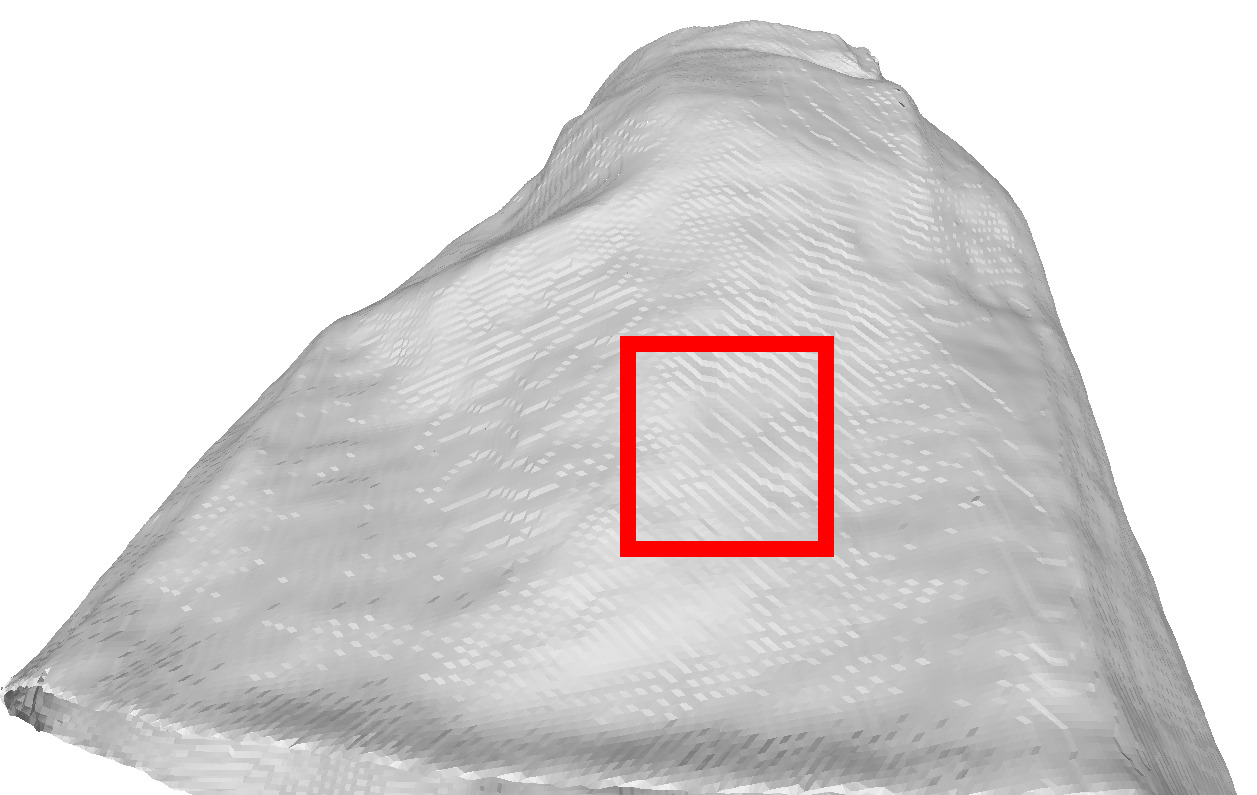} \\
        \includegraphics[width=.95in]{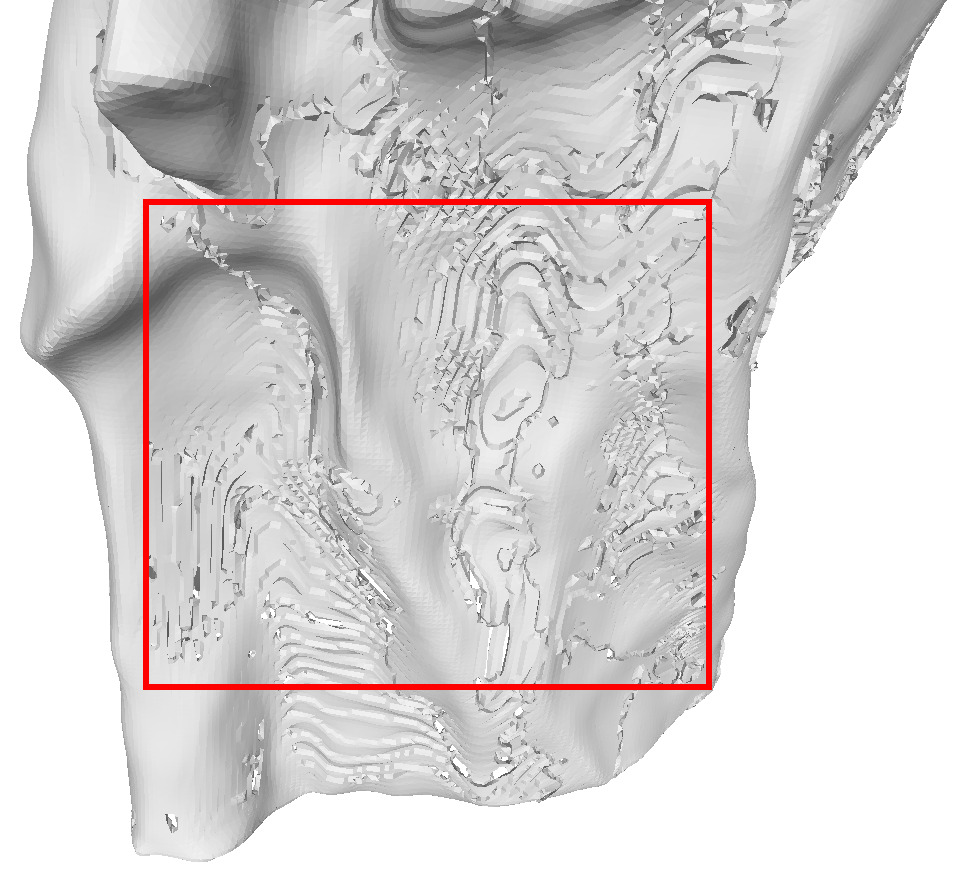} & \includegraphics[width=.65in]{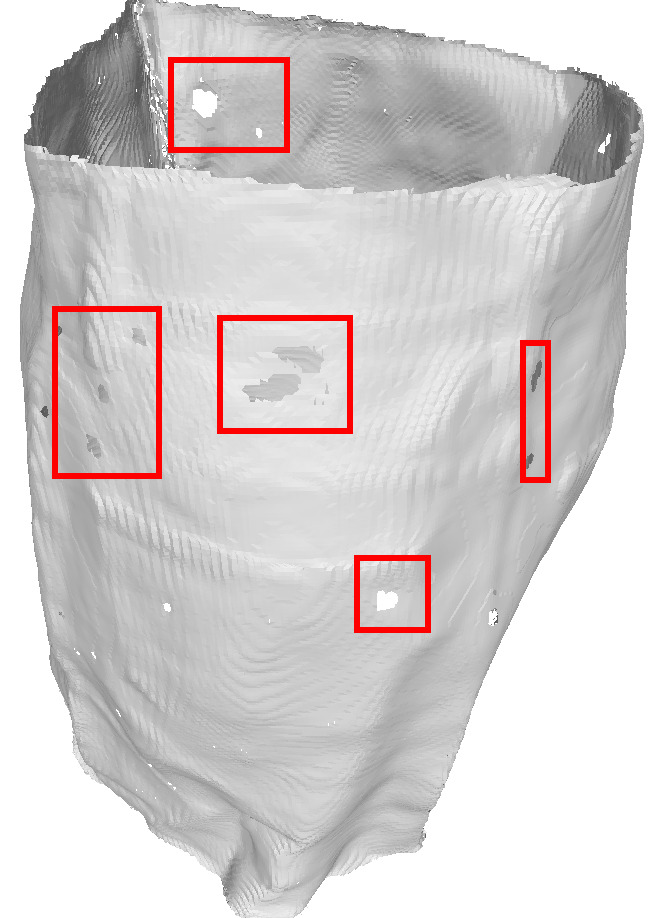} & \includegraphics[width=.95in]{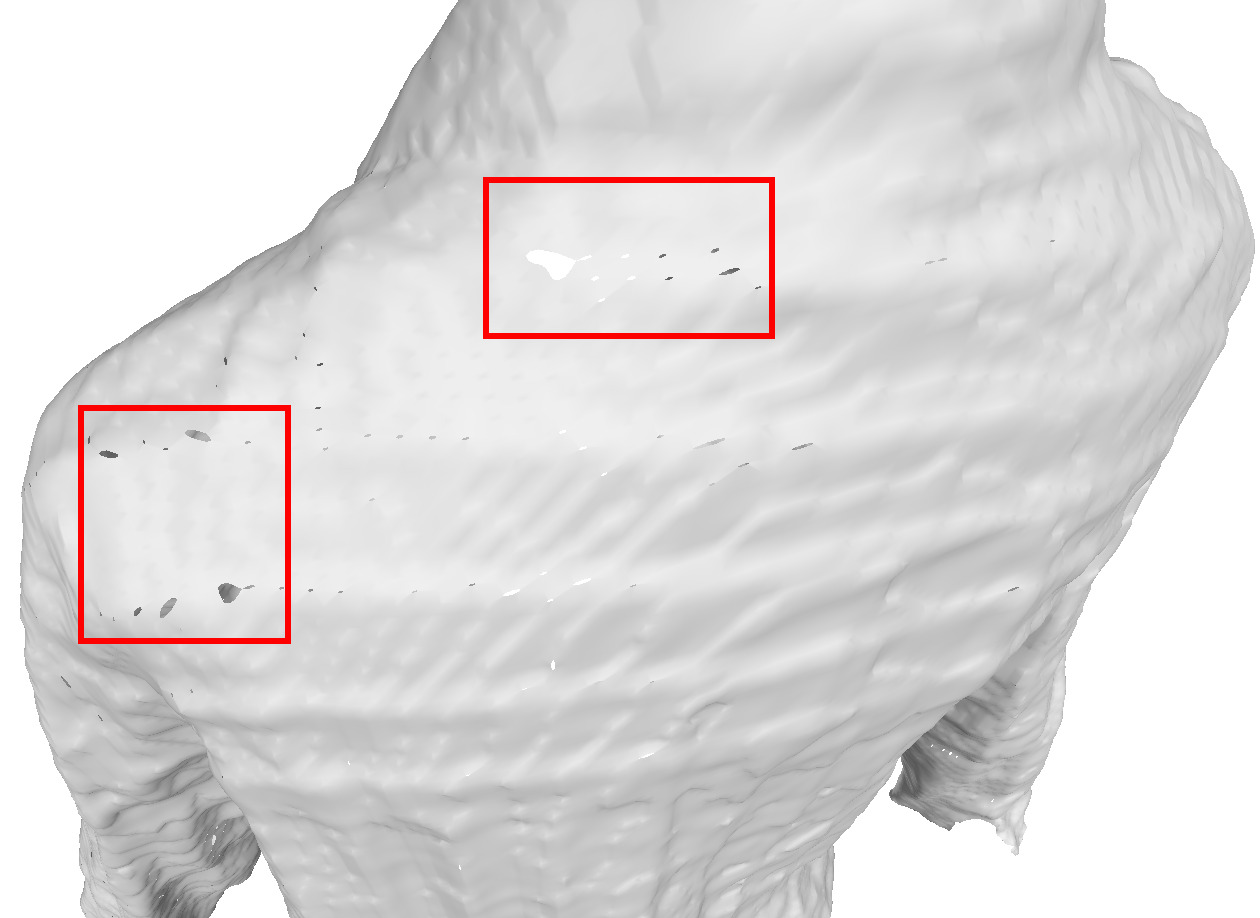} \\
        NeuralUDF~\cite{Long2023} & NeUDF~\cite{Liu2023NeUDF} & Ours \\
    \end{tabular}
    \caption{The ``staircase effect'' and hole artifacts found in extracted meshes using MeshUDF~\cite{Guillard2022}. The first row shows raw meshes that have visible ``staircases'' widely found in NeuralUDF~\cite{Long2023}, NeUDF~\cite{Liu2023NeUDF} and our method, all using MeshUDF. The second row shows the hole artifacts found in extracted meshes. These artifacts may negatively impact on visual effects.}
    \label{fig:meshudf}
\end{figure}

Unlike signed distance fields (SDFs), from which extracting a mesh is extensively studied, extracting a mesh from unsigned distance fields is still an actively developing research field with several challenges, which can lead to sub-optimal reconstruction results. MeshUDF~\cite{Guillard2022} is a UDF-mesh extraction method that has enjoyed considerable popularity, yet it still contains some limitations. Figure~\ref{fig:meshudf} showcases two common limitations of MeshUDF: the extracted mesh exhibits a visible ``staircase effect'' and hole artifacts resulting in a negative visual impact. ``Staircase effect'' and holes are pervasive across the results of NeuralUDF~\cite{Long2023}, NeUDF~\cite{Liu2023NeUDF} and our method. To eliminate these artifacts, we can use DoubleCoverUDF~\cite{Hou2023} for mesh extraction from UDF in the future, but we use MeshUDF in this work for fair comparisons.

\section{Implementation Details}
The UDF network is an MLP, consisting of 8 hidden layers, each with 256 elements. We use skip connections after every 4 hidden layers. The output of the UDF network is a single value representing the predicted UDF and a 256-dimensional feature vector used in the color network.

For the color network, we use another MLP with 4 hidden layers, each having 256 elements.
We use the coarse-to-fine strategy proposed by Park \emph{et al}.~\cite{Park2021} for position encoding, setting the maximum number of frequency bands to 16 for the UDF network and 6 for the color network. For background rendering, we use NeRF++~\cite{Zhang2020} for background prediction.
During training, we use the Adam optimizer~\cite{Kingma2015} with a global learning rate of 5e-4.
We sample 512 rays per batch and train our model for 250,000 iterations for the first stage and another 50,000 iterations for the second stage, making up a total of 300,000 iterations. We leverage MeshUDF~\cite{Guillard2022} to extract meshes from trained UDFs.

For the weights of each loss function term, we empirically set $\lambda_1=0.1$, $\lambda_2=0.01$, and $\lambda_3=0.001$, although $\lambda_2$ is occasionally set to $0.02$, and $\lambda_3$ is optional. The weight $\lambda_m$ for mask loss $\mathcal{L}_{loss}$ is set to 0.1 aligning with other works~\cite{Wang2021,Long2023}, if mask supervision is adopted.

\section{More Ablation Studies}

We conduct additional ablation studies in this section.

\begin{figure}[!ht]
    \centering
    \includegraphics[width=2.5in]{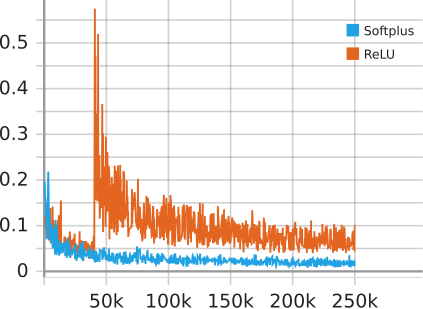}
    \caption{Ablation study on the usage of ReLU (orange)~\cite{Fukushima1975} versus softplus (blue)~\cite{Liu2023NeUDF,Dugas2000} in the MLP output layer. The former is non-differentiable at $0$ and its gradient vanishes for negative input, whereas the latter is differentiable everywhere. 
    Using ReLU after the output layer of the MLP, the network makes progress at the early stage of training, but collapses after 40K iterations, leading to a training loss reduction through the rendering of only backgrounds. In contrast, softplus leads to correct learning of both geometry and color, and consistently decreases the training loss over iterations.}
    \label{fig:vs_relu}
\end{figure}

\emph{Non-negativity.} Ensuring that the computed distances in the proposed method are non-negative is important, and can be achieved by applying either ReLU~\cite{Fukushima1975} or softplus~\cite{Dugas2000,Liu2023NeUDF} to the MLP output. However, ReLU is not differentiable at 0 and has vanishing gradients for negative inputs, which can make the network difficult to train. An ablation study confirms that training with ReLU only results in early progress, but fails to learn a valid UDF later on. See Figure~\ref{fig:vs_relu} for details.

\begin{figure}[!ht]
    \centering
    \begin{tabular}{ccc}
    \includegraphics[width=.9in]{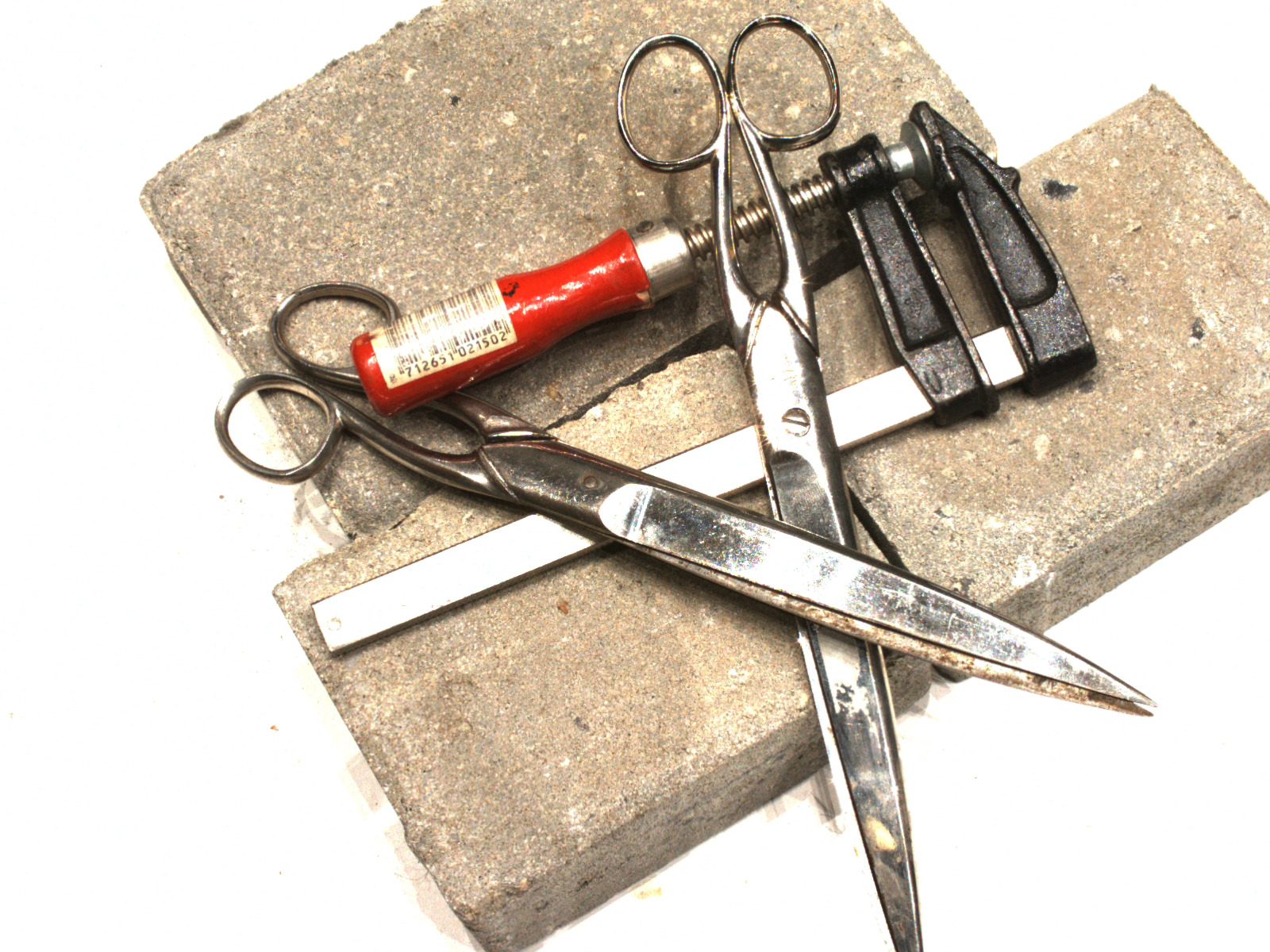} & \includegraphics[width=.9in]{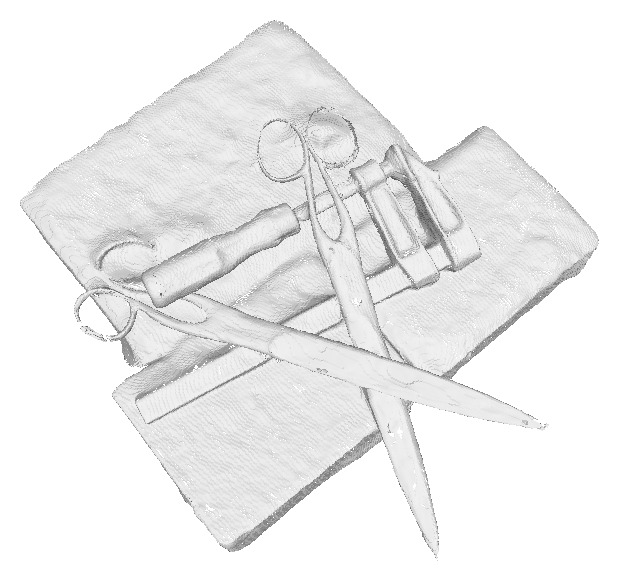} & \includegraphics[width=.9in]{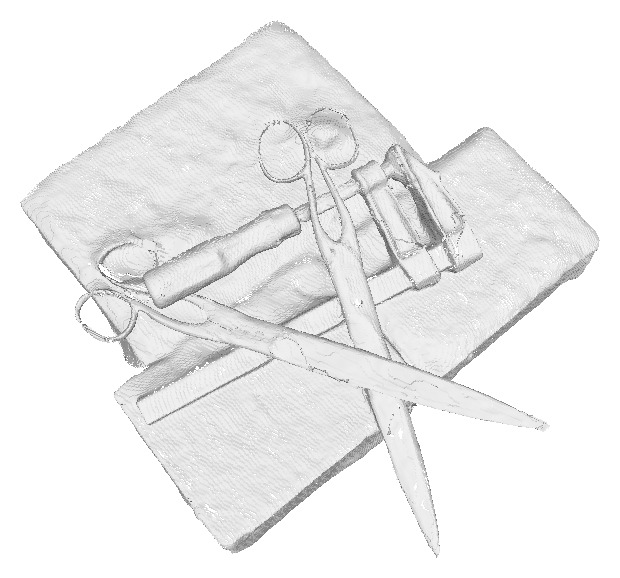} \\
    & CD=0.891 & CD=0.904 \\
    Reference Image & w/ $\mathcal{L}_s$ & w/o $\mathcal{L}_s$\\
    \end{tabular}
    \caption{Qualitative and quantitative ablation study on the s-value loss $\mathcal{L}_s$. The visual impact and the quantitative impact are both very small.}
    \label{fig:vs_sval}
\end{figure}

\emph{S-value loss.} Although $\mathcal{L}_s$ is optional, it is still important that the learned $s$ is large enough so that the model has better convergence, and the result is sharper. As shown in Figure~\ref{fig:vs_sval}, there are cases where omitting $\mathcal{L}_s$ results in a worse reconstruction result, as the Chamfer distances are higher. However, the impact is negligible both in quantitative metrics and qualitative comparisons, hinting at the optional nature of $\mathcal{L}_s$.

\begin{figure}[htp]
\centering
\setlength\tabcolsep{1pt}
\begin{tabular}{ccccc}
    & GT & Ours & NeuralUDF & NeUDF \\
    \raisebox{.24in}{\#1} & \includegraphics[width=.685in]{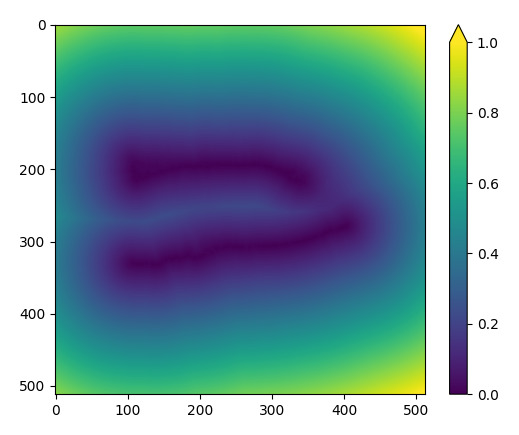} & \includegraphics[width=.685in]{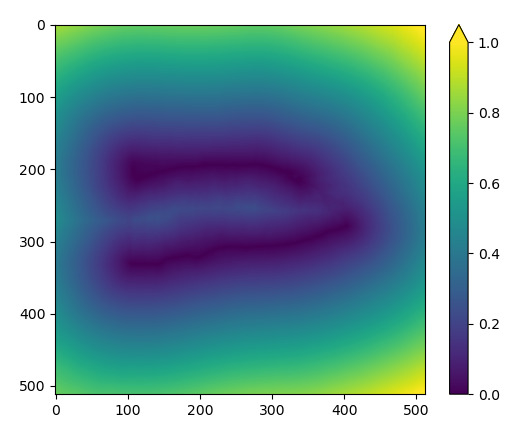} & \includegraphics[width=.685in]{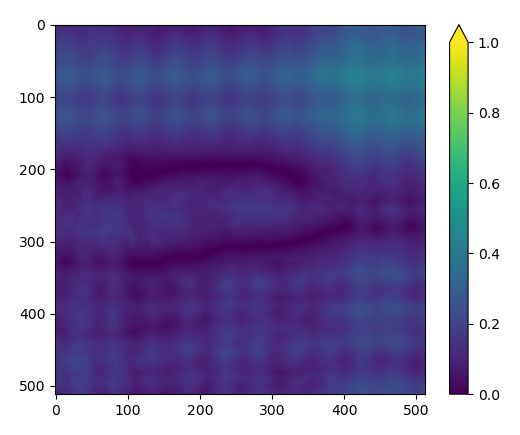} & \includegraphics[width=.685in]{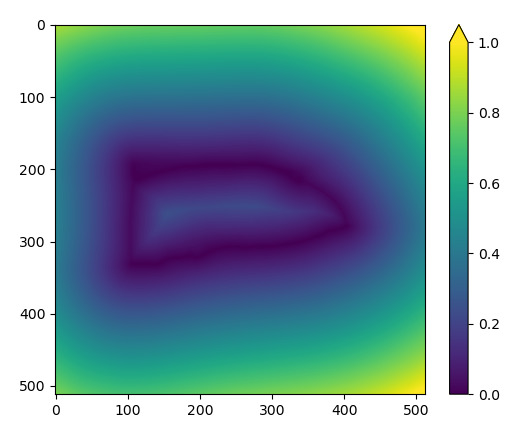} \\
    \raisebox{.24in}{\#5} & \includegraphics[width=.685in]{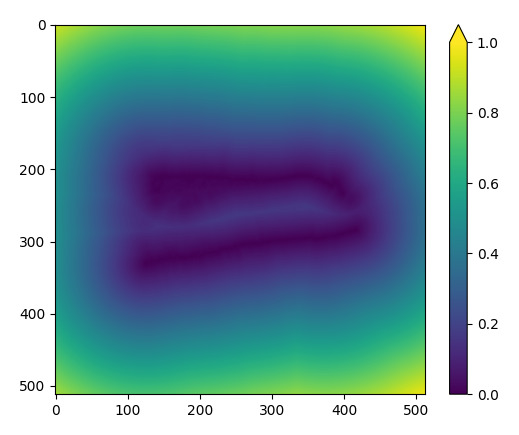} & \includegraphics[width=.685in]{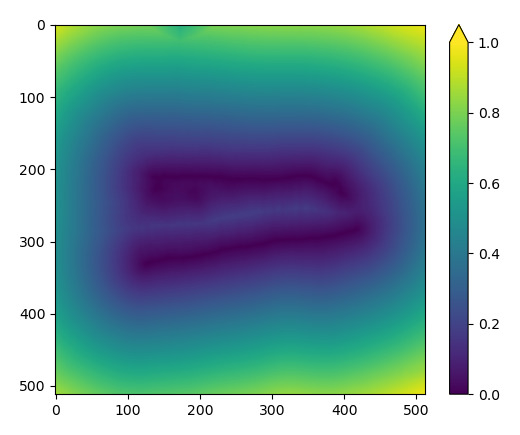} & \includegraphics[width=.685in]{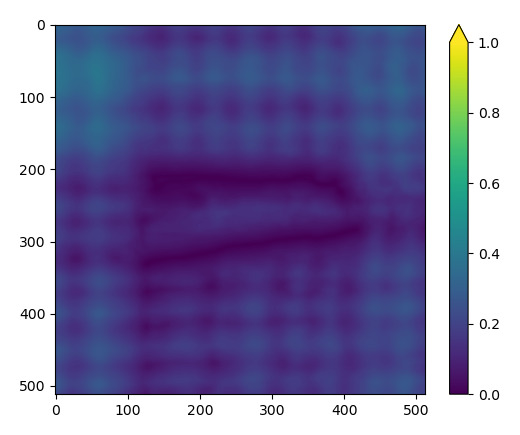} & \includegraphics[width=.685in]{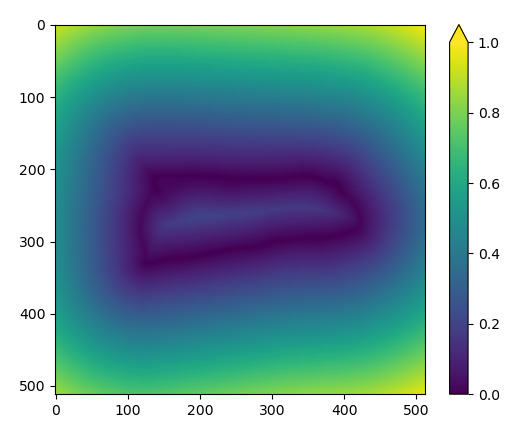} \\
    \raisebox{.24in}{\#6} & \includegraphics[width=.685in]{gt/432-1-cut} & \includegraphics[width=.685in]{ours/432-1-cut} & \includegraphics[width=.685in]{neuraludf/432-1-cut} & \includegraphics[width=.685in]{neudf/432-1-cut} \\
    \raisebox{.24in}{\#7} & \includegraphics[width=.685in]{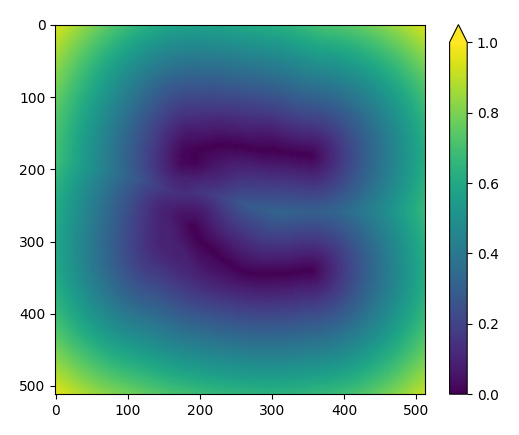} & \includegraphics[width=.685in]{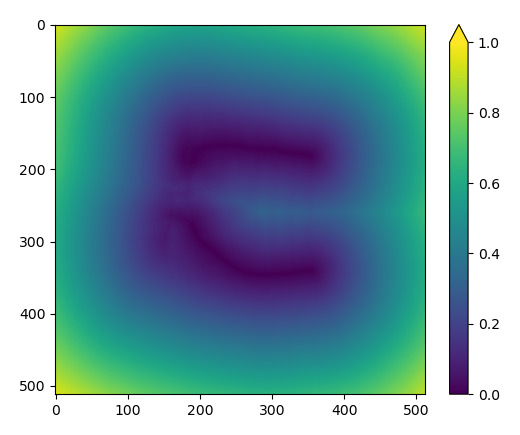} & \includegraphics[width=.685in]{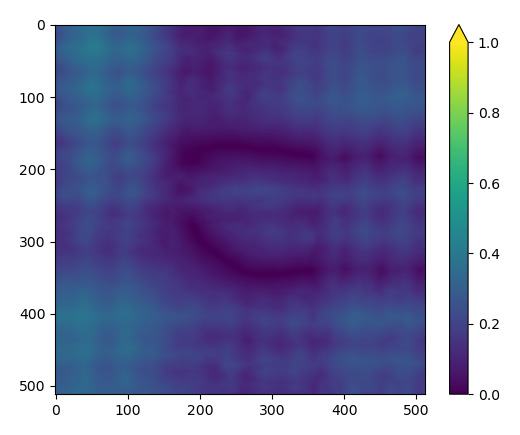} & \includegraphics[width=.685in]{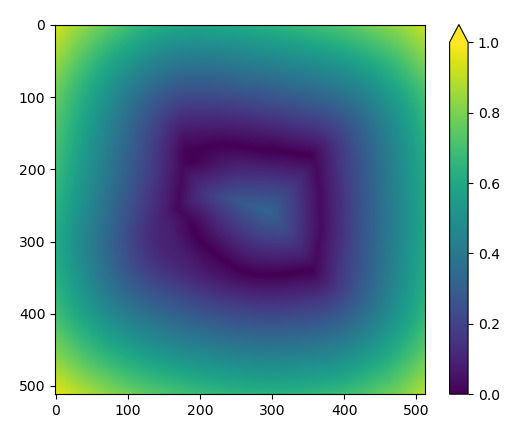} \\
    \raisebox{.24in}{\#8} & \includegraphics[width=.685in]{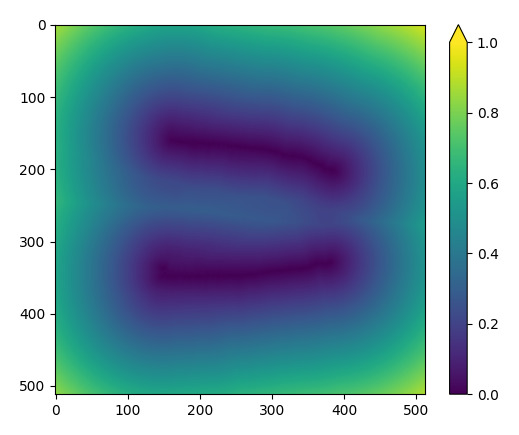} & \includegraphics[width=.685in]{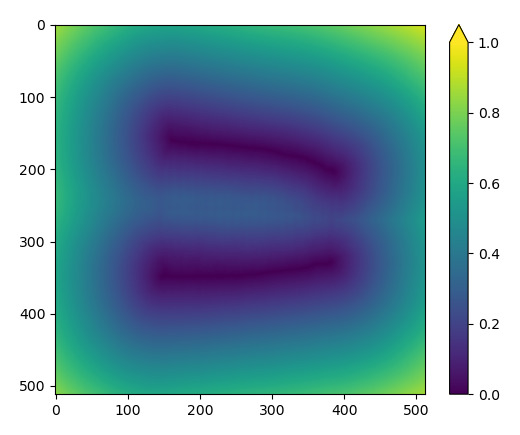} & \includegraphics[width=.685in]{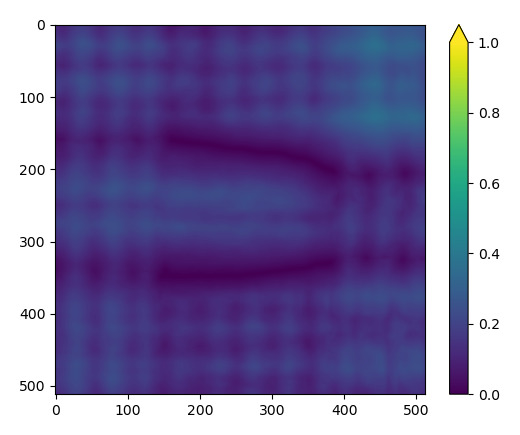} & \includegraphics[width=.685in]{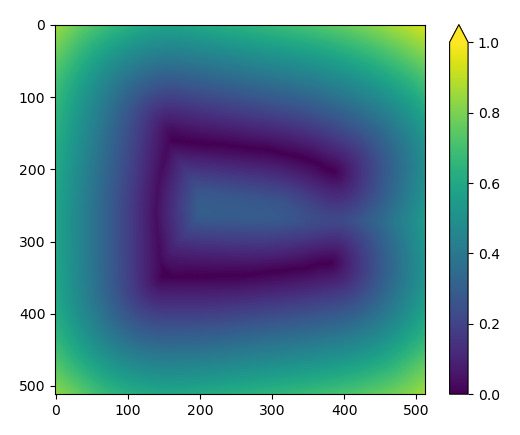} \\
    \raisebox{.24in}{\#9} & \includegraphics[width=.685in]{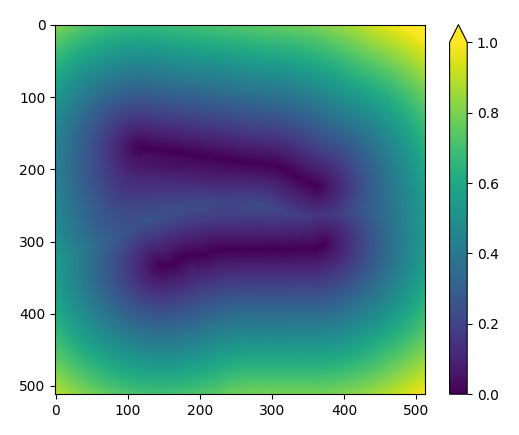} & \includegraphics[width=.685in]{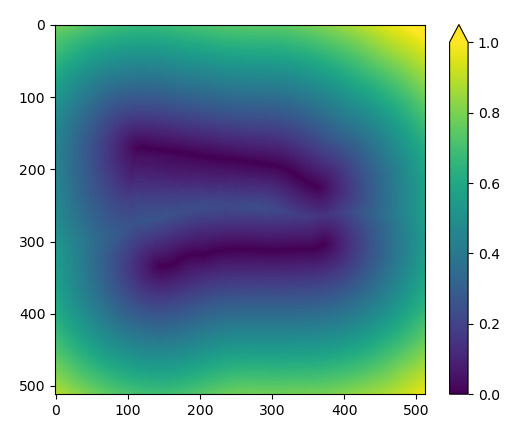} & \includegraphics[width=.685in]{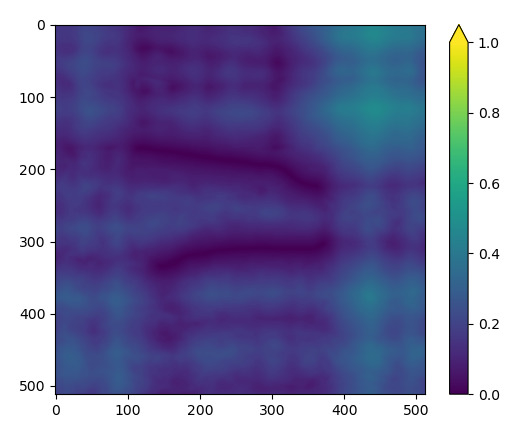} & \includegraphics[width=.685in]{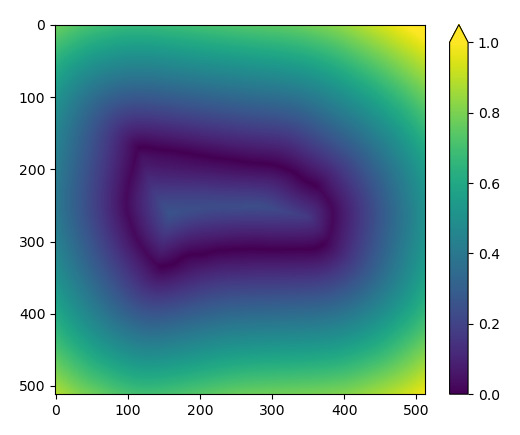} \\
    \raisebox{.24in}{LS-C0} & \includegraphics[width=.685in]{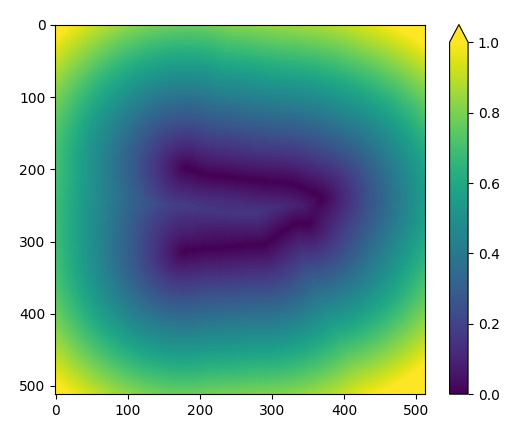} & \includegraphics[width=.685in]{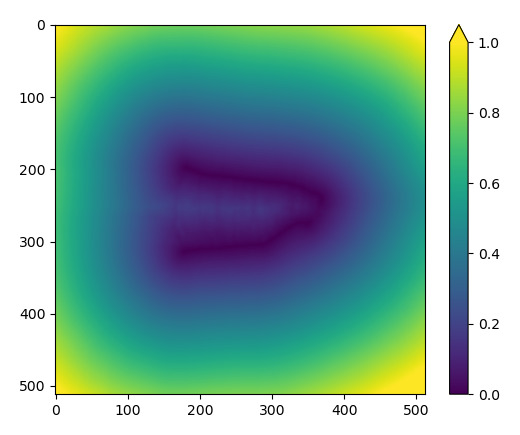} & \includegraphics[width=.685in]{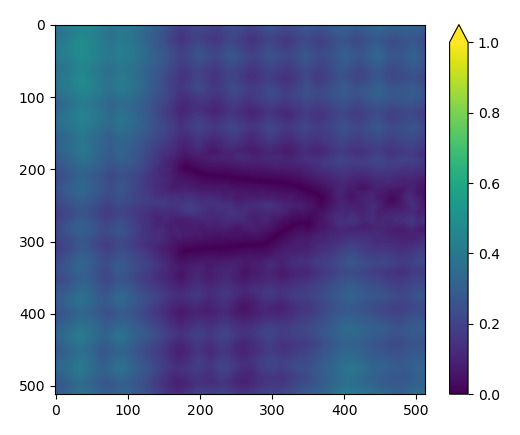} & \includegraphics[width=.685in]{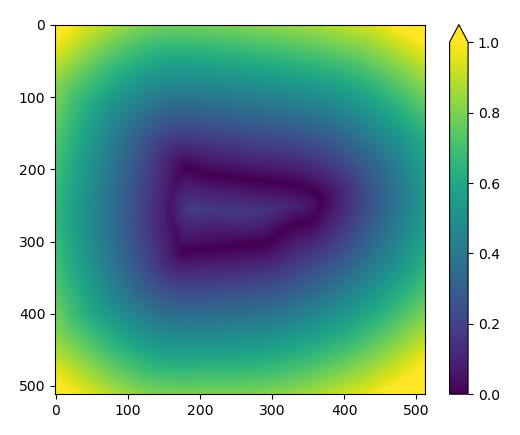} \\
    \raisebox{.24in}{SS-D0} & \includegraphics[width=.685in]{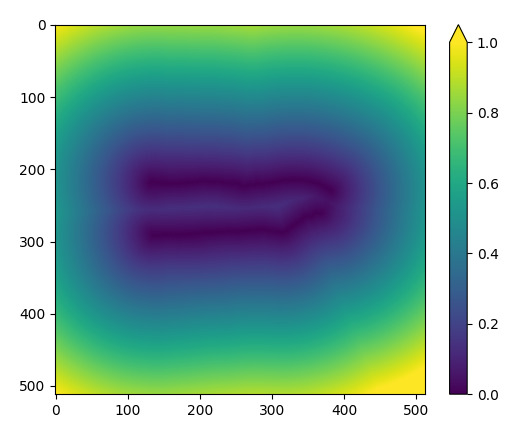} & \includegraphics[width=.685in]{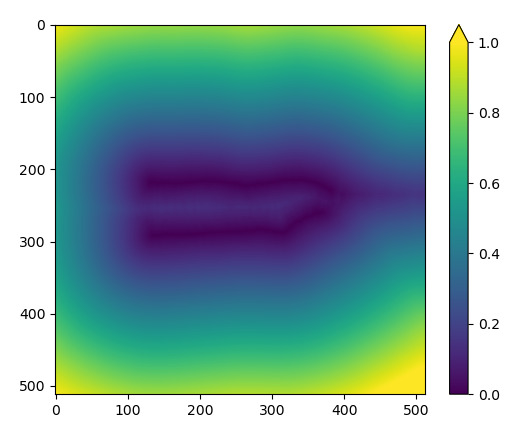} & \includegraphics[width=.685in]{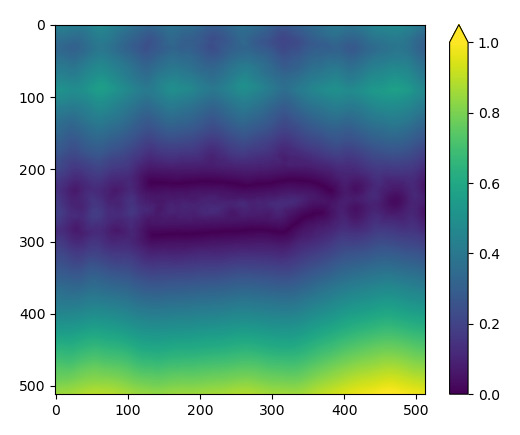} & \includegraphics[width=.685in]{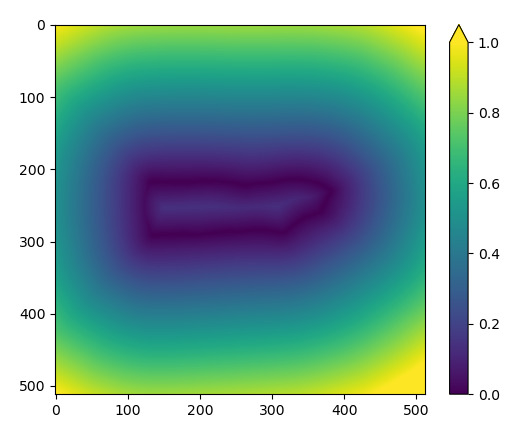} \\
    \raisebox{.24in}{NS-D1} & \includegraphics[width=.685in]{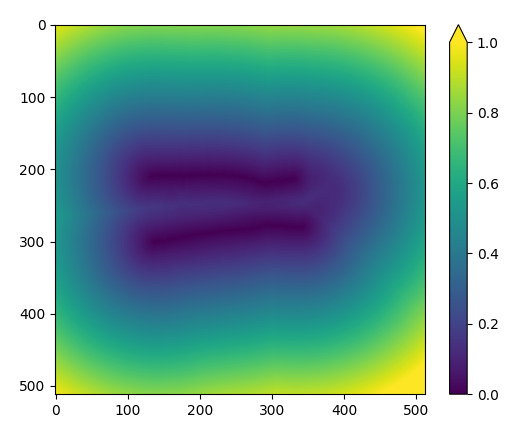} & \includegraphics[width=.685in]{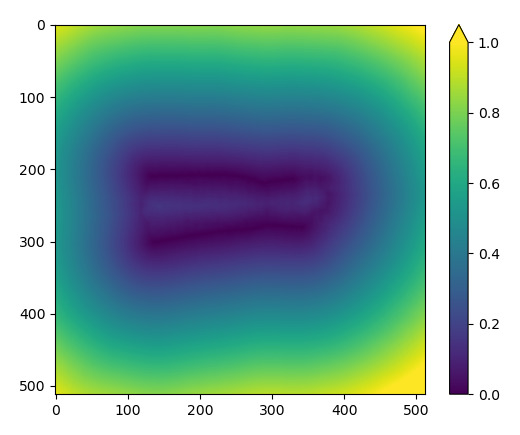} & \includegraphics[width=.685in]{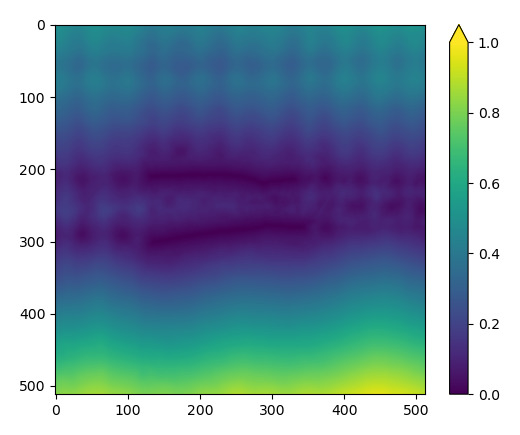} & \includegraphics[width=.685in]{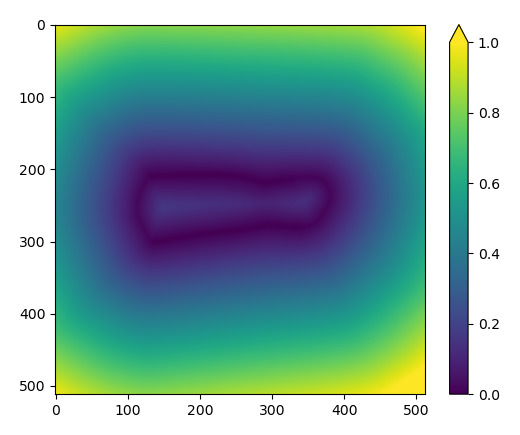} \\
    \raisebox{.24in}{LS-C1} & \includegraphics[width=.685in]{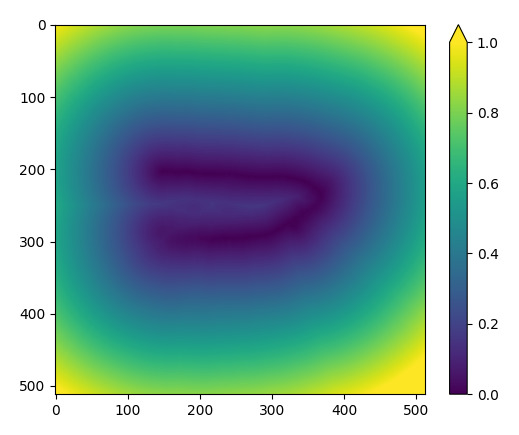} & \includegraphics[width=.685in]{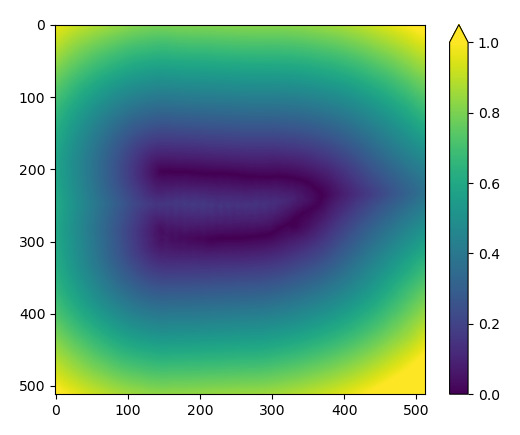} & \includegraphics[width=.685in]{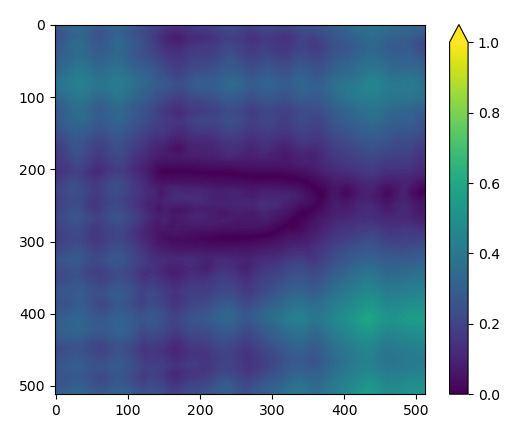} & \includegraphics[width=.685in]{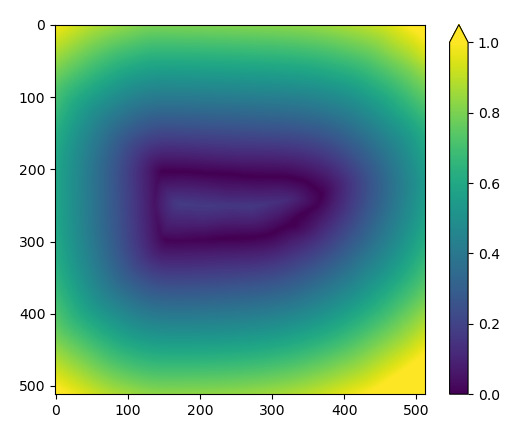} \\
    \raisebox{.24in}{Skirt1} & \includegraphics[width=.685in]{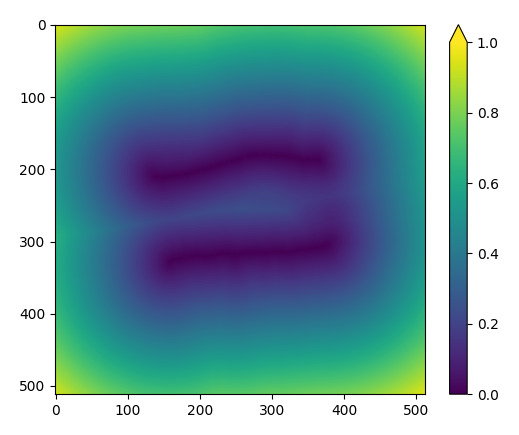} & \includegraphics[width=.685in]{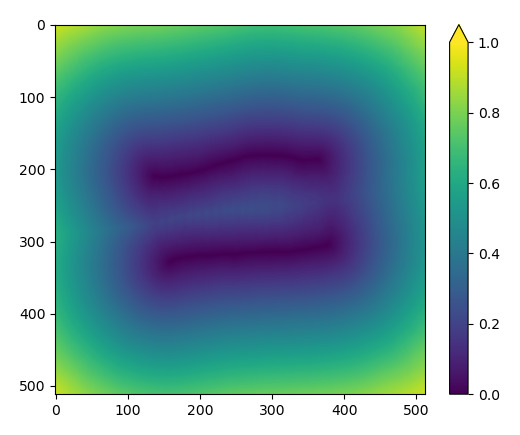} & \includegraphics[width=.685in]{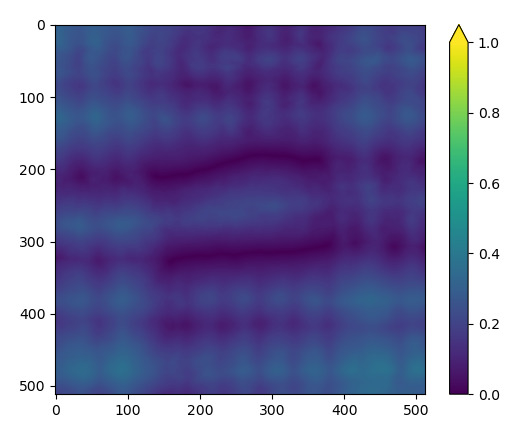} & \includegraphics[width=.685in]{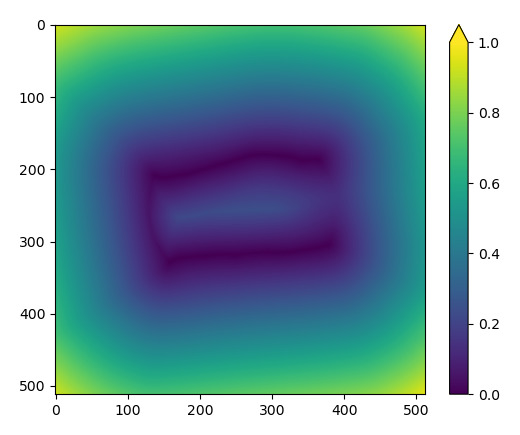} \\
    \raisebox{.24in}{SS-C0} & \includegraphics[width=.685in]{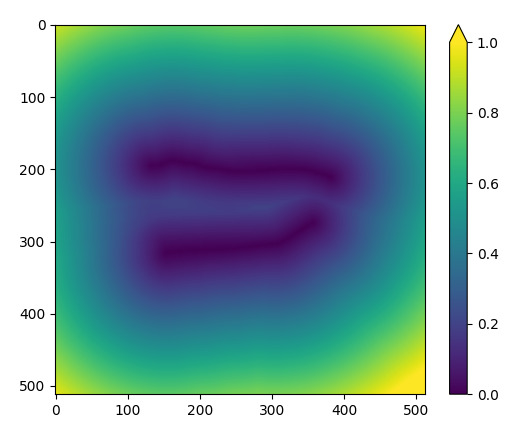} & \includegraphics[width=.685in]{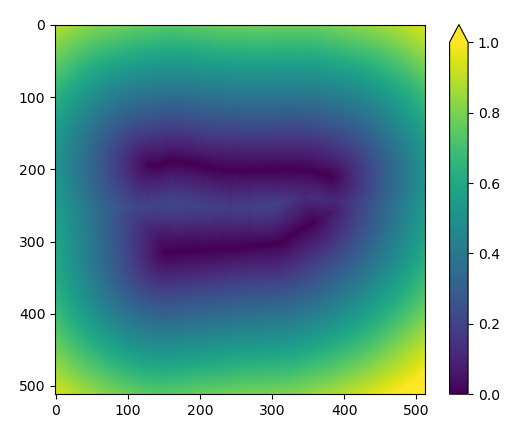} & \includegraphics[width=.685in]{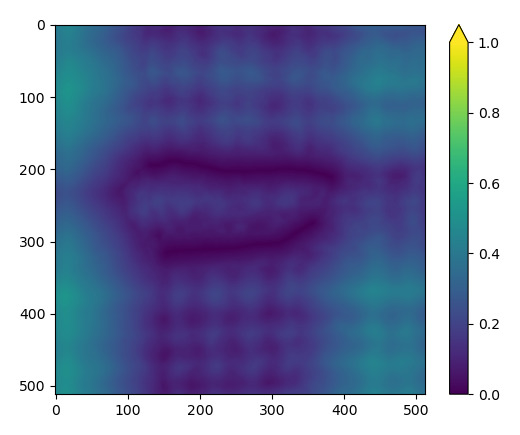} & \includegraphics[width=.685in]{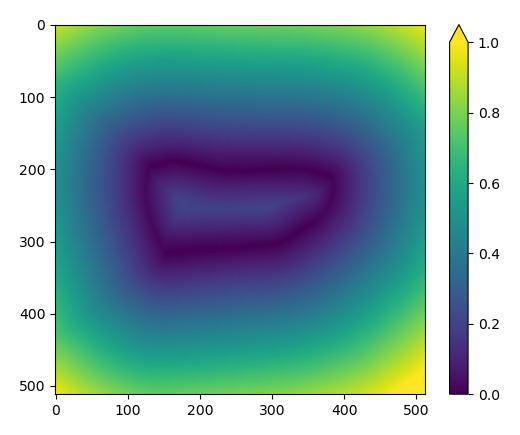} \\
\end{tabular}
\caption{Visualization of the learned UDFs on cross sections for the remaining garments from DeepFashion3D.}
\label{fig:vs_udf_more}
\end{figure}

\section{Complete Results}

\begin{figure*}[p]
\centering
\setlength\tabcolsep{0pt}
\begin{tabular}{ccccccccc}
    & Ref. Image & GT & Ours & VolSDF & NeuS & NeAT & NeuralUDF & NeUDF\\
    \raisebox{.24in}{\#1} & \includegraphics[width=.53in]{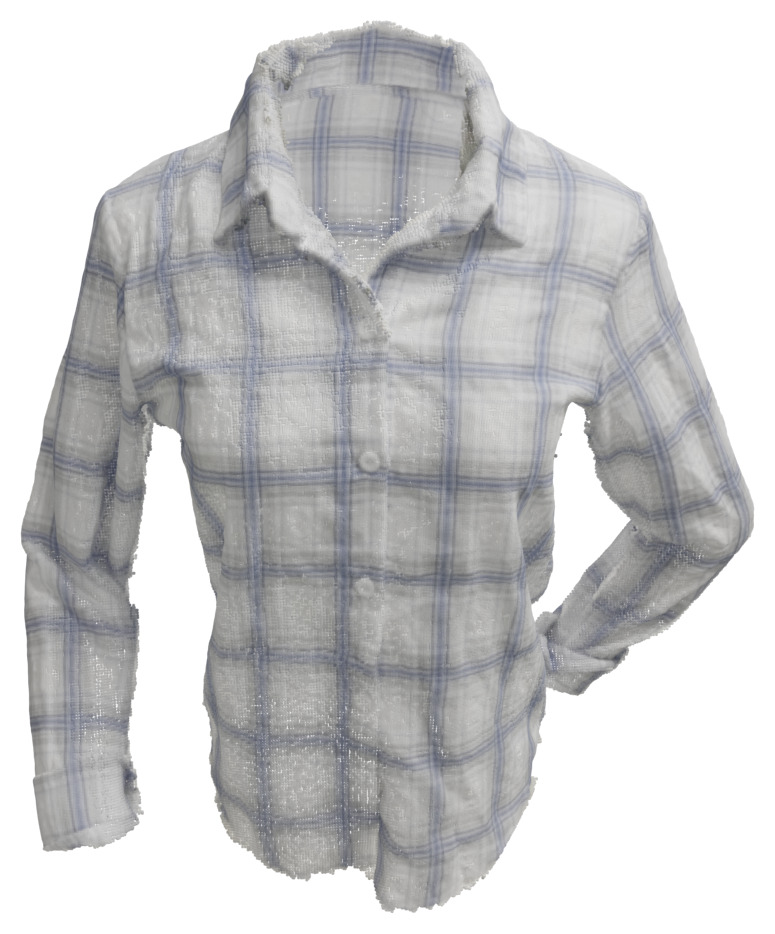}&
    \includegraphics[width=.47in]{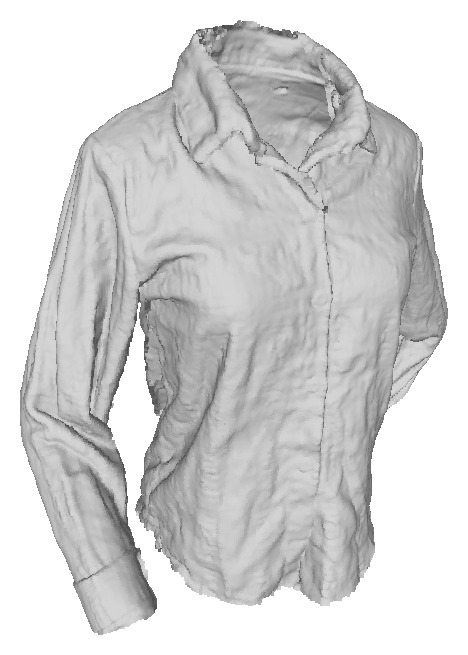}\includegraphics[width=.3in]{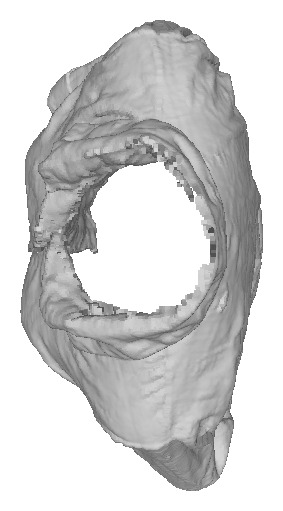}&
    \includegraphics[width=.47in]{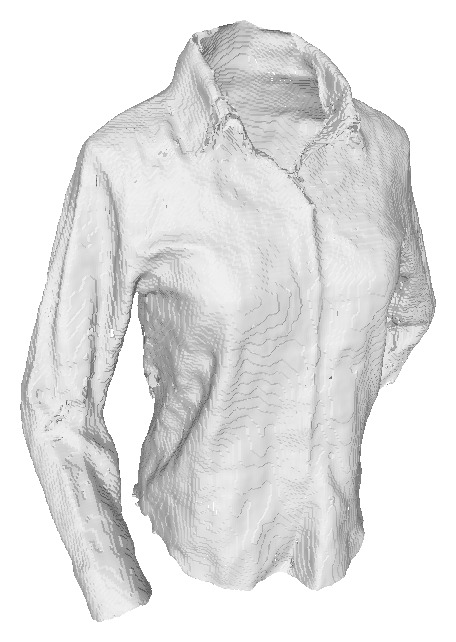}\includegraphics[width=.3in]{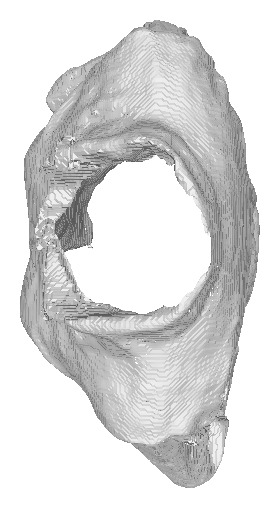}&
    \includegraphics[width=.47in]{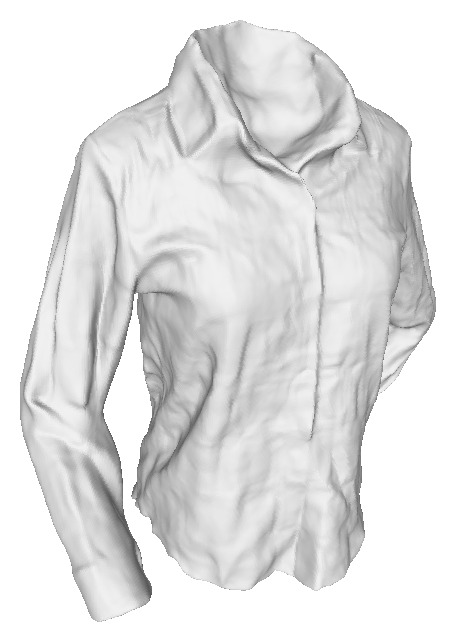}\includegraphics[width=.3in]{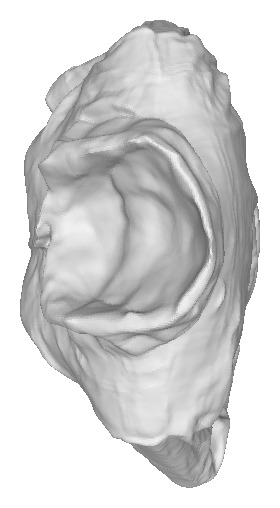}&
    \includegraphics[width=.47in]{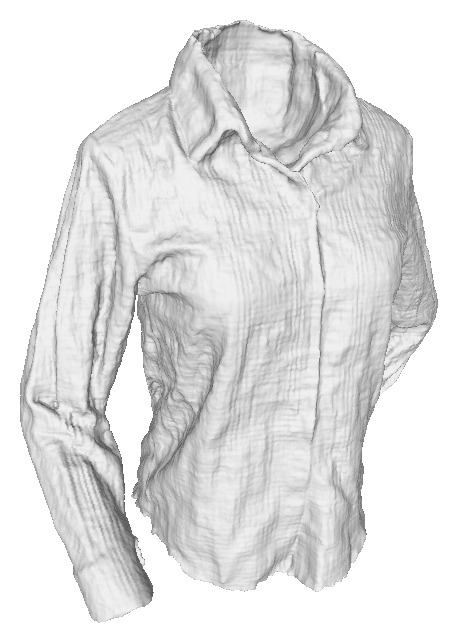}\includegraphics[width=.3in]{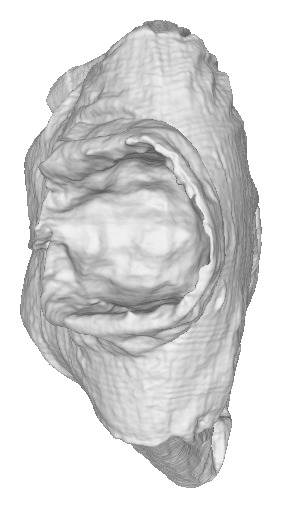}&
    \includegraphics[width=.47in]{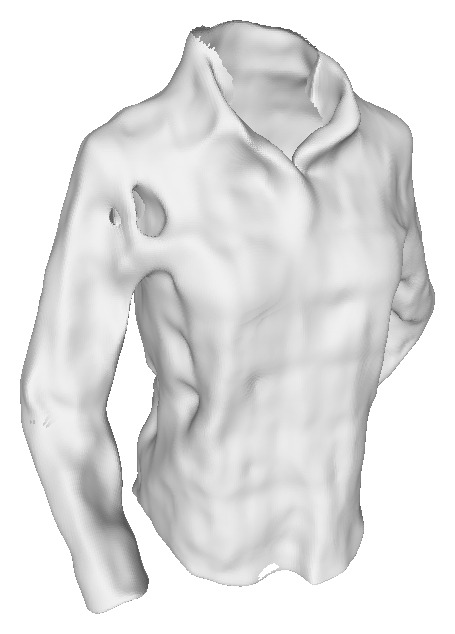}\includegraphics[width=.3in]{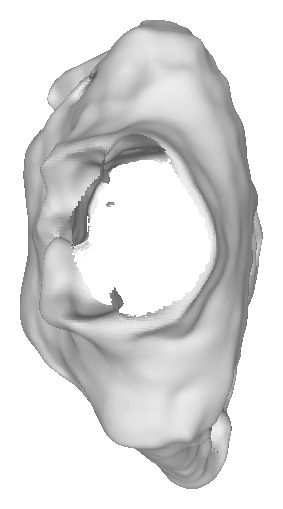}&
    \includegraphics[width=.47in]{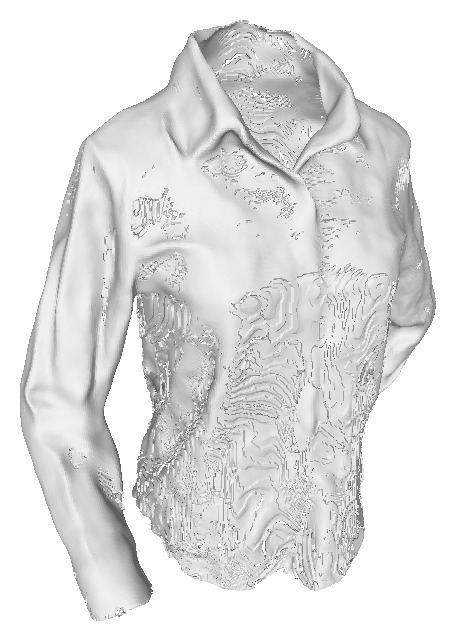}\includegraphics[width=.3in]{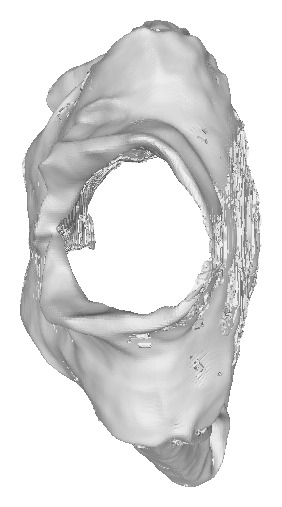}&
    \includegraphics[width=.47in]{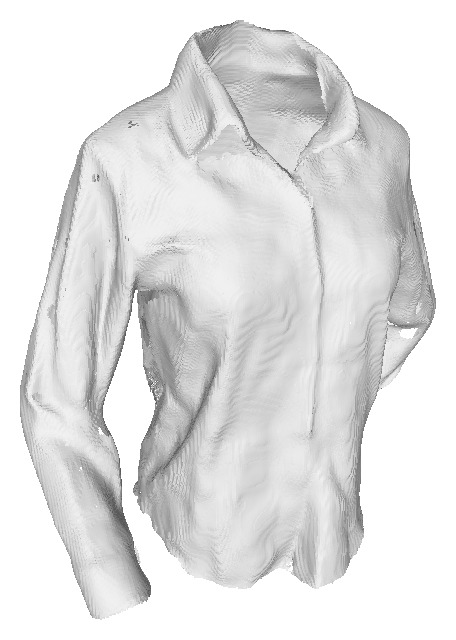}\includegraphics[width=.3in]{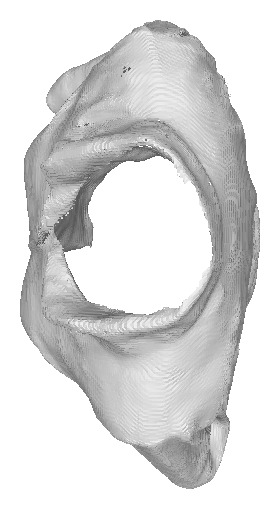}\\
    \raisebox{.28in}{\#5} & \includegraphics[width=.386in]{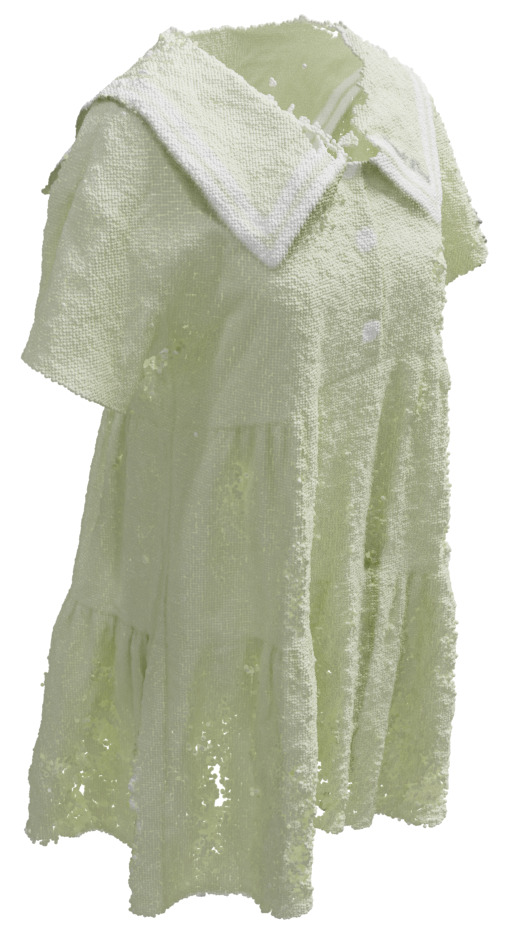}&
    \includegraphics[width=.47in]{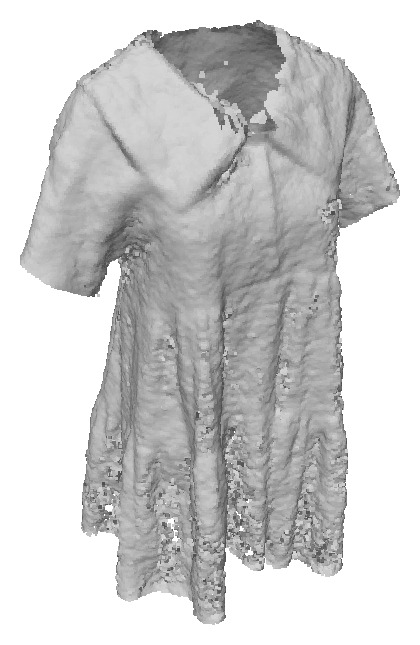}\includegraphics[width=.3in]{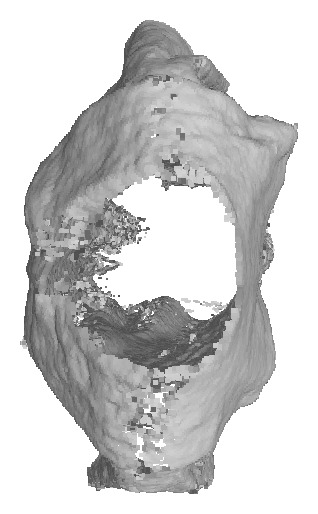}&
    \includegraphics[width=.47in]{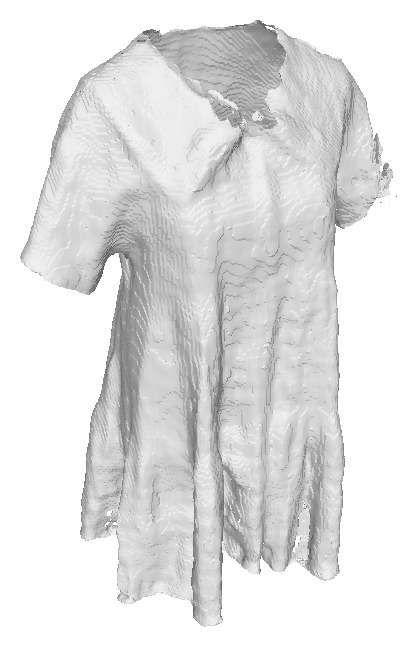}\includegraphics[width=.3in]{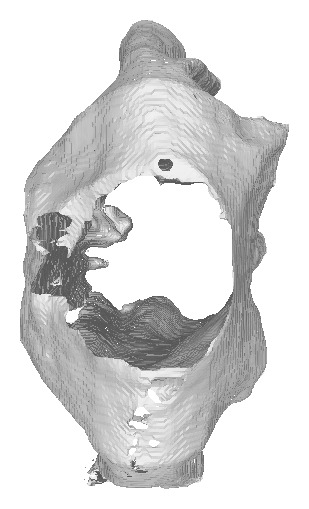}&
    \includegraphics[width=.47in]{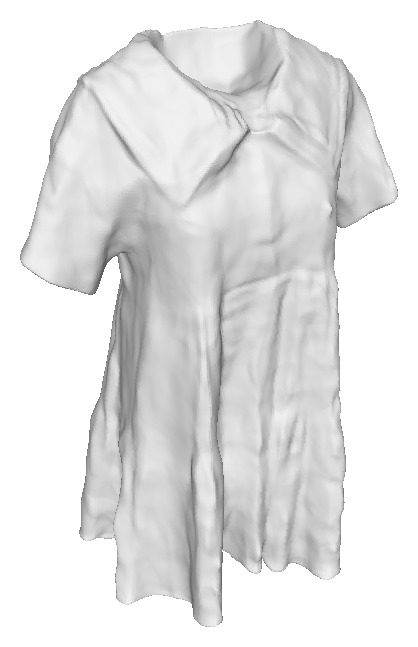}\includegraphics[width=.3in]{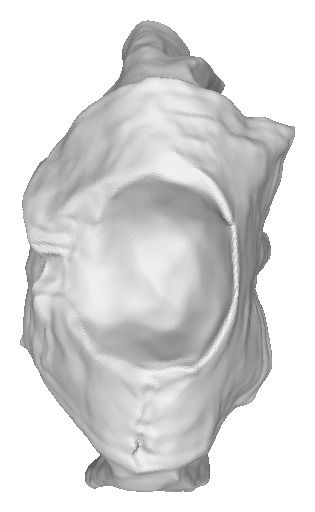}&
    \includegraphics[width=.47in]{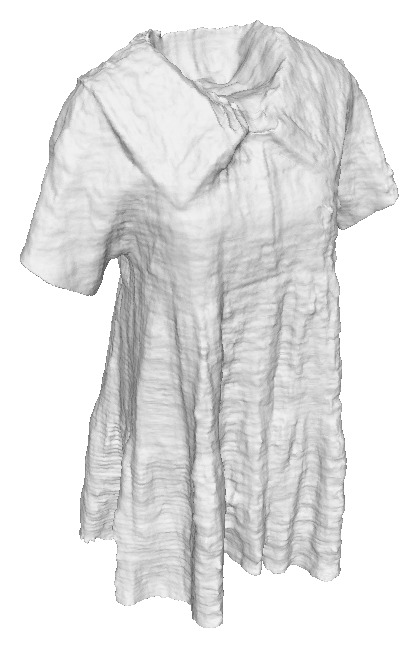}\includegraphics[width=.3in]{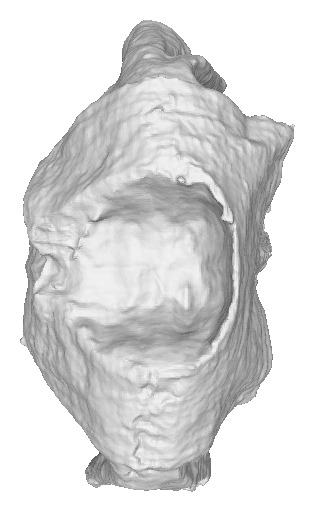}&
    \includegraphics[width=.47in]{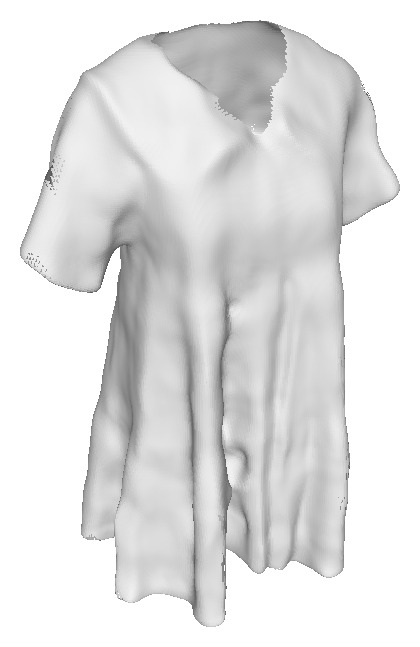}\includegraphics[width=.3in]{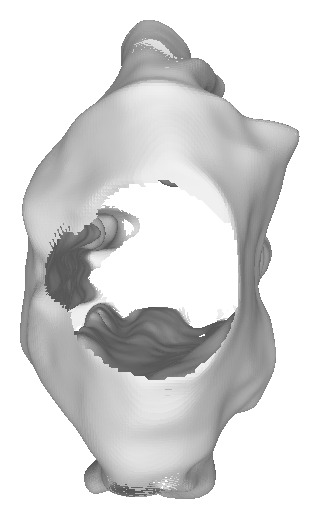}&
    \includegraphics[width=.47in]{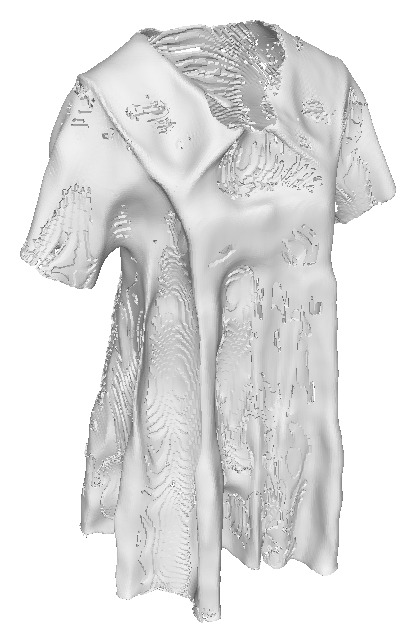}\includegraphics[width=.3in]{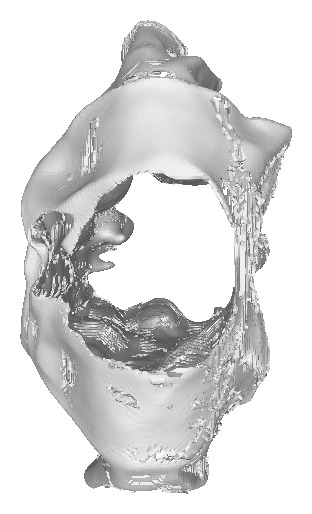}&
    \includegraphics[width=.47in]{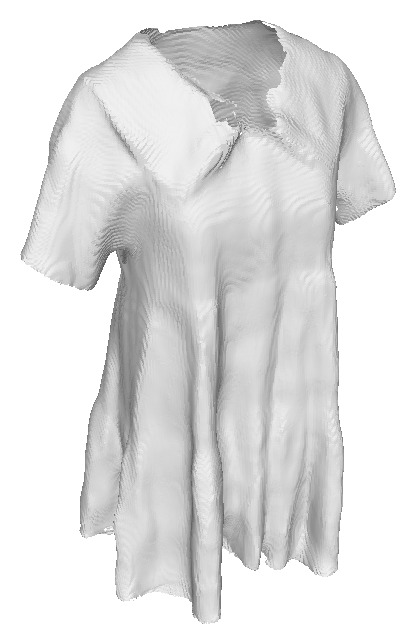}\includegraphics[width=.3in]{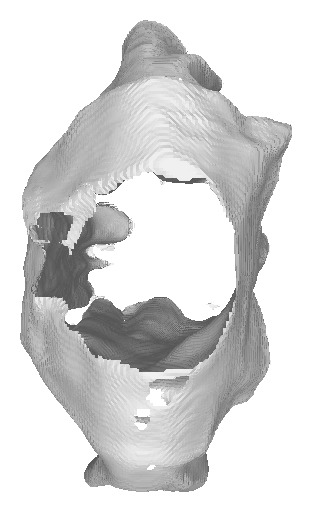}\\
    \raisebox{.28in}{\#6} & \includegraphics[width=.47in]{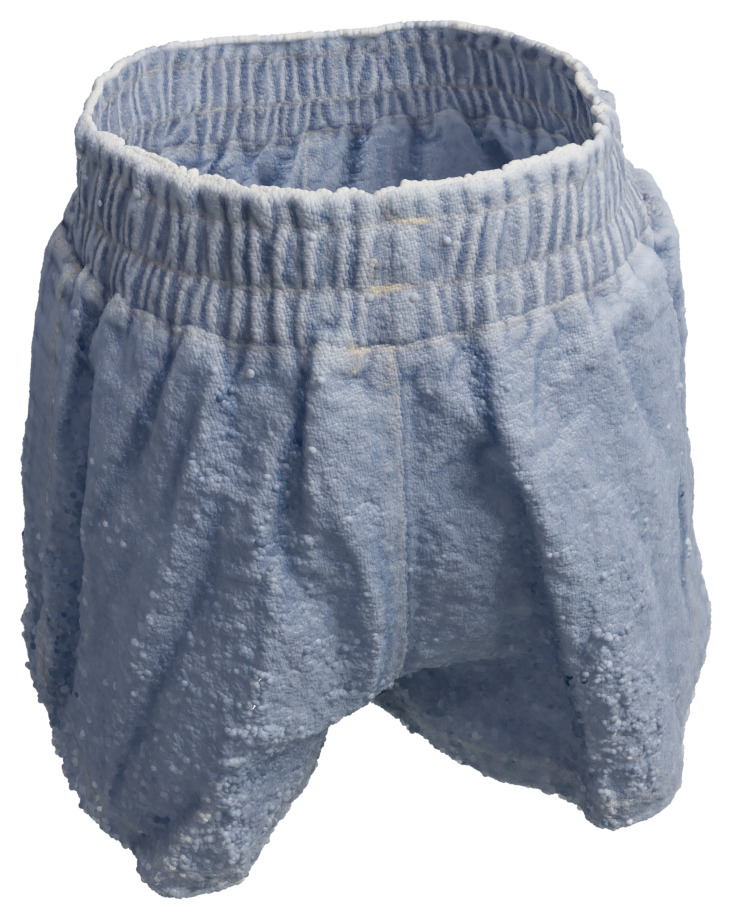}&
    \includegraphics[width=.44in]{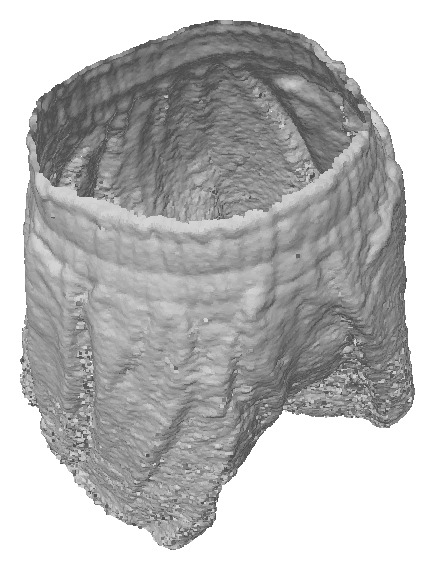}\includegraphics[width=.34in]{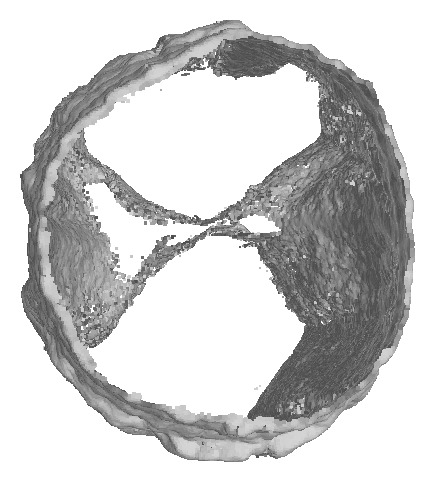}&
    \includegraphics[width=.44in]{ours/432-1}\includegraphics[width=.34in]{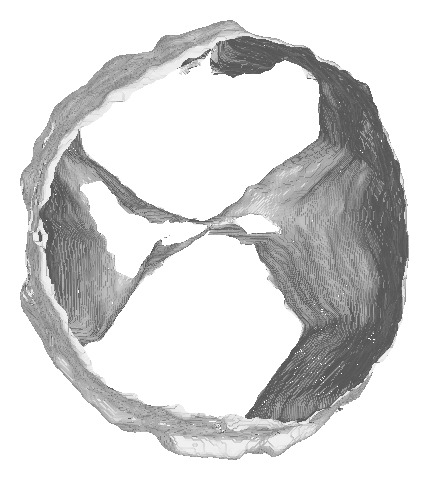}&
    \includegraphics[width=.44in]{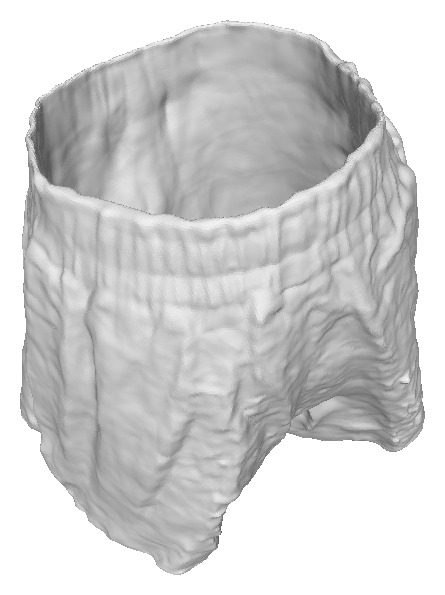}\includegraphics[width=.34in]{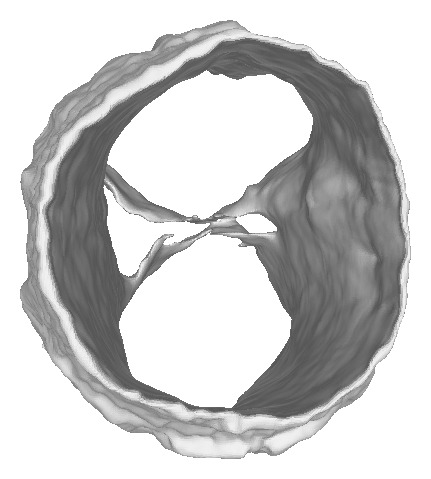}&
    \includegraphics[width=.44in]{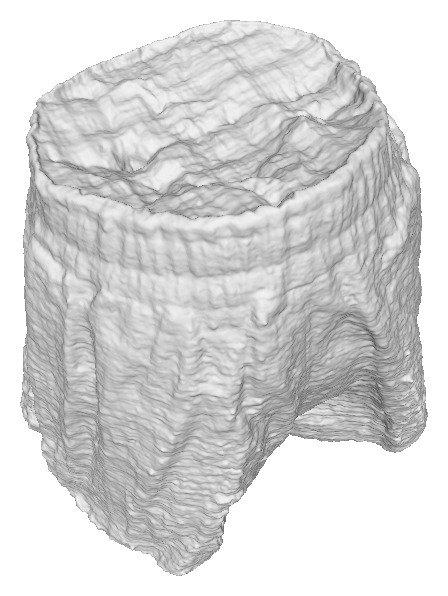}\includegraphics[width=.34in]{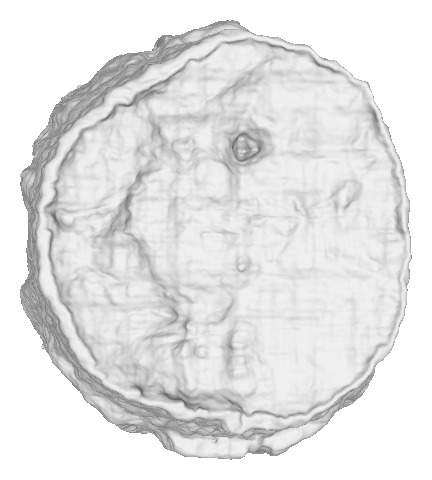}&
    \includegraphics[width=.44in]{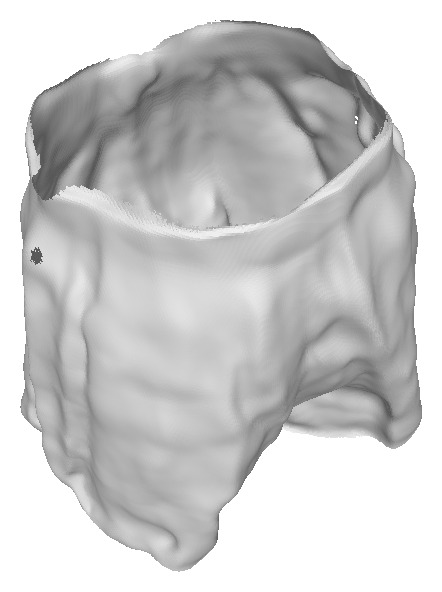}\includegraphics[width=.34in]{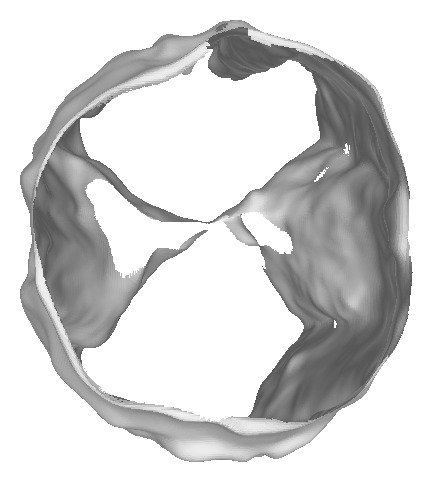}&
    \includegraphics[width=.44in]{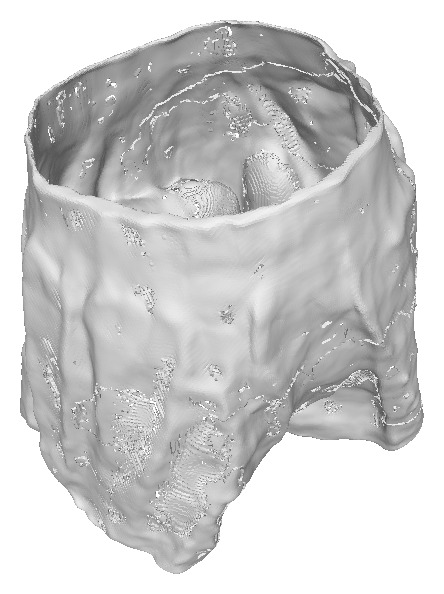}\includegraphics[width=.34in]{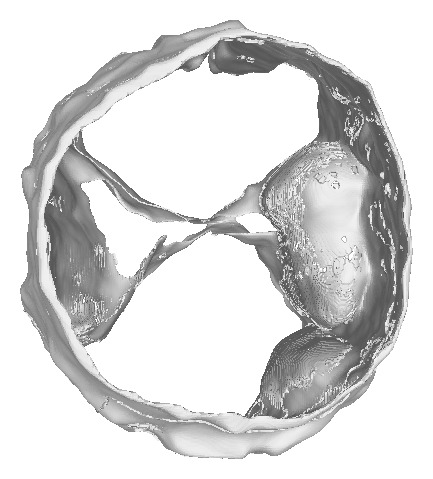}&
    \includegraphics[width=.44in]{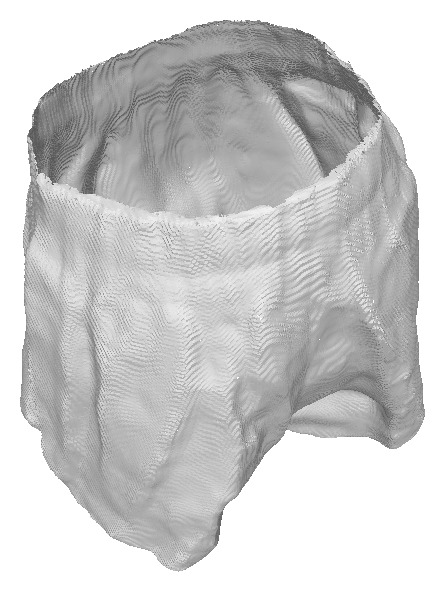}\includegraphics[width=.34in]{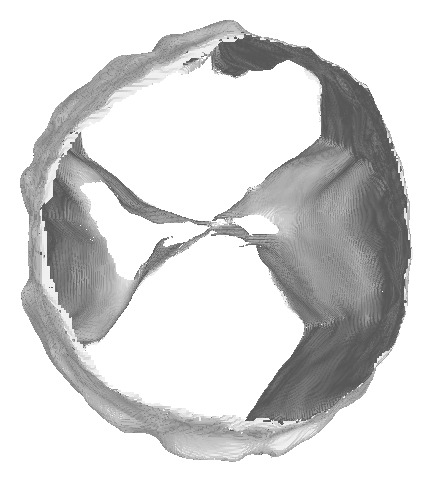}\\
    \raisebox{.24in}{\#7} & \includegraphics[width=.47in]{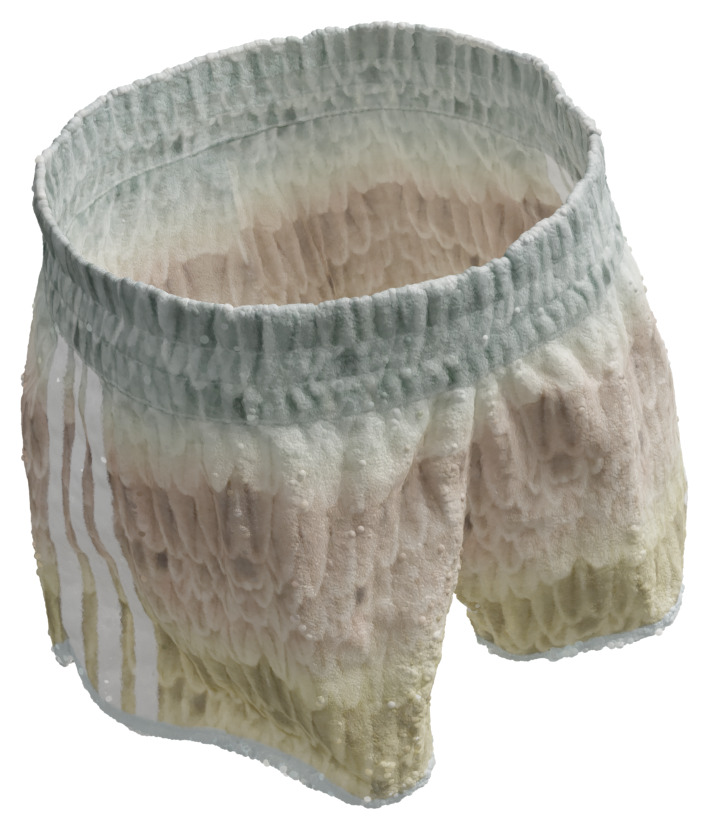}&
    \includegraphics[width=.47in]{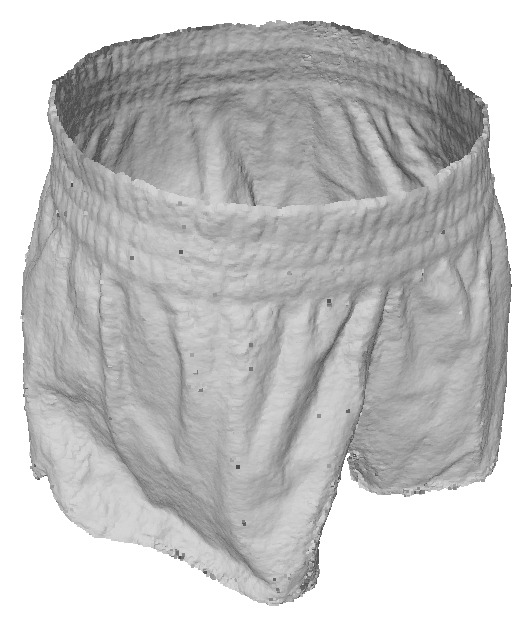}\includegraphics[width=.343in]{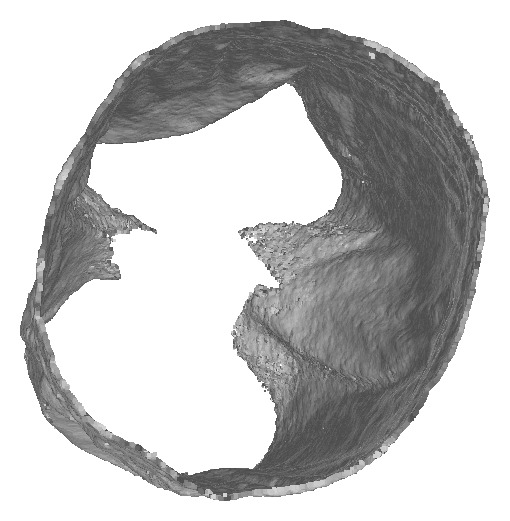}&
    \includegraphics[width=.47in]{ours/448-2}\includegraphics[width=.343in]{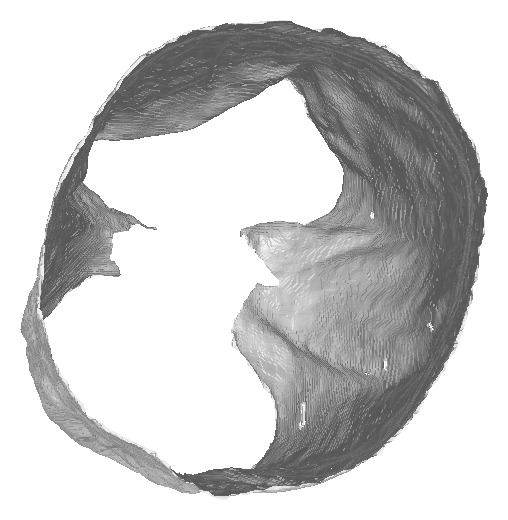}&
    \includegraphics[width=.47in]{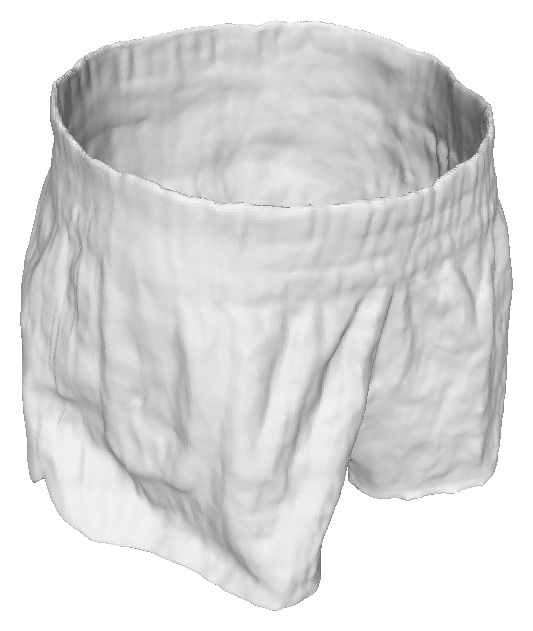}\includegraphics[width=.343in]{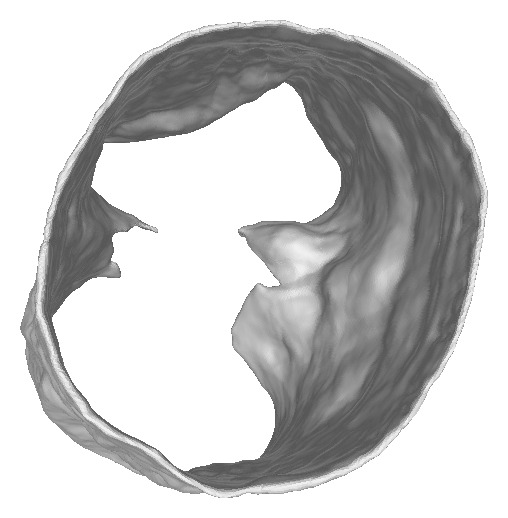}&
    \includegraphics[width=.47in]{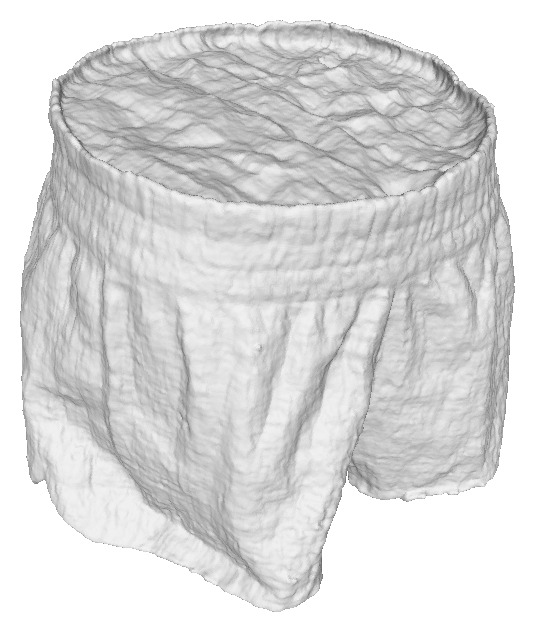}\includegraphics[width=.343in]{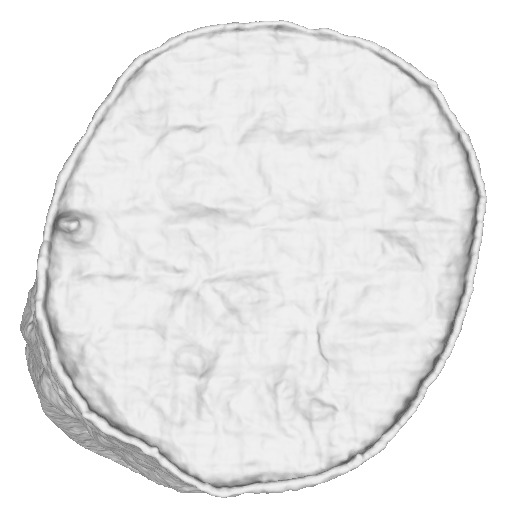}&
    \includegraphics[width=.47in]{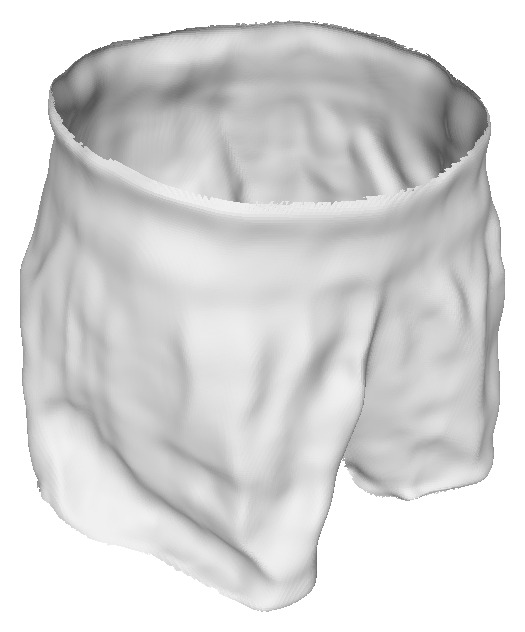}\includegraphics[width=.343in]{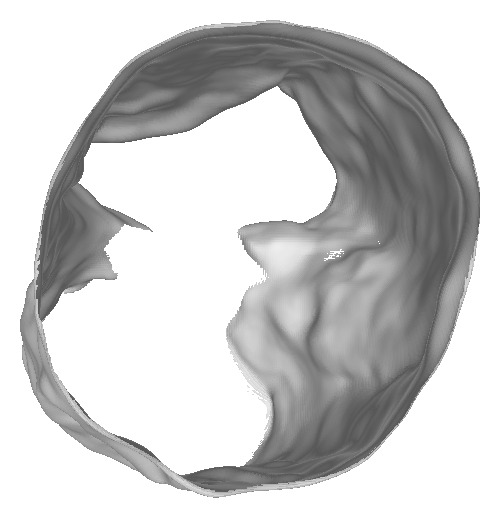}&
    \includegraphics[width=.47in]{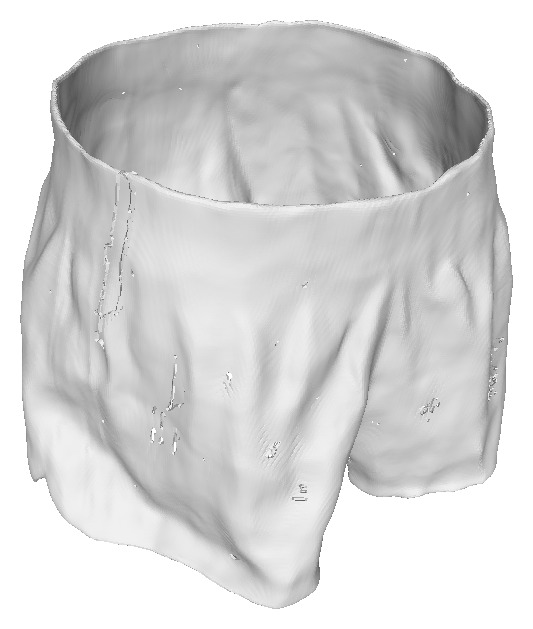}\includegraphics[width=.343in]{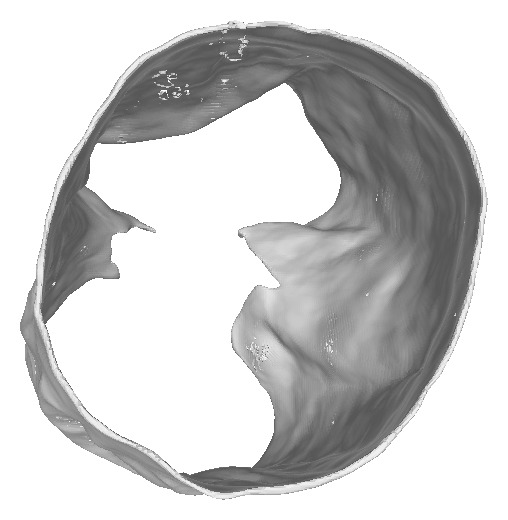}&
    \includegraphics[width=.47in]{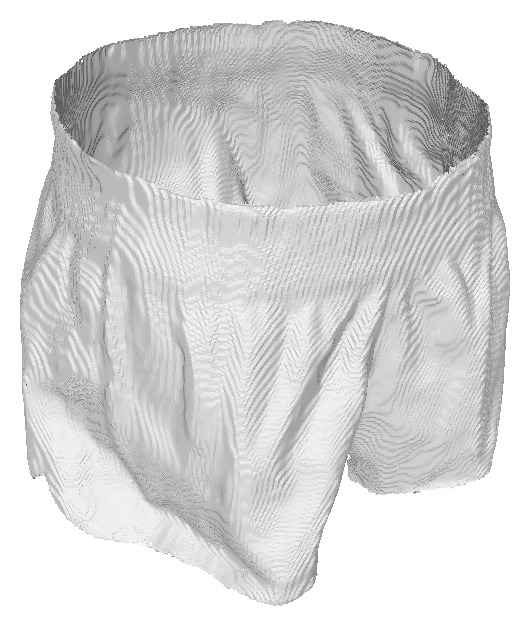}\includegraphics[width=.343in]{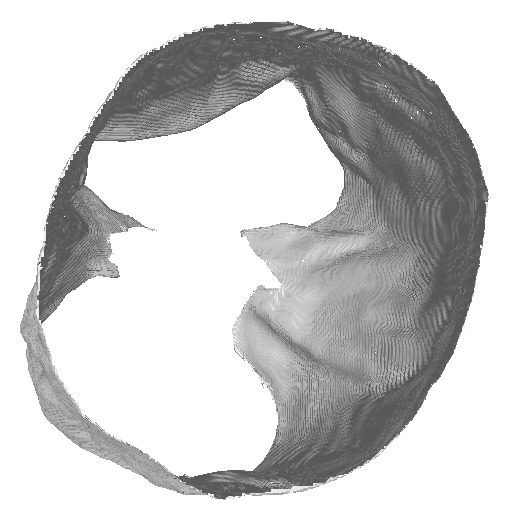}\\
    \raisebox{.26in}{\#8} & \includegraphics[width=.423in]{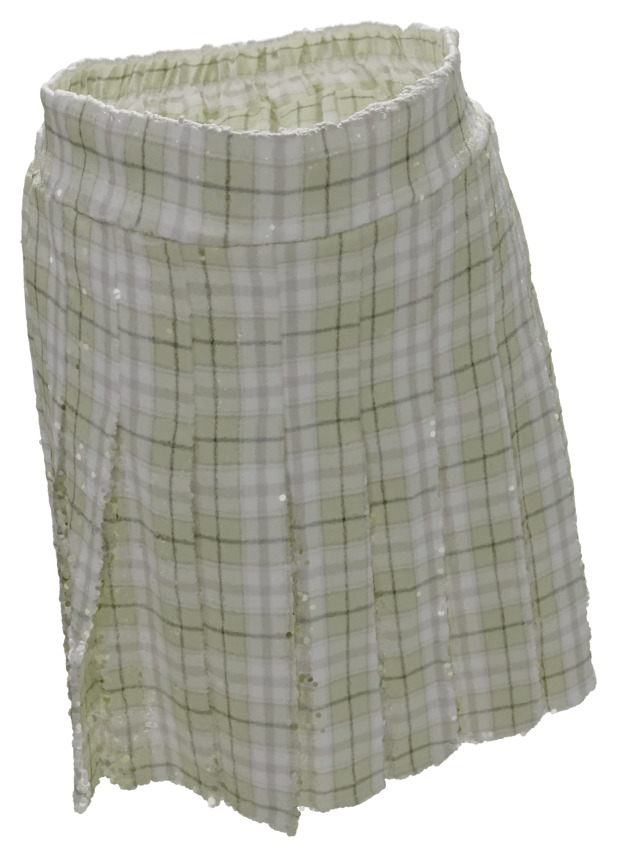}&
    \includegraphics[width=.423in]{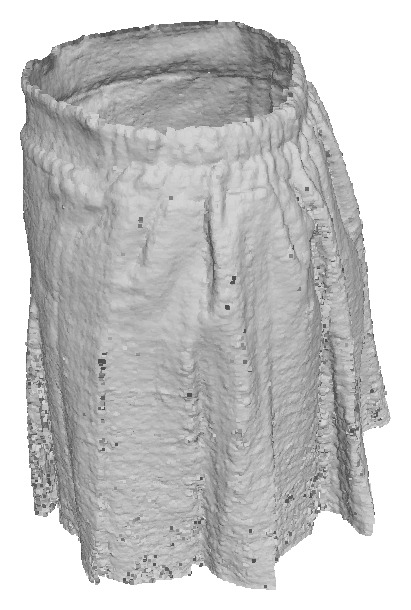}\includegraphics[width=.386in]{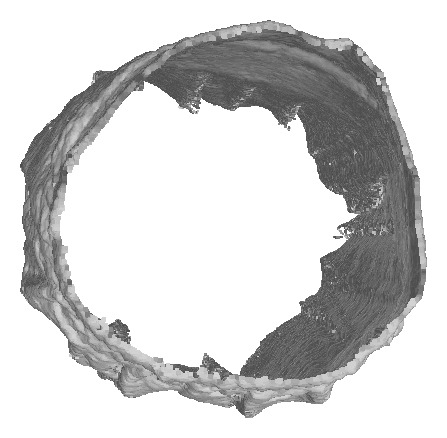}&
    \includegraphics[width=.423in]{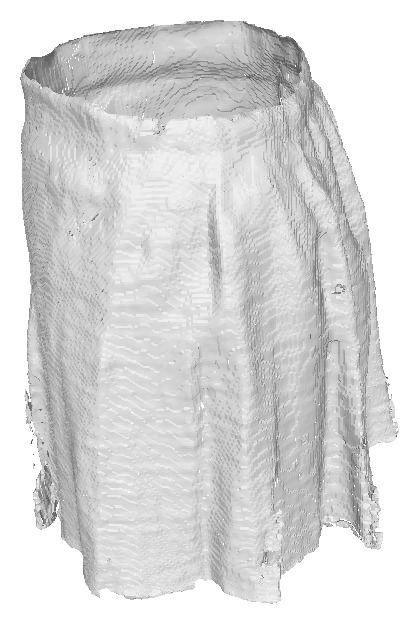}\includegraphics[width=.386in]{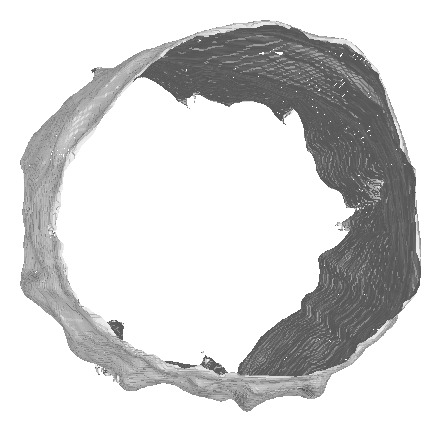}&
    \includegraphics[width=.423in]{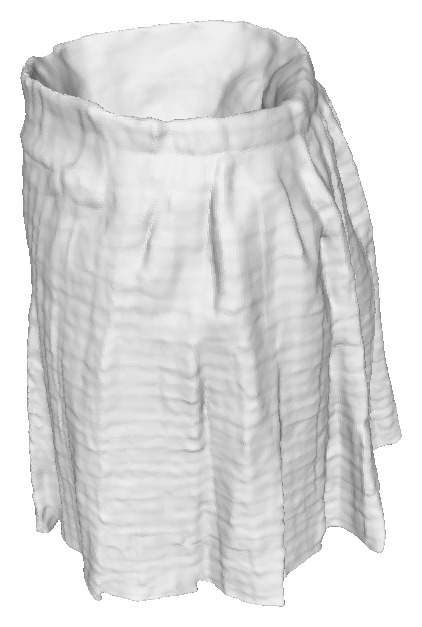}\includegraphics[width=.386in]{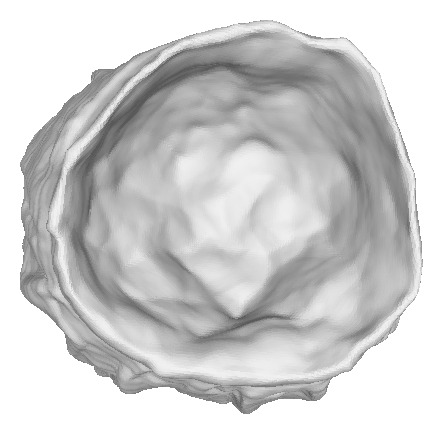}&
    \includegraphics[width=.423in]{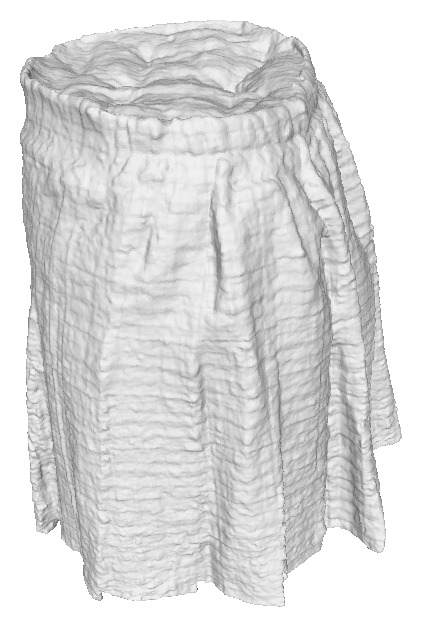}\includegraphics[width=.386in]{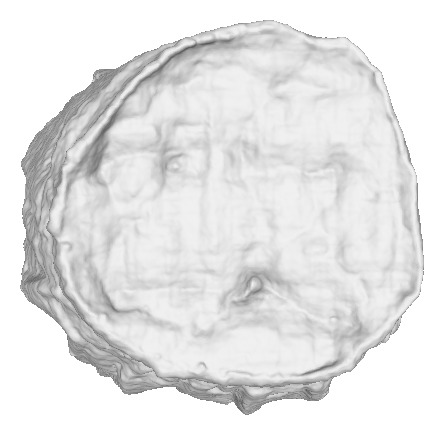}&
    \includegraphics[width=.423in]{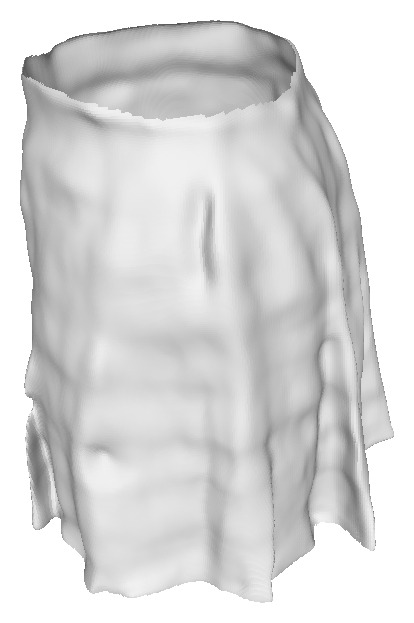}\includegraphics[width=.386in]{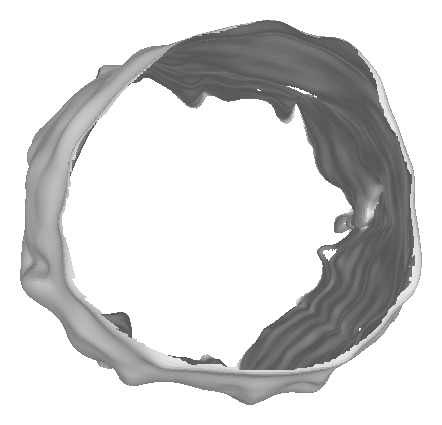}&
    \includegraphics[width=.423in]{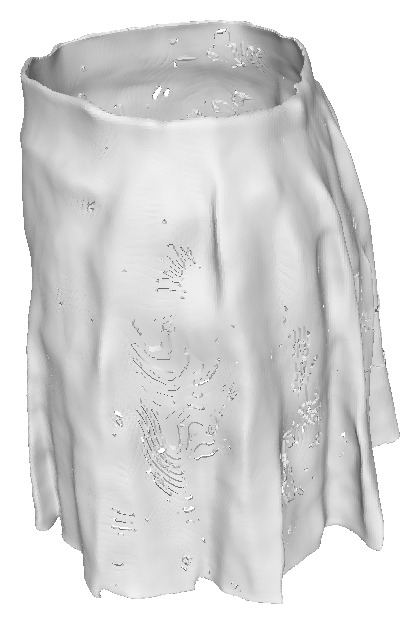}\includegraphics[width=.386in]{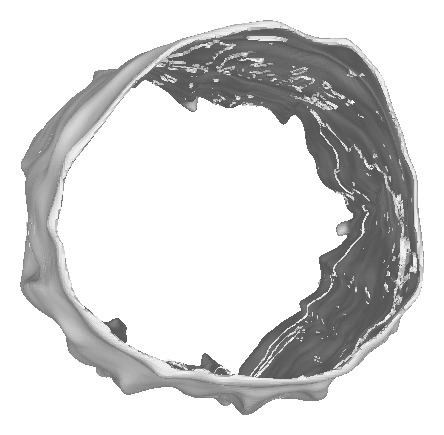}&
    \includegraphics[width=.423in]{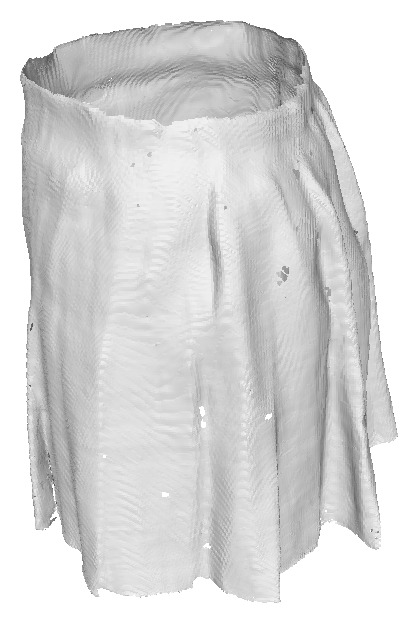}\includegraphics[width=.386in]{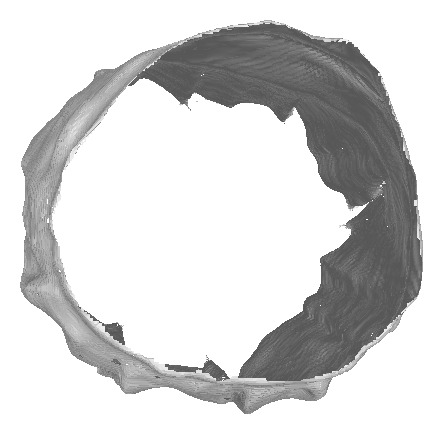}\\
    \raisebox{.25in}{\#9} & \includegraphics[width=.497in]{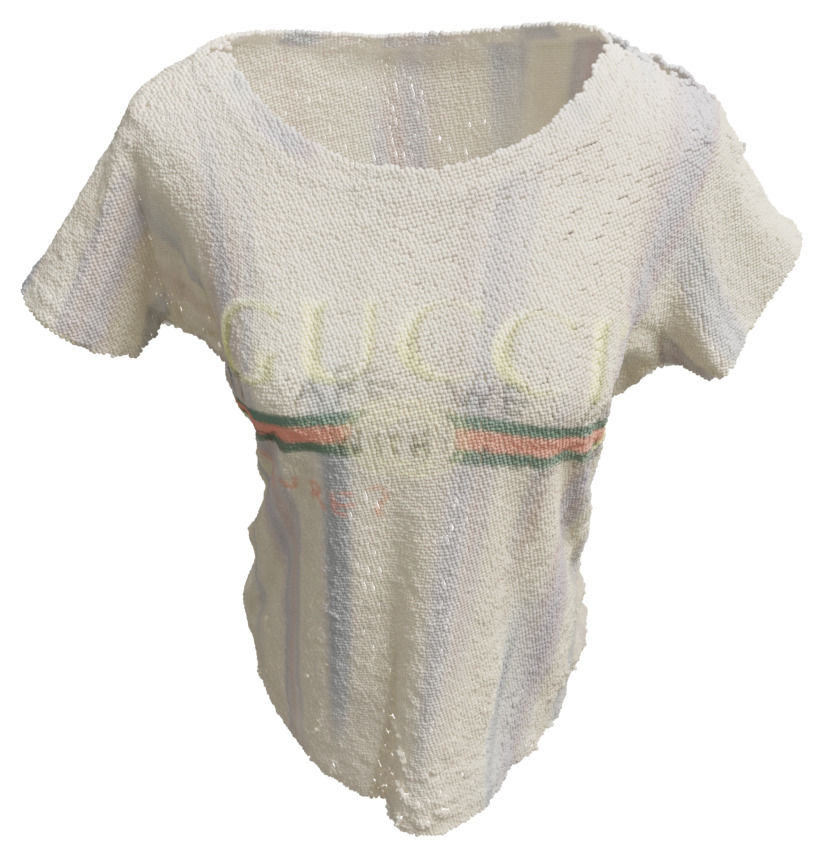}&
    \includegraphics[width=.47in]{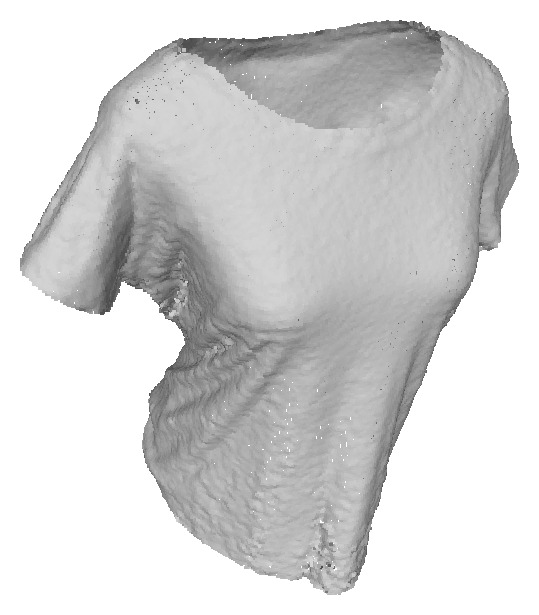}\includegraphics[width=.257in]{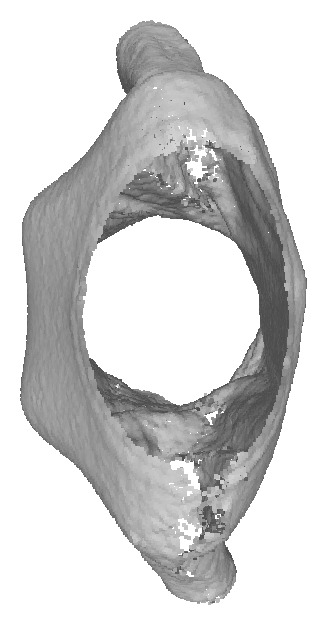}&
    \includegraphics[width=.47in]{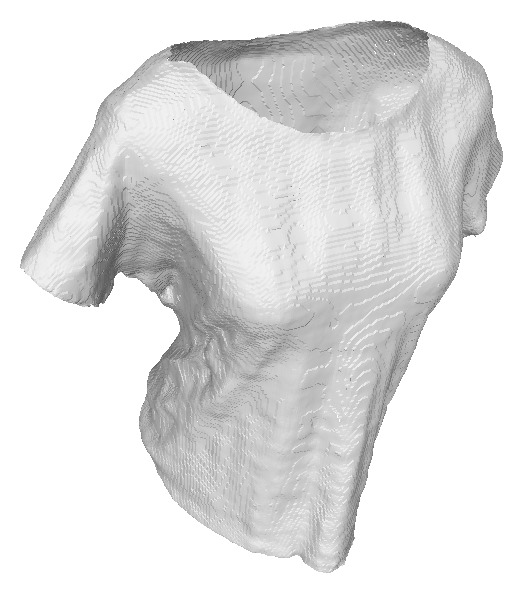}\includegraphics[width=.257in]{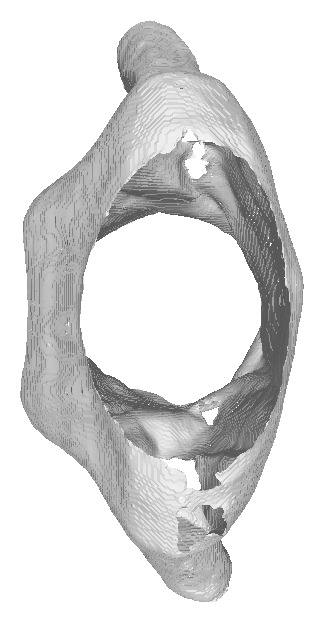}&
    \includegraphics[width=.47in]{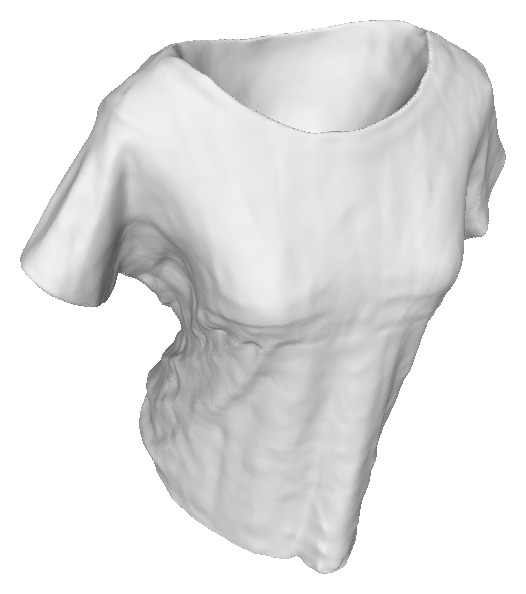}\includegraphics[width=.257in]{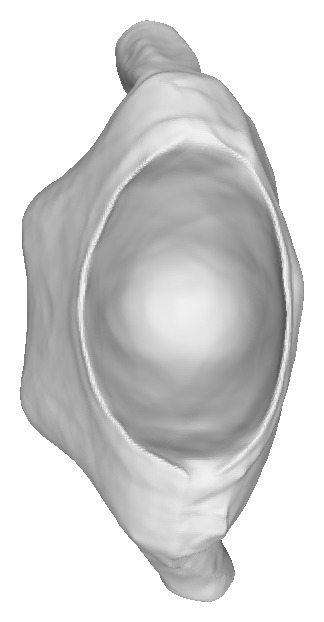}&
    \includegraphics[width=.47in]{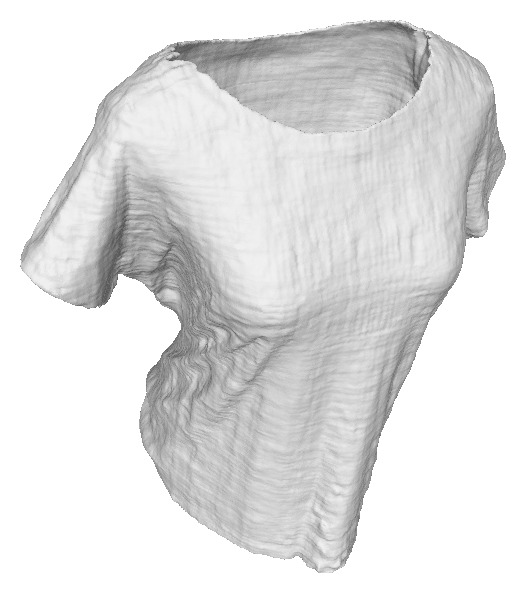}\includegraphics[width=.257in]{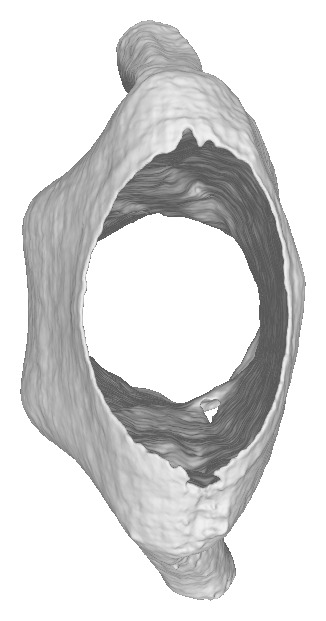}&
    \includegraphics[width=.47in]{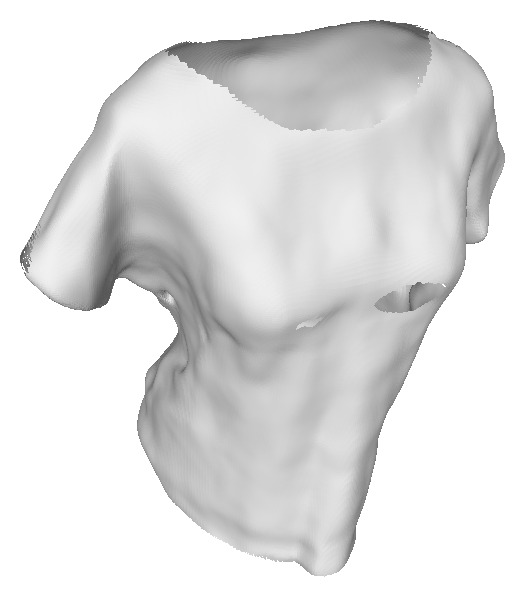}\includegraphics[width=.257in]{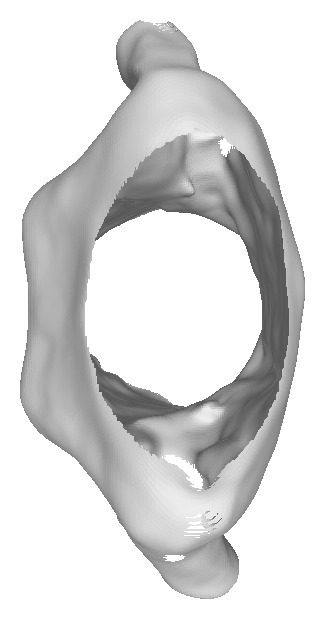}&
    \includegraphics[width=.47in]{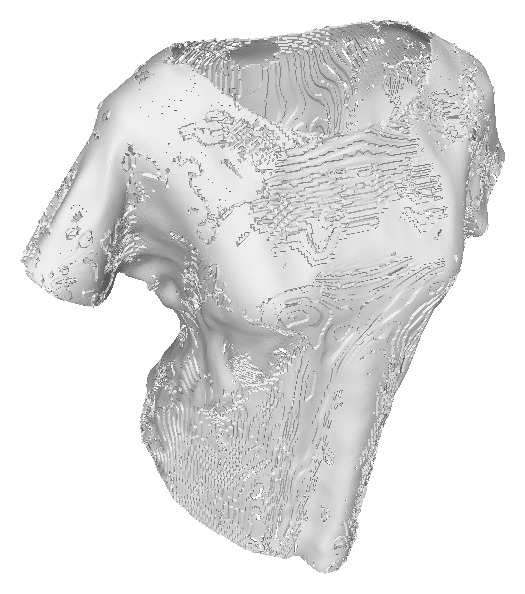}\includegraphics[width=.257in]{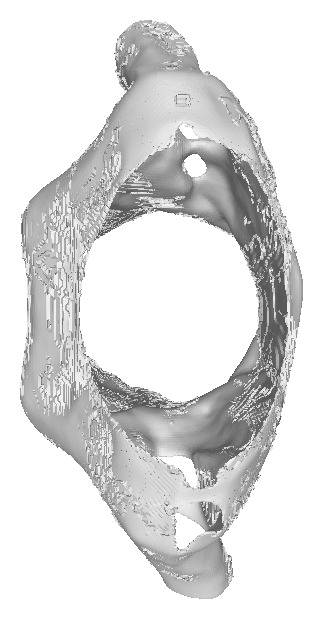}&
    \includegraphics[width=.47in]{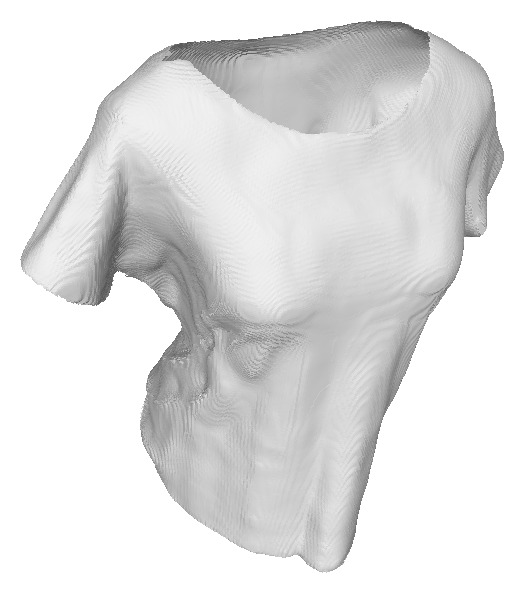}\includegraphics[width=.257in]{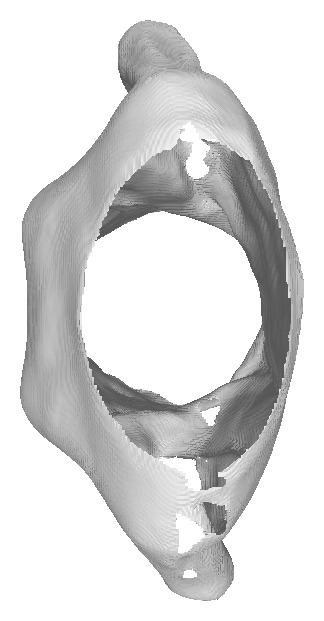}\\
    \raisebox{.26in}{LS-C0} & \includegraphics[width=.46in]{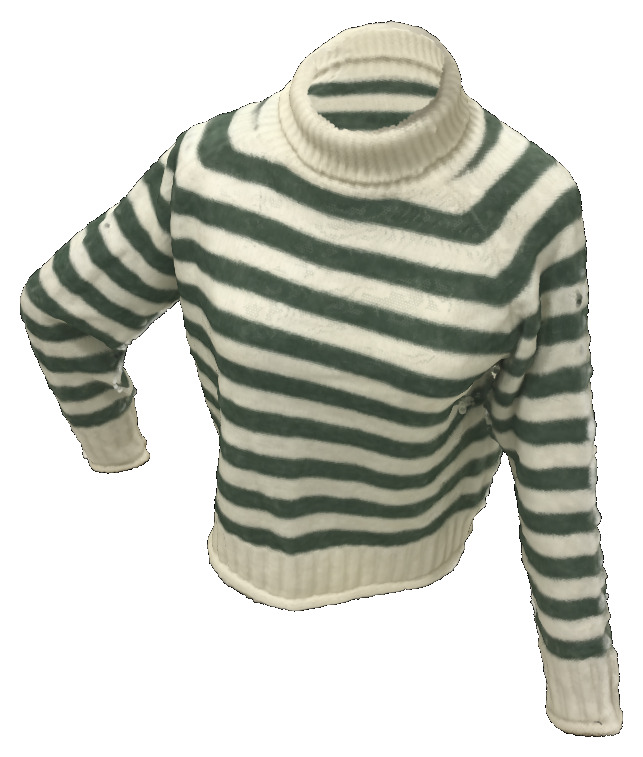}&
    \includegraphics[width=.425in]{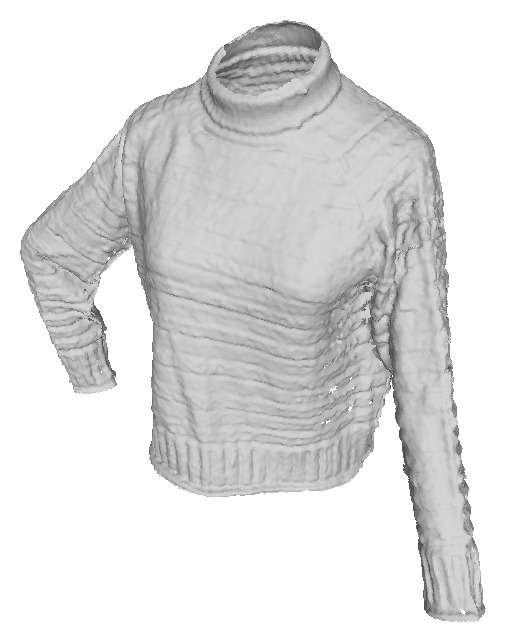}\includegraphics[width=.3in]{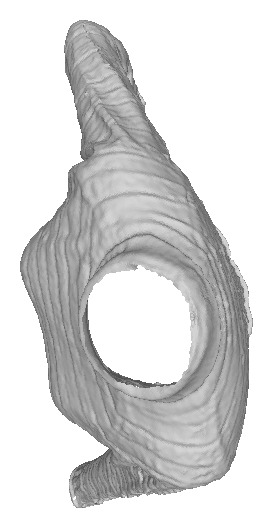}&
    \includegraphics[width=.425in]{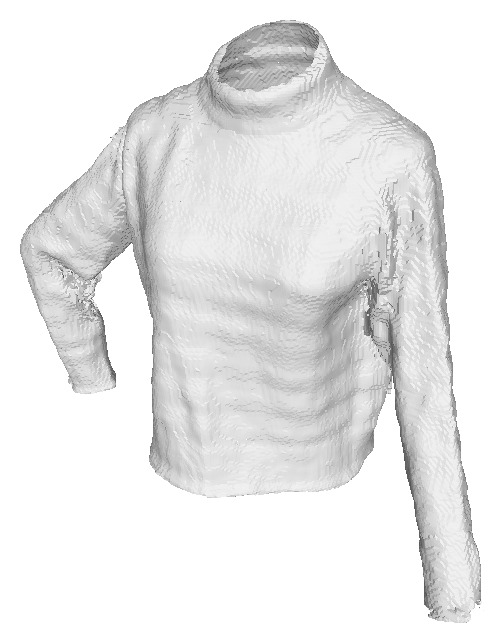}\includegraphics[width=.3in]{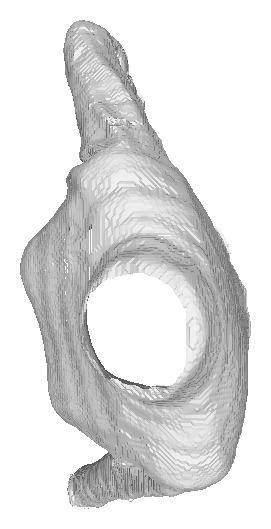}&
    \includegraphics[width=.425in]{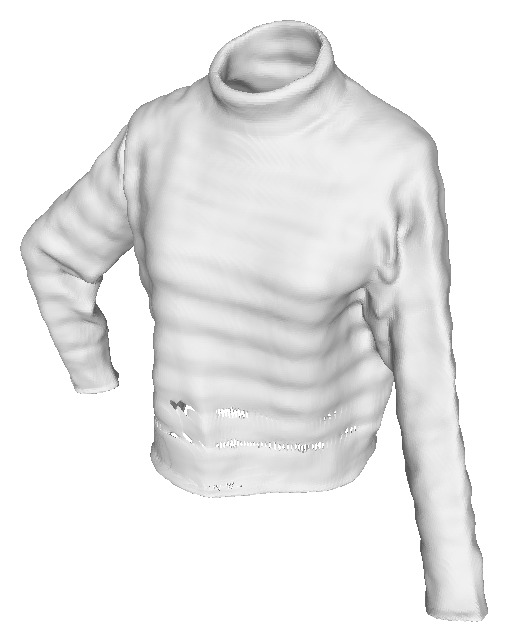}\includegraphics[width=.3in]{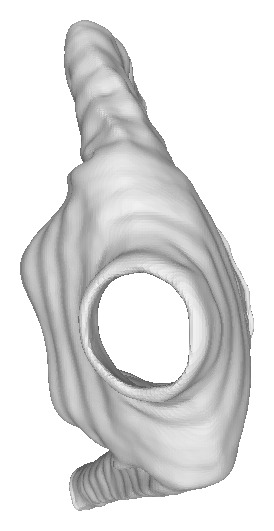}&
    \includegraphics[width=.425in]{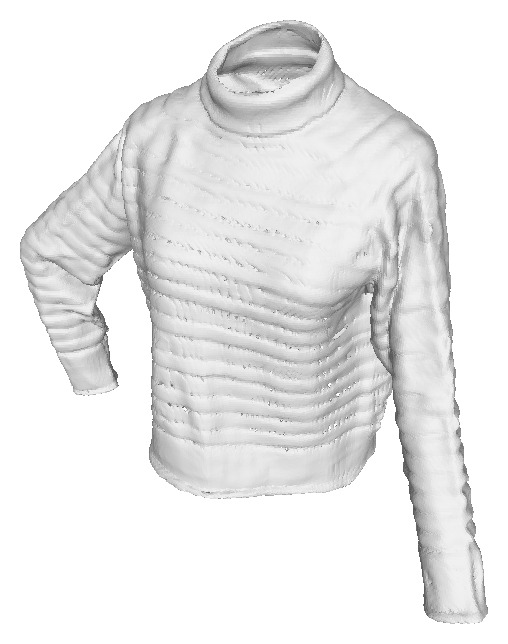}\includegraphics[width=.3in]{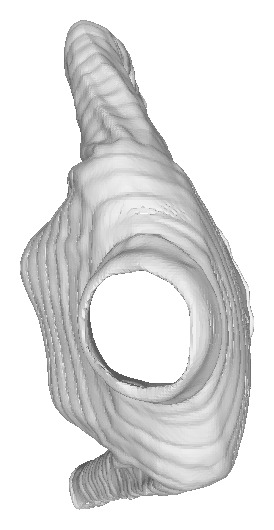}&
    \includegraphics[width=.425in]{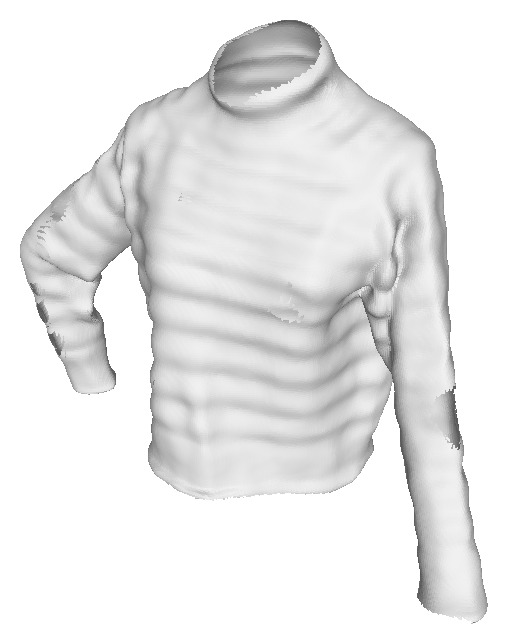}\includegraphics[width=.3in]{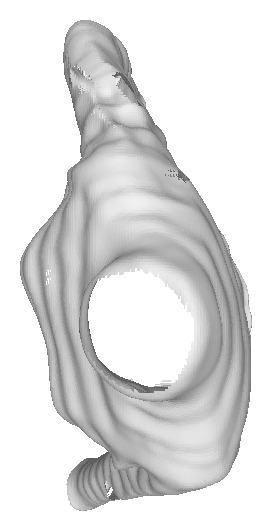}&
    \includegraphics[width=.425in]{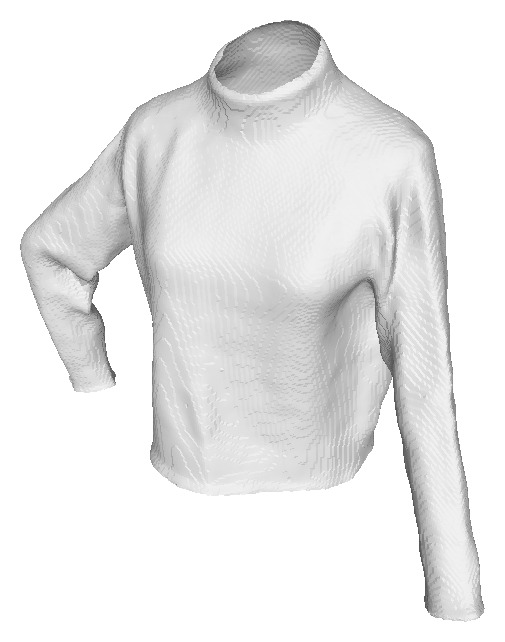}\includegraphics[width=.3in]{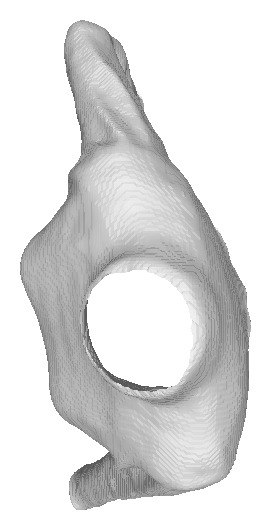}&
    \includegraphics[width=.425in]{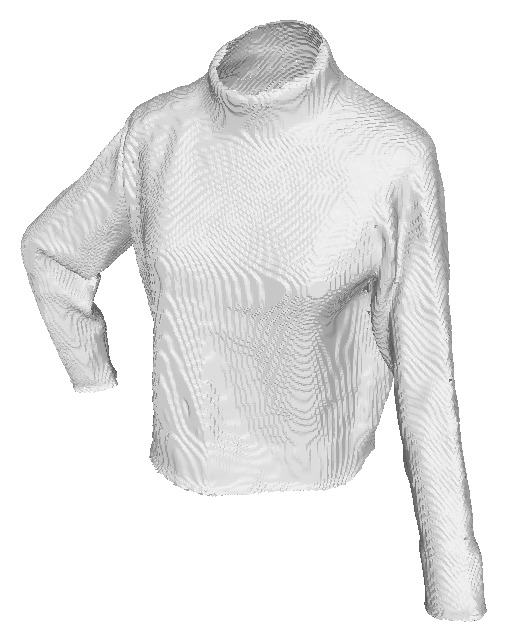}\includegraphics[width=.3in]{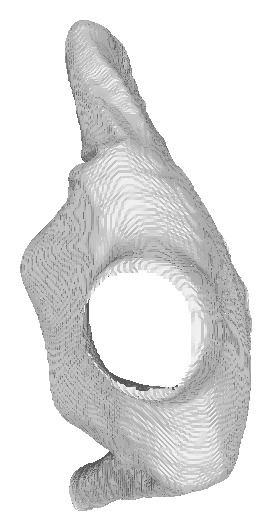} \\
    \raisebox{.26in}{SS-D0} & \includegraphics[width=.369in]{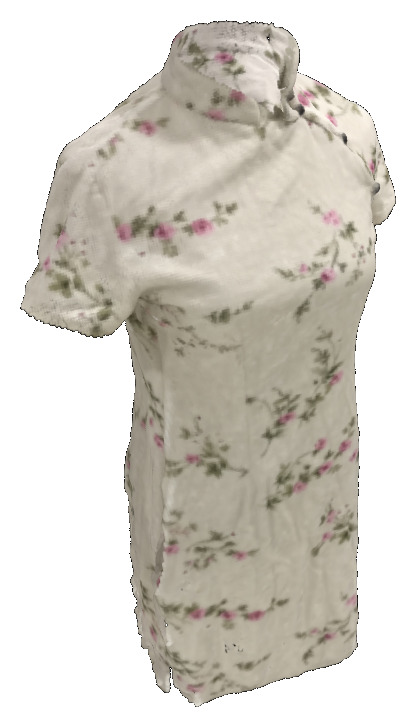}&
    \includegraphics[width=.386in]{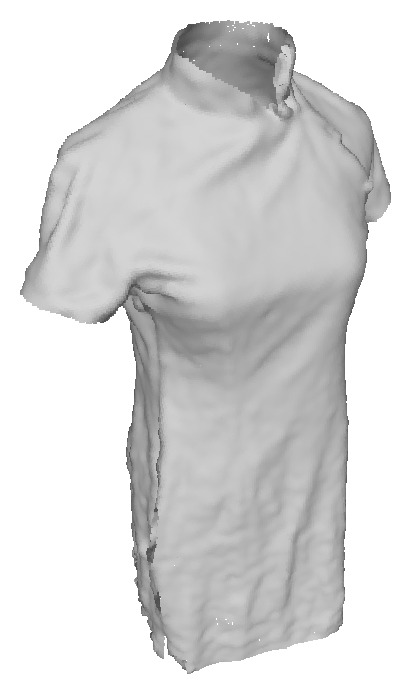}\includegraphics[width=.3in]{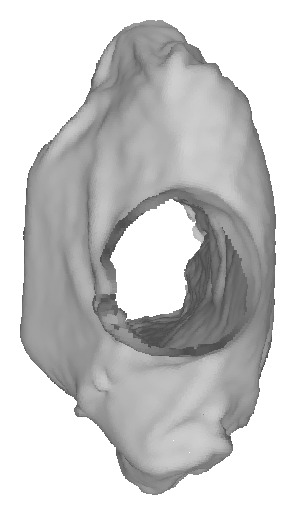}&
    \includegraphics[width=.386in]{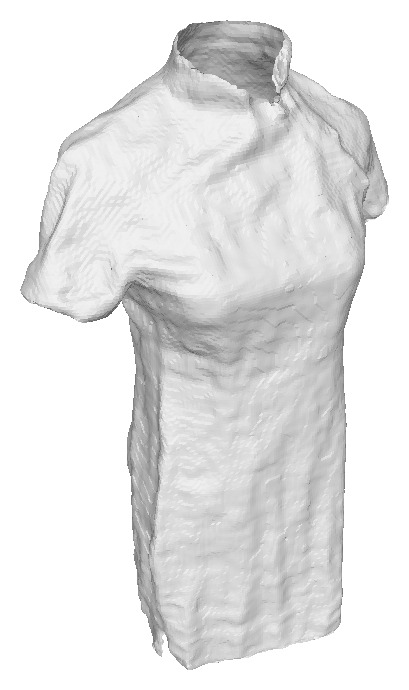}\includegraphics[width=.3in]{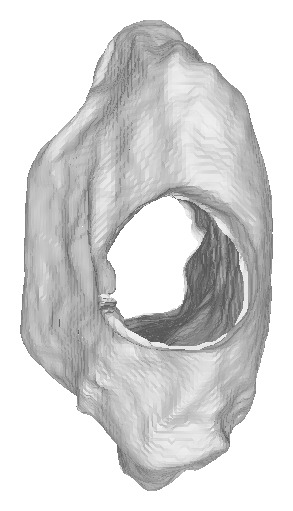}&
    \includegraphics[width=.386in]{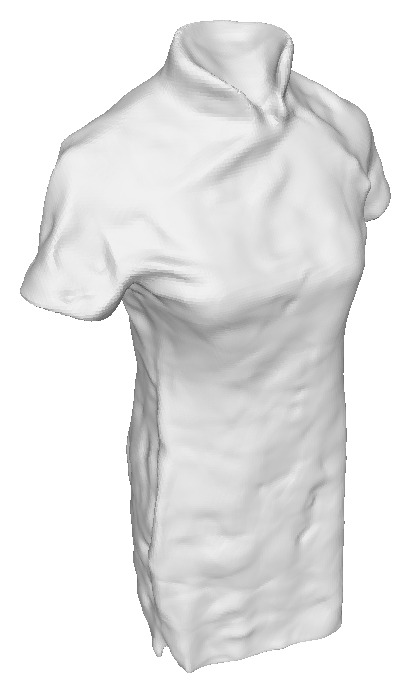}\includegraphics[width=.3in]{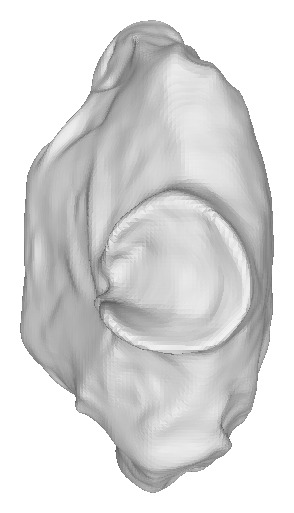}&
    \includegraphics[width=.386in]{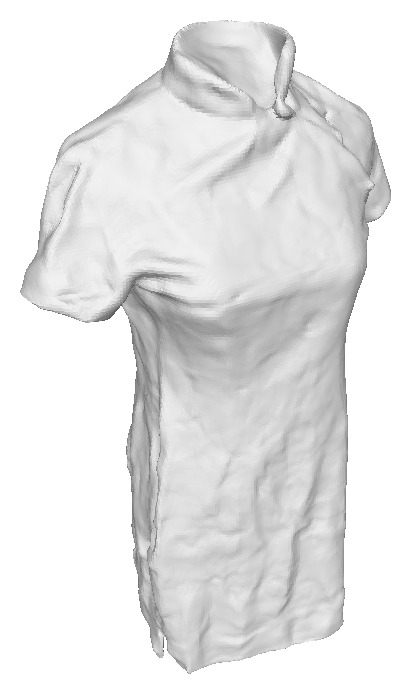}\includegraphics[width=.3in]{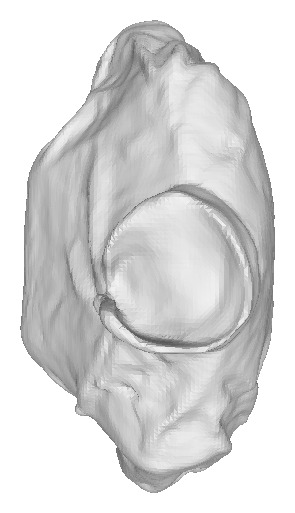}&
    \includegraphics[width=.386in]{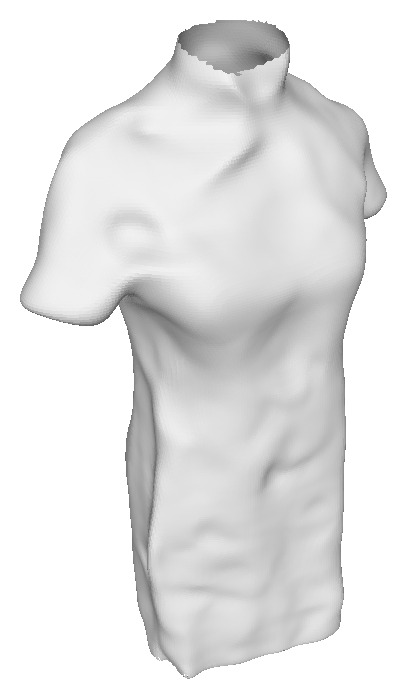}\includegraphics[width=.3in]{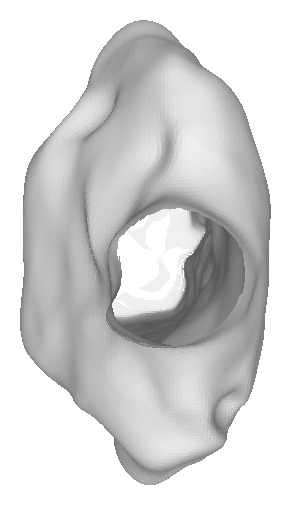}&
    \includegraphics[width=.386in]{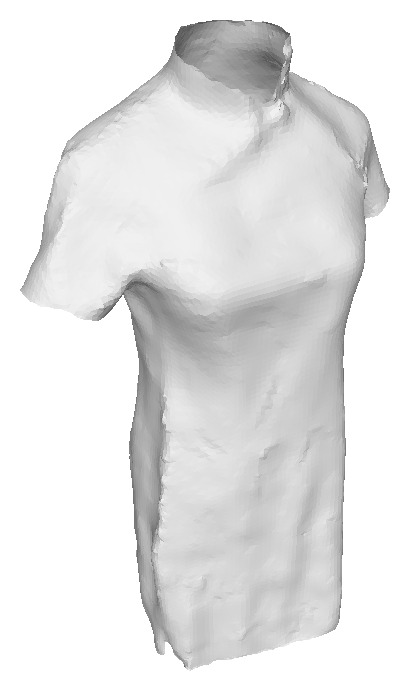}\includegraphics[width=.3in]{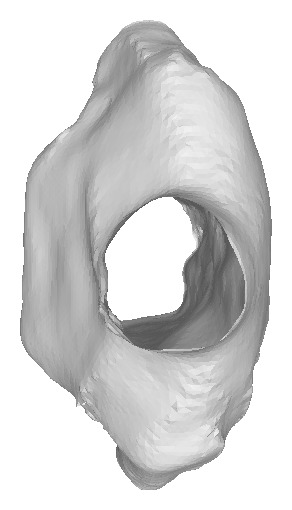}&
    \includegraphics[width=.386in]{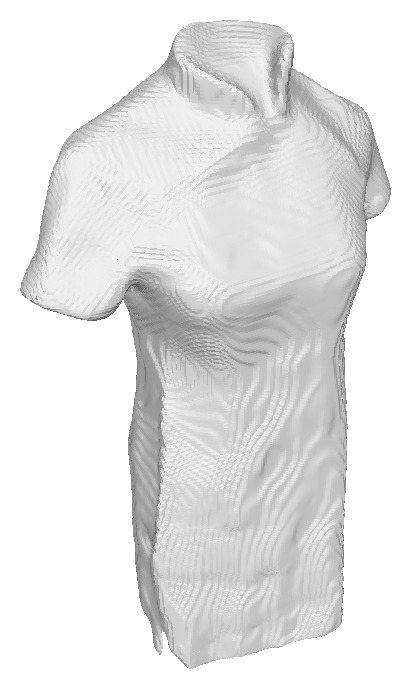}\includegraphics[width=.3in]{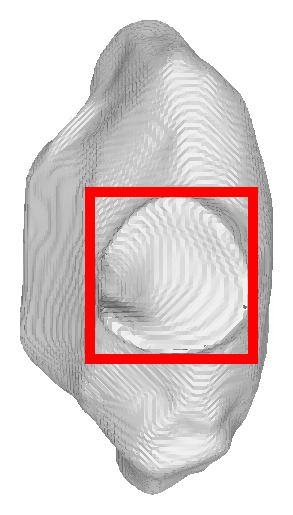} \\
    \raisebox{.28in}{NS-D1} & \includegraphics[width=.386in]{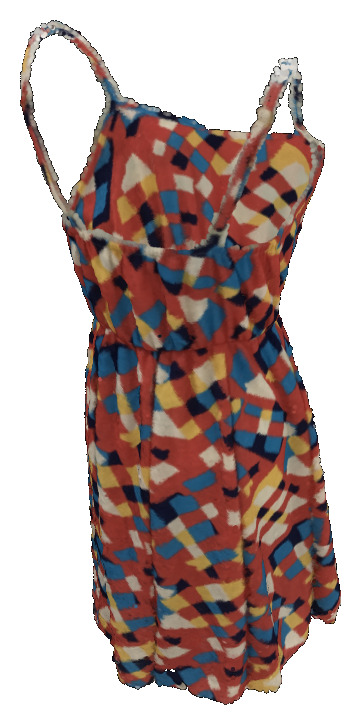}&
    \includegraphics[width=.36in]{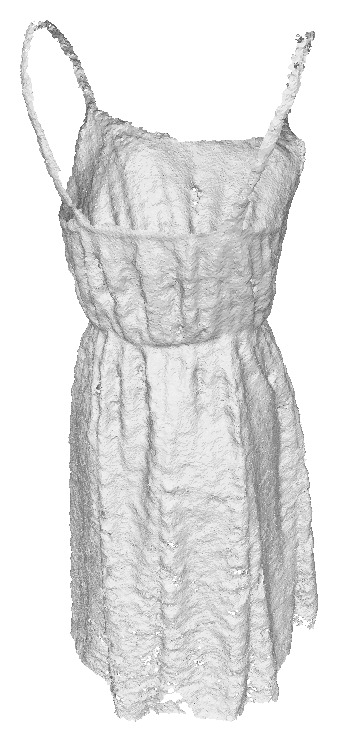}\includegraphics[width=.317in]{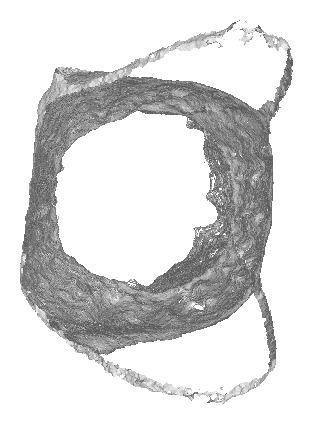}&
    \includegraphics[width=.36in]{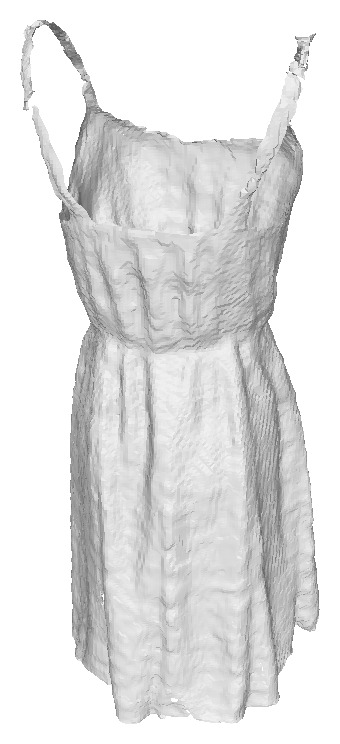}\includegraphics[width=.317in]{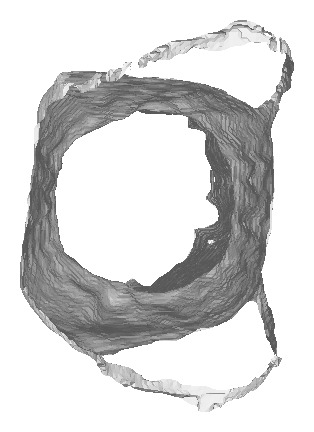}&
    \includegraphics[width=.36in]{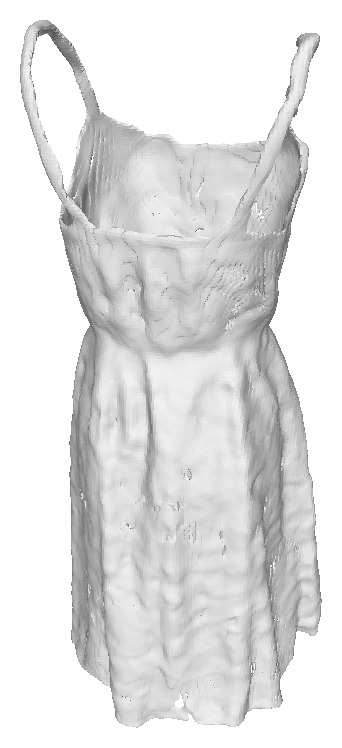}\includegraphics[width=.317in]{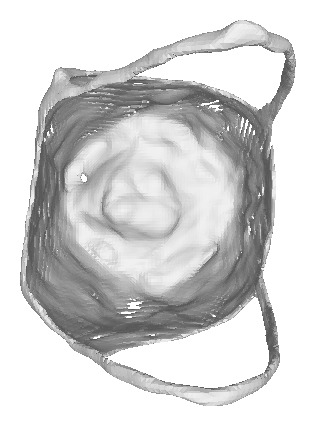}&
    \includegraphics[width=.36in]{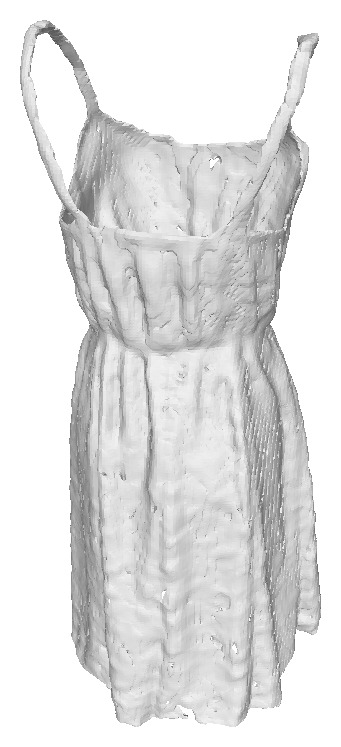}\includegraphics[width=.317in]{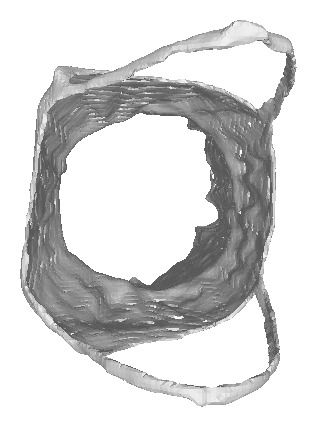}&
    \includegraphics[width=.36in]{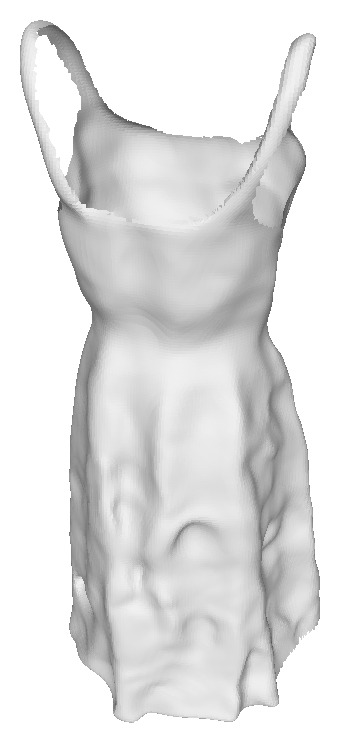}\includegraphics[width=.317in]{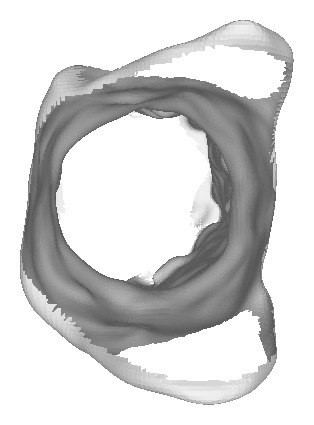}&
    \includegraphics[width=.36in]{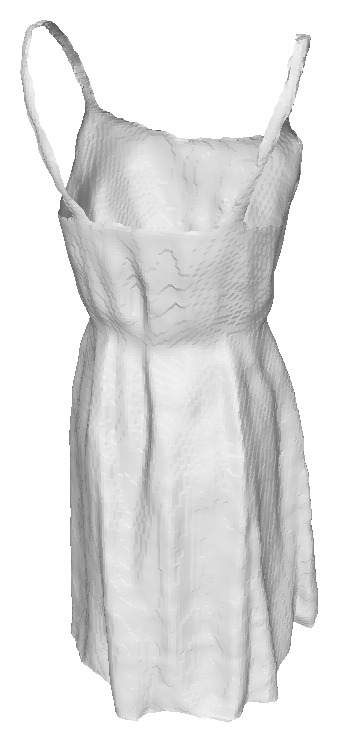}\includegraphics[width=.317in]{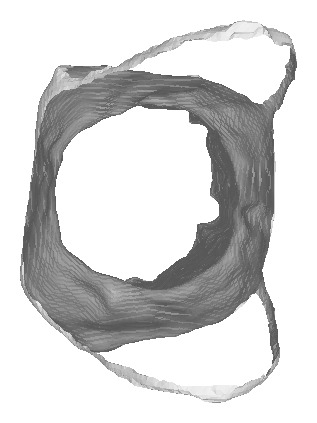}&
    \includegraphics[width=.36in]{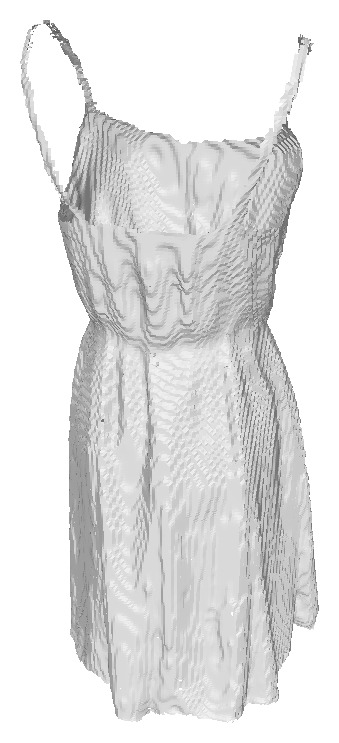}\includegraphics[width=.317in]{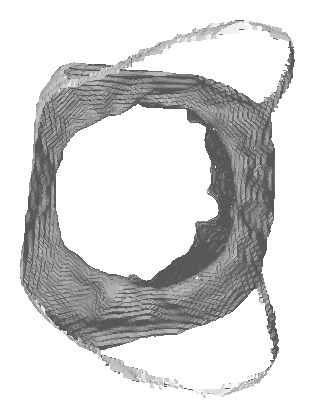} \\
    \raisebox{.22in}{LS-C1} & \includegraphics[width=.514in]{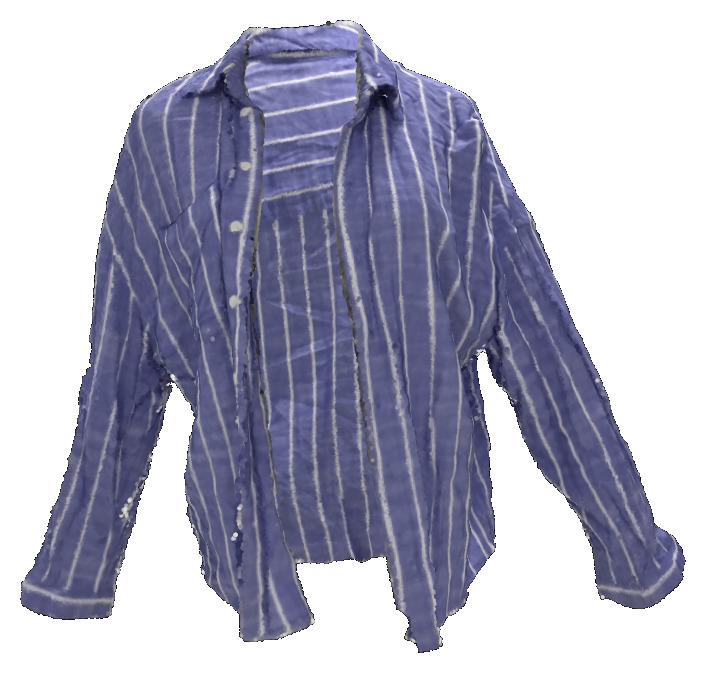}&
    \includegraphics[width=.514in]{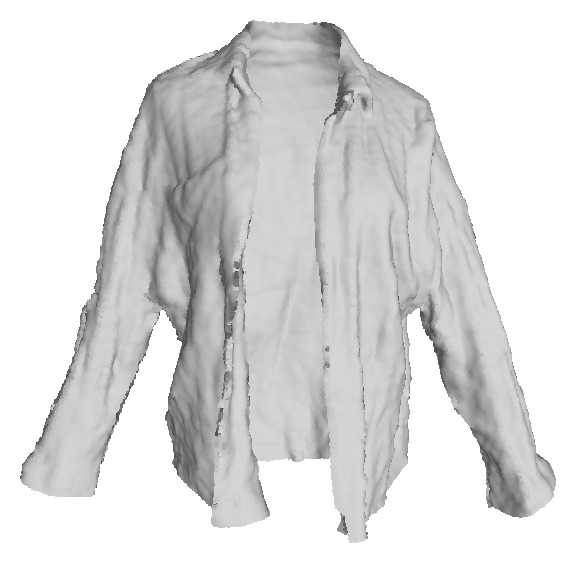}\includegraphics[width=.257in]{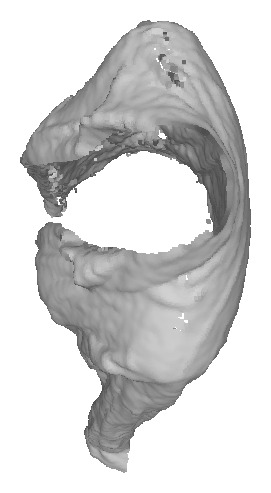}&
    \includegraphics[width=.514in]{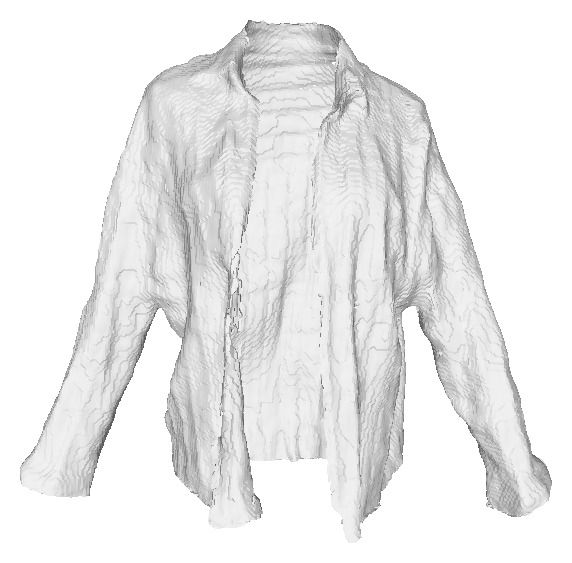}\includegraphics[width=.257in]{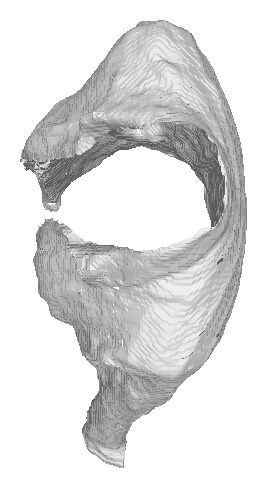}&
    \includegraphics[width=.514in]{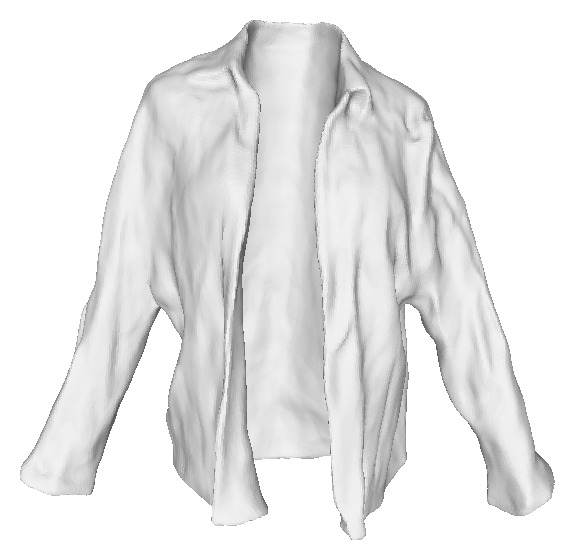}\includegraphics[width=.257in]{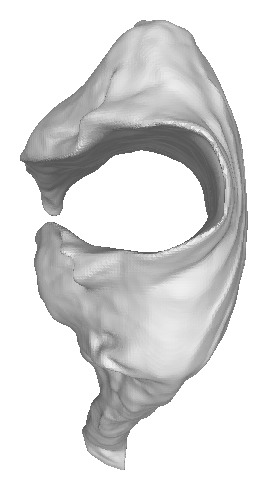}&
    \includegraphics[width=.514in]{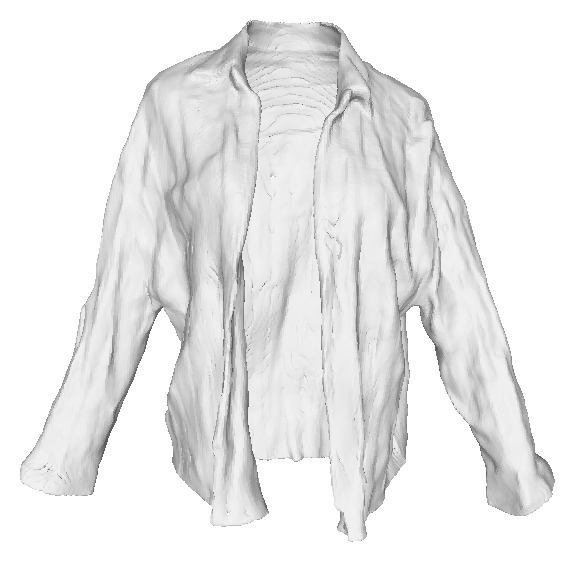}\includegraphics[width=.257in]{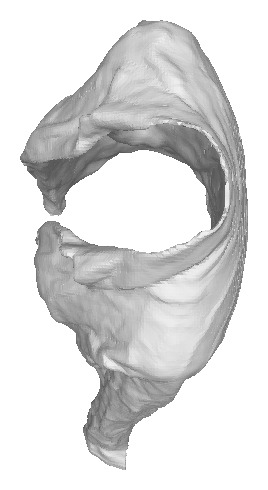}&
    \includegraphics[width=.514in]{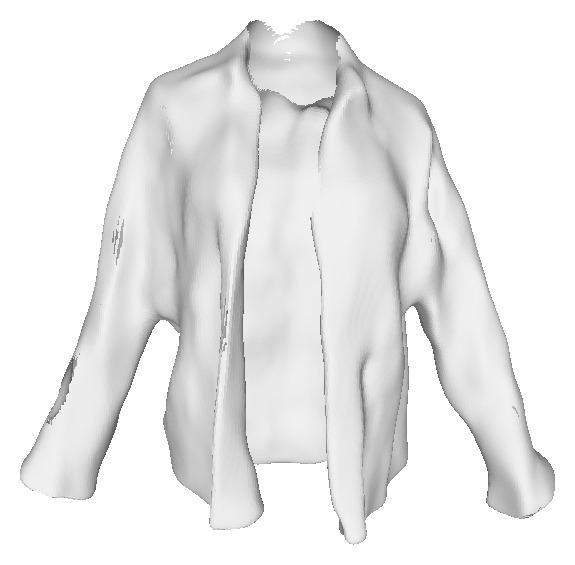}\includegraphics[width=.257in]{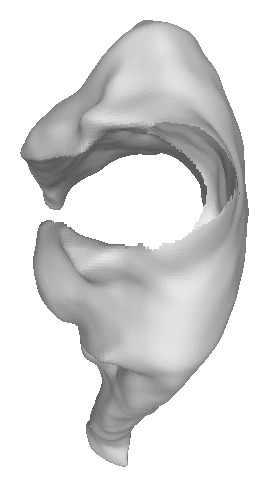}&
    \includegraphics[width=.514in]{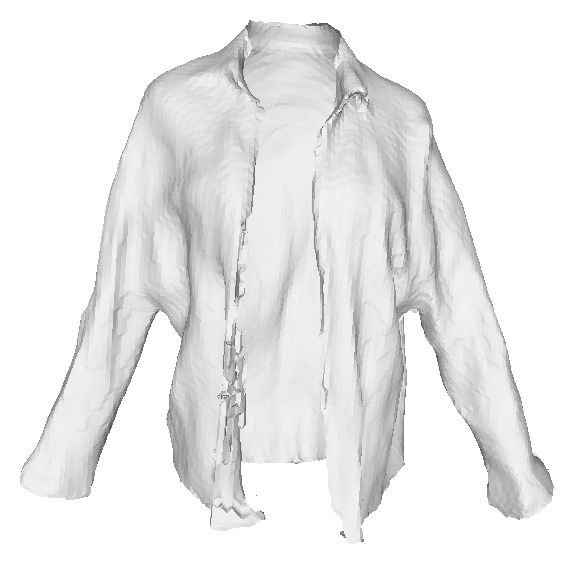}\includegraphics[width=.257in]{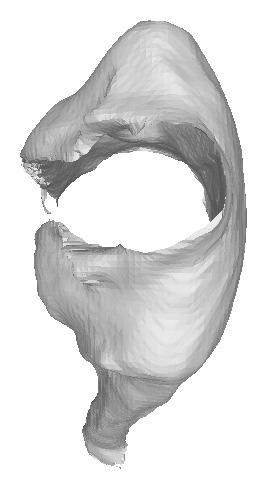}&
    \includegraphics[width=.514in]{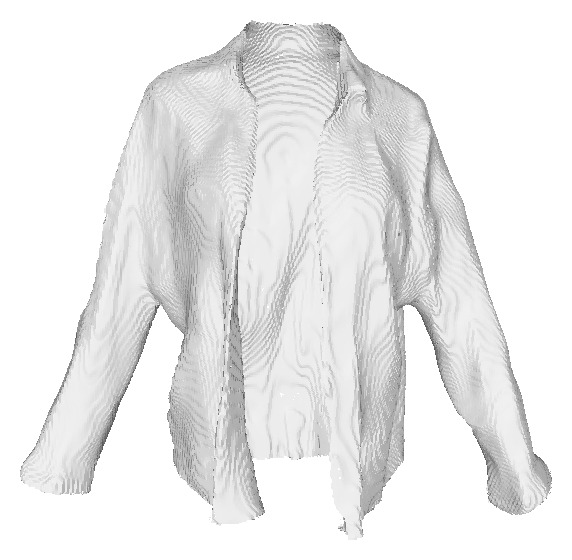}\includegraphics[width=.257in]{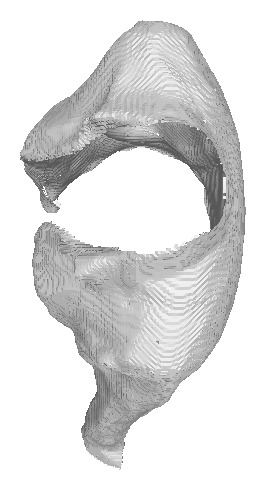} \\
    \raisebox{.20in}{Skirt1} & \includegraphics[width=.386in]{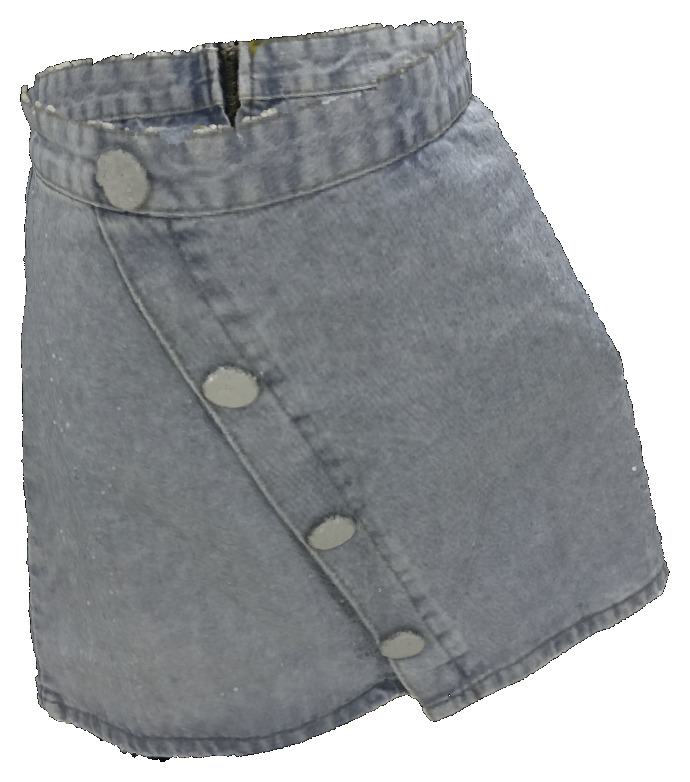}&
    \includegraphics[width=.386in]{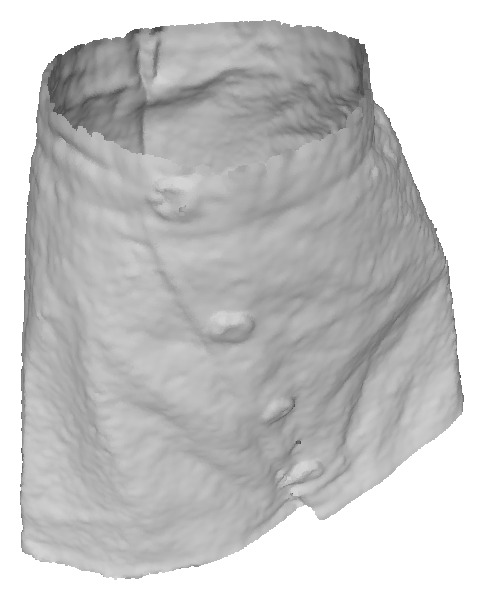}\includegraphics[width=.343in]{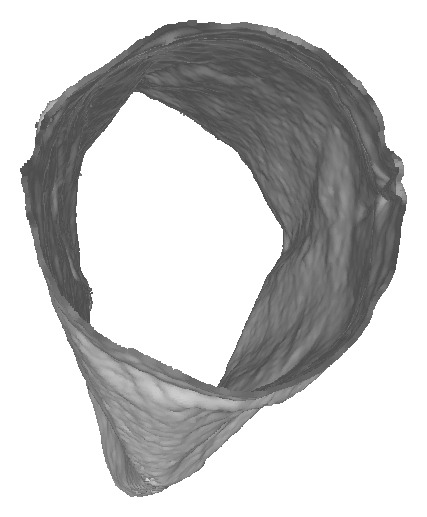}&
    \includegraphics[width=.386in]{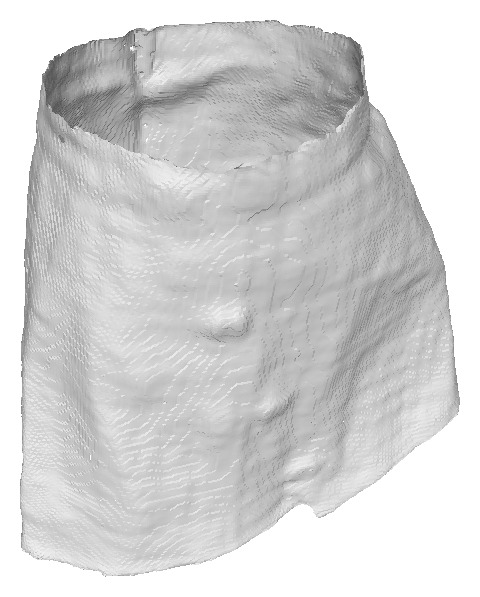}\includegraphics[width=.343in]{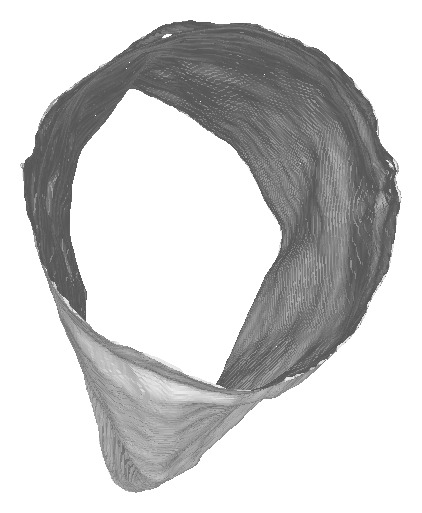}&
    \includegraphics[width=.386in]{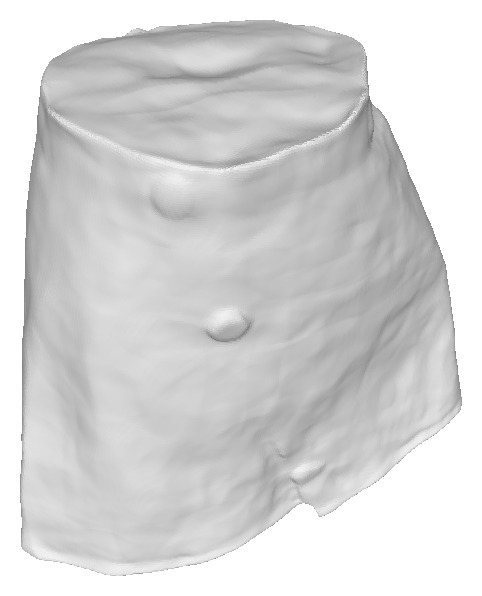}\includegraphics[width=.343in]{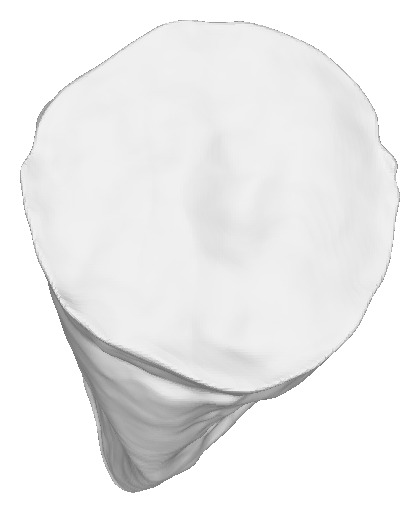}&
    \includegraphics[width=.386in]{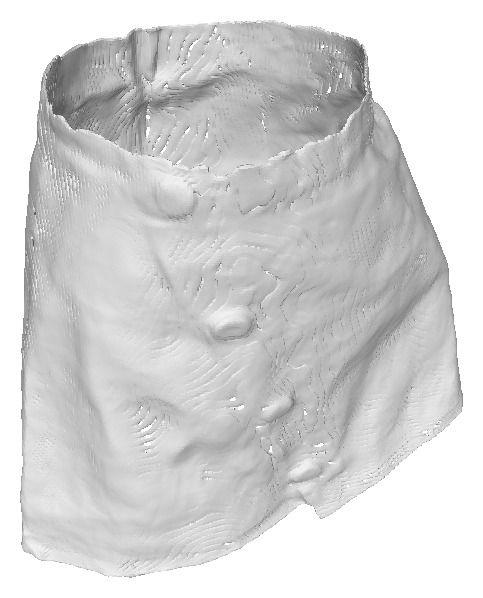}\includegraphics[width=.343in]{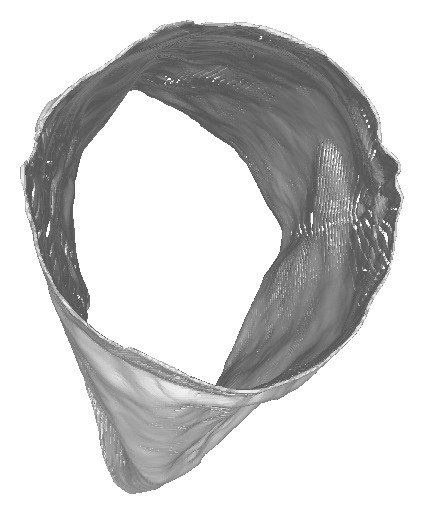}&
    \includegraphics[width=.386in]{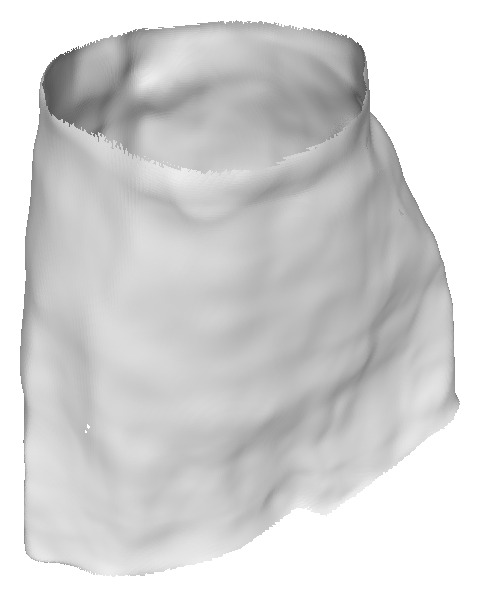}\includegraphics[width=.343in]{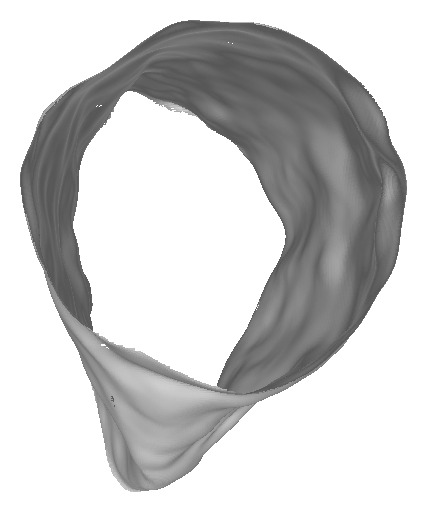}&
    \includegraphics[width=.386in]{neuraludf/591}\includegraphics[width=.343in]{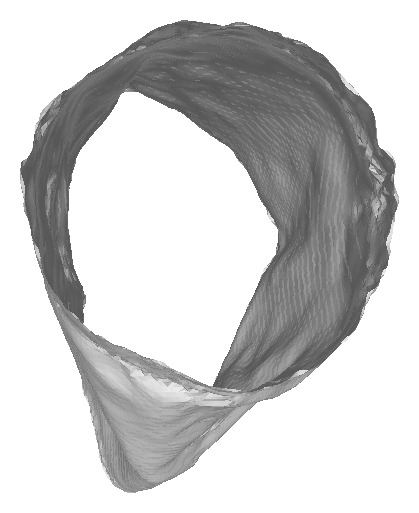}&
    \includegraphics[width=.386in]{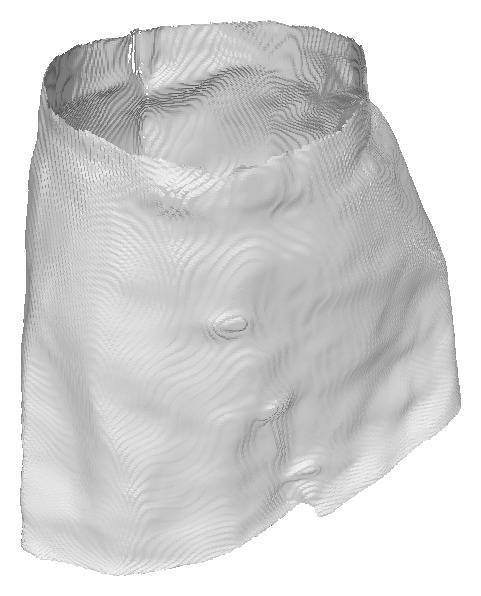}\includegraphics[width=.343in]{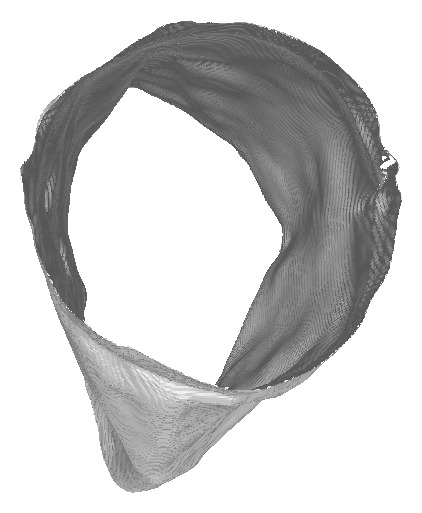} \\
    \raisebox{.24in}{SS-C0} & \includegraphics[width=.42in]{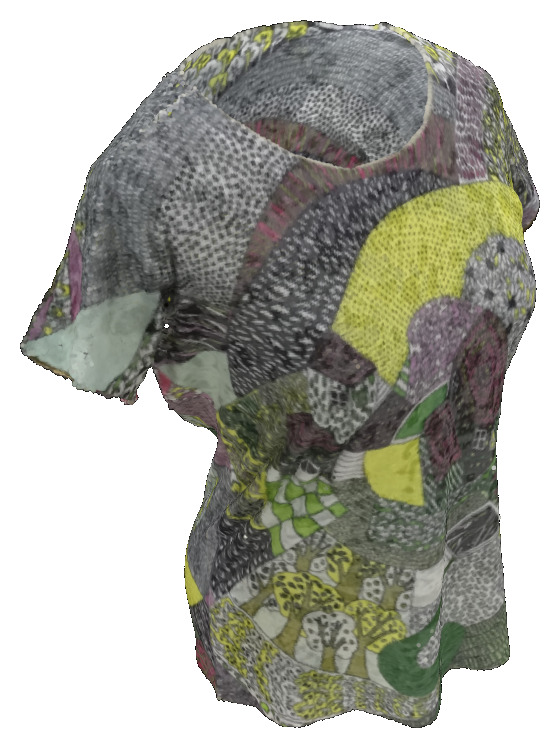}&
    \includegraphics[width=.446in]{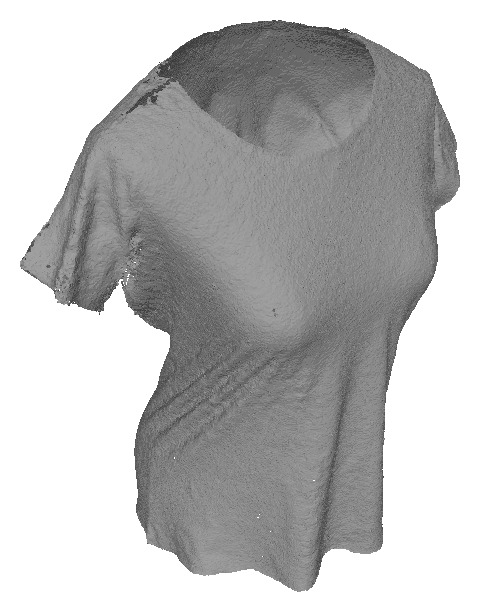}\includegraphics[width=.257in]{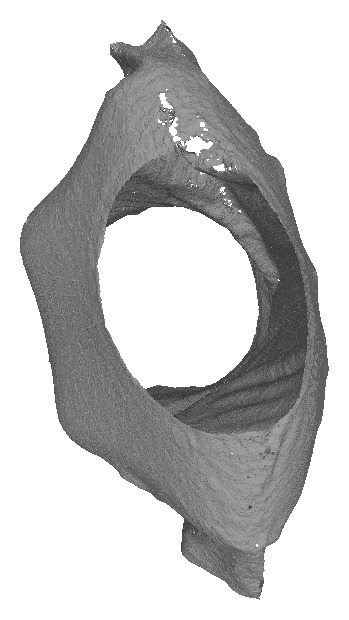}&
    \includegraphics[width=.446in]{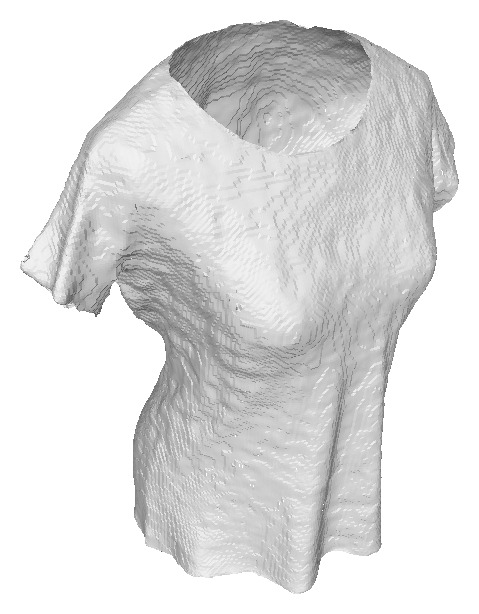}\includegraphics[width=.257in]{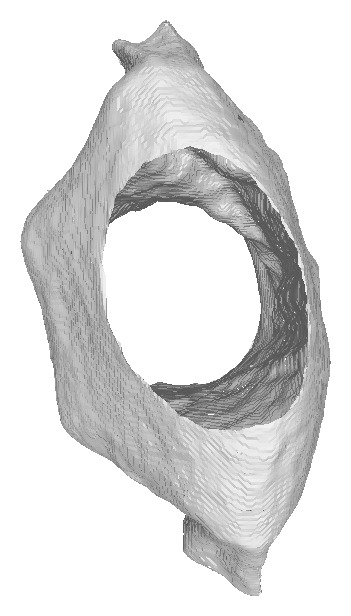}&
    \includegraphics[width=.446in]{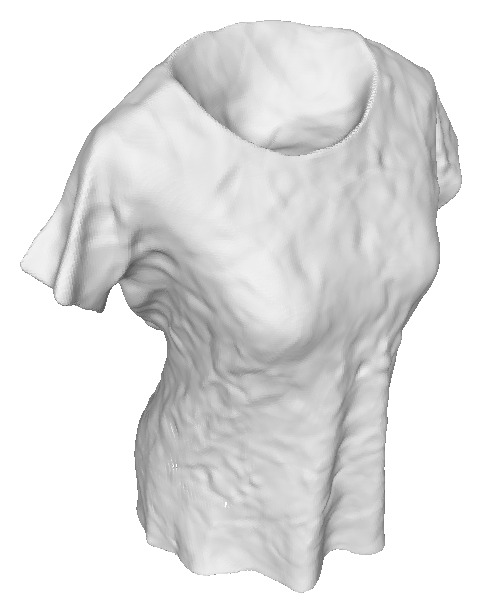}\includegraphics[width=.257in]{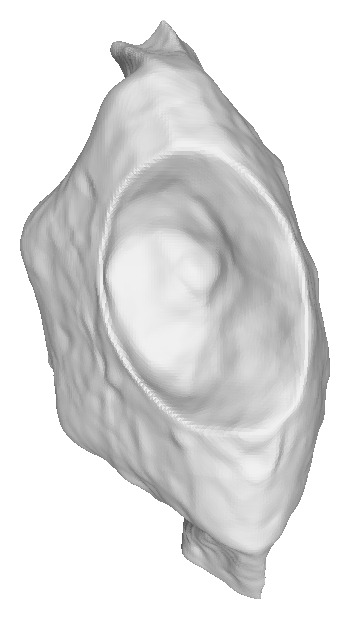}&
    \includegraphics[width=.446in]{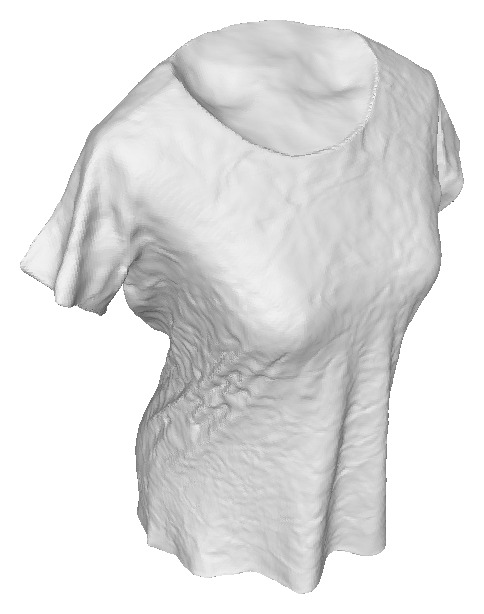}\includegraphics[width=.257in]{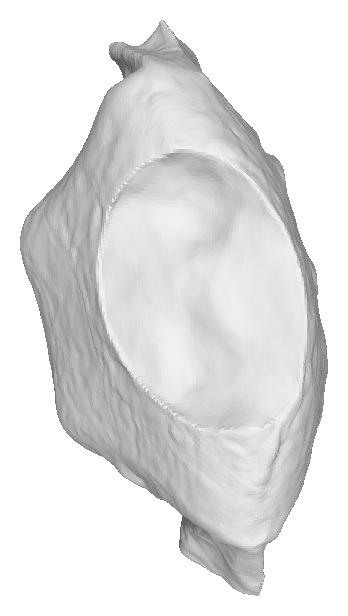}&
    \includegraphics[width=.446in]{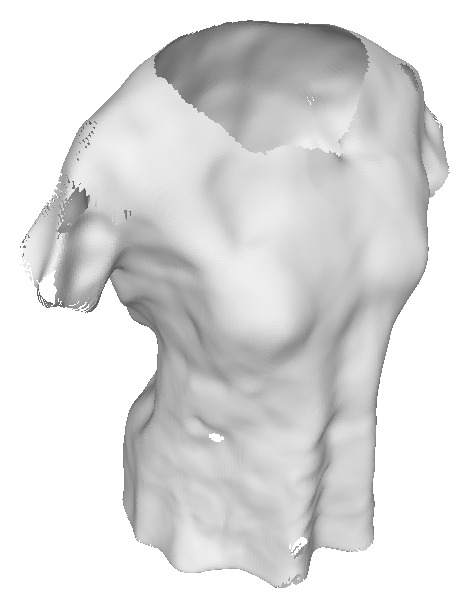}\includegraphics[width=.257in]{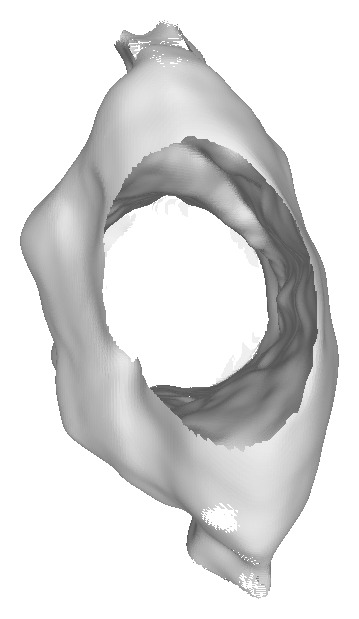}&
    \includegraphics[width=.446in]{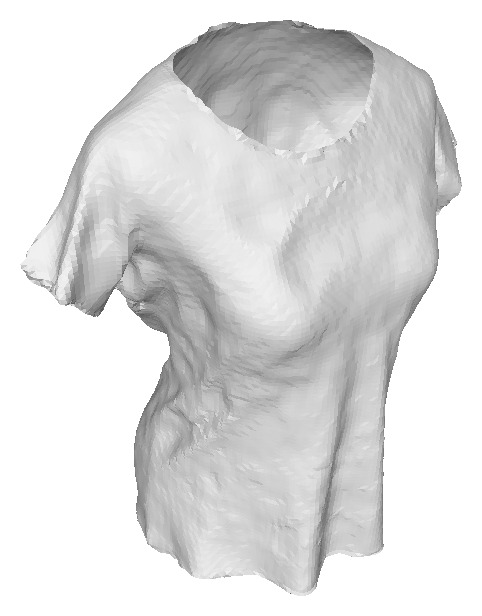}\includegraphics[width=.257in]{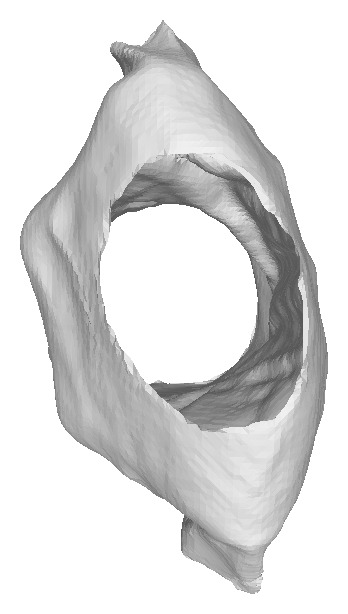}&
    \includegraphics[width=.446in]{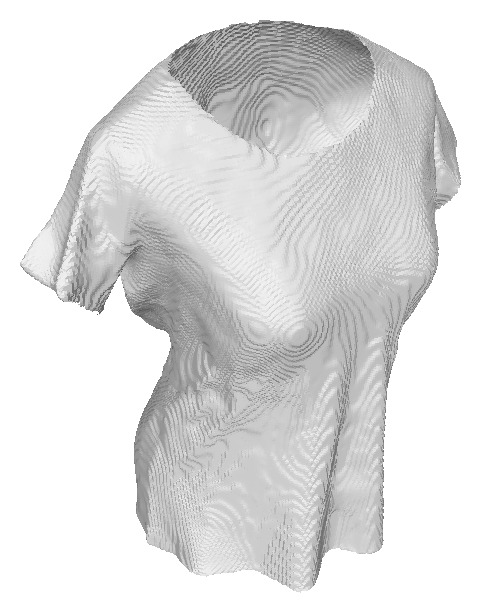}\includegraphics[width=.257in]{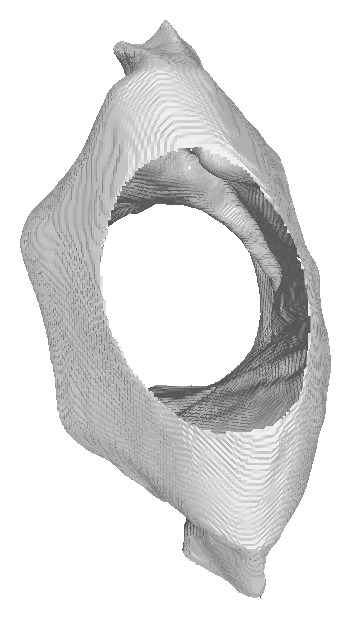} \\
\end{tabular}
\caption{The remaining qualitative comparisons with VolSDF~\cite{Yariv2021}, NeuS~\cite{Wang2021}, NeAT~\cite{Meng2023} (with mask supervision), NeuralUDF~\cite{Long2023} and NeUDF~\cite{Liu2023NeUDF} on the DeepFashion3D~\cite{Zhu2020} dataset.}

\label{fig:df3d-more}
\end{figure*}

We present the remaining results on DeepFashion3D~\cite{Zhu2020} dataset in Figure~\ref{fig:vs_udf_more} for UDFs and Figure~\ref{fig:df3d-more} for reconstructed models.
The UDFs of NeuralUDF exhibit apparent oscillation. The UDFs of NeUDF are nearly closed possibly resulting in watertight models. In contrast, our learned UDFs are closest to the ground truth.

NeuralUDF~\cite{Long2023} performs poorly on some cases in Figure~\ref{fig:df3d-more}, possibly due to its complicated visibility indicator function. SDF-based methods such as VolSDF~\cite{Yariv2021} and NeuS~\cite{Wang2021} produce closed or double-cover models, leading to large reconstruction loss. Note that the UDF-based method NeUDF~\cite{Liu2023NeUDF} also fails to learn open models in case SS-D0. The reason is that the learned UDF of NeUDF is usually nearly closed, so it is liable to generate watertight models.

\begin{figure*}[p]
\centering
\setlength\tabcolsep{3pt}
\begin{tabular}{cccccc}
    & Reference Image & Ours & NeUDF & NeuralUDF & NeAT \\
    \raisebox{.3in}{37} & \includegraphics[width=1in]{gt/dtu_scan37} &
    \includegraphics[width=1in]{ours/dtu_scan37} &
    \includegraphics[width=1in]{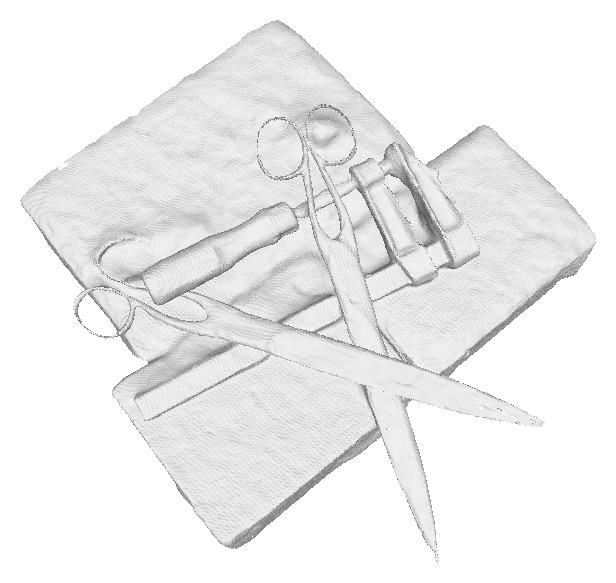} &
    \includegraphics[width=1in]{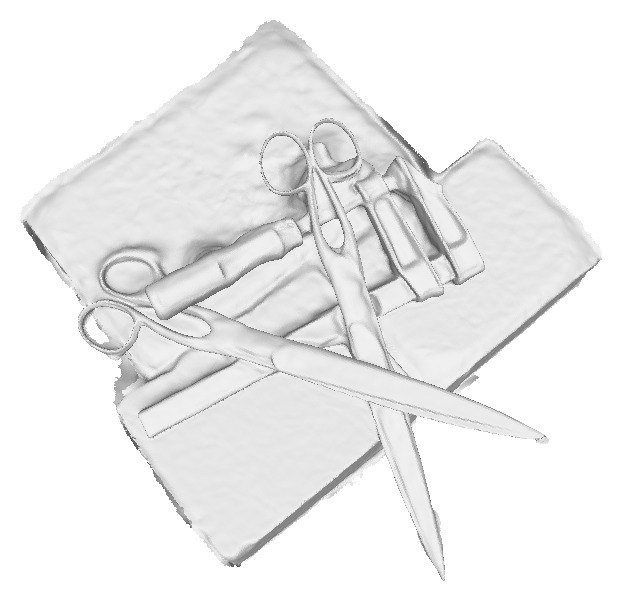} &
    \includegraphics[width=1in]{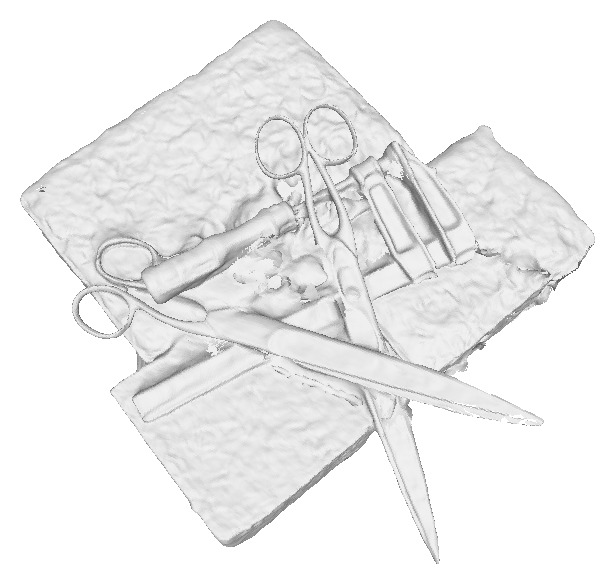} \\
    \raisebox{.3in}{69} & \includegraphics[width=1in]{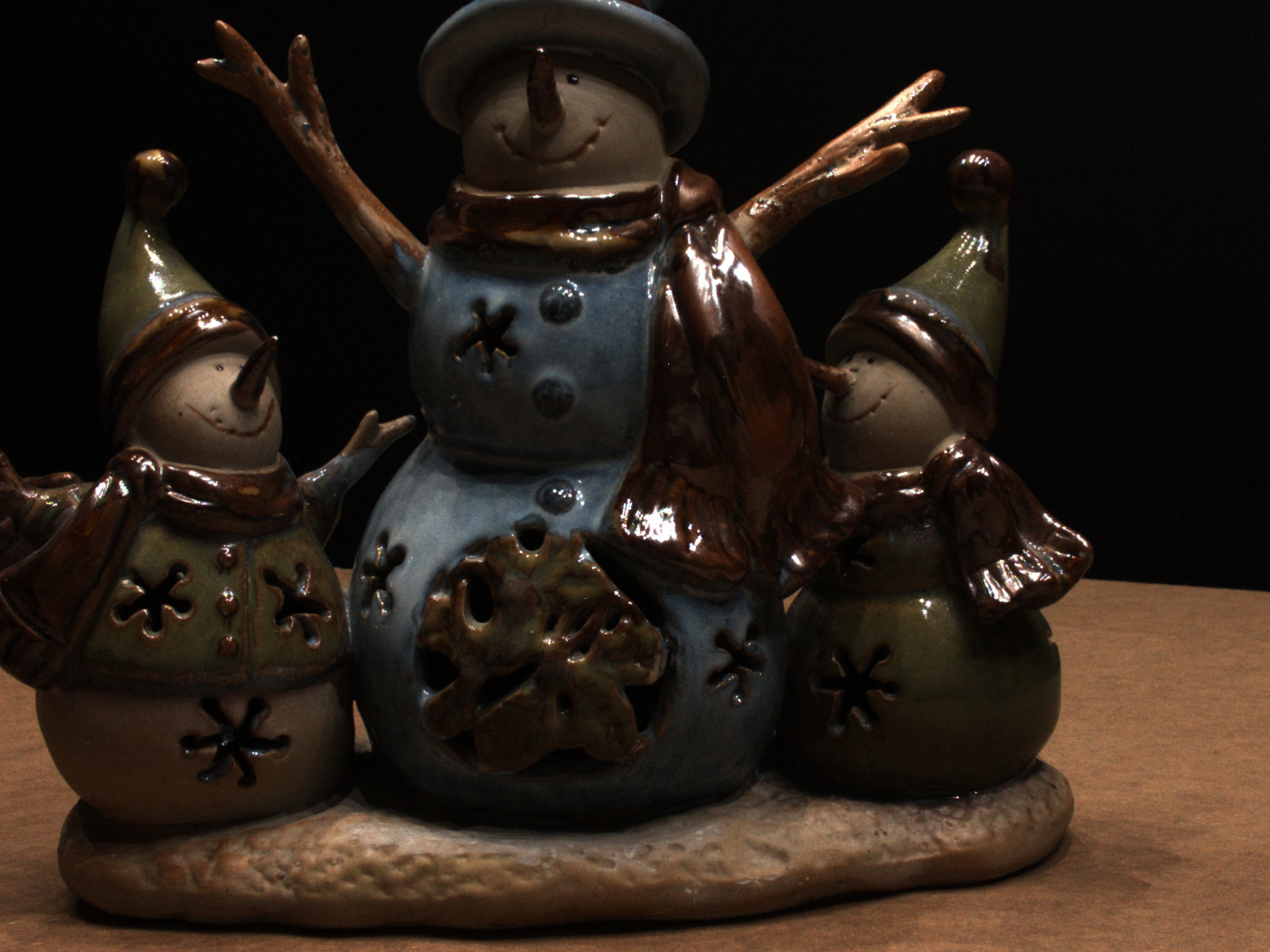} &
    \includegraphics[width=.83in]{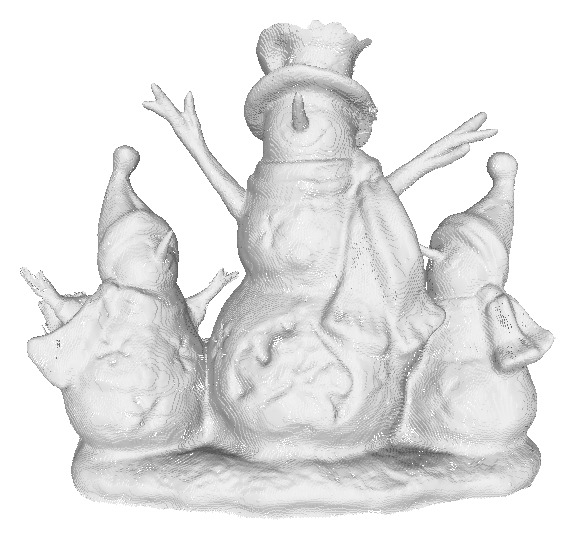} &
    \includegraphics[width=.83in]{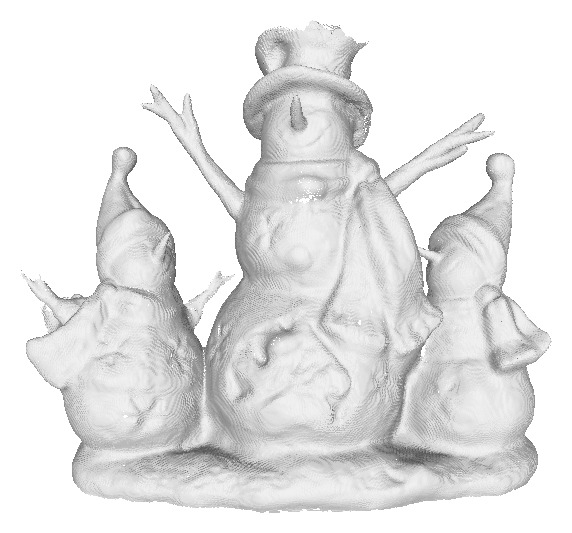} &
    \includegraphics[width=.83in]{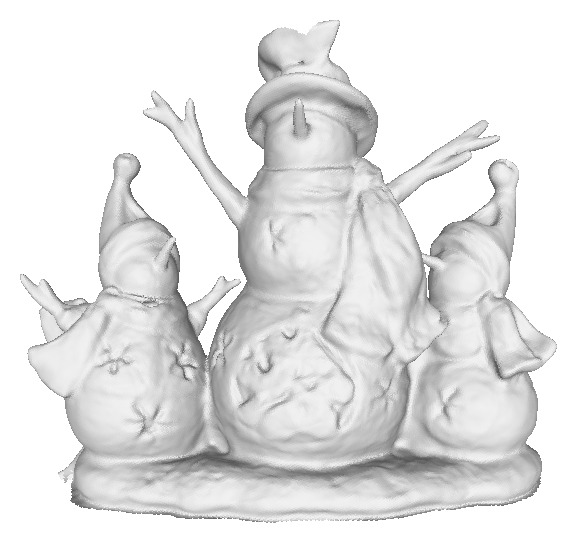} &
    \includegraphics[width=.83in]{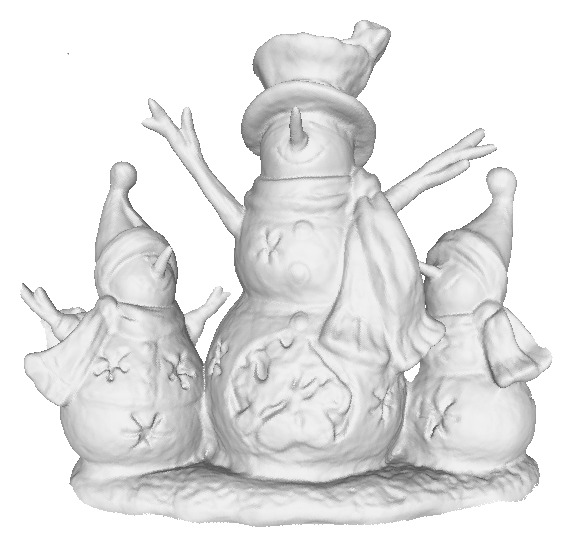} \\
    \raisebox{.3in}{97} & \includegraphics[width=1in]{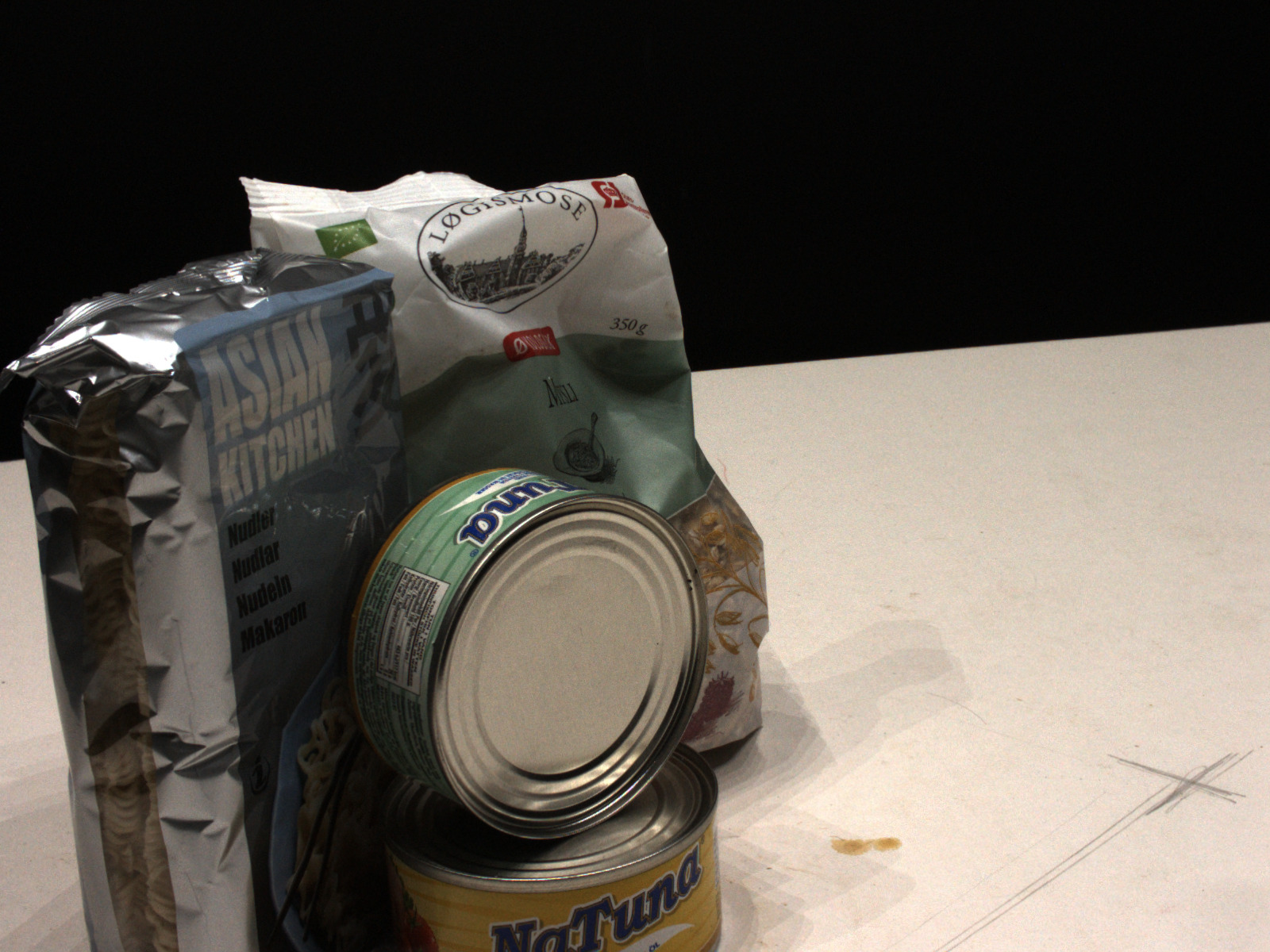} &
    \includegraphics[width=.6in]{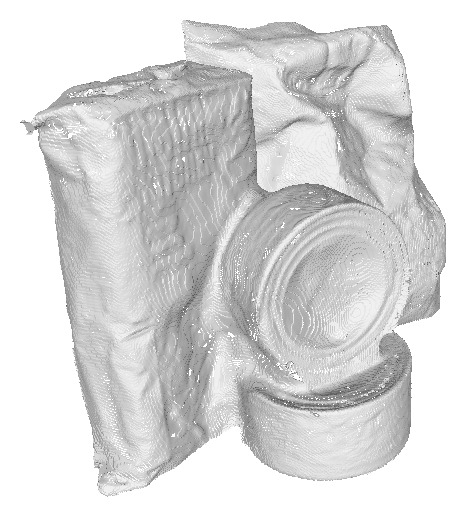} &
    \includegraphics[width=.6in]{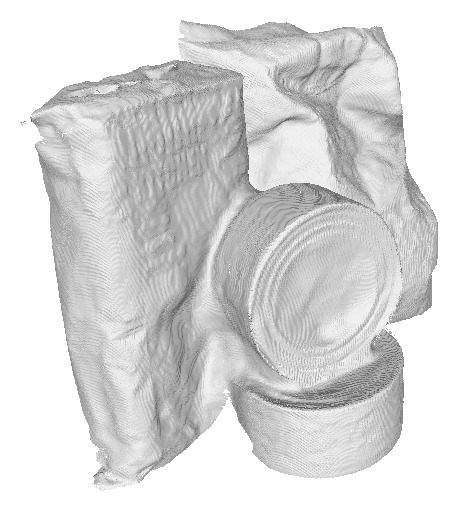} &
    \includegraphics[width=.6in]{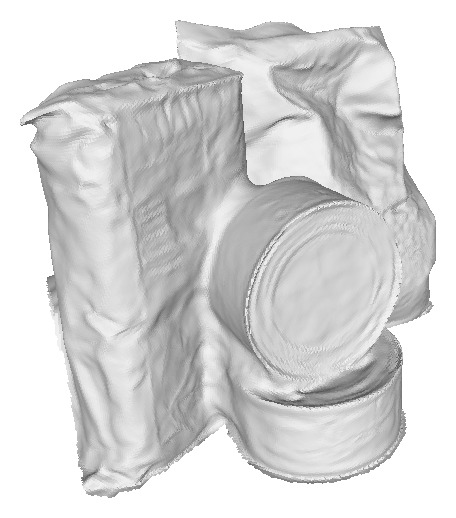} &
    \includegraphics[width=.6in]{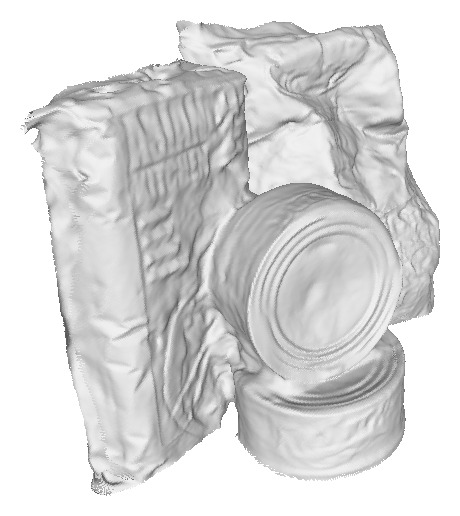} \\
    \raisebox{.3in}{105} & \includegraphics[width=1in]{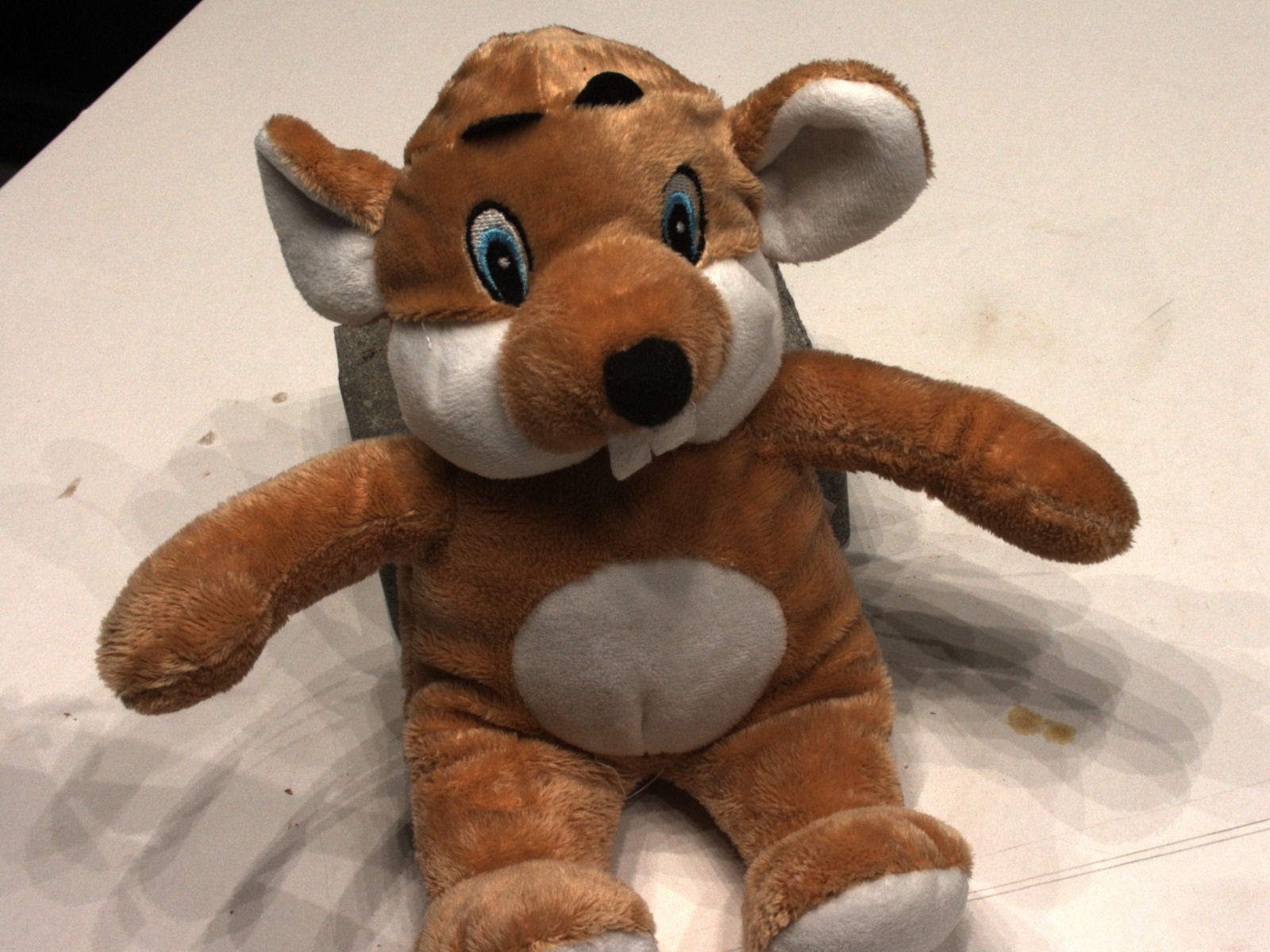} &
    \includegraphics[width=.83in]{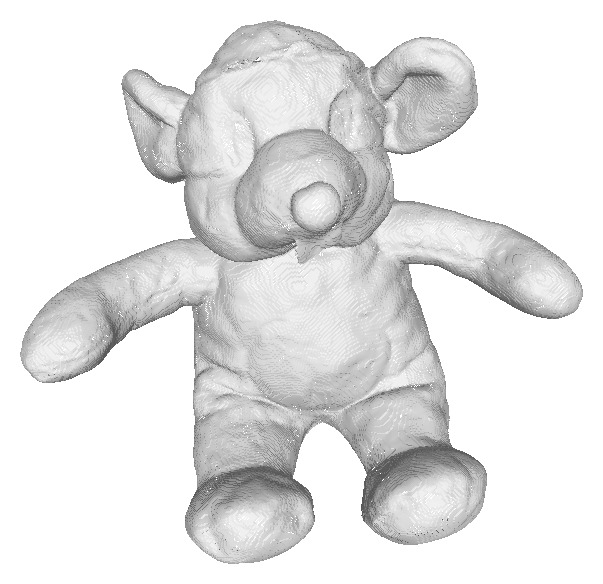} &
    \includegraphics[width=.83in]{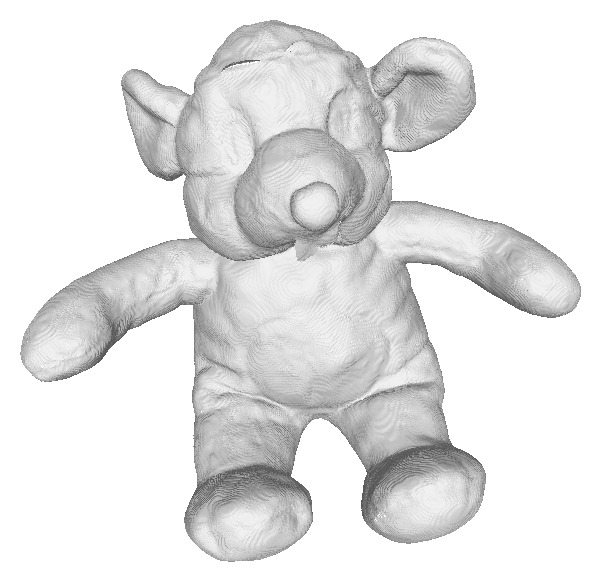} &
    \includegraphics[width=.83in]{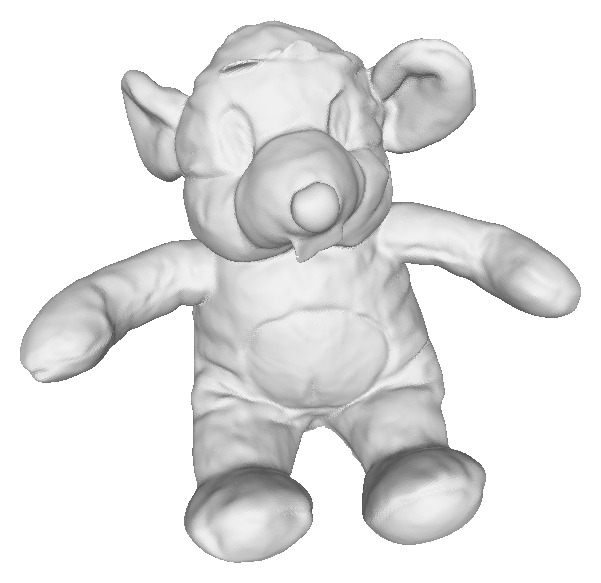} &
    \includegraphics[width=.83in]{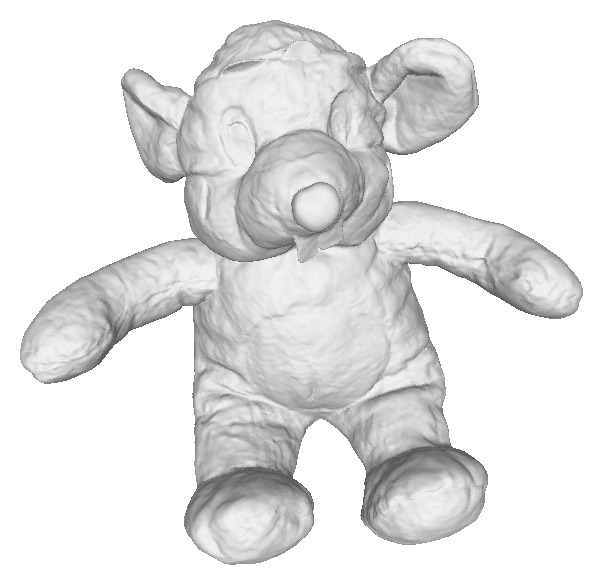} \\
    \raisebox{.3in}{106} & \includegraphics[width=1in]{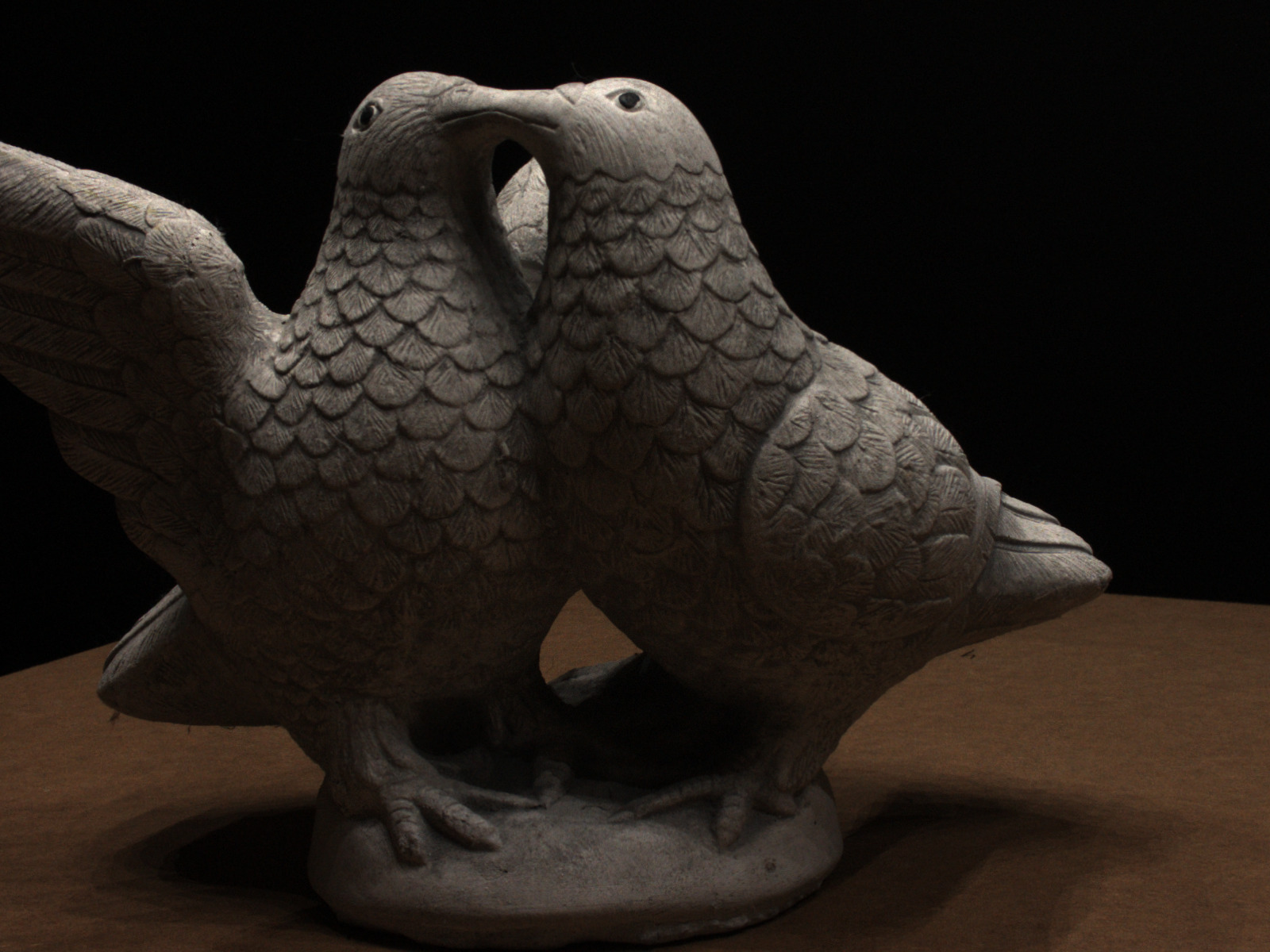} &
    \includegraphics[width=1in]{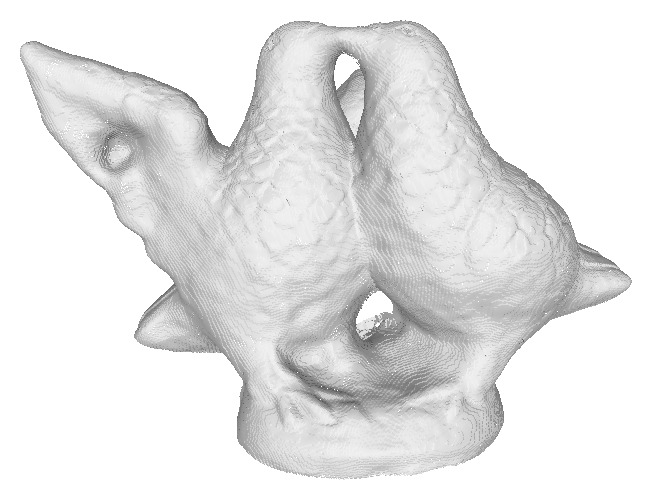} &
    \includegraphics[width=1in]{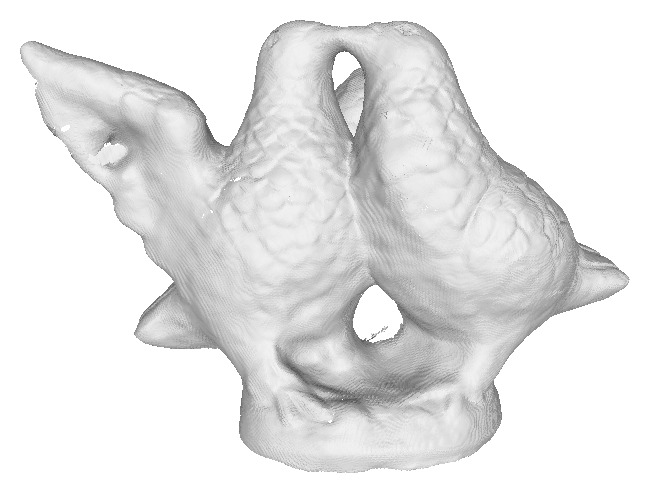} &
    \includegraphics[width=1in]{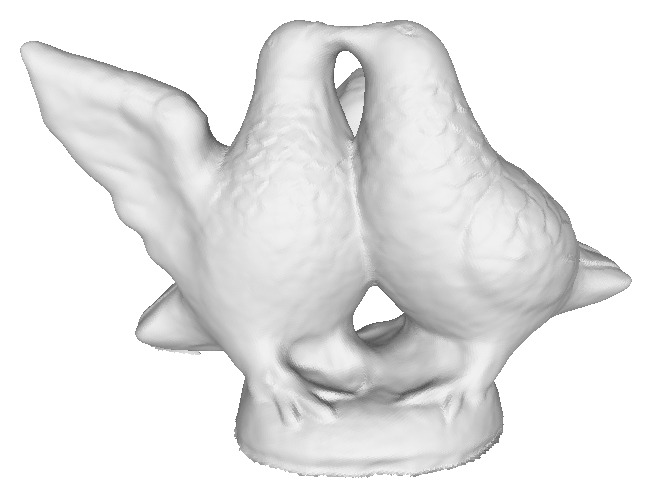} &
    \includegraphics[width=1in]{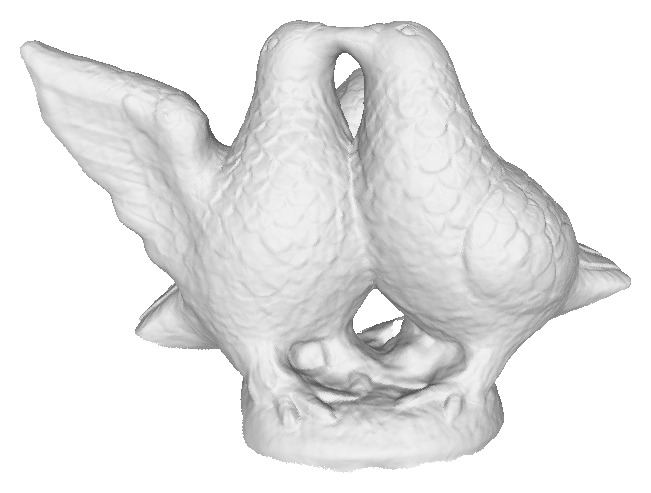} \\
    \raisebox{.3in}{114} & \includegraphics[width=1in]{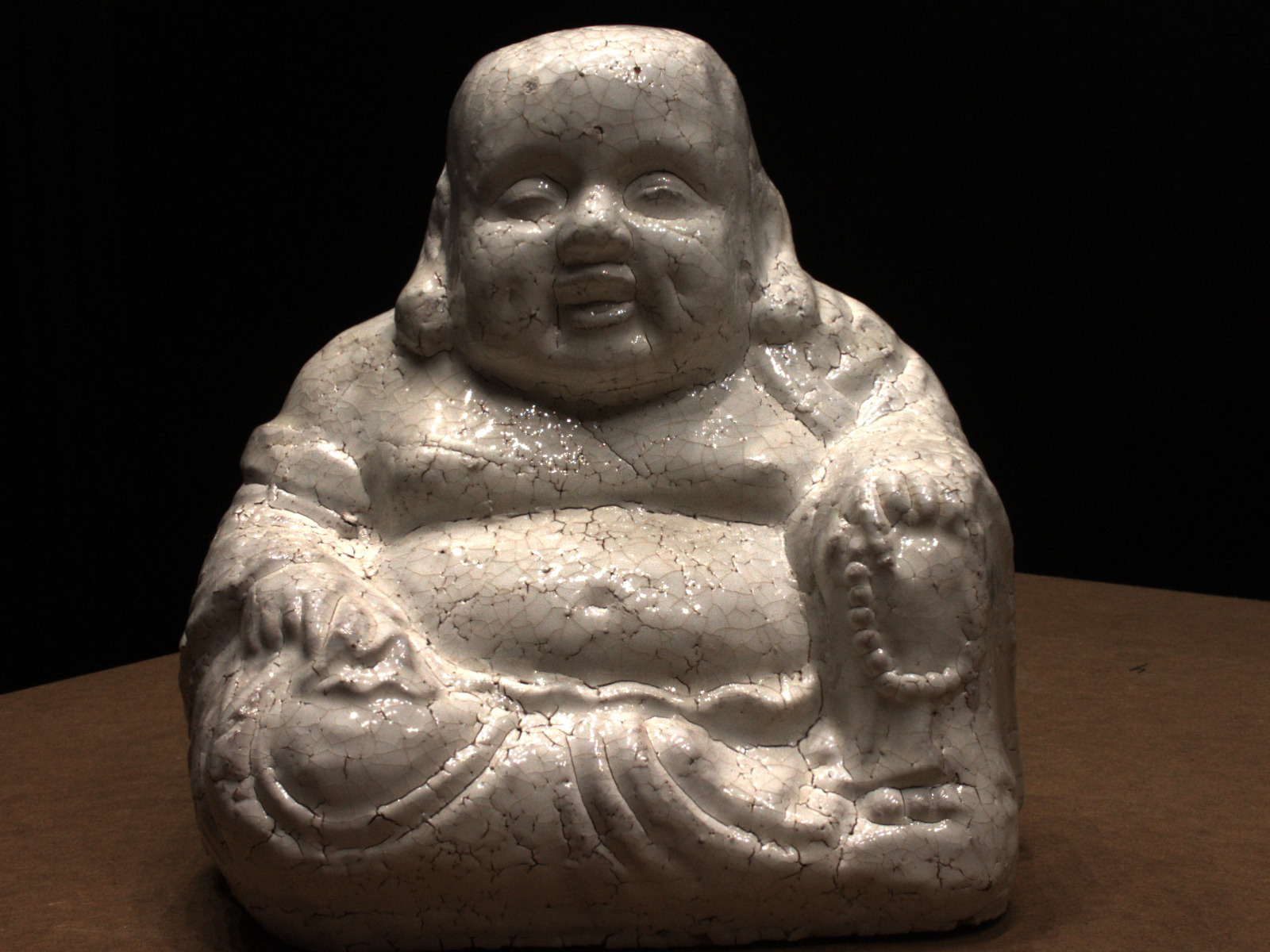} &
    \includegraphics[width=.7in]{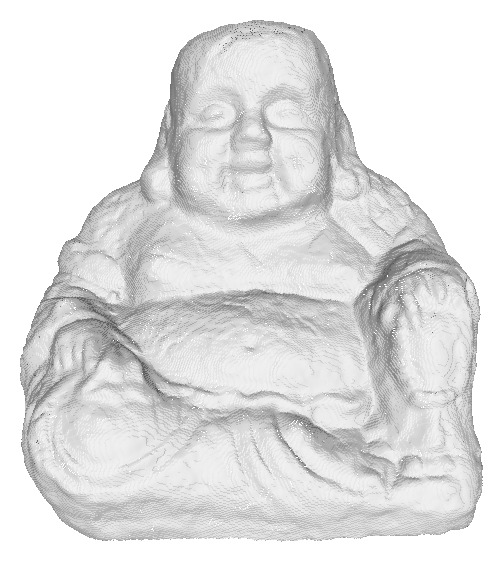} &
    \includegraphics[width=.7in]{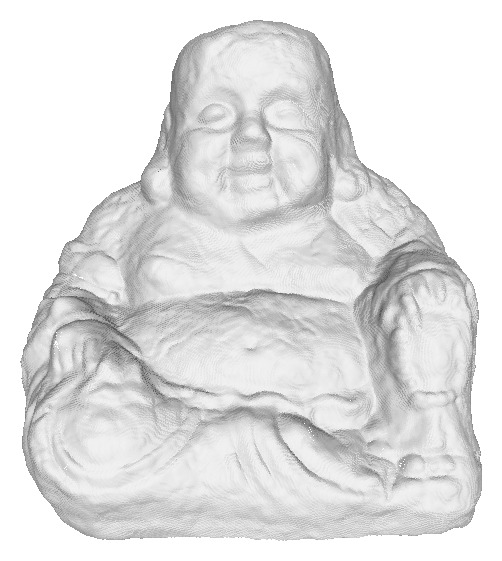} &
    \includegraphics[width=.7in]{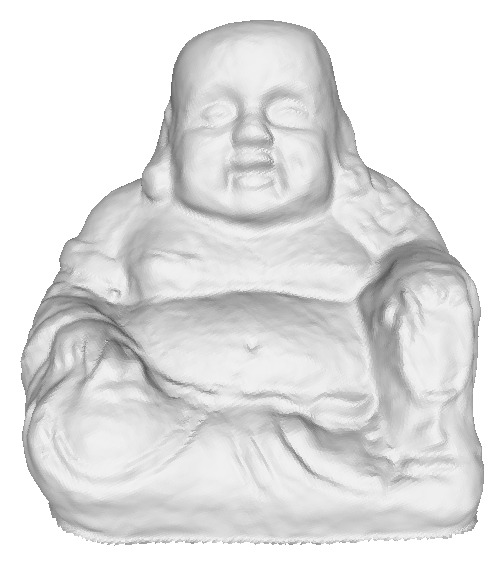} &
    \includegraphics[width=.7in]{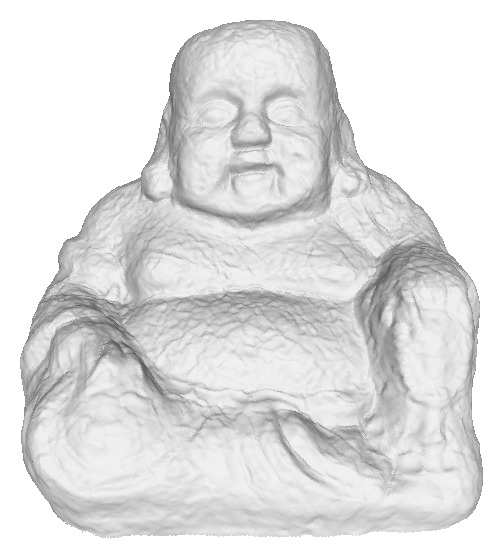} \\
    \raisebox{.42in}{bear} & \includegraphics[width=.75in]{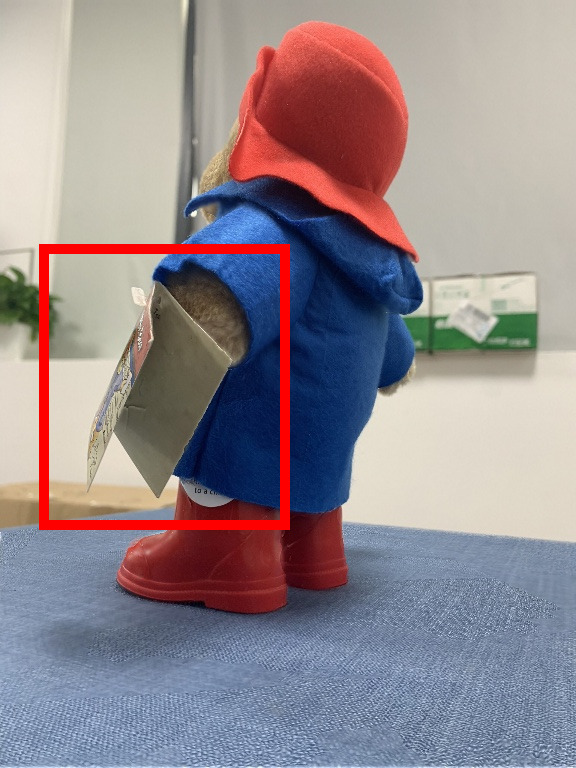} &
    \includegraphics[width=.52in]{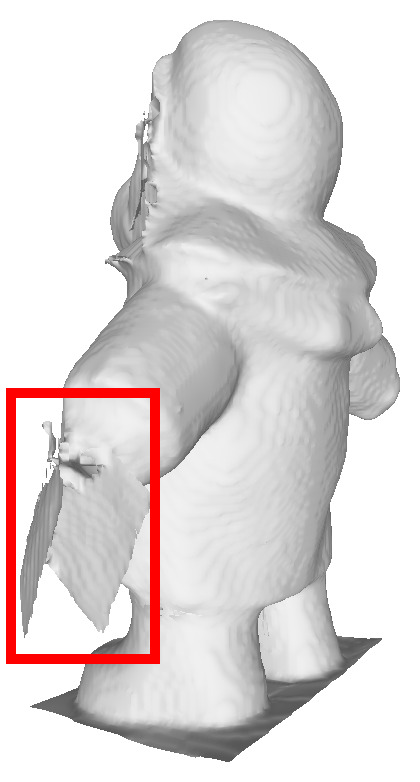} &
    \includegraphics[width=.52in]{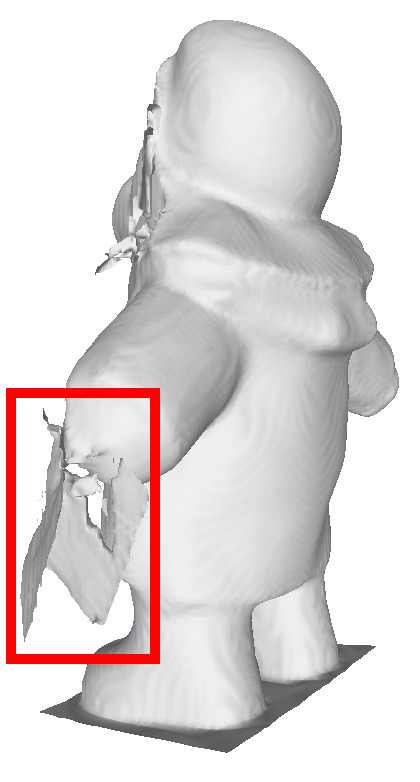} &
    \includegraphics[width=.52in]{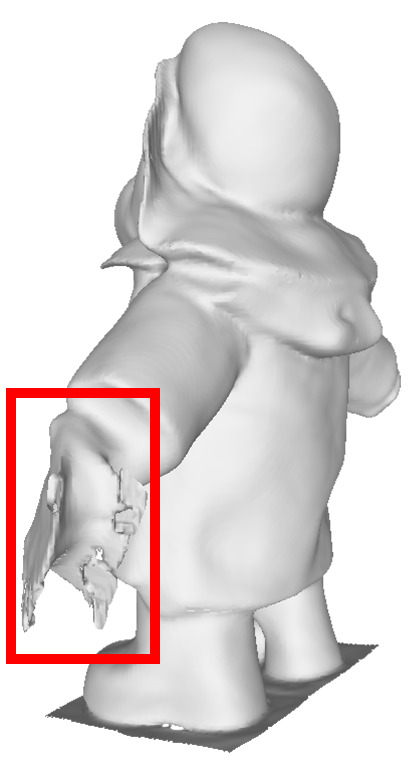} &
    \includegraphics[width=.52in]{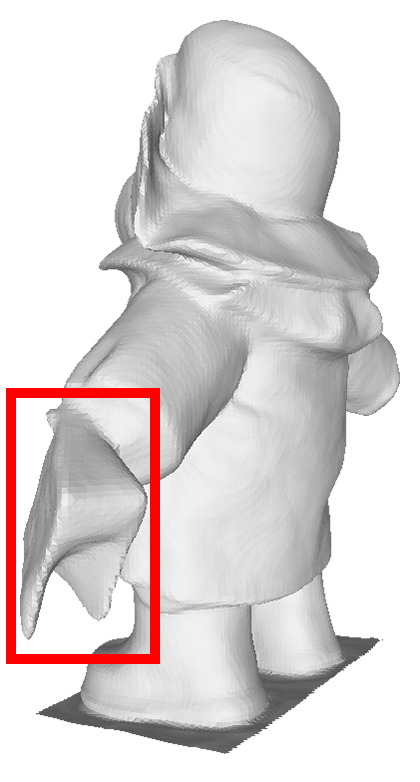} \\
    \raisebox{.3in}{man} & \includegraphics[width=1in]{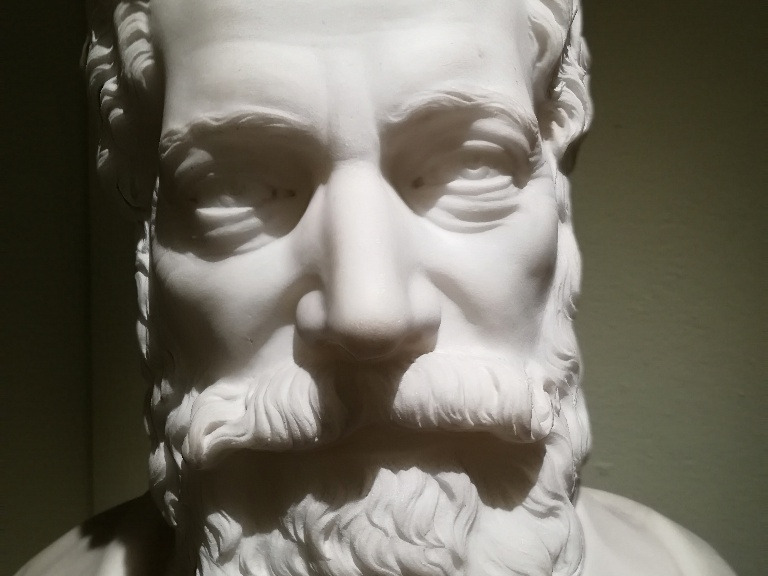} &
    \includegraphics[width=.45in]{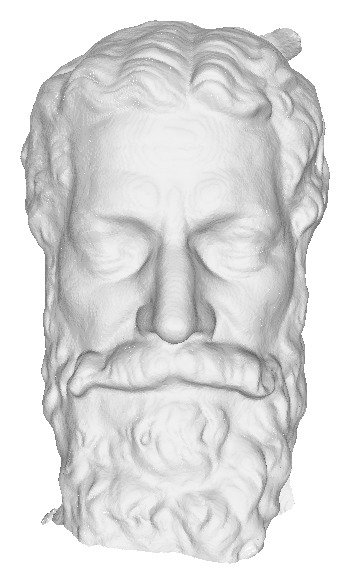} &
    \includegraphics[width=.45in]{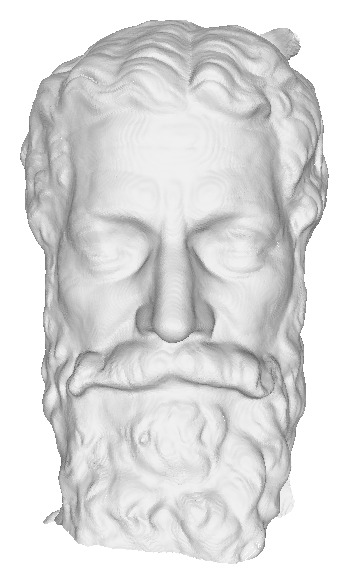} &
    \includegraphics[width=.45in]{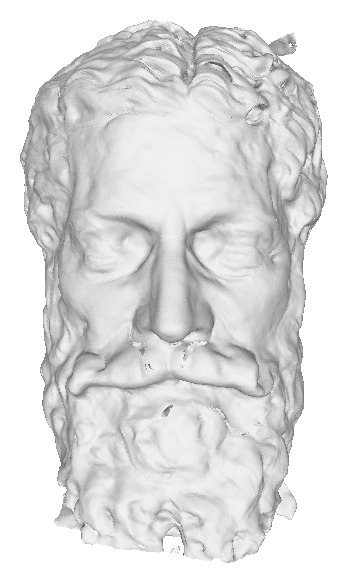} &
    \includegraphics[width=.45in]{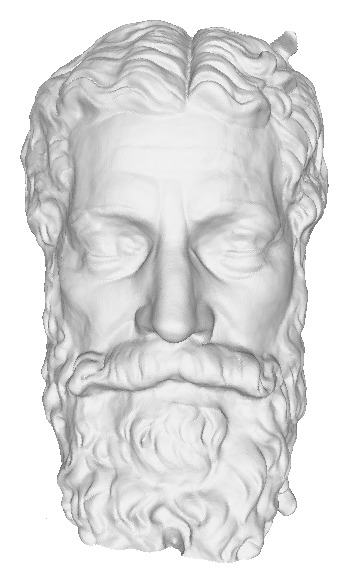} \\
    \raisebox{.3in}{sculpture} & \includegraphics[width=1in]{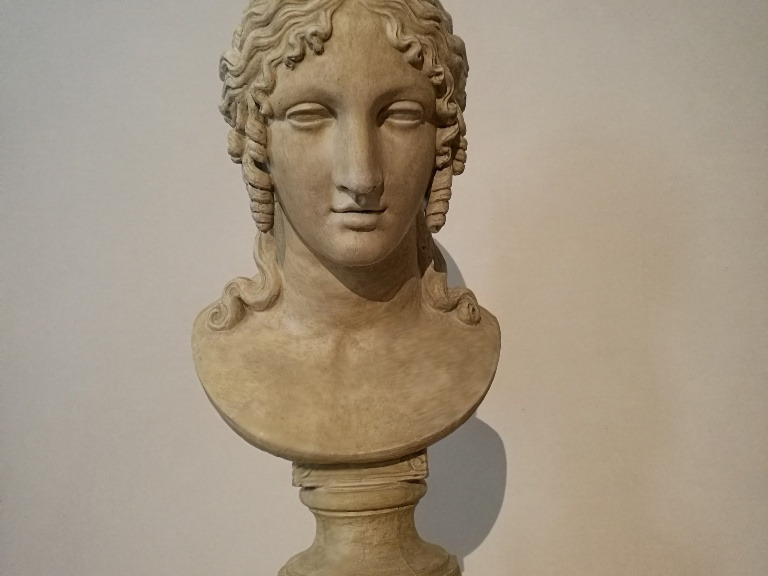} &
    \includegraphics[width=.35in]{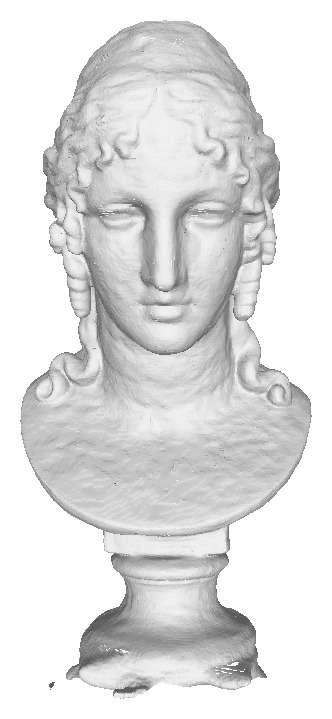} &
    \includegraphics[width=.35in]{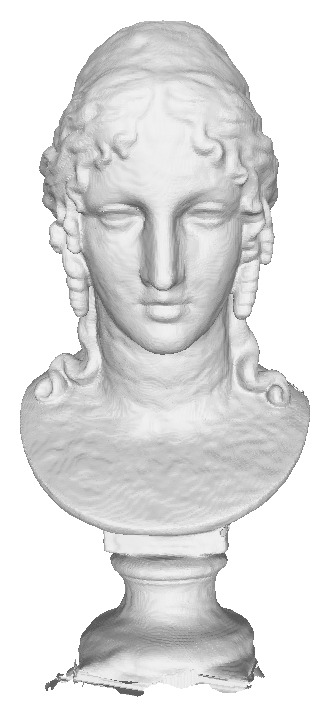} &
    \includegraphics[width=.35in]{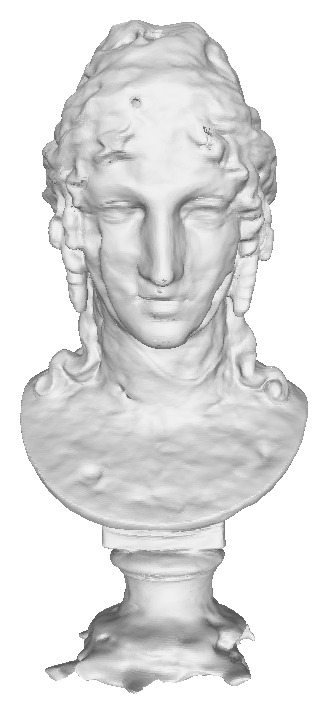} &
    \includegraphics[width=.35in]{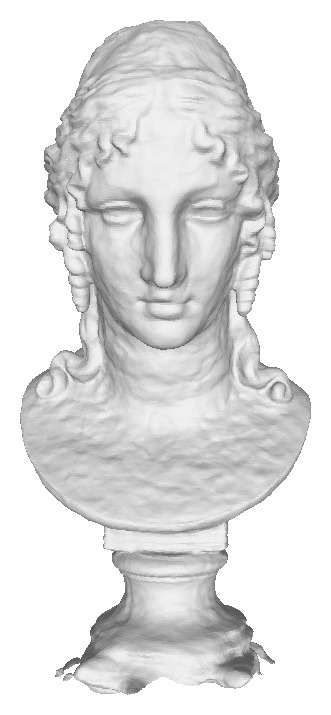} \\
    
\end{tabular}
\caption{Qualitative comparisons with NeAT~\cite{Meng2023}, NeuralUDF~\cite{Long2023} and NeUDF~\cite{Liu2023NeUDF} on the DTU~\cite{Jensen2014} dataset and BlendedMVS~\cite{Yao_2020_CVPR} dataset.}

\label{fig:dtu-blendedmvs-more}
\end{figure*}

We also present the results on DTU~\cite{Jensen2014} dataset and BlendedMVS~\cite{Yao_2020_CVPR} in Figure~\ref{fig:dtu-blendedmvs-more}. For DTU dataset where quantitative comparisons are feasible, our Stage 2 optimization generally improves the reconstruction results (measured by Chamfer distances) of NeUDF~\cite{Liu2023NeUDF} by around 10\%. The reason is presented in the main text. For BlendedMVS dataset, we encourage readers to focus on the ``bear'' data. The brochure held by the bear (marked in red box) is an open part of the model. NeuralUDF~\cite{Long2023} and NeAT~\cite{Meng2023}, both of which use SDF implicitly or explicitly, as explained in the main text, fail to reconstruct the open brochure. NeUDF~\cite{Liu2023NeUDF} correctly reconstructs the brochure as a single-layer open surface but with large holes. Our method can generate a visually better open surface for such parts in real-life captured data.

\end{document}